\documentclass[10pt,twocolumn,letterpaper]{article}
\usepackage{titling}

\usepackage{iccv}
\usepackage{times}
\usepackage{epsfig}
\usepackage{graphicx}
\usepackage{amsmath}
\usepackage{amssymb}
\usepackage[accsupp]{axessibility}
\usepackage{epstopdf}
\usepackage{epsfig}
\usepackage{stfloats}
\usepackage{url}
\usepackage{times}
\usepackage{amsthm}
\usepackage{cite}
\usepackage{color}
\usepackage{float}
\usepackage{caption}
\usepackage{collcell}
\usepackage{xstring}
\usepackage{multirow}
\usepackage{arydshln}
\usepackage{grffile}
\usepackage{setspace}

\makeatletter
\@namedef{ver@everyshi.sty}{}
\makeatother
\usepackage{tikz}

\usepackage[pagebackref=true,breaklinks=true,letterpaper=true,colorlinks,bookmarks=false]{hyperref}
\usepackage[capitalize]{cleveref}

\iccvfinalcopy 

\def\iccvPaperID{10190} 

\ificcvfinal\fi

\begin{document}

\title{Multi-Directional Subspace Editing in Style-Space}
\date{\vspace{-3ex}}
\author{Chen Naveh \hspace{5em} Yacov Hel-Or\\
School of Computer Science, Reichman University\\
{\tt\small navehchen1@gmail.com \hspace{5em} toky@runi.ac.il}
}

\newcommand*{\imwidth}{0.16}

\twocolumn[{
\maketitle
\begin{center}
    \captionsetup{type=figure[htb]}
    \begin{tabular}{cc|cccc}
         \rotatebox{90}{\hspace{0.29in}Gender} &
         \includegraphics[width=\imwidth\linewidth]{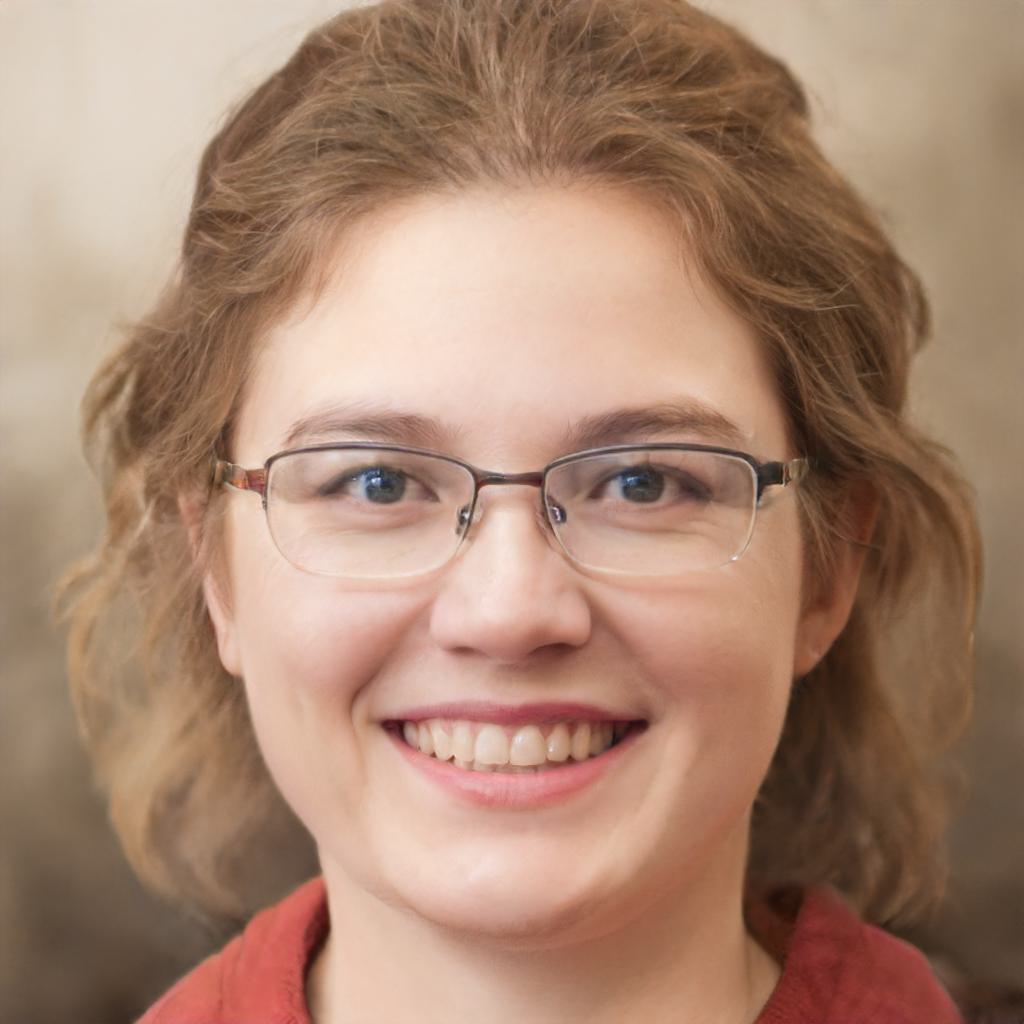} &
         \includegraphics[width=\imwidth\linewidth]{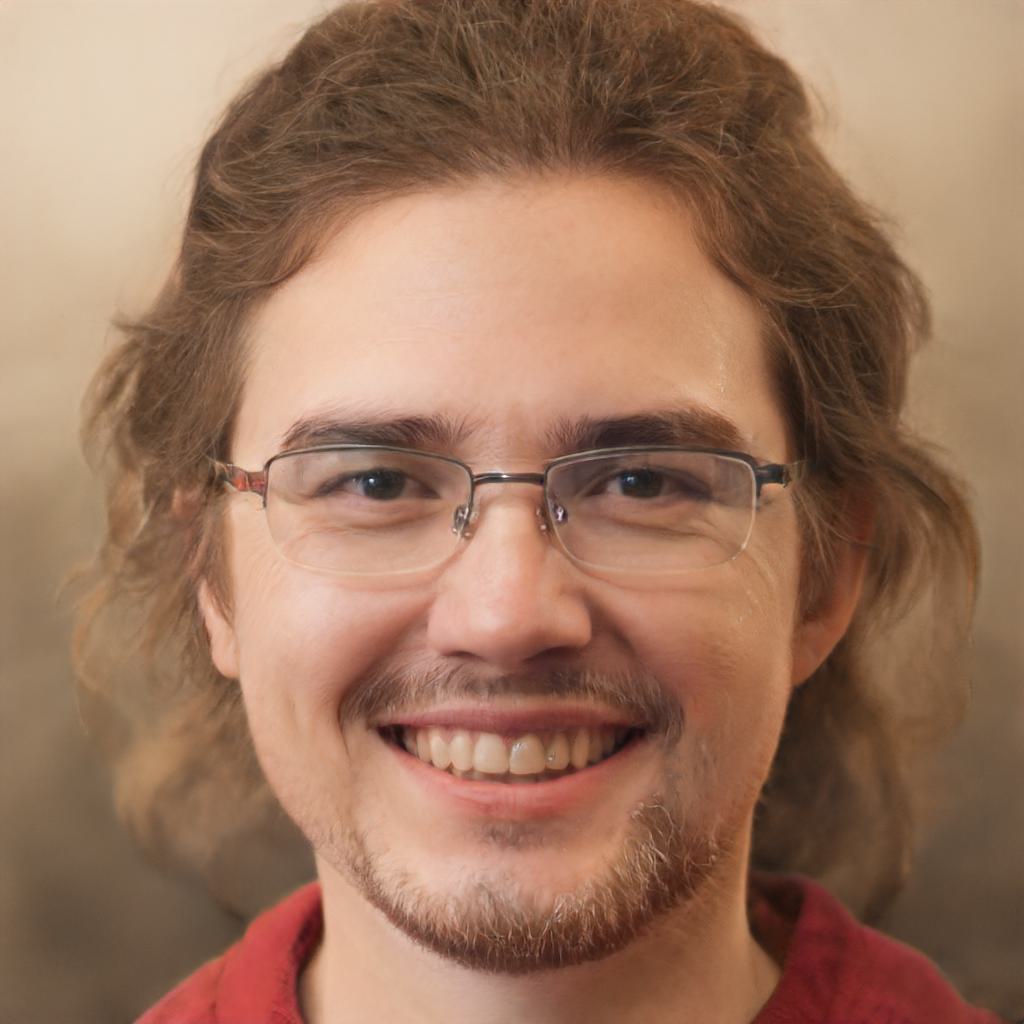} &
         \includegraphics[width=\imwidth\linewidth]{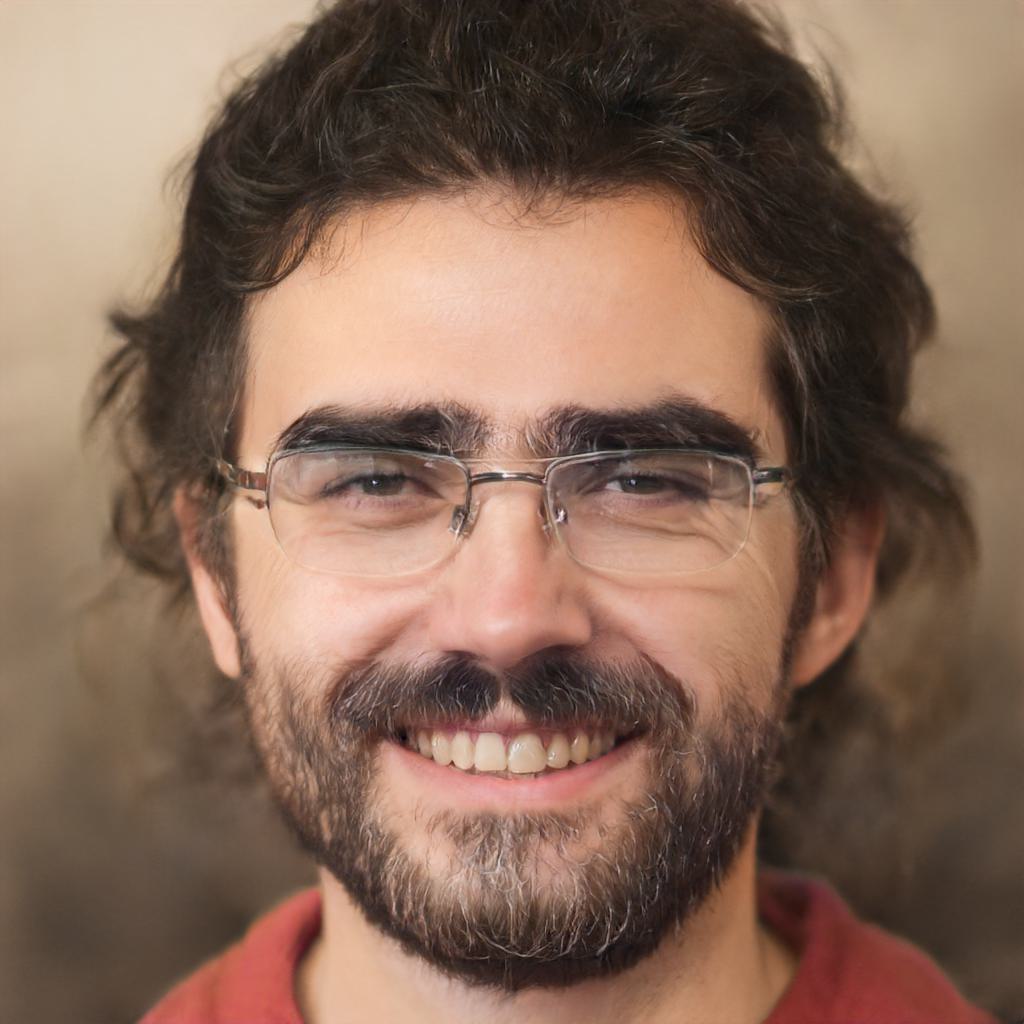} &
         \includegraphics[width=\imwidth\linewidth]{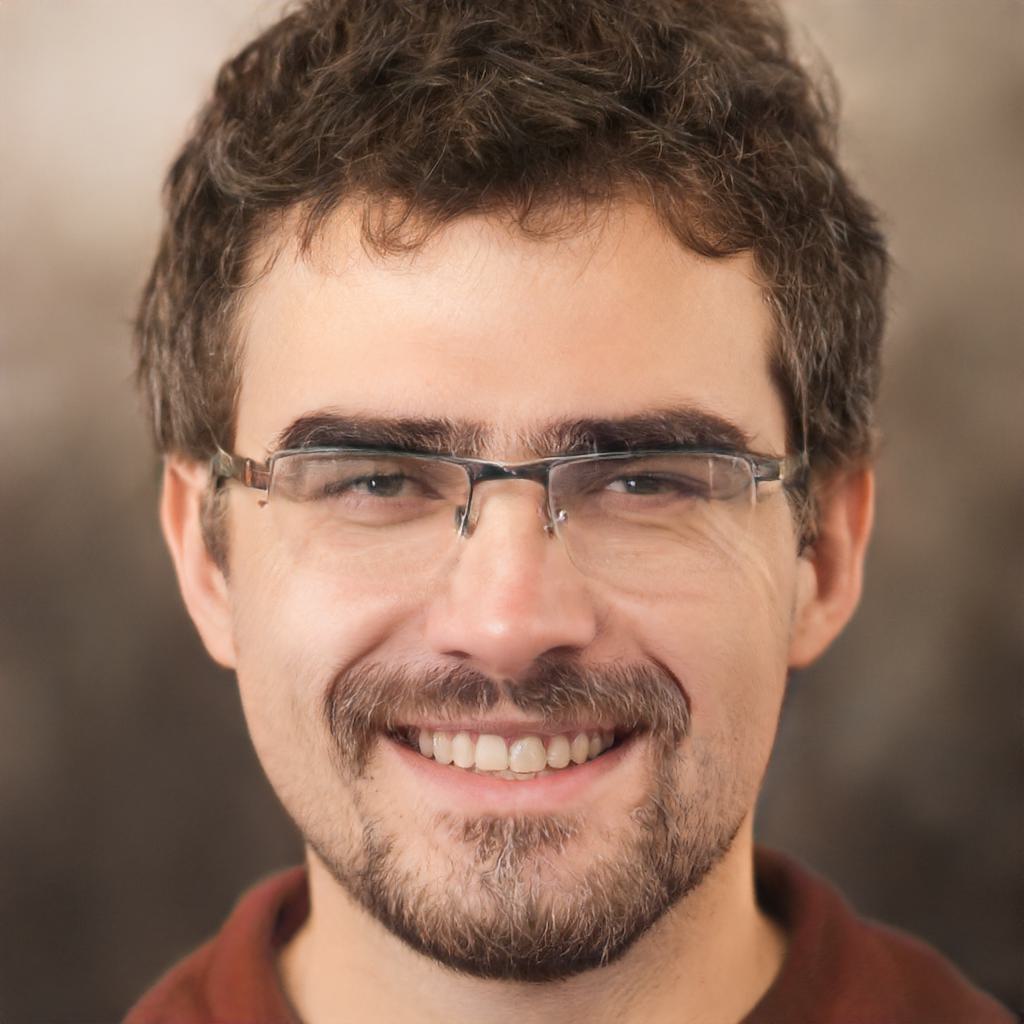} &
         \includegraphics[width=\imwidth\linewidth]{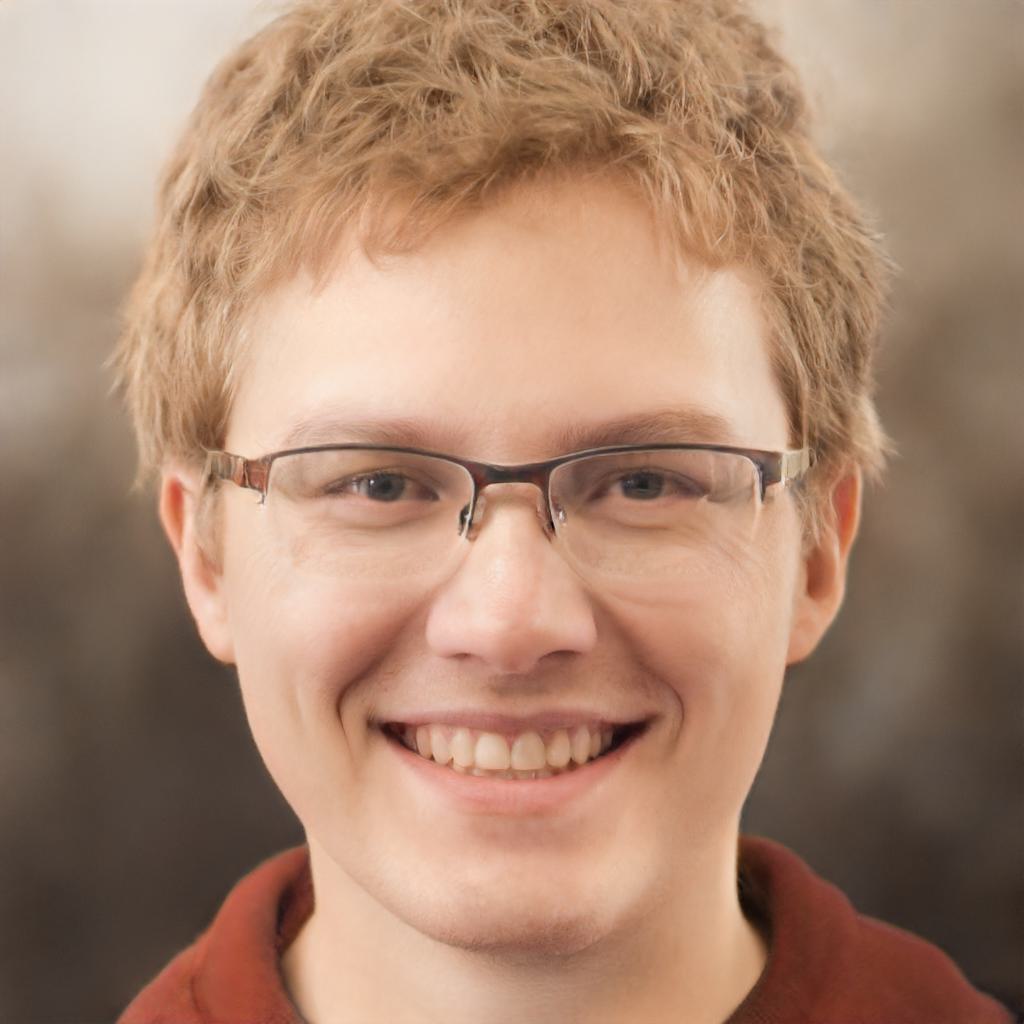} \\
         \rotatebox{90}{\hspace{0.29in}Age} &
         \includegraphics[width=\imwidth\linewidth]{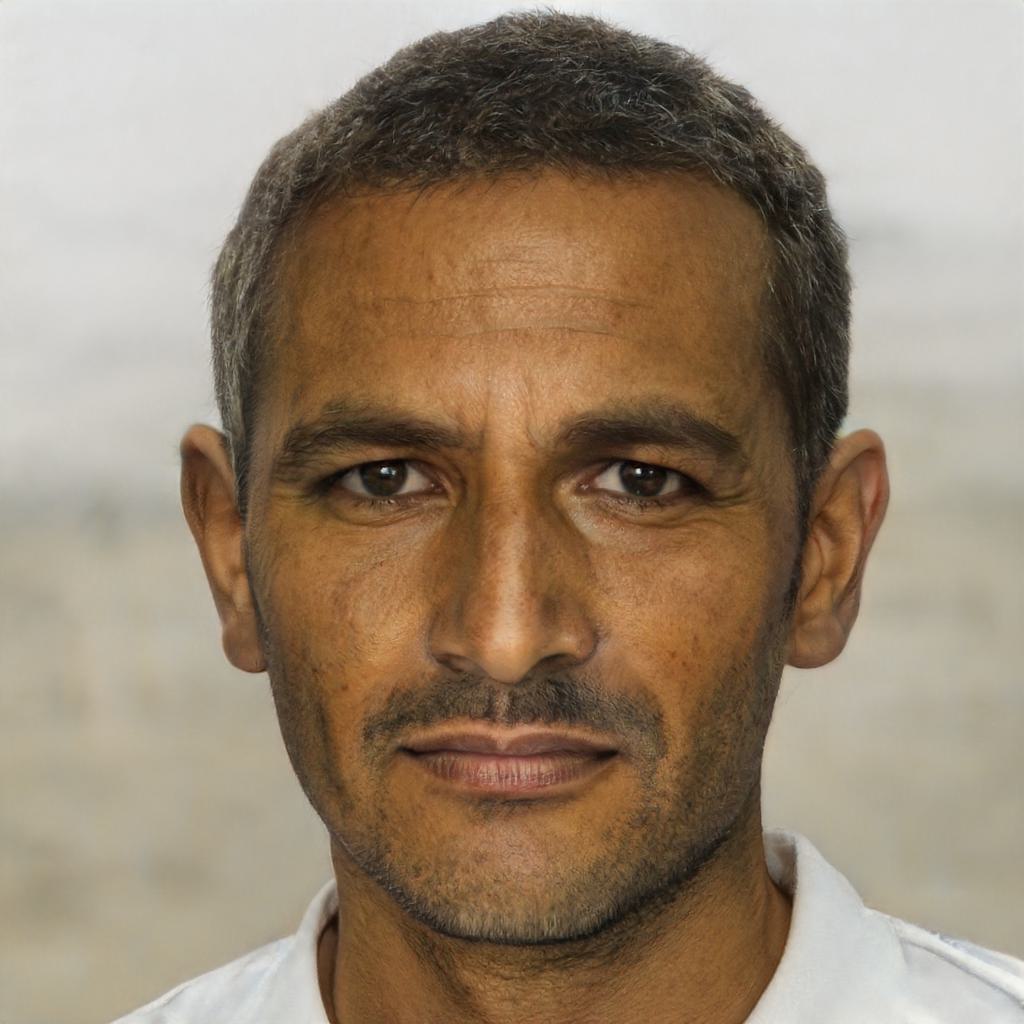} &
         \includegraphics[width=\imwidth\linewidth]{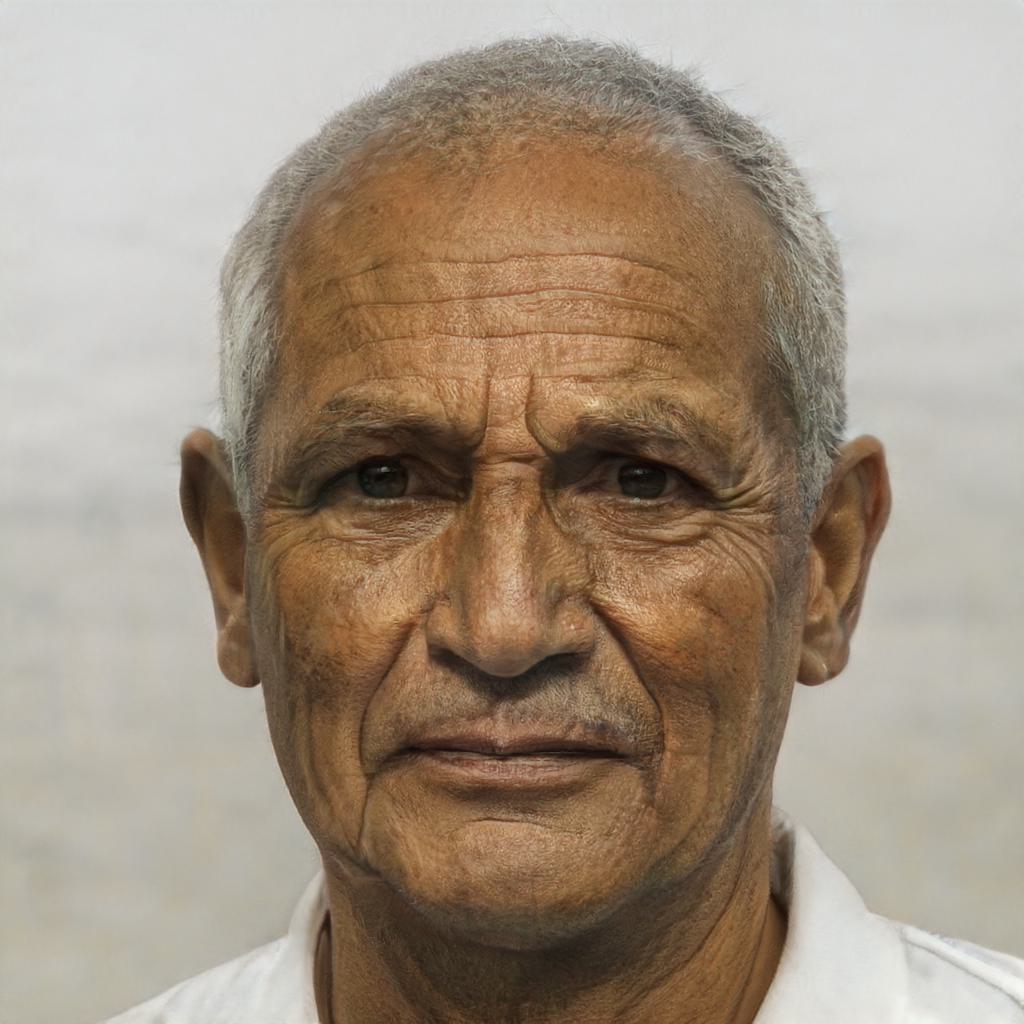} &
         \includegraphics[width=\imwidth\linewidth]{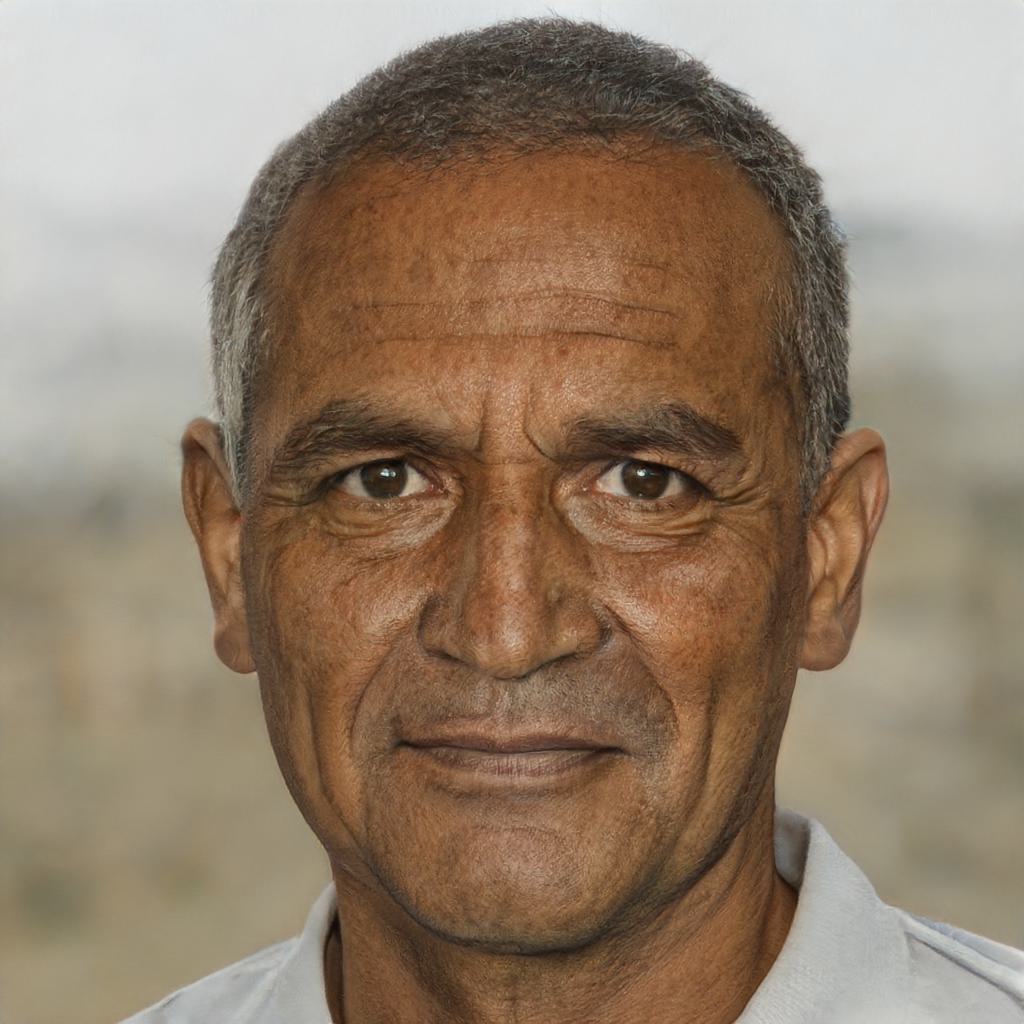} &
         \includegraphics[width=\imwidth\linewidth]{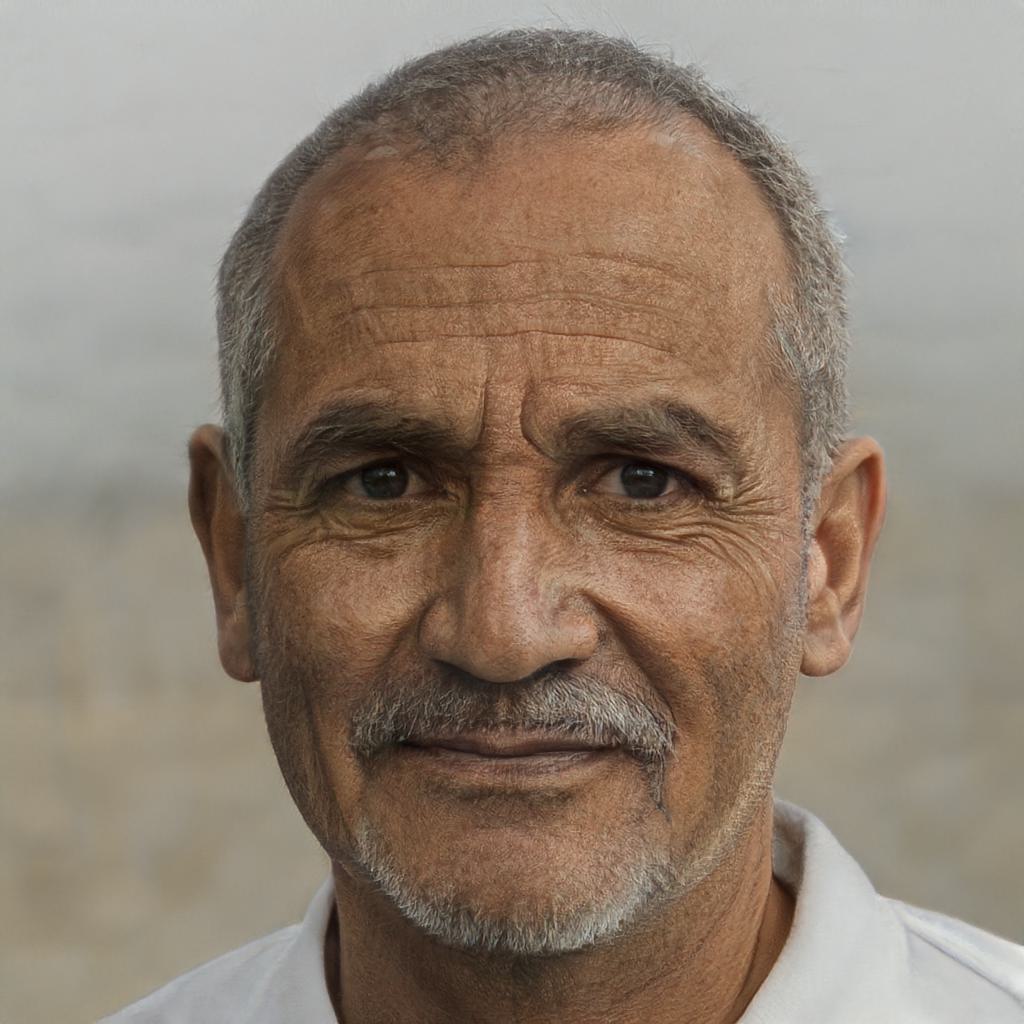} &
         \includegraphics[width=\imwidth\linewidth]{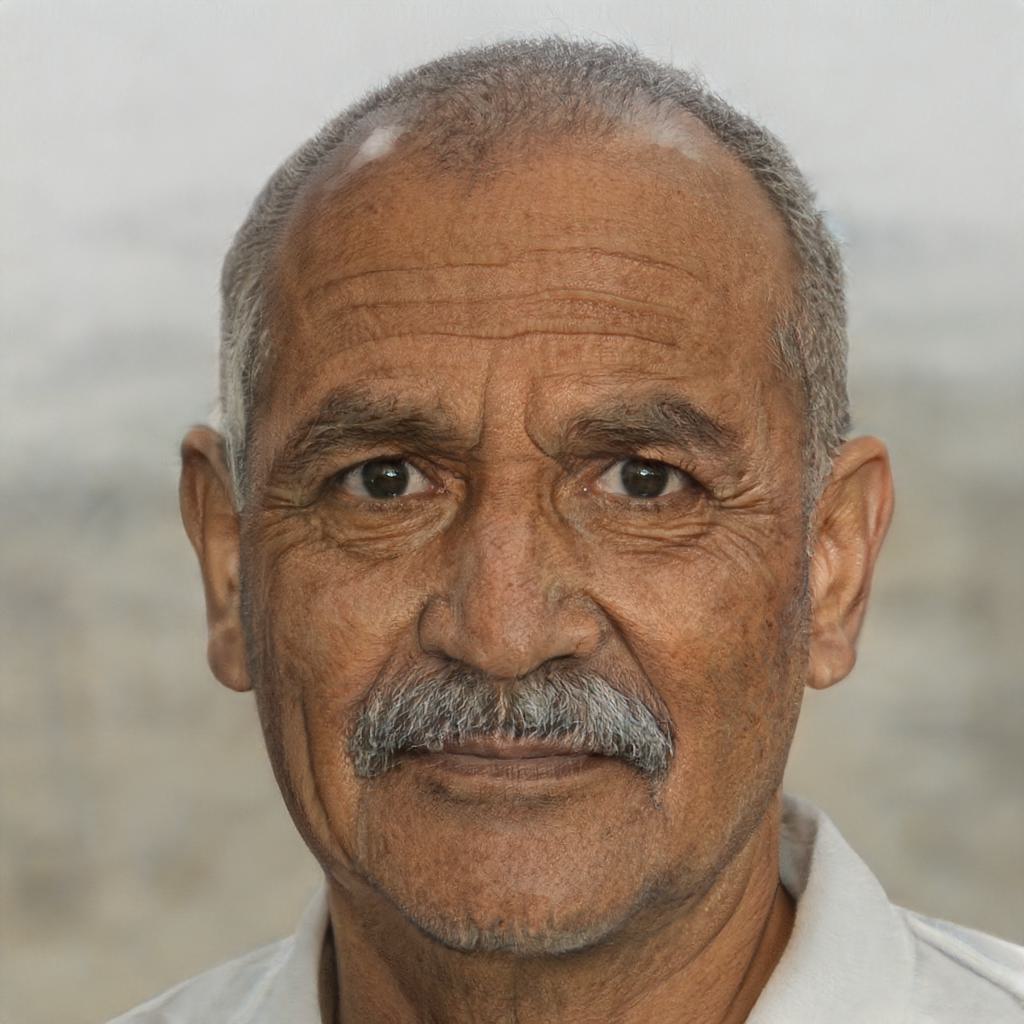} \\
         \rotatebox{90}{\hspace{0.29in}Race} &
         \includegraphics[width=\imwidth\linewidth]{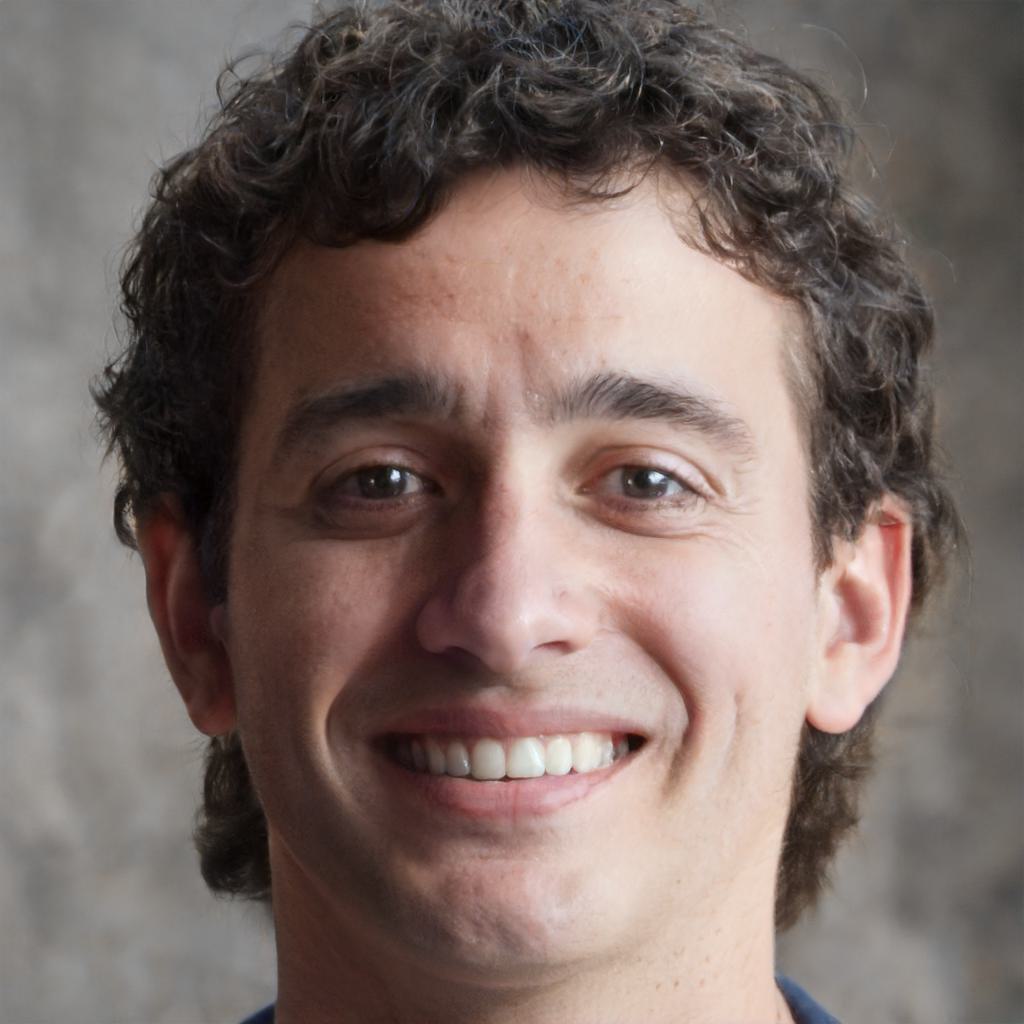} &
         \includegraphics[width=\imwidth\linewidth]{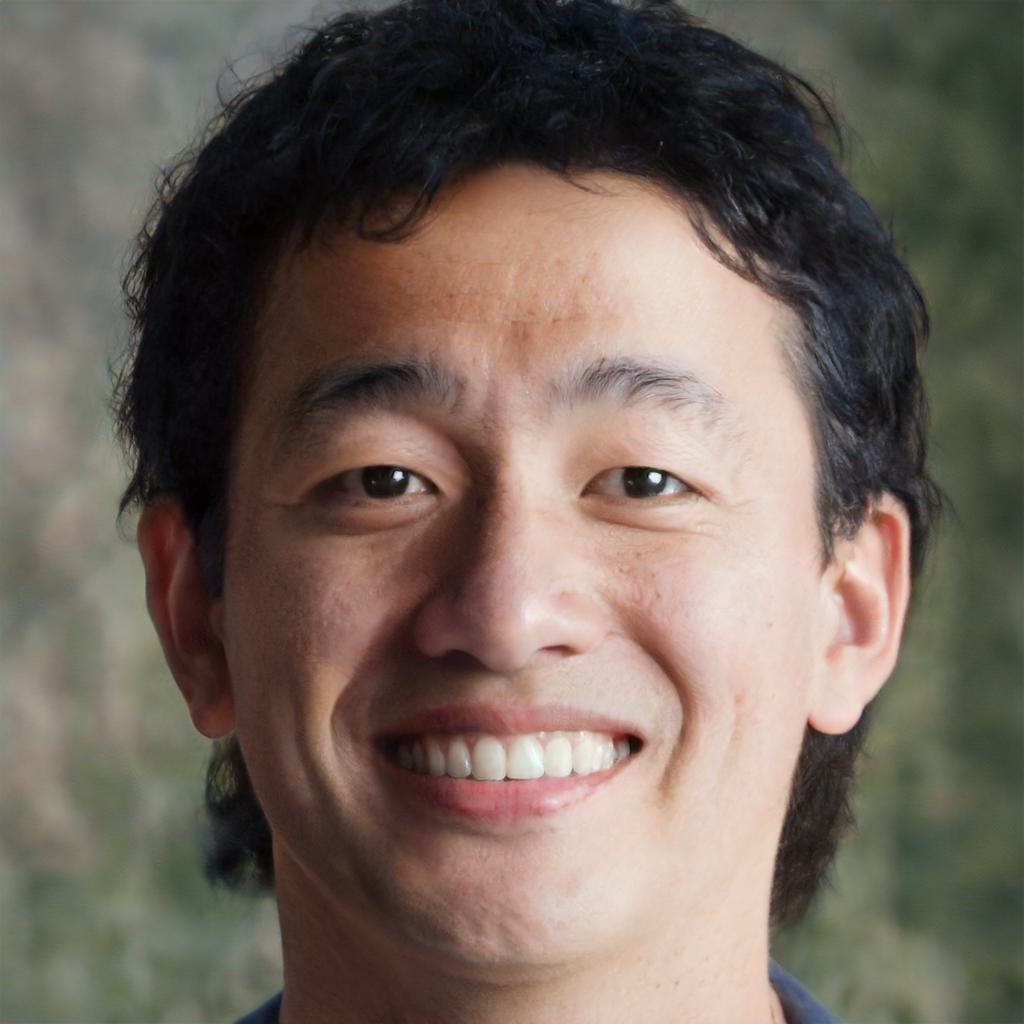} &
         \includegraphics[width=\imwidth\linewidth]{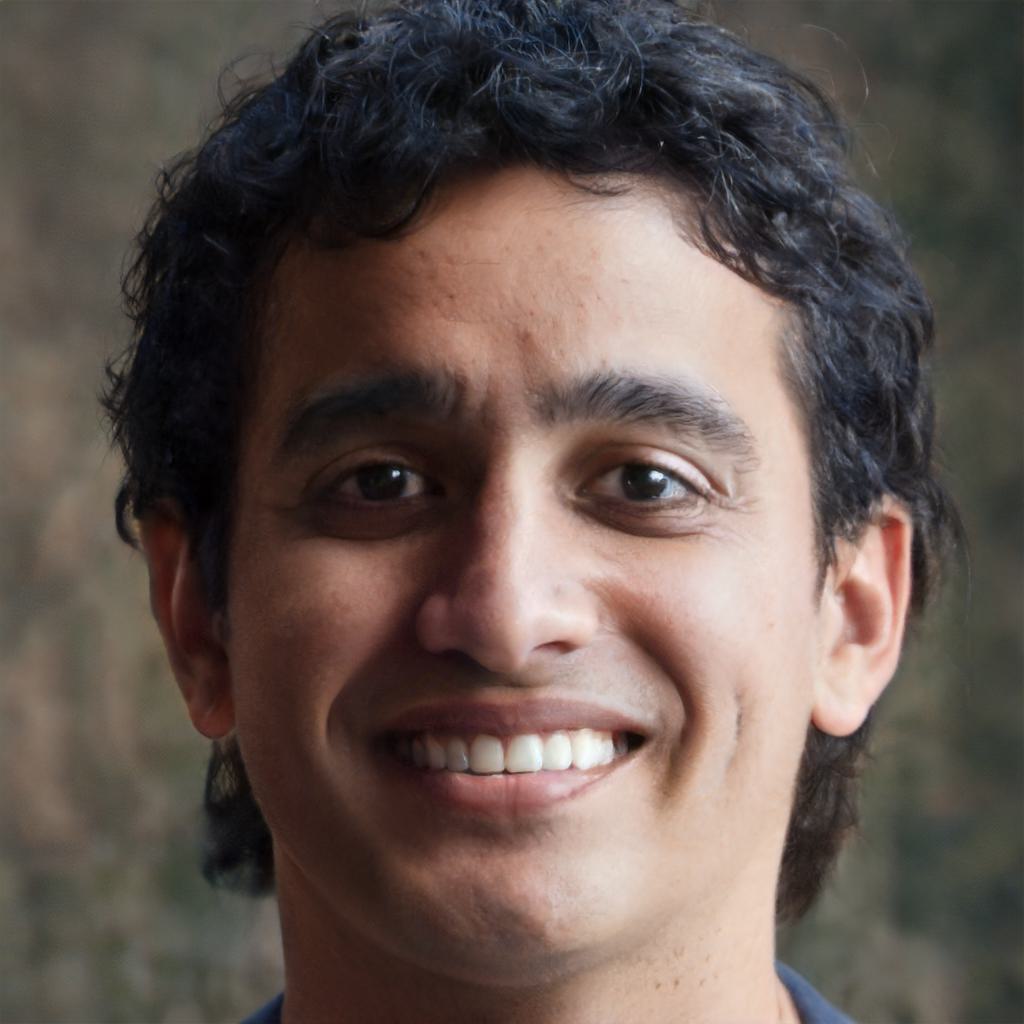} &
         \includegraphics[width=\imwidth\linewidth]{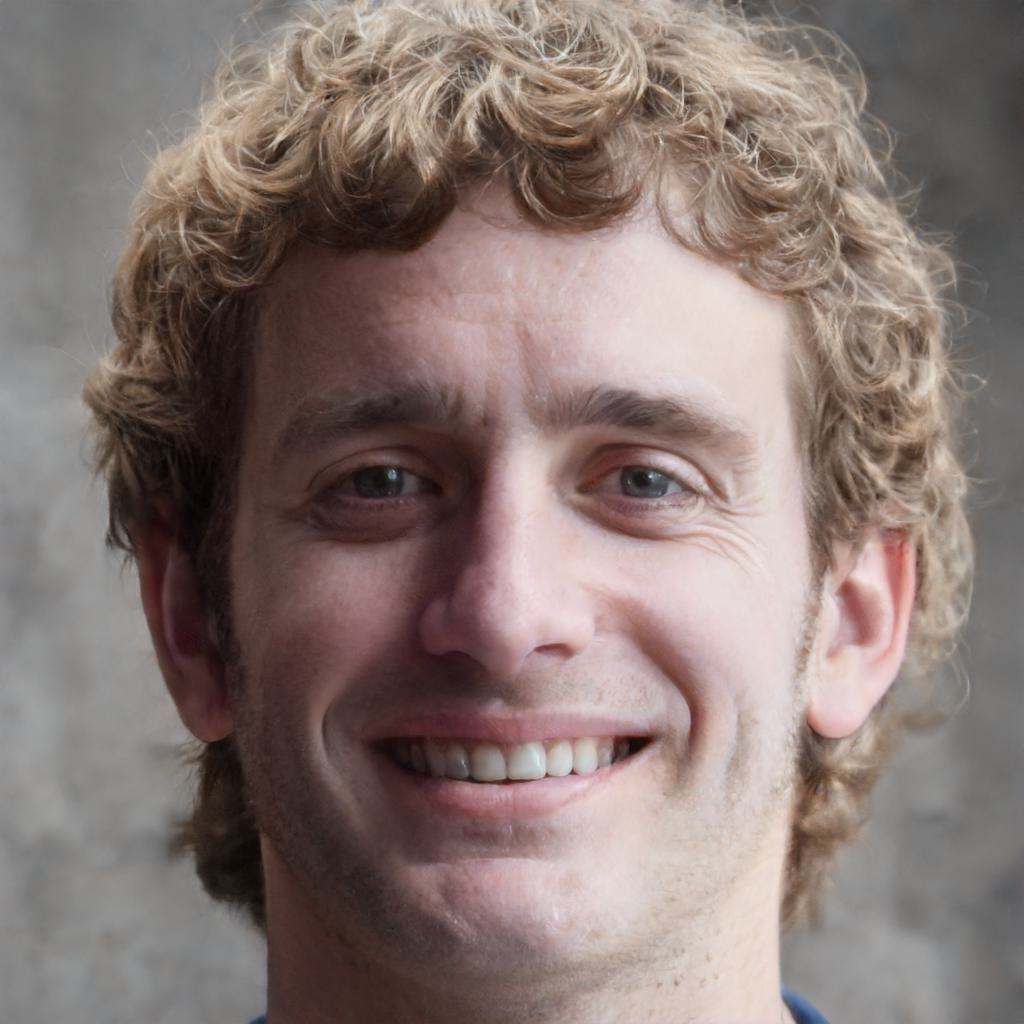} &
         \includegraphics[width=\imwidth\linewidth]{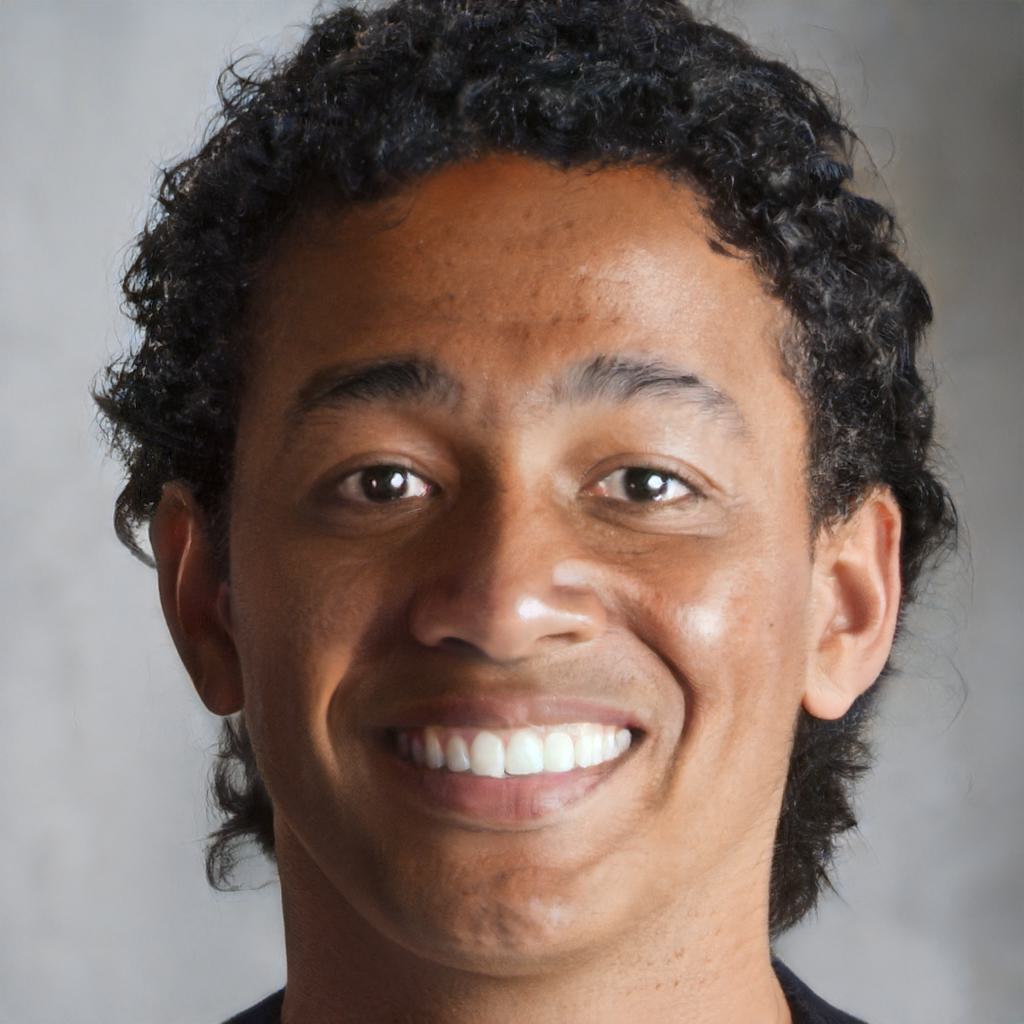} \\
         & Input & \multicolumn{4}{c}{Edits}
         
    \end{tabular}

    \captionof{figure}{Multi-directional human face editing. Each row indicates changes in a subspace associated with a particular attribute: (top to bottom) gender, age and race.}
    \label{fig:fig1}
\end{center}
}]

\ificcvfinal\thispagestyle{empty}\fi
\begin{abstract}
This paper describes a new technique for finding disentangled semantic directions in the latent space of StyleGAN. Our method identifies meaningful orthogonal subspaces that allow editing of one human face attribute, while minimizing undesired changes in other attributes. Our model is capable of editing a single attribute in multiple directions, resulting in a range of possible generated images. We compare our scheme with three state-of-the-art models and show that our method outperforms them in terms of face editing and disentanglement capabilities. Additionally, we suggest quantitative measures for evaluating attribute separation and disentanglement, and exhibit the superiority of our model with respect to those measures\footnotemark{}.
\end{abstract}

\footnotetext{Project page and code are available at \url{https://chennaveh.github.io/MDSE/}}
\section{Introduction}

Recent developments in computer vision have enabled the generation of photorealistic, high-resolution synthetic images. The most notable technique is Generative Adversarial Networks (GANs)\cite{goodfellow2014generative}, which have advanced the progress of many applications, including image generation, super-resolution, image inpainting, and more. In particular, the generator of StyleGAN\cite{abdal2019image2stylegan}, one of the most notable GAN models, has been extensively explored by researchers. 

StyleGAN samples a latent vector $\mathbf{z} \in \mathbb{R}^{512}$ from a Gaussian distribution $\mathcal{N}(\textbf{0}, \textbf{I})$, and maps it to an intermediate vector $\mathbf{w} \in \mathcal{W} = \mathbb{R}^{512}$ using a mapping network. This vector is then used to generate a 1024x1024 RGB image. The vector $\mathbf{w}$ is inserted into 18 multi-resolution style blocks that control various characteristics of the synthesized image. Vectors at lower resolutions determine high-level features such as pose and hair. Intermediate resolutions affect facial expressions, while vectors at higher resolutions dictate fine details like colors and texture. The original StyleGAN model uses the $\mathcal{W}$ latent space and the same $\mathbf{w}$ vector for all 18 style blocks. Subsequent studies\cite{abdal2019image2stylegan, abdal2020image2stylegan++, richardson2021encoding, tov2021designing} utilize the $\mathcal{W^+} = \mathbb{R}^{18\cdot 512}$ space, which extends the $\mathcal{W}$ space, and applies different $\mathbf{w}$ vectors to different style blocks. This allows for enhanced control over the resulting image. The $\mathcal{W^+}$ latent space is commonly used for inverse mapping, converting an image into $\mathbf{w} \in \mathcal{W}^+$.

Recent works have explored methods for gaining control over the synthesized images by altering the latent space vectors\cite{radford2015unsupervised, bojanowski2017optimizing, shen2020interfacegan,harkonen2020ganspaceli2021dystyle,shen2021closed, abdal2021styleflow}. For instance, shifting the latent vector of a generated image towards the direction corresponding to ``smile" can progressively amplify the smile in the output image. While these methods showcase impressive and realistic editing capabilities, pinpointing the attribute directions remains a challenge. One approach is to use vectors in latent space sampled from the GAN model combined with attribute ground truth (either manually labeled or determined using a pre-trained attribute estimator)\cite{shen2020interfacegan, goetschalckx2019ganalyze, jahanian2019steerability}. A linear classifier in the latent space is then employed. The normal to the classification boundary corresponds to the most significant direction of the attribute. These models often suffer from issues of entangled data and biases in the training set. Such biases might result in attribute correlations, like glasses with age or beard with gender. Consequently, adding glasses to a face may inadvertently age it. Other methods use unsupervised techniques\cite{harkonen2020ganspaceli2021dystyle, shen2021closed}, such as PCA, to find meaningful orthogonal directions in the latent space. However, these directions typically require subjective human post-annotation to link directions to attributes in the synthesized images. All these models share the notion of a singular direction for each attribute, despite the possibility of some attributes defined by multiple dimensions (\eg, age might be affected by factors like hair color and skin wrinkles, see \cref{fig:fig1}).

In this work, we present MDSE, an acronym for \textit{Multi-Directional Subspace Editing}. This framework aims to identify orthogonal semantics within the latent space of a pre-trained StyleGAN. Our goal is to discover meaningful subspaces that are mutually orthogonal, with each subspace controlling specific facial attributes. ``Multi-directional editing" means altering a latent vector within a particular subspace in various directions. This approach enables the generation of a wide range of images, each varying in a specific attribute. Furthermore, because the subspaces are mutually orthogonal, modifications in one attribute result in minimal changes to others. We explore the disentanglement capabilities of our model and quantitatively assess its performance in comparison to leading image editing techniques.
\section{Related Work}
\subsection{Generative Adversarial Networks (GANs)}
Generative models have revolutionized the generation of vast amounts of visual content. They can produce visually compelling content with remarkable fidelity, making them indispensable for image editing applications. GAN models learn a mapping from a specific distribution (usually
Gaussian or uniform) to the target distribution. The generator $G(\cdot)$ is trained to fool an adversary $D(\cdot)$, which aims to differentiate between the generated images and the real ones. Both models $G(\cdot)$ and $D(\cdot)$ are trained concurrently using the min-max loss\cite{goodfellow2014generative}. The GAN approach is widely recognized and often outperforms other techniques, such as variational autoencoders (VAEs)\cite{kingma2013auto}. Recently, denoising diffusion probabilistic models (DDPM)\cite{ho2020denoising} have risen in prominence and showcased superior results. However, despite their leading performance in image generation, their potential in image editing and semantic interpretation remains relatively unexplored. In this work, our attention is centered on a GAN-based generator, specifically the StyleGAN\cite{karras2019style, karras2020analyzing, karras2021alias}, which exhibits exceptional capability in the human face domain.

\textbf{GAN Latent Space.} Given a pre-trained generator $\textbf{x}=G(\textbf{w})$, GAN inversion $\hat{\textbf{w}} = G^{-1}(\textbf{x})$ is the process of inverting a specified image $\textbf{x}$ into a latent vector $\hat{\textbf{w}}$ that faithfully reconstructs the image using the generator. In other words, $G(\hat{\textbf{w}}) \approx \textbf{x}$. Most of the common methods either train an encoder\cite{tov2021designing, richardson2021encoding, pidhorskyi2020adversarial} from the image space into the latent space or optimize the latent vector, using backpropagation, directly until it generates the desired image\cite{lipton2017precise,abdal2019image2stylegan, abdal2020image2stylegan++}. Recent approaches that modify the generator's weight have gained traction, resulting in enhanced image reconstruction\cite{roich2021pivotal,alaluf2021hyperstyle}. Previous works show that the latent space of a pre-trained GAN model encodes image semantics in a meaningful structure. Hence, latent codes corresponding to images with similar semantics tend to cluster closely in the latent space\cite{shen2020interfacegan}. For instance, within the domain of human face generation, faces sharing characteristics (\eg, being young and blond) will have nearby latent vectors. This foundational insight drives modern image editing techniques.

\subsection{Multi-Directional Manipulation in Latent Space}
The idea of editing an image through latent space manipulation has been extensively explored in recent years\cite{radford2015unsupervised, bojanowski2017optimizing}. Such approaches aim to harness the strong capabilities of the generator. Therefore, they keep the generator network static and conduct vector operations solely in the latent space. Recent works primarily rely on a linear latent space assumption and demonstrate pleasing-looking results of image editing by adding or subtracting vectors within the latent space\cite{shen2020interfacegan,harkonen2020ganspaceli2021dystyle,shen2021closed}. InterFaceGAN\cite{shen2020interfacegan} employs pre-trained binary classifiers to label images with facial features, such as young-old, male-female, and with/without glasses, among others. It then trains a linear classifier (specifically, a Support Vector Machine or SVM) on the respective latent vectors to derive a classifying hyperplane. The normal to this hyperplane serves as the 1D direction for editing the relevant attribute.

GANSpace\cite{harkonen2020ganspaceli2021dystyle} adopts a data-driven approach, applying PCA on a set of vectors sampled from $\mathcal{W}$ to find meaningful directions. The eigenvectors associated with the largest eigenvalues serve as the editing directions. Some derived directions can be entangled, such as head rotation and gender. To address this issue, edits are restricted to a specifically selected subset of the generator's layers. 

Another work, SeFa\cite{shen2021closed}, avoids the image sampling pre-processing step. Instead, it determines a closed-form factorization of the latent space by utilizing an eigen-decomposition of the generator's weights. For StyleGAN, the style affine transformation matrices are employed for vector decomposition. The eigenvectors corresponding to the greatest variations are selected to define the semantic directions. However, this unsupervised method faces two main issues. Firstly, there's a subjective post-annotation process to match the directions with relevant semantics. Secondly, the directions that are derived often mix several semantics (\eg, age and glasses), a result of biases present in the training data.

A distinct set of studies moves away from the linear approach, opting instead to learn non-linear functions to manipulate latent vectors. StyleFlow\cite{abdal2021styleflow} employs continuous normalizing flows conditioned on the image attribute vector. This attribute vector must be provided using pre-trained networks before every edit. Another non-linear approach is StyleRig\cite{tewari2020stylerig}, which incorporates a 3D semantic network and harnesses it to perform rigid edits such as pose and light. Although non-linear approaches generally perform better, our research primarily emphasizes the more comprehensive linear approach.

All previous methods that manipulate the latent space of StyleGAN define a semantic as a singular direction. Both linear and non-linear approaches ultimately derive positive and negative directions for editing a desired attribute. We believe that this oversimplification may restrict the editing capabilities and limit the diversity of the generated images. For example, imagine we want to edit the gender of a given face. One person may argue that hairstyle is a distinguishing factor between men and women, while another might think of facial structure or the presence of facial hair as a more prominent factor. In this work, we overcome this variability by associating an attribute with an $n$-dim \textbf{subspace} where $n \geq 1$, containing more than a single direction, in the semantic space. We call this a multi-directional subspace, as different vectors within the subspace can affect the resulting image differently, but all affect the associated attribute (see examples in \cref{fig:fig1}). We also require different attributes to be associated with mutually orthogonal subspaces, to promote disentanglement between attributes.

We outline our contributions as follows:
\begin{itemize}
\item We propose MDSE to extend the concept of meaningful latent directions to multi-directional subspaces, thereby diversifying and expanding the capabilities of the editing  process.
\item MDSE identifies orthogonal directions in StyleGAN's latent space, to promote disentanglement. We introduce an orthogonality loss in the model training and demonstrate its significance in preserving some image attributes while editing others. We visualize the results and demonstrate improved results in consecutive edits of multiple attributes (refer to \cref{fig:fig4}).
\item We develop a new metric for evaluating the disentanglement properties of image editing. In addition to the conventional approaches of visual comparisons, we suggest using our evaluation metrics as a quantitative measurement of disentanglement capabilities.
\end{itemize}

\section{Method}
Given a set of attributes, $B \triangleq \left \{gender, glasses, \cdots \right \}$, we aim to find a subspace for each attribute $b_i \in B$ so that editing within this subspace leads to changes in the associated attribute $b_i$. Different directions in each subspace are responsible for distinct visual semantics. To achieve disentanglement, we desire that each subspace solely influences its respective attribute.

Certain attributes, such as glasses and age, have been observed to correlate with each other\cite{shen2020interfacegan}, making disentangled editing challenging. To overcome this problem, our model seeks to decompose the latent space $\mathcal{W^+}$ into multiple orthogonal subspaces each of which is associated with a single attribute. This requirement has two main outcomes. Firstly, editing within a subspace facilitates multi-directional edits of a single attribute. Secondly, the orthogonality ensures that altering a specific facial attribute doesn't affect other attributes.

\subsection{Latent Space Decomposition}
We decompose $\mathcal{W^+}$ into $N+1$ orthogonal subspaces $\left \{ S_i \right \}_{i=0}^N$ and define each subspace $S_i$ as 
\begin{equation}
    S_i \triangleq span \left \{ \textbf{p}_i^1, \textbf{p}_i^2, \cdots, \textbf{p}_i^{n_i} \right \}, n_i < dim(\mathcal{W^+})
\end{equation}
where $\textbf{p}_i^j \in \mathcal{W^+}$, and $\{ \textbf{p}_i^j\}_{j=1}^{n_i}$ are linearly independent vectors. Therefore, $\{ \textbf{p}_i^j \}_{j=1}^{n_i}$ form a basis for subspace $S_i$ with cardinality $n_i$. There are $N+1$ subspaces, where $N=|B|$ is the size of the attribute set $B$. Each $S_i$ corresponds to $b_i$ respectively, while $S_0$ is associated with all other information that is not labeled in $B$. This may include other semantics, \eg, clothing and image background. The correspondence of $S_i$ and $b_i$  will be elaborated on later in the paper.

To ensure orthogonality, we require the following conditions:

\begin{equation}\label{eq4}
    S_i \subset \mathcal{W^+} \;, \forall i \in \left \{ {0,\cdots,N} \right \}
\end{equation}
\begin{equation}\label{eq5}
    \bigoplus_{i=0}^N S_i = \mathcal{W^+}
\end{equation}
\begin{equation}\label{eq6}
    S_i\perp S_j \;, \forall i\neq j
\end{equation}
where $\bigoplus$ denotes direct sum, and $\perp$ signifies orthogonal subspaces. Notice that the entire vector set $\{ \textbf{p}_i^j \}$ forms a basis for $\mathcal{W^+}$. As a result, we can uniquely express every vector $\textbf{w} \in \mathbb{R}^{18 \cdot 512}$ with a set of scalars $\{ a_i^j \}$ and represent $\textbf{w}$ as the following linear combination:

\begin{equation}\label{eq8}
    \textbf{w} = \sum_{i=0}^{N} \sum_{j=1}^{n_i} a_i^j \textbf{p}_i^j
\end{equation}
This expression can also be represented in matrix form:

\begin{equation}
\textbf{w} = 
\begin{bmatrix}
    \underbrace{\begin{bmatrix}
        \vert & & \vert \\ 
        \textbf{p}_0^1 & \cdots & \textbf{p}_0^{n_0} \\ 
        \vert & & \vert
    \end{bmatrix}}_{\textbf{P}_0}
    \cdots
    \underbrace{\begin{bmatrix}
        \vert & & \vert \\ 
        \textbf{p}_N^1 & \cdots & \textbf{p}_N^{n_N} \\ 
        \vert & & \vert
    \end{bmatrix}}_{\textbf{P}_N}
\end{bmatrix}
\begin{bmatrix}
    \begin{bmatrix}
        a_0^1  \\
        \vdots \\
        a_0^{n_0}  \\
    \end{bmatrix}
    \\
    \vdots
    \\
    \begin{bmatrix}
        a_N^1  \\
        \vdots \\
        a_N^{n_N}  \\
    \end{bmatrix}
\end{bmatrix}
\end{equation}
\begin{equation}
    \textbf{w} = \sum_{i=0}^N \mathbf{P}_i\textbf{a}_i = \\
    \underbrace{\begin{bmatrix}
            \mathbf{P}_0, \cdots, \mathbf{P}_N \\ 
    \end{bmatrix}}_{\mathbf{P}}
    \underbrace{\begin{bmatrix}
            \textbf{a}_0 \\ \vdots \\ \textbf{a}_N  \\
    \end{bmatrix}}_{\mathbf{a}}
\end{equation}
where $\mathbf{P}$ is a matrix defined by:
\begin{equation}
    \mathbf{P} = \left [ \textbf{p}_0^1 \cdots \textbf{p}_0^{n_0} \cdots \textbf{p}_N^1 \cdots \textbf{p}_N^{n_N} \right ]
\end{equation}
and $\mathbf{a}^T = [\textbf{a}^T_0, \dots, \textbf{a}^T_N]$ is a vector of coefficients. Finding $\mathbf{P}$ that satisfies \cref{eq4}-(\ref{eq6}) is the core part of our framework.


\subsection{Training Procedure}
\label{sec:train}
Our dataset consists of a set of 2,000 vector pairs $\left \{ (\textbf{w}^{(i)}, \textbf{y}^{(i)}) \right \}_{i=1}^{2000}$. The latent vectors $\left \{ \textbf{w}^{(i)} \right \}$ are generated from random vectors sampled from a Gaussian distribution $\textbf{z}^{(i)} \sim \mathcal{N}(0, 1)$ and then mapped to $\mathcal{W^+}$ space using the StyleGAN mapping function: $\textbf{w}^{(i)}=M(\textbf{z}^{(i)})$. Each sample $\textbf{w}^{(i)} \in \mathbb{R}^{18 \cdot 512}$ is mapped onto the image space using the StyleGAN generator, $\textbf{x}^{(i)}=G(\textbf{w}^{(i)})$ and then annotated using pre-trained classifiers to determine an attribute score vector:
\begin{equation}
    \textbf{y}^{(i)} \triangleq (y^{(i)}_1, \cdots, y^{(i)}_N) = \left ( \mathcal{C}_1(\textbf{x}^{(i)}), \cdots, \mathcal{C}_N(\textbf{x}^{(i)}) \right )
\end{equation}
Here, each $y^{(i)}_k$ denotes the $b_k$ attribute score for sample $i$. Depending on the attribute, $y^{(i)}_k$ is either a discrete or continuous number. $\mathcal{C}_k$ represents the pre-trained classifier for attribute $k$. For age, smile, gender, and glasses, we used the face attribute classifier from\cite{he2018harnessing} trained on the FFHQ dataset\cite{karras2019style}. For the pose attribute, we utilized img2pose\cite{albiero2021img2pose} for face estimation.  Additionally, we employed a race classifier\cite{karkkainenfairface} trained on the Yahoo YFCC100M dataset\cite{thomee2016yfcc100m}.

The primary goal during the training phase is to determine the matrix $\mathbf{P}$  and use it to reconstruct all samples from the training set. To satisfy \cref{eq8}, we jointly learn a vector $\textbf{a}^{(i)}$ for each vector $\textbf{w}^{(i)}$ such that $\textbf{w}^{(i)} = \mathbf{P}\textbf{a}^{(i)}$. We then introduce the following loss:

\begin{equation}
\mathcal{L}^{(i)}_{rec} = \left \| \textbf{w}^{(i)} - \mathbf{P}\textbf{a}^{(i)} \right \|_1
\end{equation}
where $\mathcal{L}_{rec}$ represents the reconstruction loss. Instead of pixel-based image reconstructions, we use a $L_1$ loss in the latent space using the original vector $\textbf{w}^{(i)}=M(\textbf{z}^{(i)})$ as the target value.

Additionally, to enforce disentanglement, as referred in \cref{eq6}, we introduce an orthogonality loss:
\begin{equation}
\mathcal{L}_{orth} = \sum_{i \neq j} \left \| \mathbf{P}_i^T \mathbf{P}_j \right \|_2^2
\end{equation}
and require that the columns of matrices $\mathbf{P}_i, \mathbf{P}_j$ be orthogonal for $i \neq j$, where $\left \| \cdot  \right \|_2$ is the element-wise Frobenius-norm.


Since learning disentangled representations is fundamentally impossible without a supervised inductive bias on the data\cite{locatello2019challenging}, we leverage the attribute vector $\textbf{y}^{(i)}$ and introduce a mixing loss to establish the association between $S_k$ and $b_k$. Given a vector $\textbf{w}^{(i)}$, we can import attribute $b_k$ from a randomly chosen vector $\textbf{w}^{(j)}$, to generate $\textbf{w}^{(i)}_{mix}$:

\begin{equation}
\textbf{w}^{(i)}_{mix} = \mathbf{P}_k\textbf{a}^{(j)}_k + \sum_{l\neq k}  \mathbf{P}_l\textbf{a}^{(i)}_l
\end{equation}
and its corresponding image $\textbf{x}^{(i)}_{mix} \triangleq G(\textbf{w}^{(i)}_{mix})$. For the modified image we require $\mathcal{C}_k(\textbf{x}^{(i)}_{mix})$ to be similar to $y^{(j)}_k$ while keeping the other attributes unchanged. Thus a mixing loss is added to the network to force changes only in $b_k$:

\begin{equation}
\mathcal{L}^{(i)}_{mixing} = L_k \left ( \mathcal{C}_k(\textbf{x}^{(i)}_{mix}), y_k^{(j)} \right ) 
+ \sum_{l \neq k} L_l \left ( \mathcal{C}_l(\textbf{x}^{(i)}_{mix}), y_l^{(i)} \right )
\end{equation}

$L_i$ are loss functions that vary based on the attribute types. For categorical labels, we use softmax with cross-entropy loss, whereas continuous labels are optimized using the $L_1$ loss. In practice, we mix all attributes together from randomly chosen vectors. This introduces complex changes to the image and encourages the association of a single subspace with a single attribute.

Finally, our model is trained using an objective function comprised of the three losses:
\begin{equation}
\mathcal{L} = \lambda_{orth}\mathcal{L}_{orth} + \frac{1}{n} \sum_{i} \mathcal{L}^{(i)}_{rec} + \lambda_{mixing}\mathcal{L}^{(i)}_{mixing}
\end{equation}
where $\lambda_{orth}$, $\lambda_{mixing}$ are hyperparameters.


\section{Experiments}
In this section, we evaluate the performance of our model compared to state-of-the-art image editing models, all of which utilize the StyleGAN generator for image synthesis.

To test our model, we generated a source and target latent vector, projected them onto a subspace $S_k$, and replaced the projections of the source vector with those from the target. Since our subspaces are orthogonal, we expect to see changes only in the corresponding attribute $b_k$. Example results are presented in \cref{fig:fig2}. Notably, if the source face wears glasses, these remain post-edit since we only incorporate pose, smile, and race from the target face. 

\newcommand*{\fsize}{0.2}
\begin{figure}
\centering
\begin{tabular}{c|ccc}
\hspace{0.0in}Source \hspace{0.0in} \hspace{0.05in}\rotatebox[origin=l]{90}{\hspace{0.1in}Target} & \includegraphics[width=\fsize\linewidth]{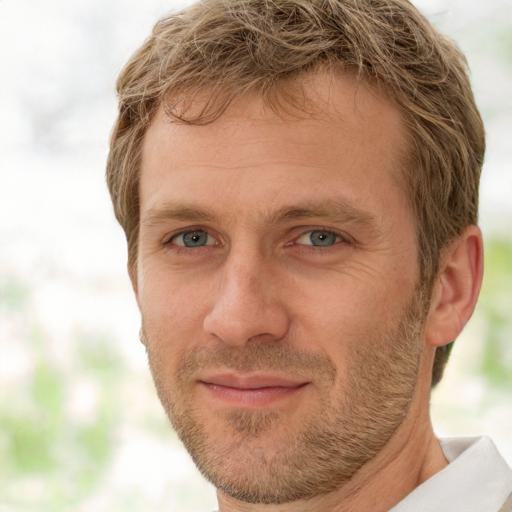} & \includegraphics[width=\fsize\linewidth]{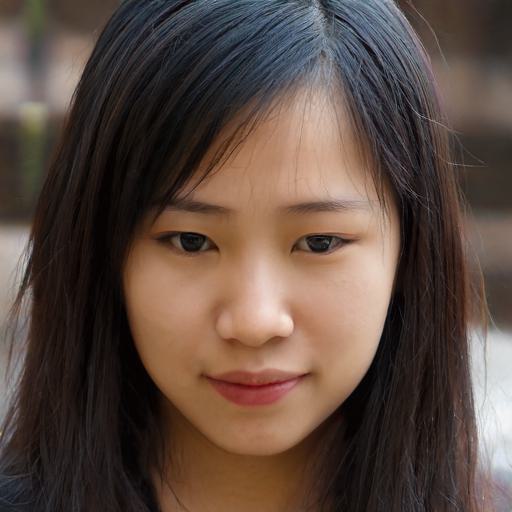} & \includegraphics[width=\fsize\linewidth]{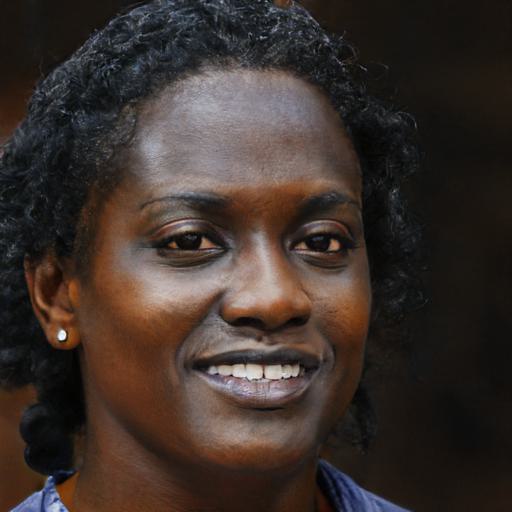}\\
\hline \\
\includegraphics[width=\fsize\linewidth]{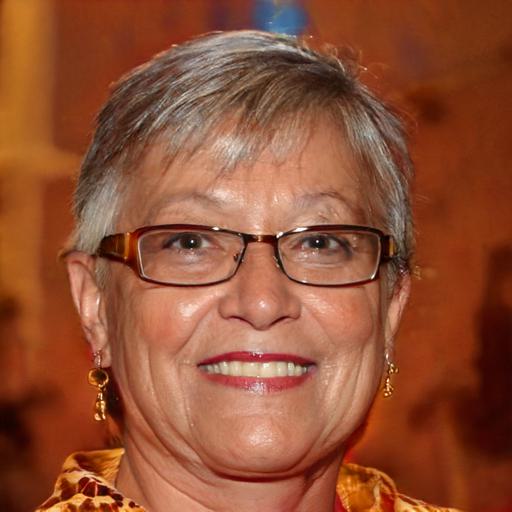} & \includegraphics[width=\fsize\linewidth]{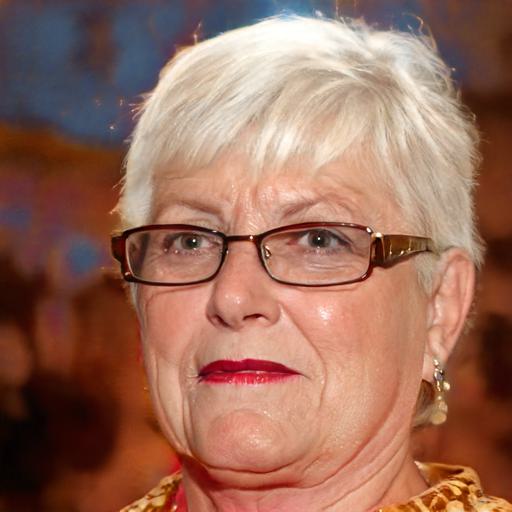} &
\includegraphics[width=\fsize\linewidth]{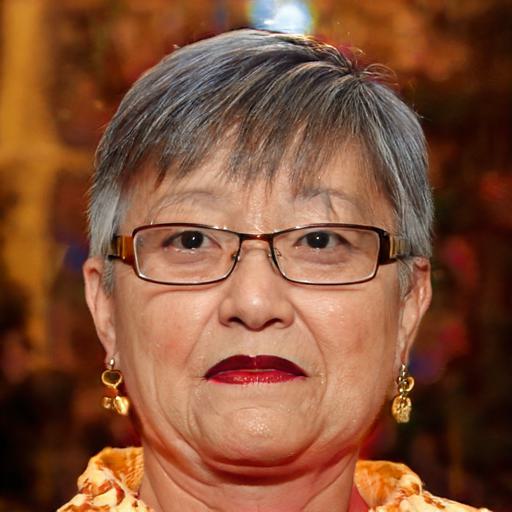} &
\includegraphics[width=\fsize\linewidth]{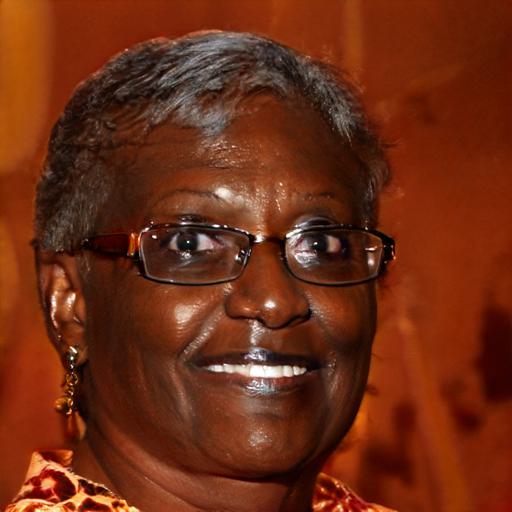}\\
\includegraphics[width=\fsize\linewidth]{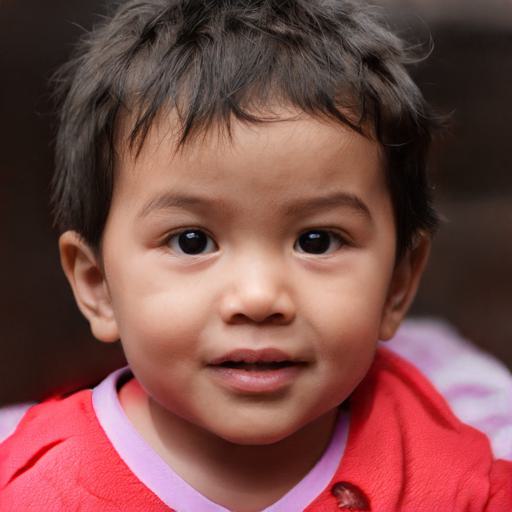} & \includegraphics[width=\fsize\linewidth]{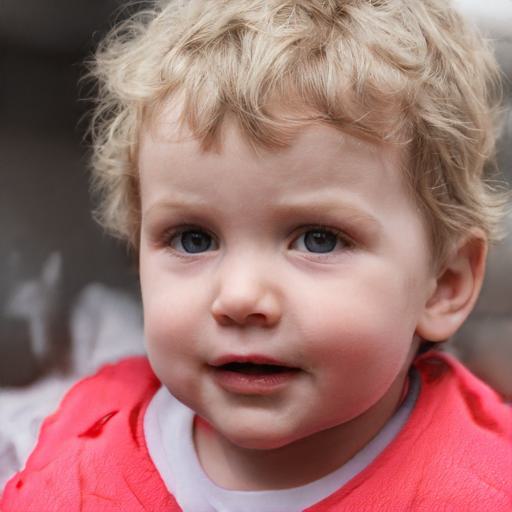} &
\includegraphics[width=\fsize\linewidth]{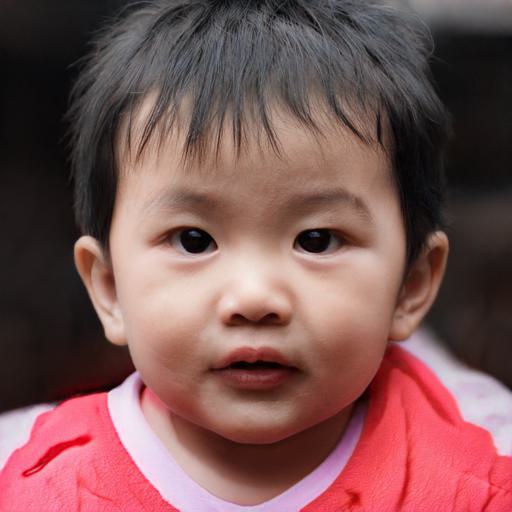} &
\includegraphics[width=\fsize\linewidth]{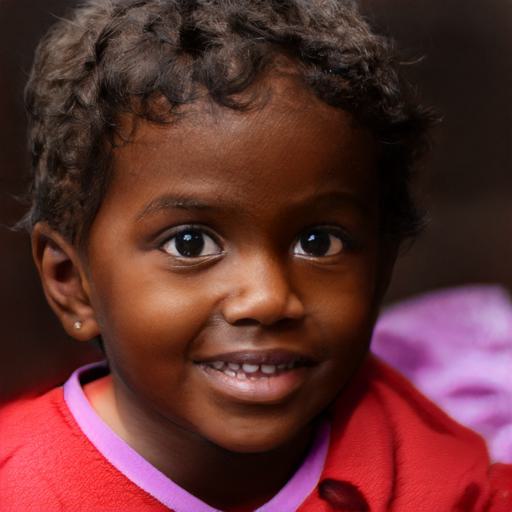}\\
\includegraphics[width=\fsize\linewidth]{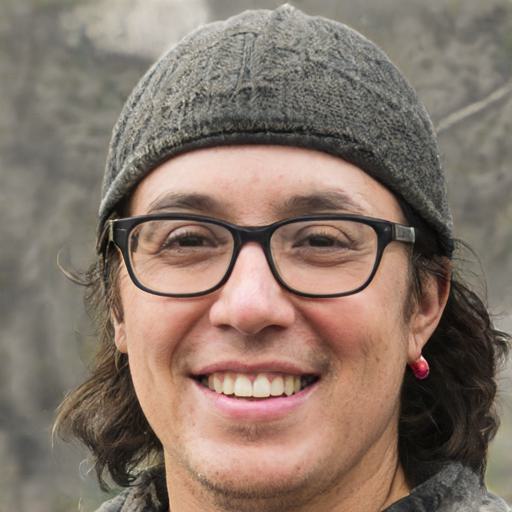} & \includegraphics[width=\fsize\linewidth]{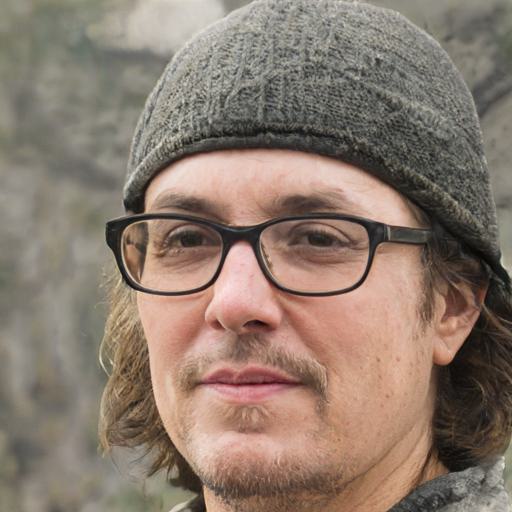} &
\includegraphics[width=\fsize\linewidth]{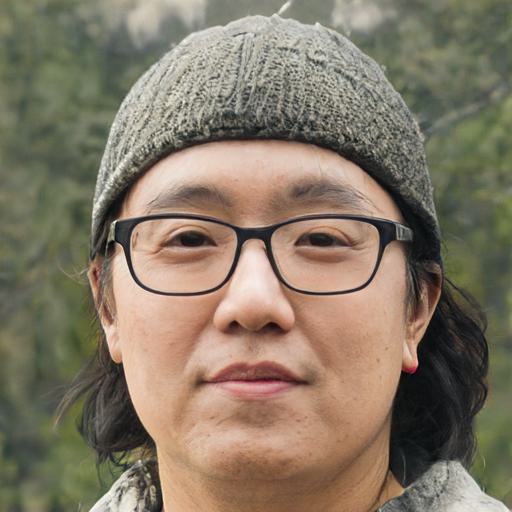} &
\includegraphics[width=\fsize\linewidth]{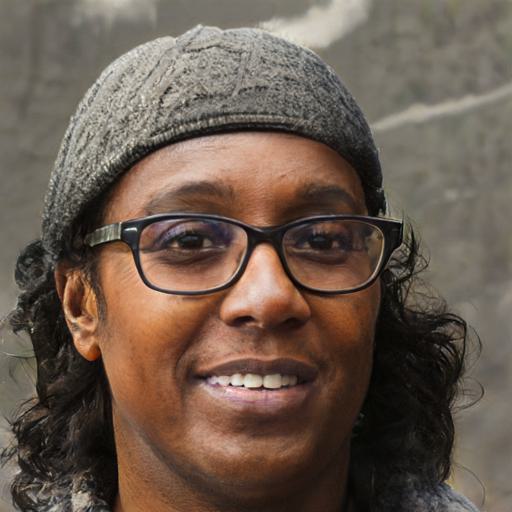}
\end{tabular}
\caption{Image editing capabilities using source images and target attribute images. The attributes derived from the target images include pose, smile, and race.}
\label{fig:fig2}
\end{figure}


\subsection{Comparison with Previous Methods}
\label{sec:comp}
We compared our model with three previous image editing methods: InterFaceGAN\cite{shen2020interfacegan}, StyleFlow\cite{abdal2021styleflow}, and SeFa\cite{shen2021closed}. We evaluated the editing and disentanglement qualities of these models for a common set of attributes supported by all of them, such as age, gender, smile, pose, and glasses. Additionally, since our method discovers a subspace rather than a single direction, to ensure a fair comparison, we trained a linear SVM inside each subspace. We then used the normal to the hyperplane as the direction for editing. This model is used in \cref{sec:qualitative} and \cref{sec:quantitative}.

\subsection{Qualitative Comparison}
\label{sec:qualitative}
\Cref{fig:fig3} offers a comparative assessment of the quality of real image editing. We inverted the images into StyleGAN's latent space using HyperStyle\cite{alaluf2022hyperstyle}. Our observations indicate that multiple edits on a single image can significantly diminish its quality, making it more prone to visual artifacts. Moreover, due to biases in the FFHQ dataset\cite{karras2019style}, some attributes are more correlated in StyleGAN's latent space than others (\eg, younger individuals are less likely to wear glasses). As depicted in \cref{fig:fig3}, while most models can effectively modify a single attribute, our model outperforms others in terms of accuracy when changing all attributes simultaneously and better maintains each individual attribute edit. Sequential editing is illustrated in \cref{fig:fig4}, showing that as edits progress, the images retain previous attributes without a loss in quality.

In \cref{fig:fig1} we visualize our model's capability of generating diverse results for different attributes due to its multi-directional nature. The images are generated by shifting the latent vectors to different directions inside the relevant subspace. We found that some attributes (\eg, smile, pose) behave like binary attributes, meaning they contain most of the information in a single direction. Others (\eg, gender, age), however, can be edited in multiple directions resulting in various images. Additional results can be found in the supplementary material.

\newcommand*{\imsize}{0.14}
\begin{figure*}
\centering
\setlength{\tabcolsep}{1pt}
\hspace*{-2.0cm}
\begin{tabular}{cccccccc}
 & Reconstruction & Smile & Gender & Pose & Age & Glasses & All \\
 
 \hspace{0.7in}\rotatebox{90}{\hspace{0.26in}Sefa} &
 \includegraphics[width=\imsize\linewidth]{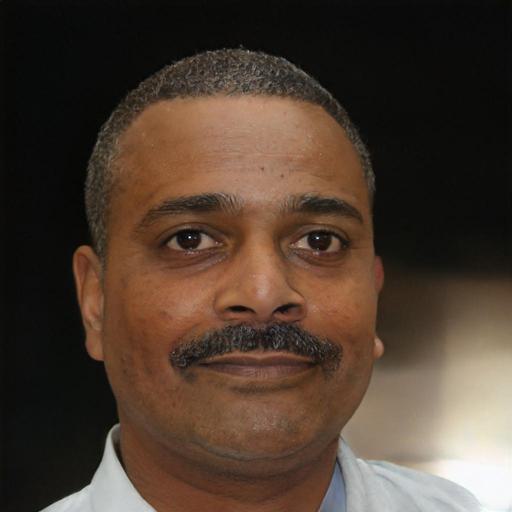} &
 \includegraphics[width=\imsize\linewidth]{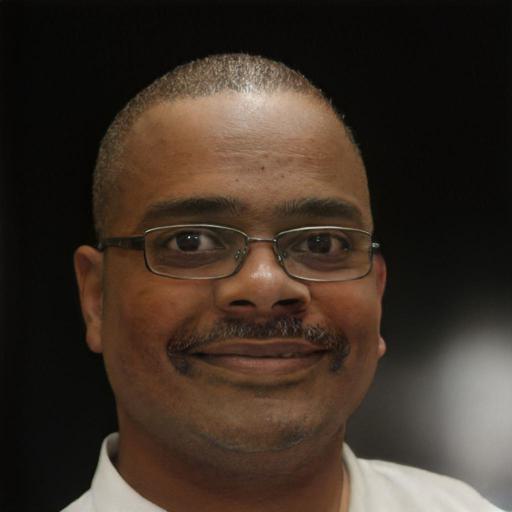} &
 \includegraphics[width=\imsize\linewidth]{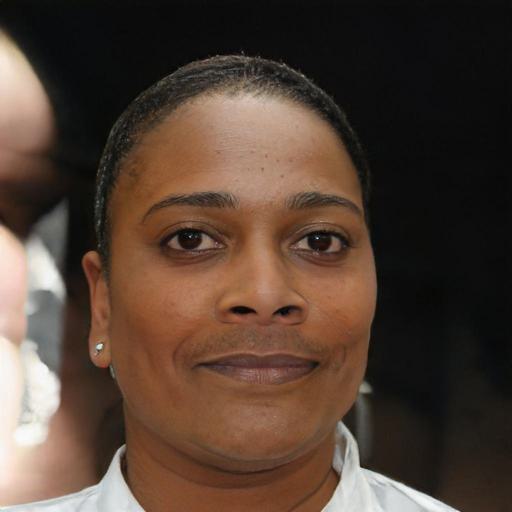} &
 \includegraphics[width=\imsize\linewidth]{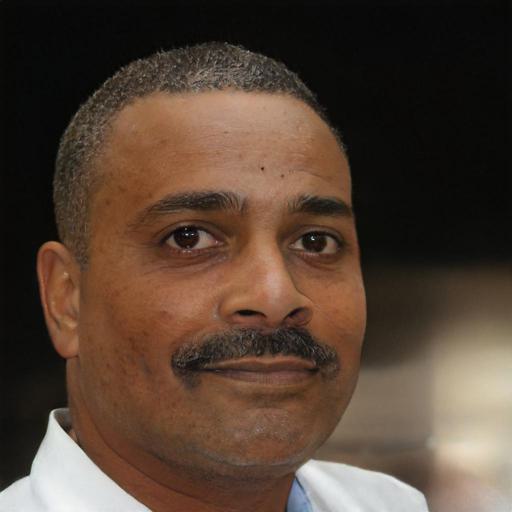} &
 \includegraphics[width=\imsize\linewidth]{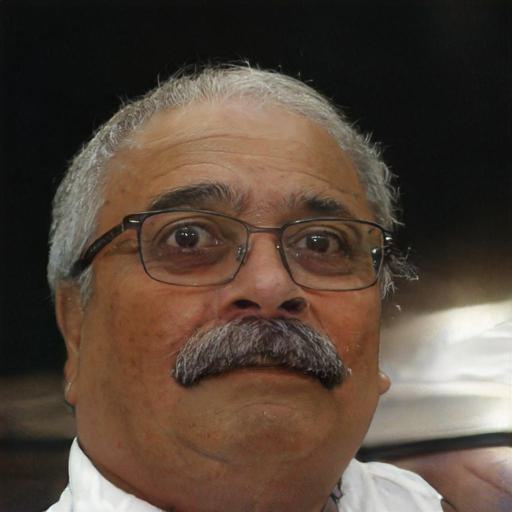} &
 \includegraphics[width=\imsize\linewidth]{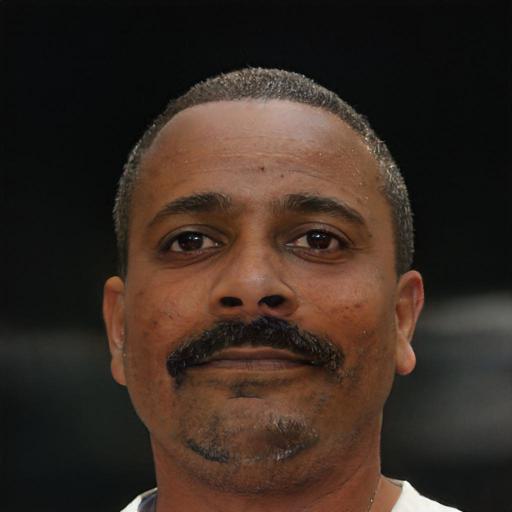} &
 \includegraphics[width=\imsize\linewidth]{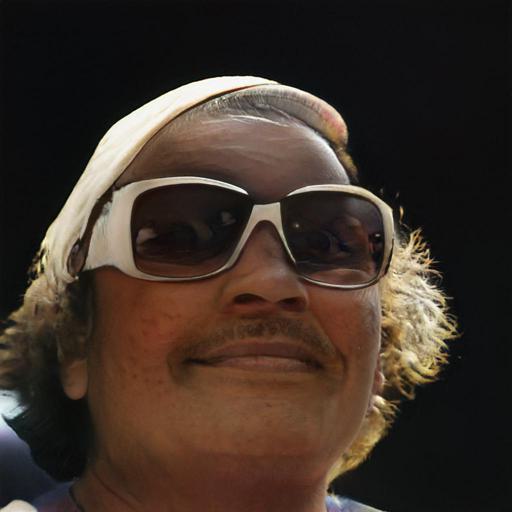} \\ 
 
 \hspace{0.7in}\rotatebox{90}{\hspace{0.0in}InterfaceGan} &
 \includegraphics[width=\imsize\linewidth]{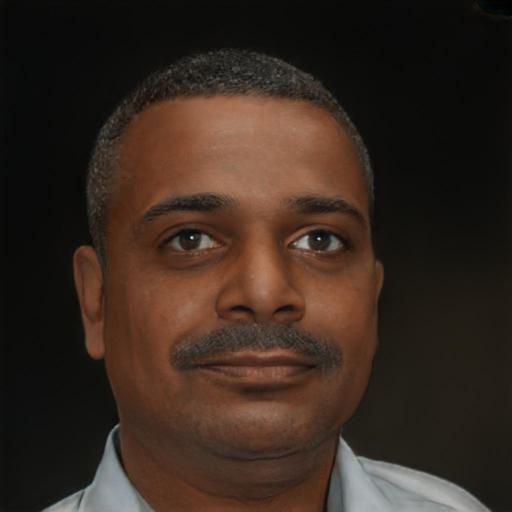} &
 \includegraphics[width=\imsize\linewidth]{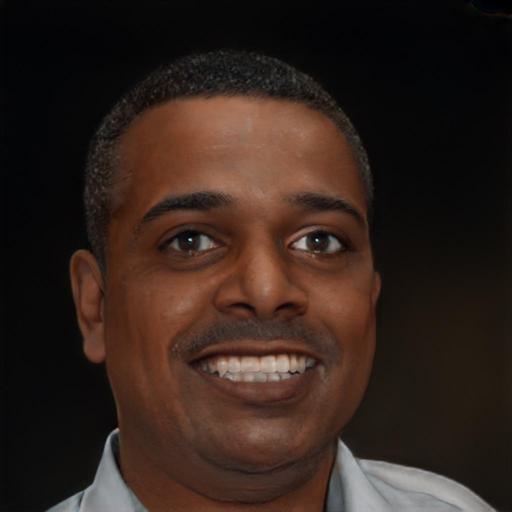} &
 \includegraphics[width=\imsize\linewidth]{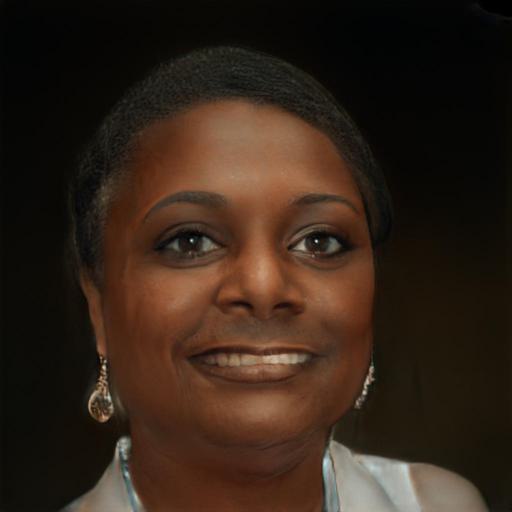} &
 \includegraphics[width=\imsize\linewidth]{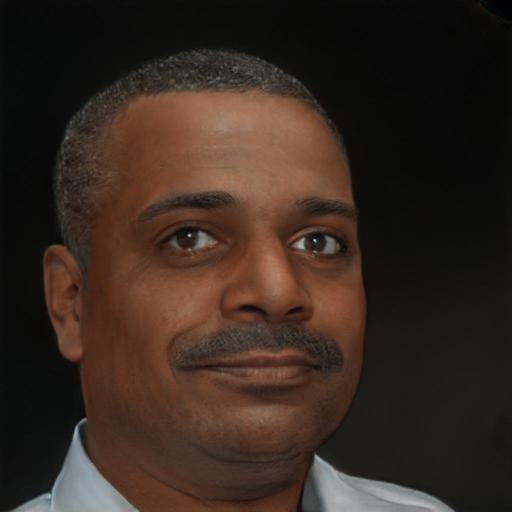} &
 \includegraphics[width=\imsize\linewidth]{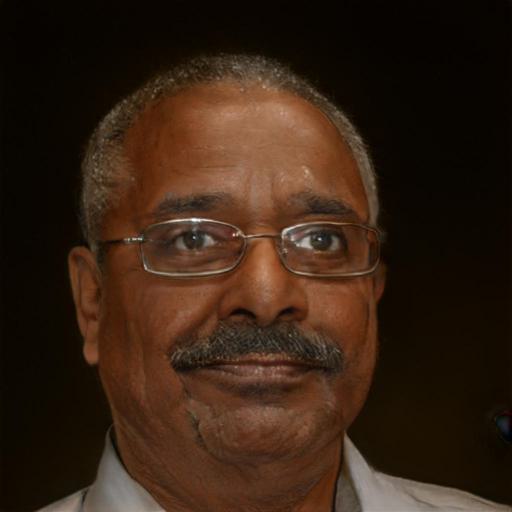} &
 \includegraphics[width=\imsize\linewidth]{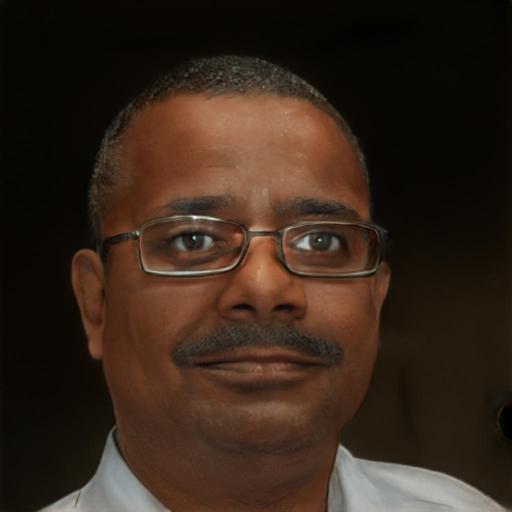} &
 \includegraphics[width=\imsize\linewidth]{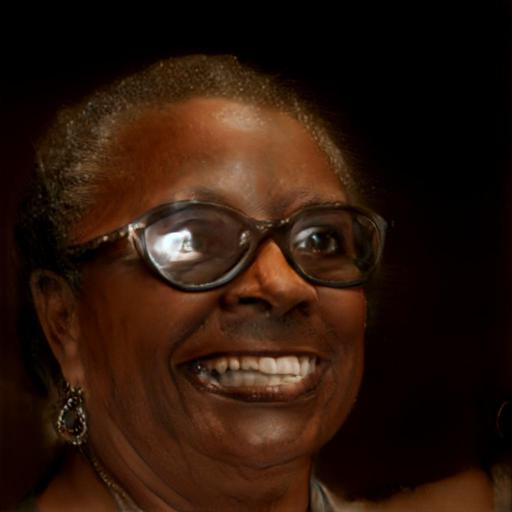} \\
 
   \hspace{0.7in}\rotatebox{90}{\hspace{0.12in}StyleFlow} &
 \includegraphics[width=\imsize\linewidth]{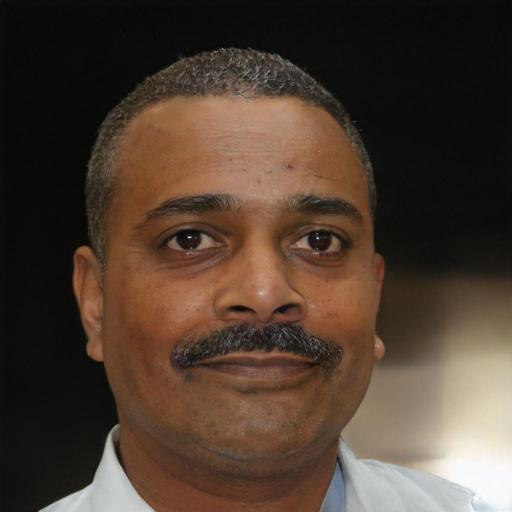} &
 \includegraphics[width=\imsize\linewidth]{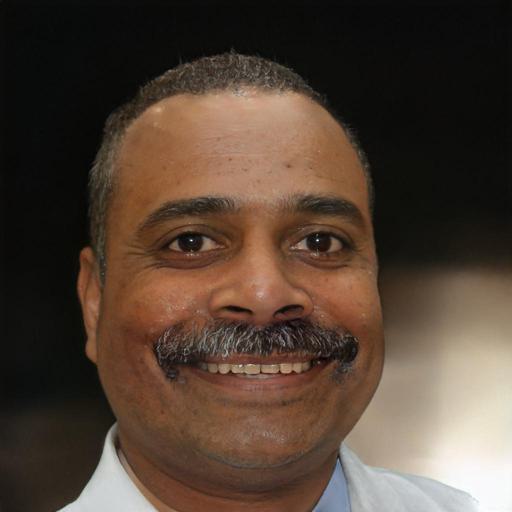} &
 \includegraphics[width=\imsize\linewidth]{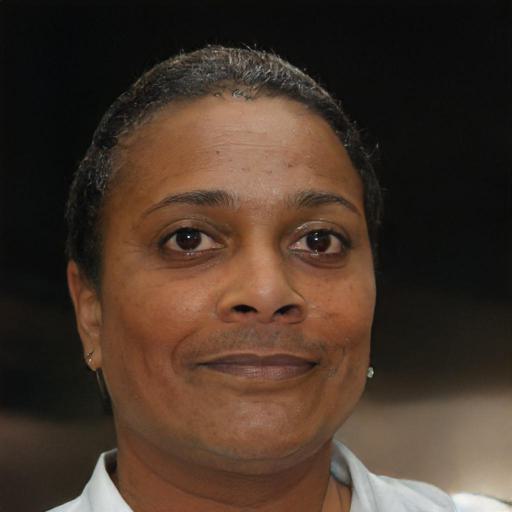} &
 \includegraphics[width=\imsize\linewidth]{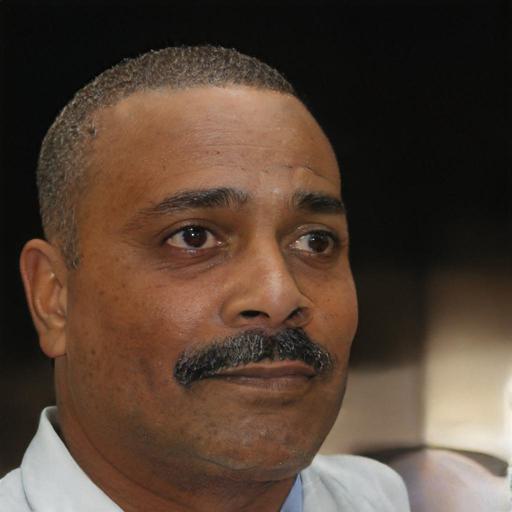} &
 \includegraphics[width=\imsize\linewidth]{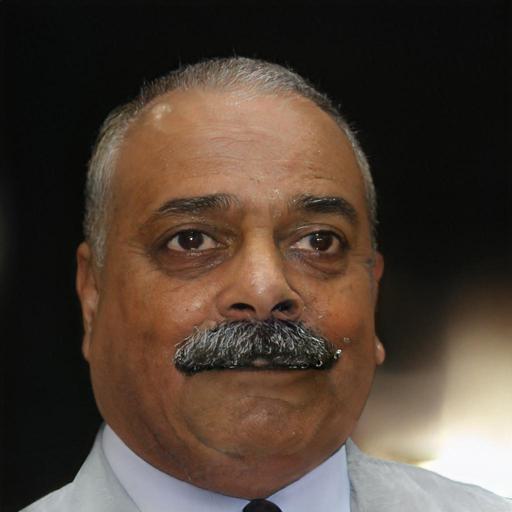} &
 \includegraphics[width=\imsize\linewidth]{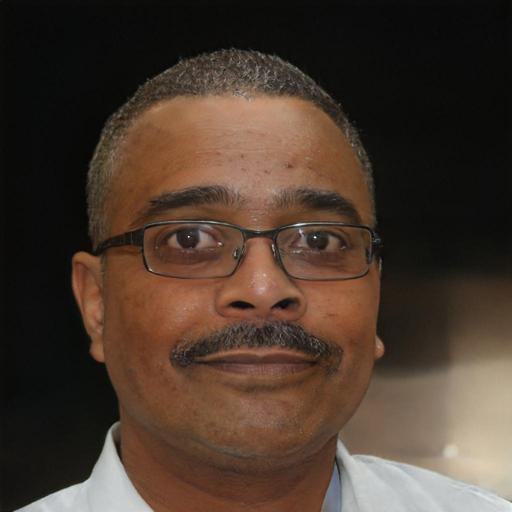} &
 \includegraphics[width=\imsize\linewidth]{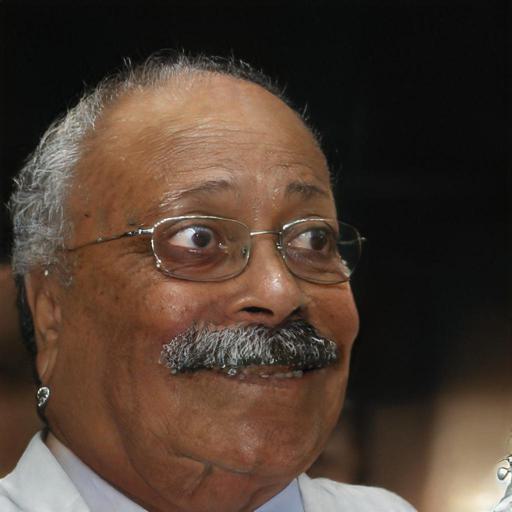} \\ 
 
  \hspace{0.7in}\rotatebox{90}{\hspace{0.29in}Ours} &
 \includegraphics[width=\imsize\linewidth]{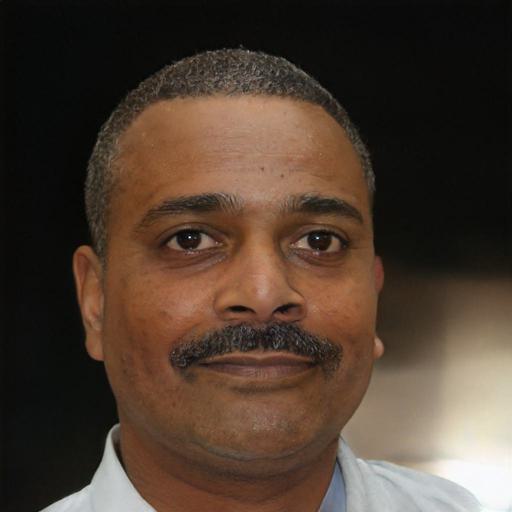} &
 \includegraphics[width=\imsize\linewidth]{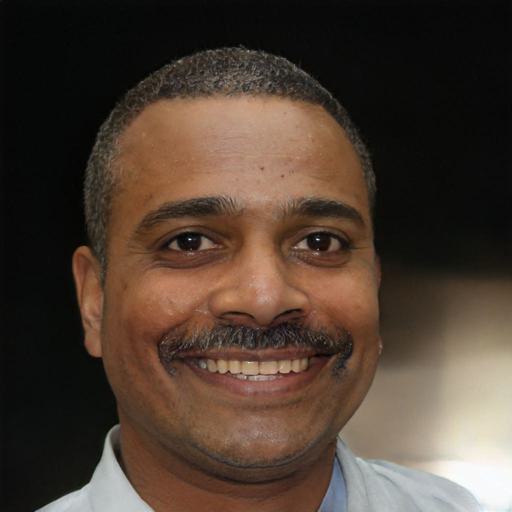} &
 \includegraphics[width=\imsize\linewidth]{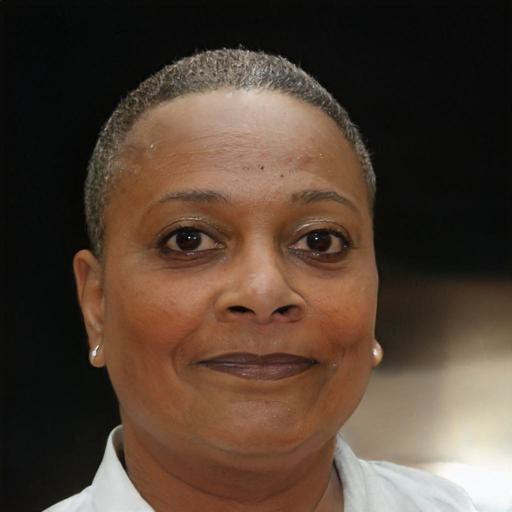} &
 \includegraphics[width=\imsize\linewidth]{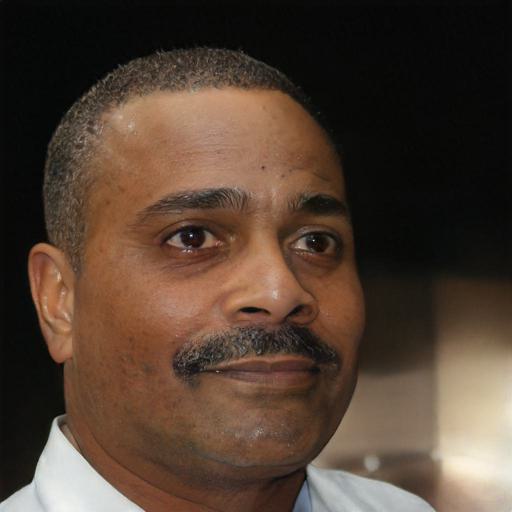} &
 \includegraphics[width=\imsize\linewidth]{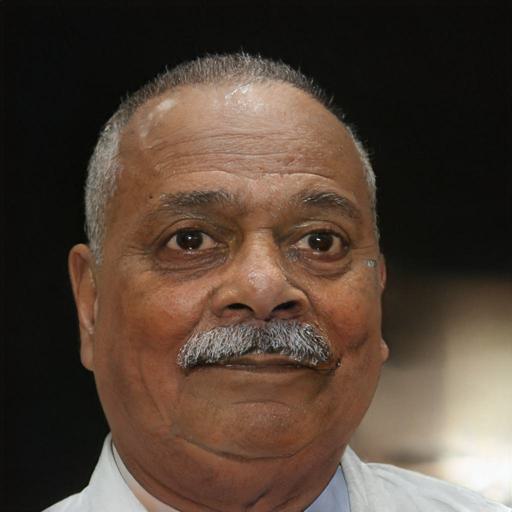} &
 \includegraphics[width=\imsize\linewidth]{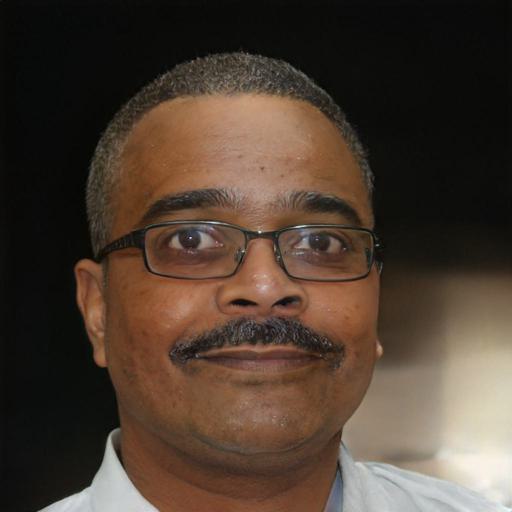} &
 \includegraphics[width=\imsize\linewidth]{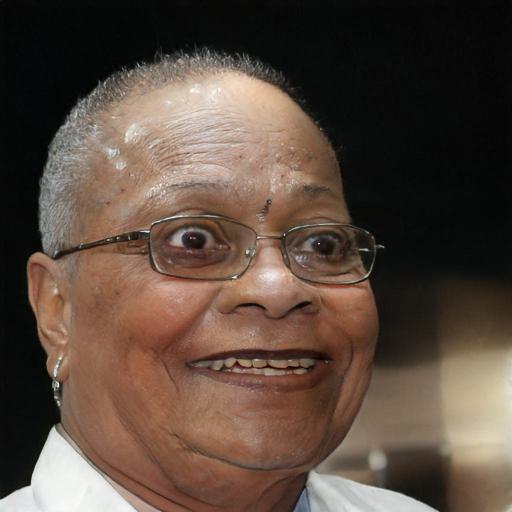} \\
 
  \hspace{0.7in}\rotatebox{90}{\hspace{0.26in}Sefa} &
 \includegraphics[width=\imsize\linewidth]{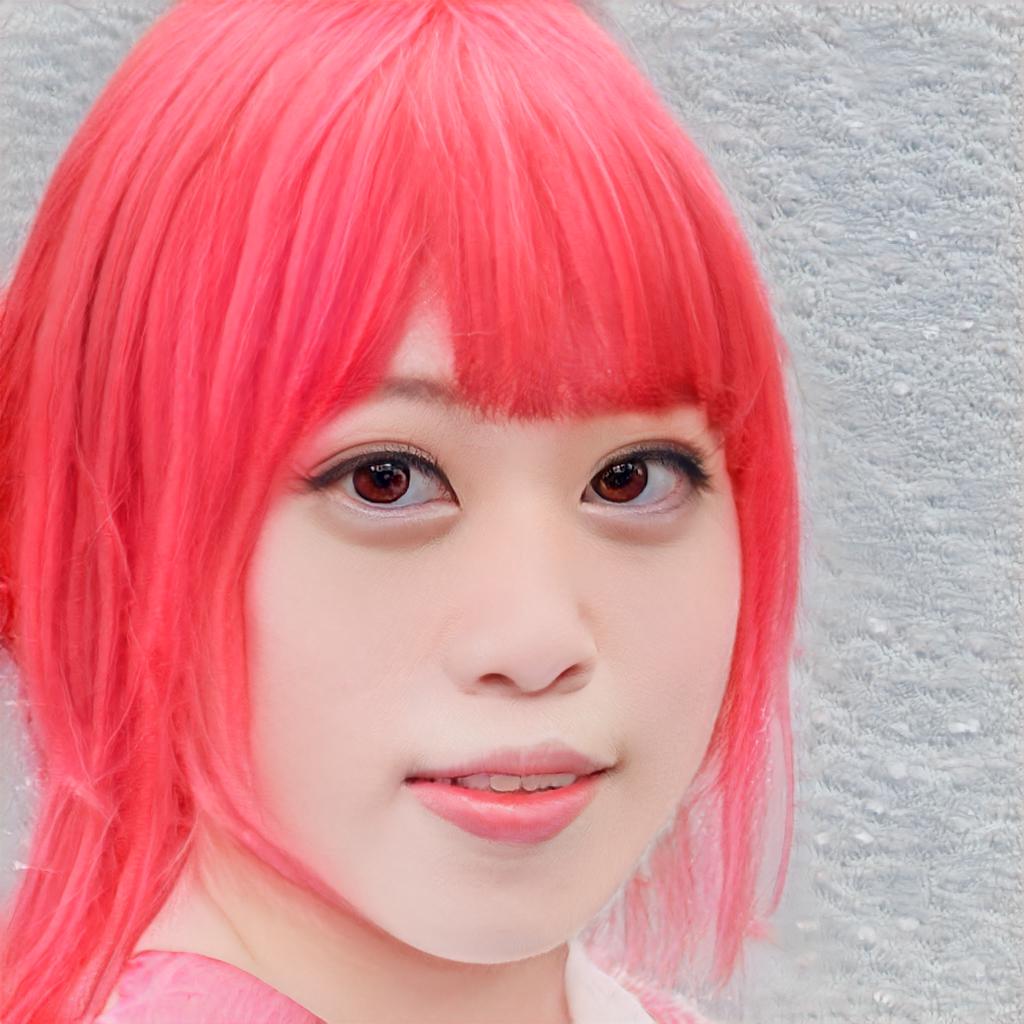} &
 \includegraphics[width=\imsize\linewidth]{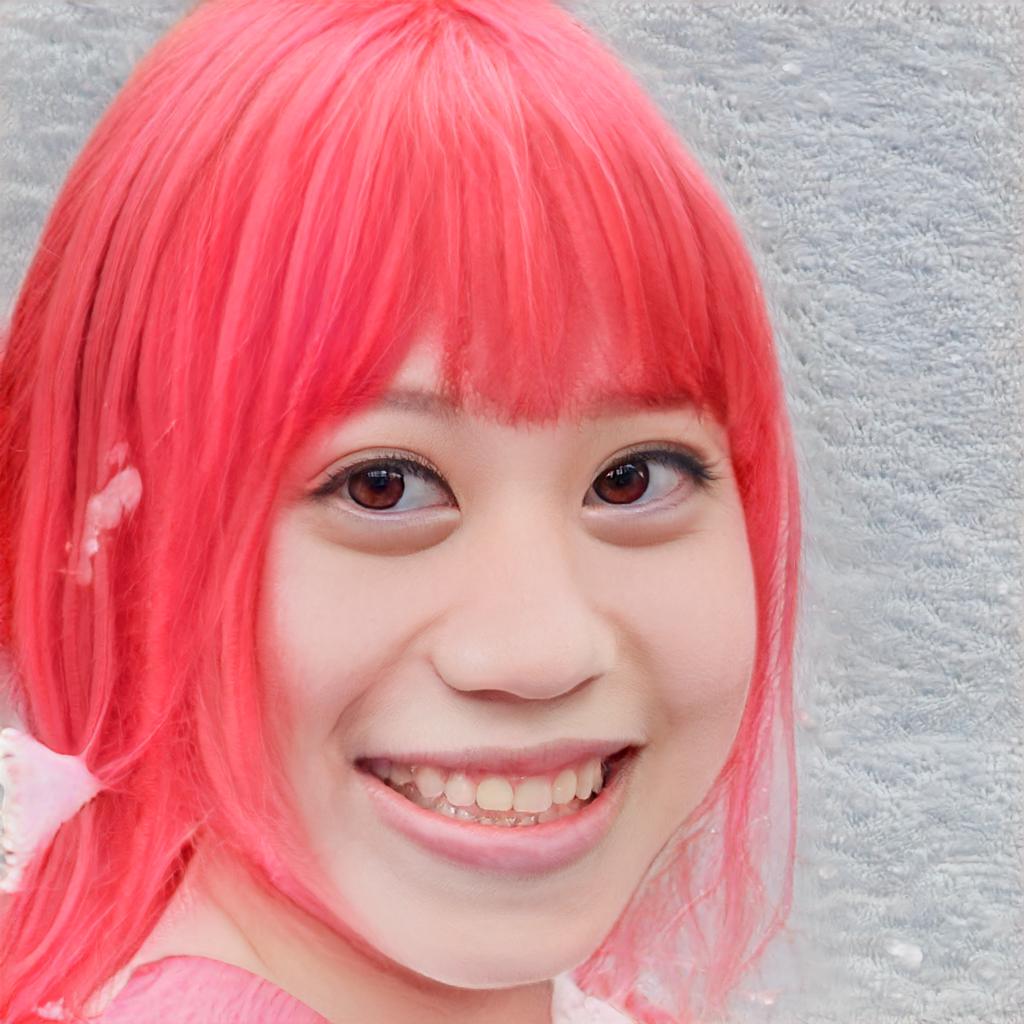} &
 \includegraphics[width=\imsize\linewidth]{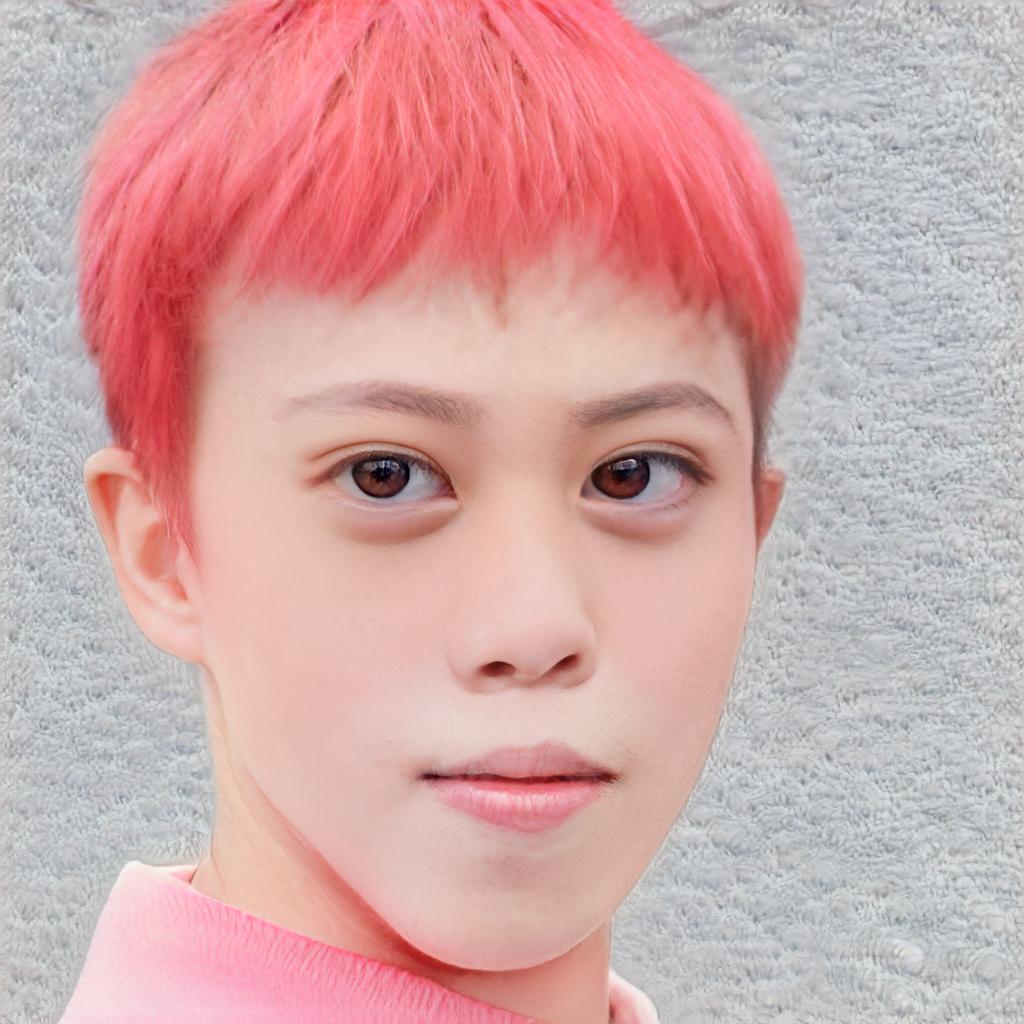} &
 \includegraphics[width=\imsize\linewidth]{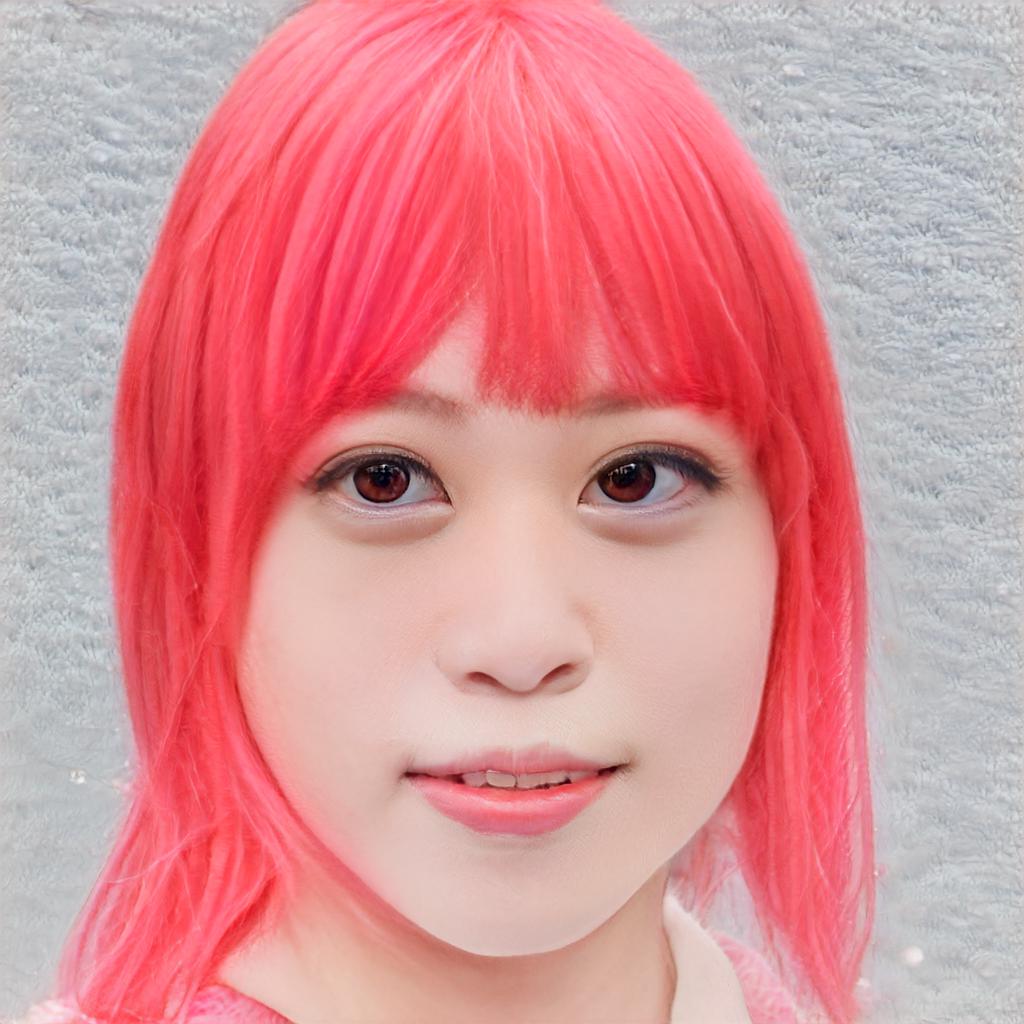} &
 \includegraphics[width=\imsize\linewidth]{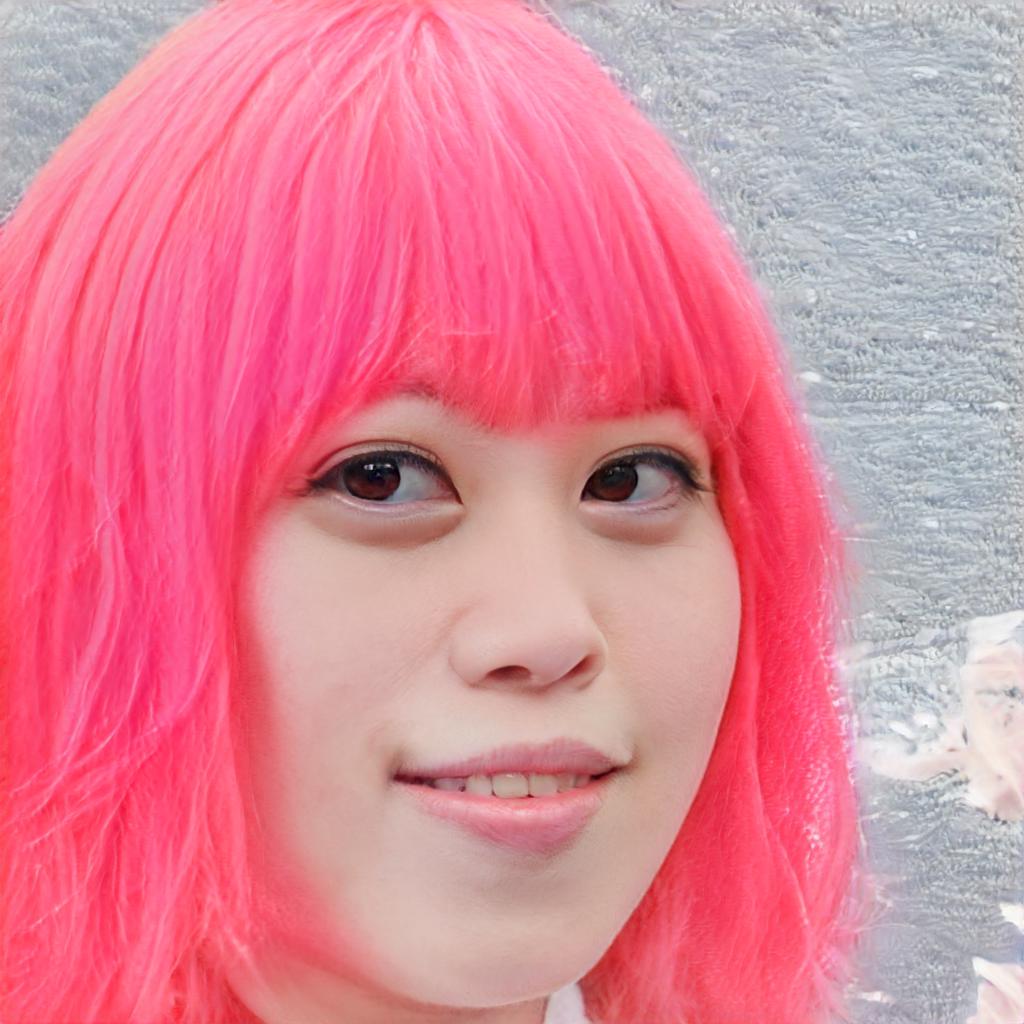} &
 \includegraphics[width=\imsize\linewidth]{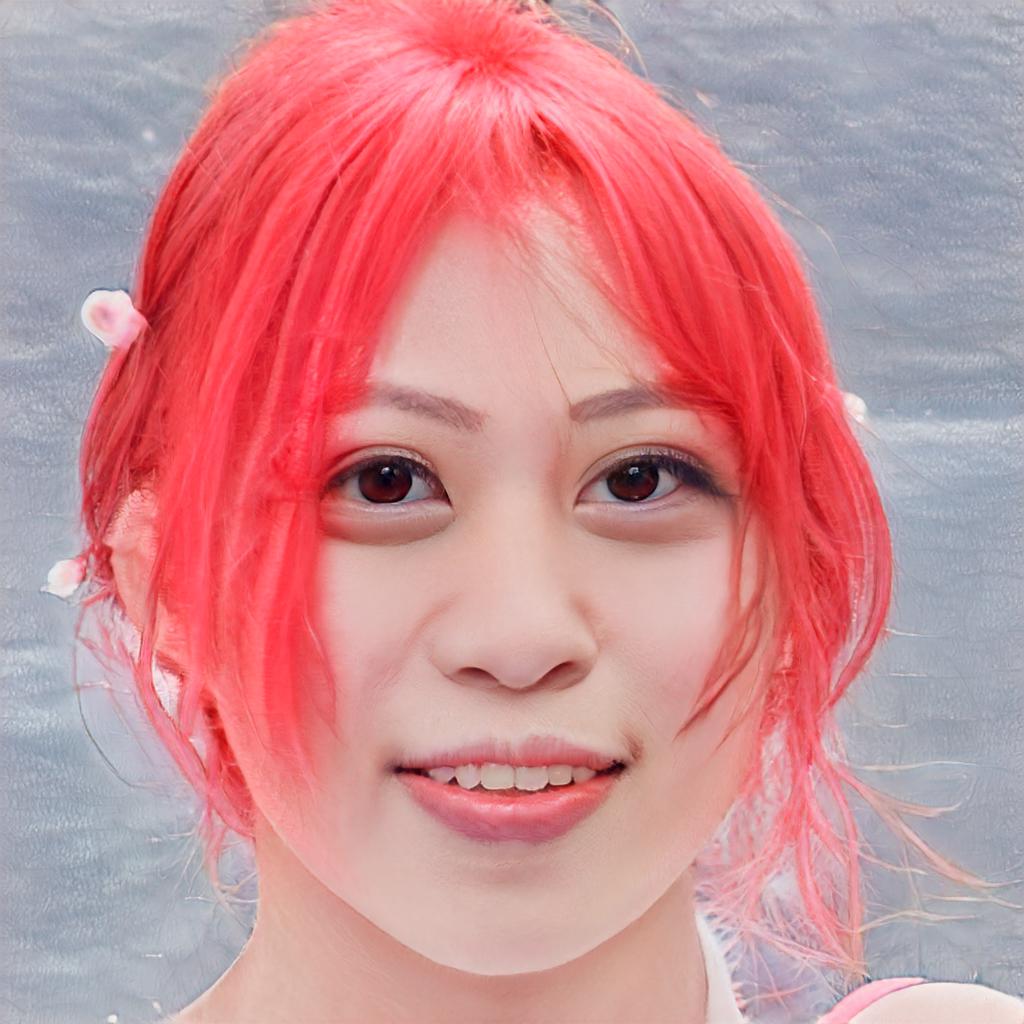} &
 \includegraphics[width=\imsize\linewidth]{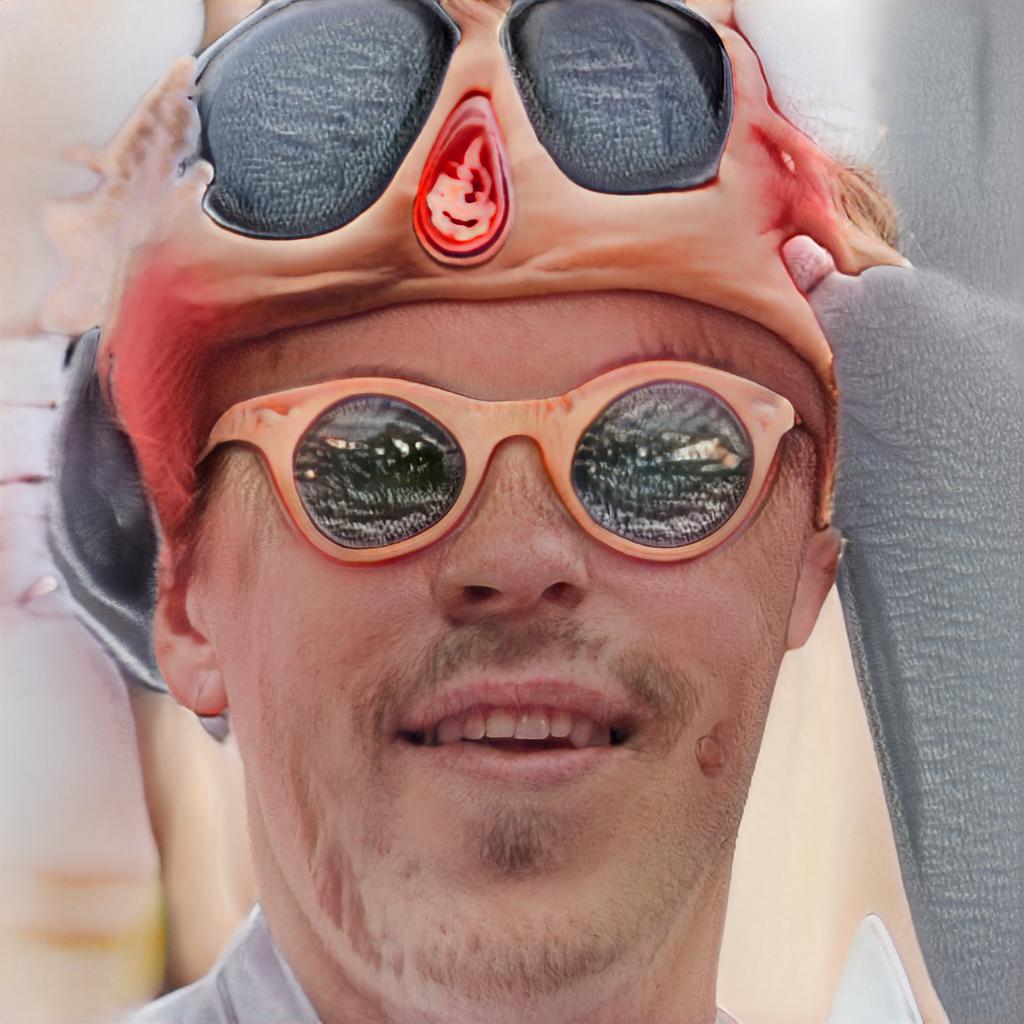} \\ 
 
    \hspace{0.7in}\rotatebox{90}{\hspace{0.0in}InterfaceGan} &
 \includegraphics[width=\imsize\linewidth]{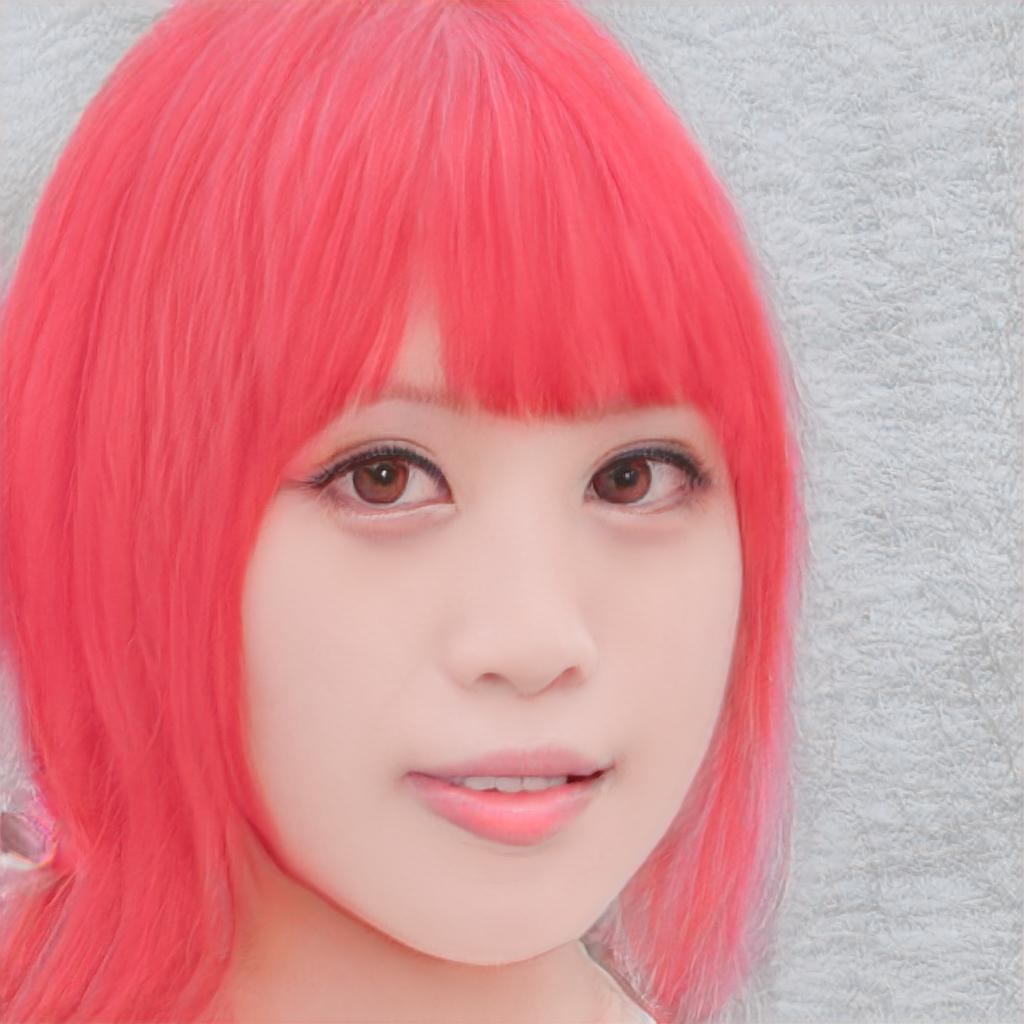} &
 \includegraphics[width=\imsize\linewidth]{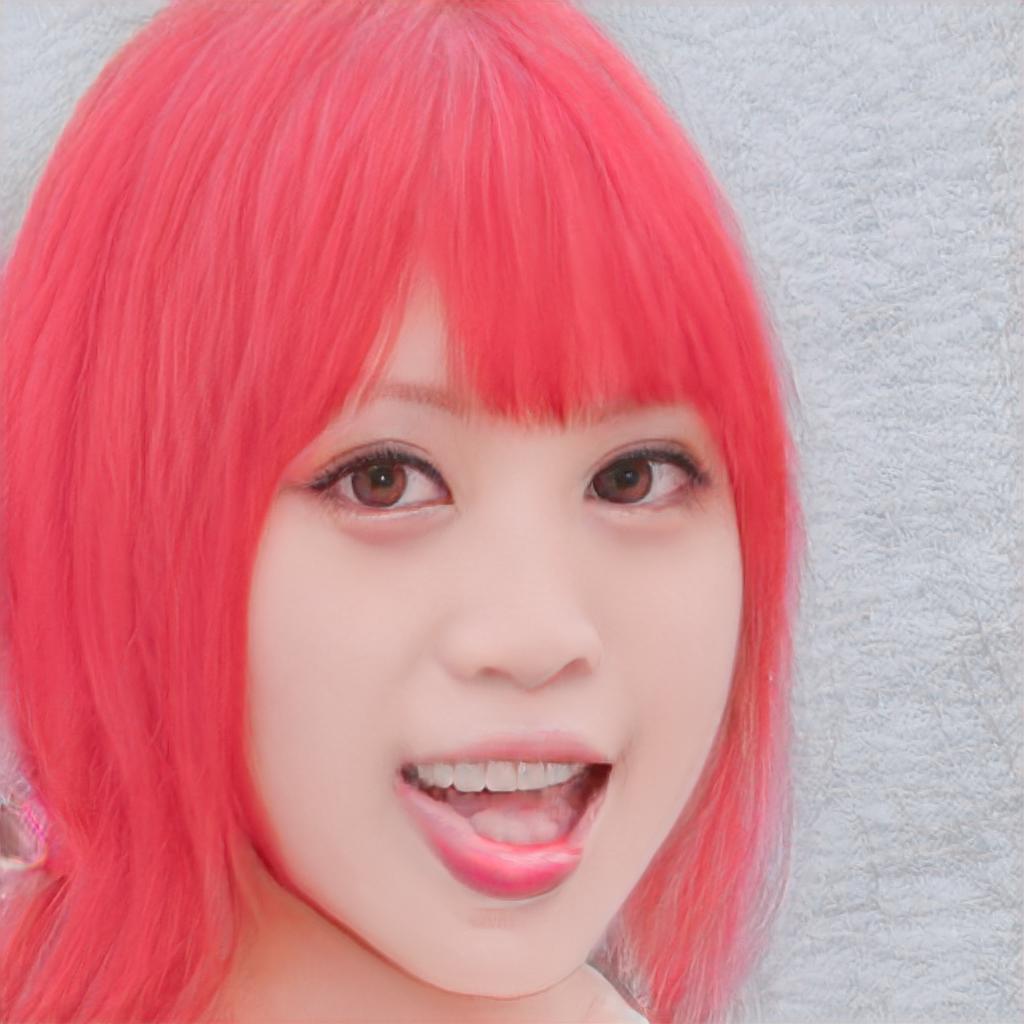} &
 \includegraphics[width=\imsize\linewidth]{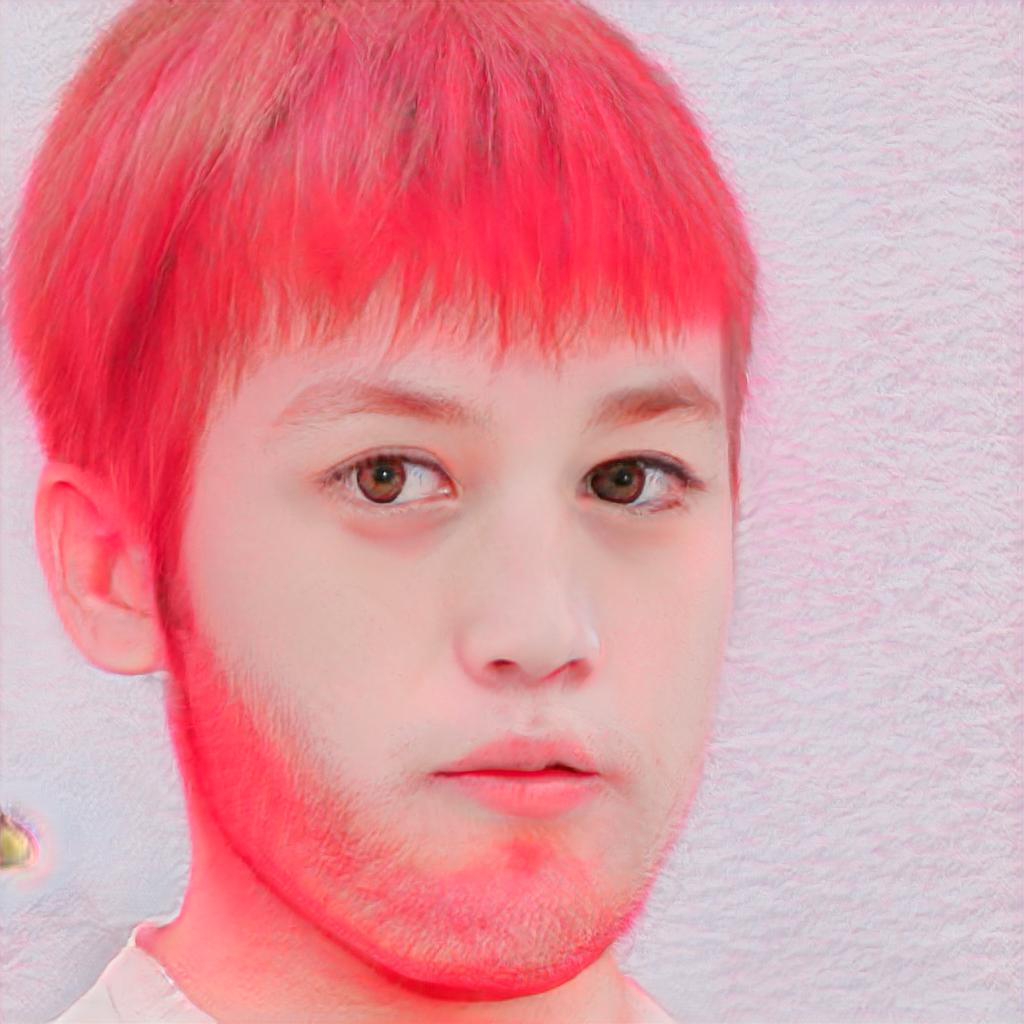} &
 \includegraphics[width=\imsize\linewidth]{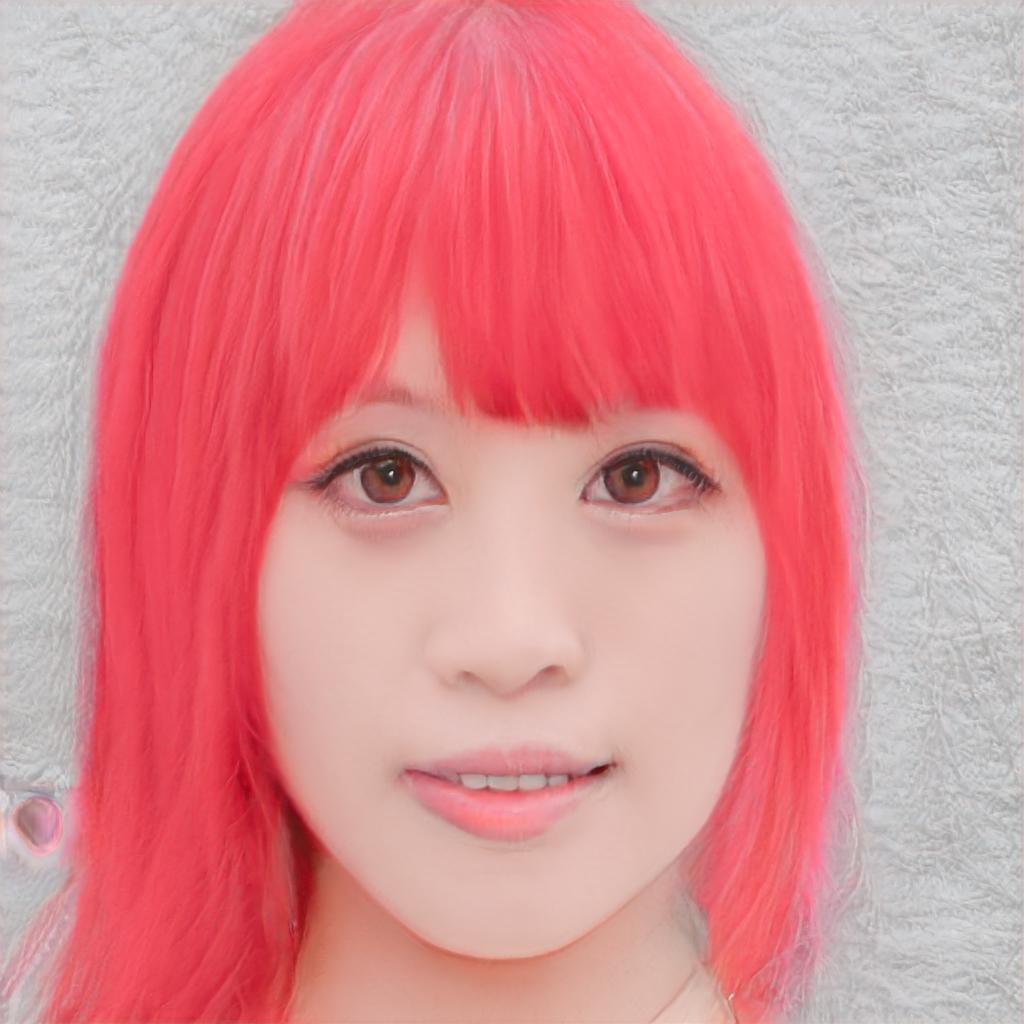} &
 \includegraphics[width=\imsize\linewidth]{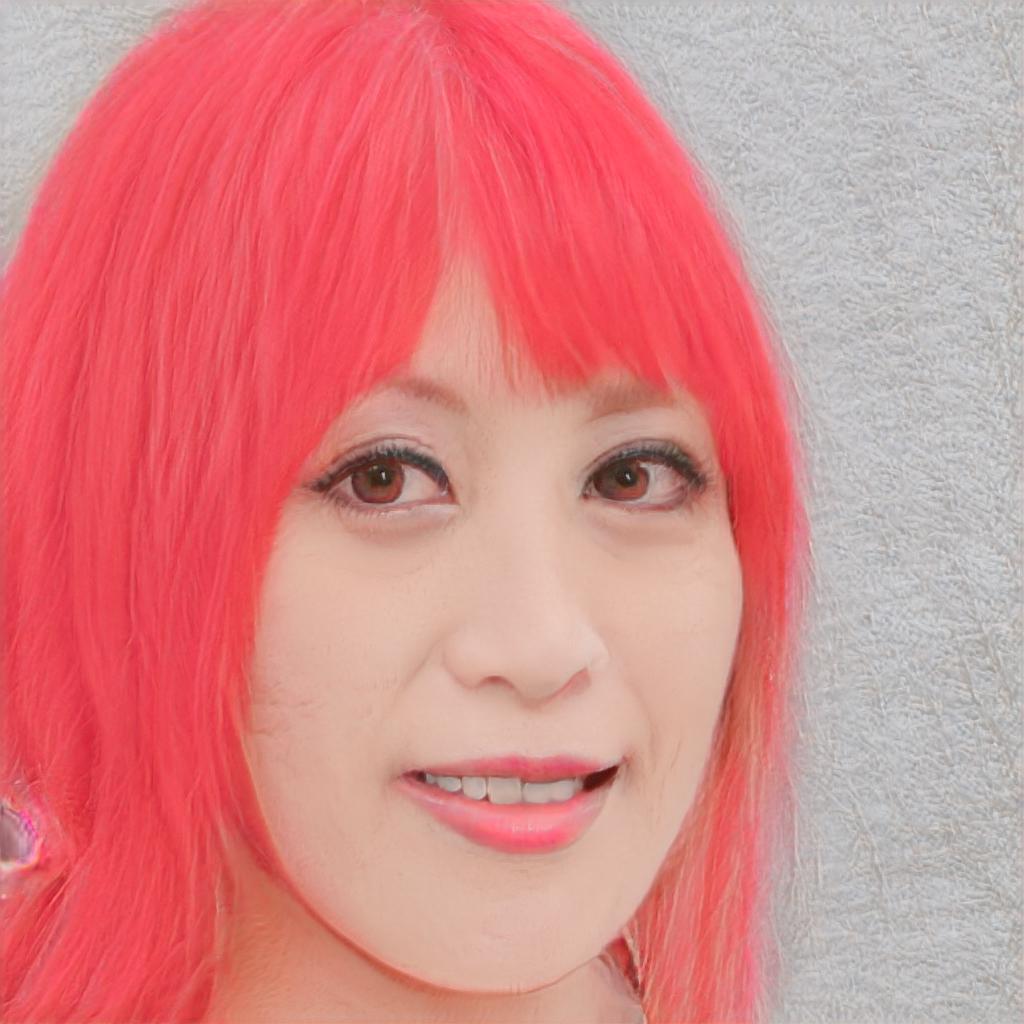} &
 \includegraphics[width=\imsize\linewidth]{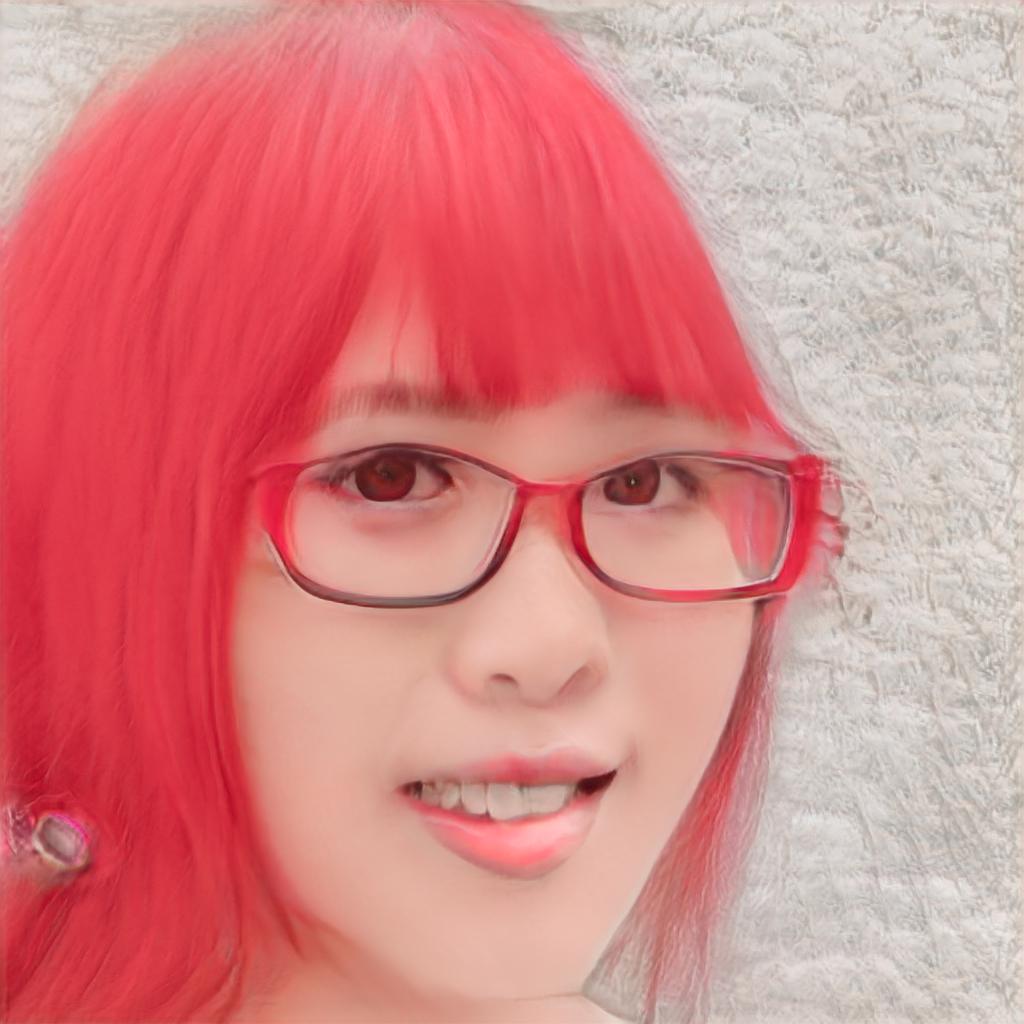} &
 \includegraphics[width=\imsize\linewidth]{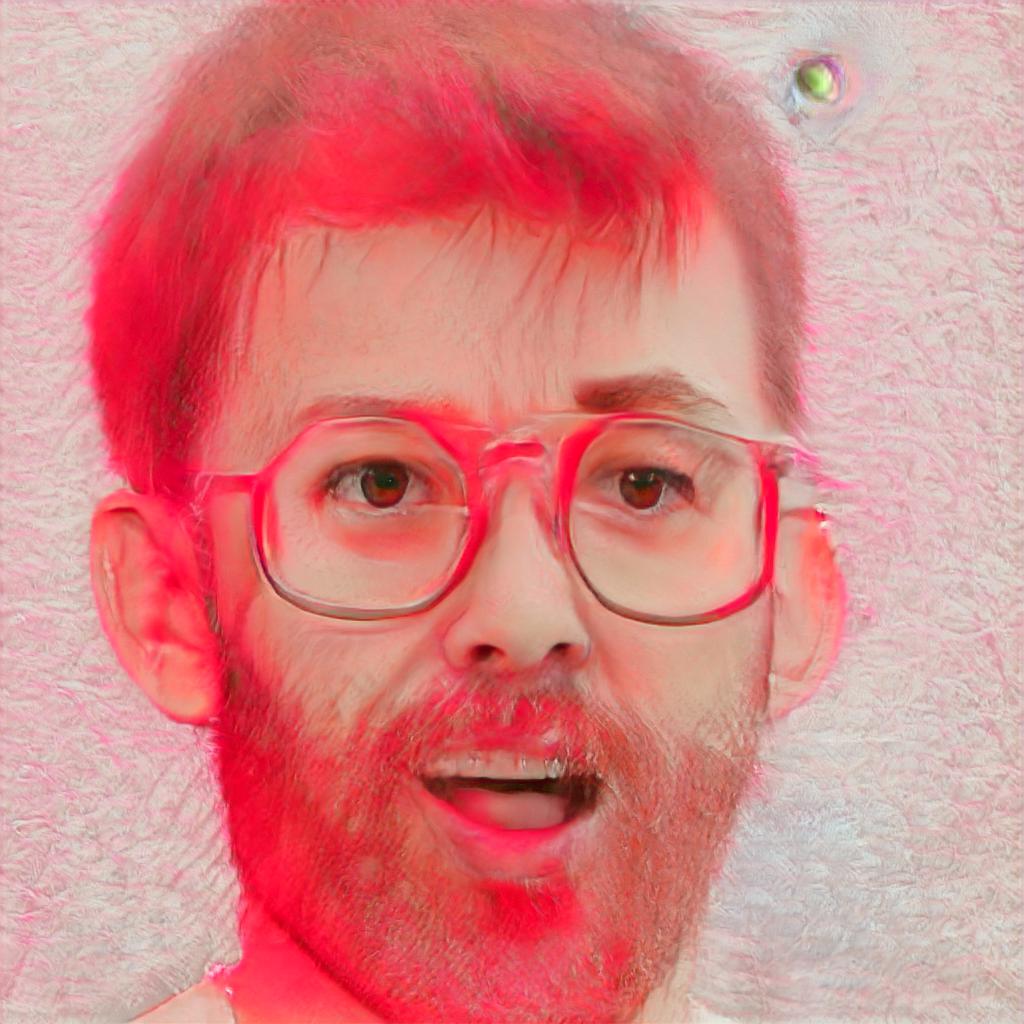} \\

  \hspace{0.7in}\rotatebox{90}{\hspace{0.12in}StyleFlow} &
 \includegraphics[width=\imsize\linewidth]{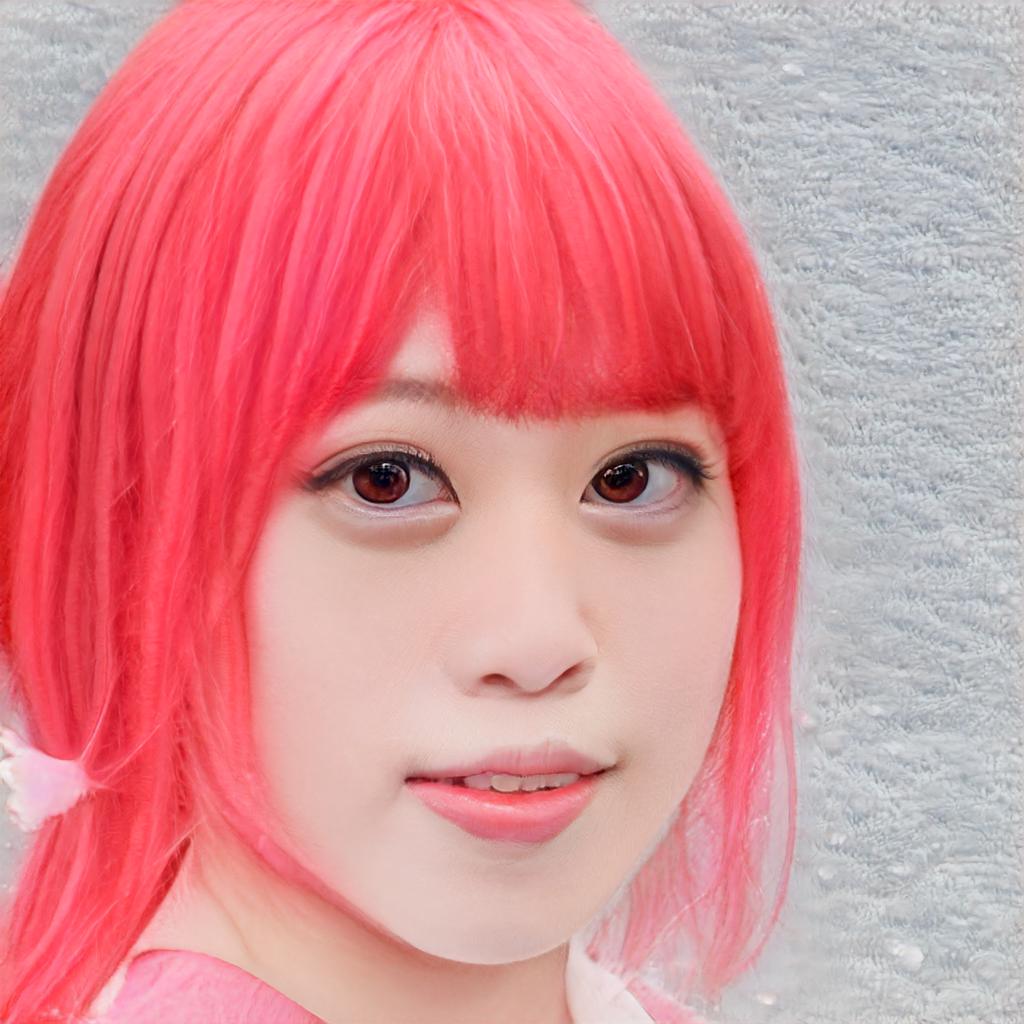} &
 \includegraphics[width=\imsize\linewidth]{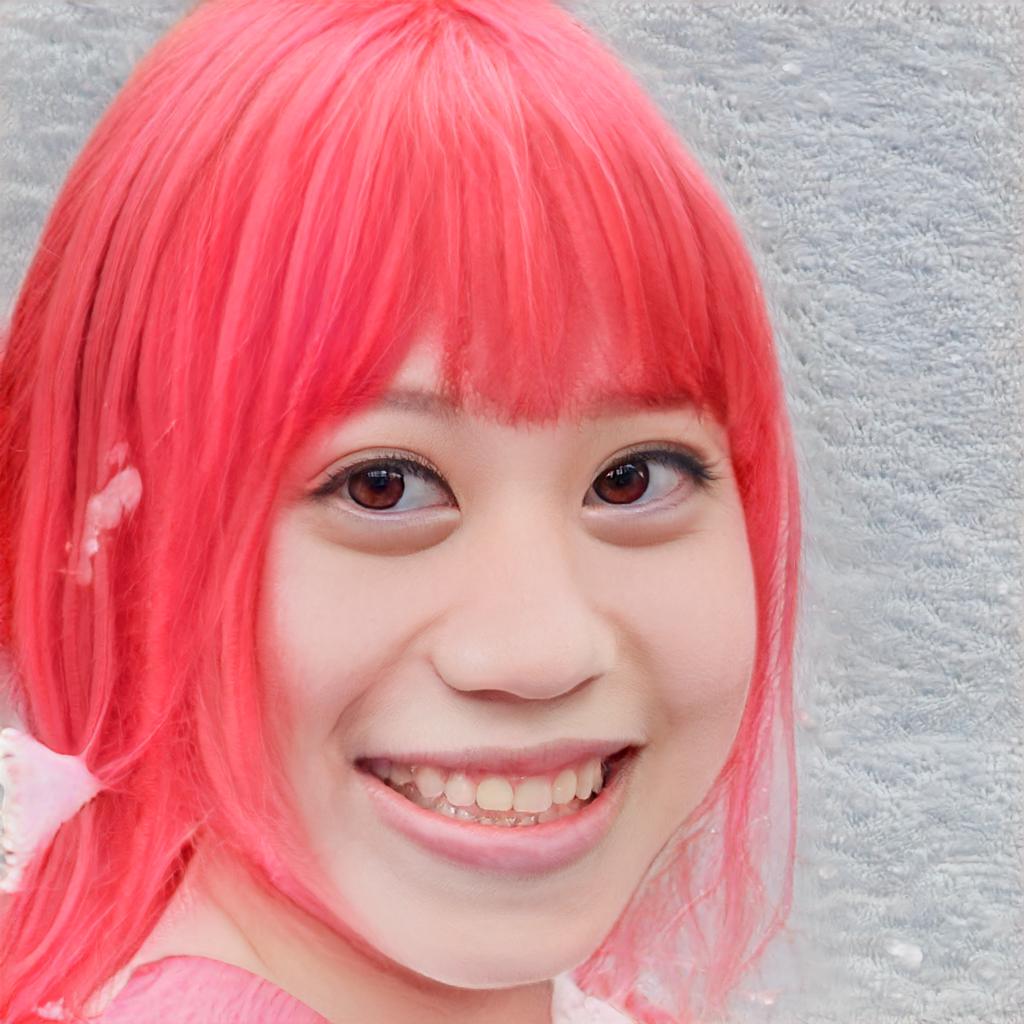} &
 \includegraphics[width=\imsize\linewidth]{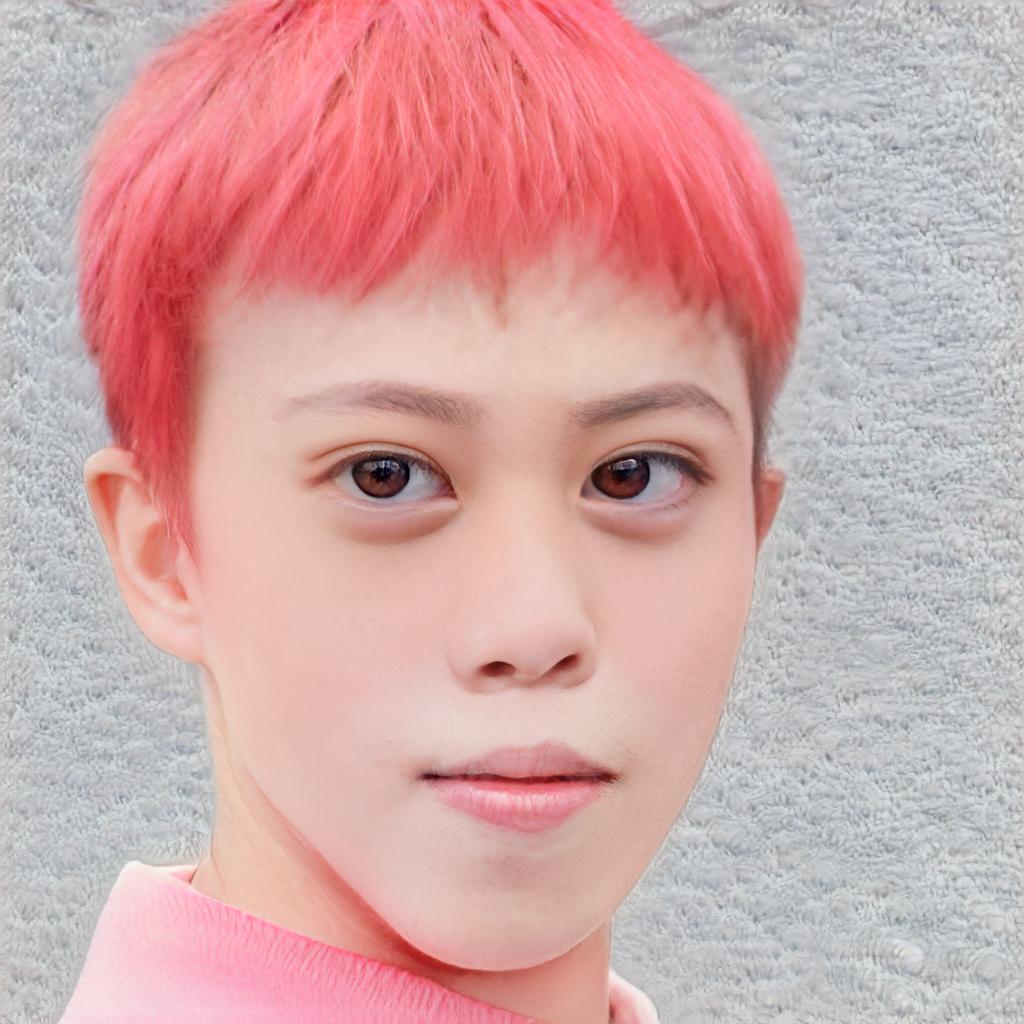} &
 \includegraphics[width=\imsize\linewidth]{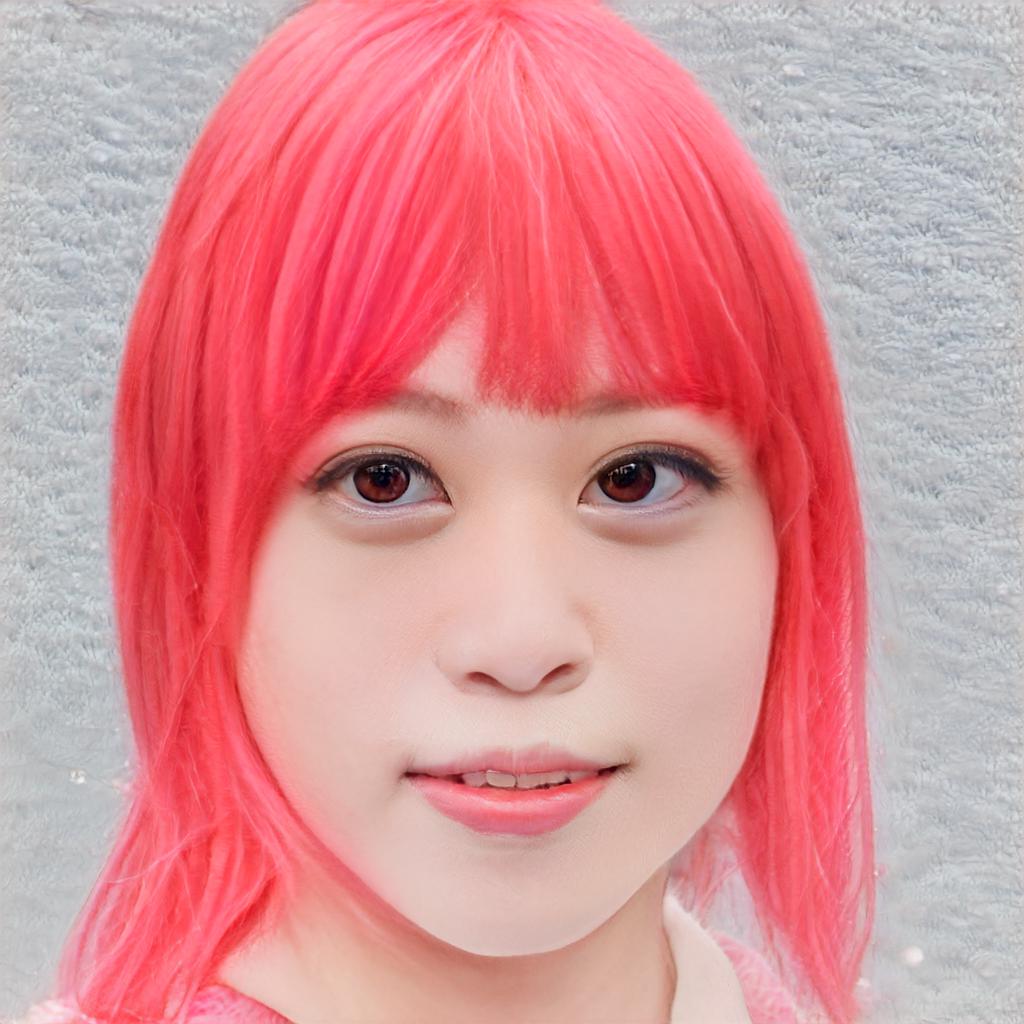} &
 \includegraphics[width=\imsize\linewidth]{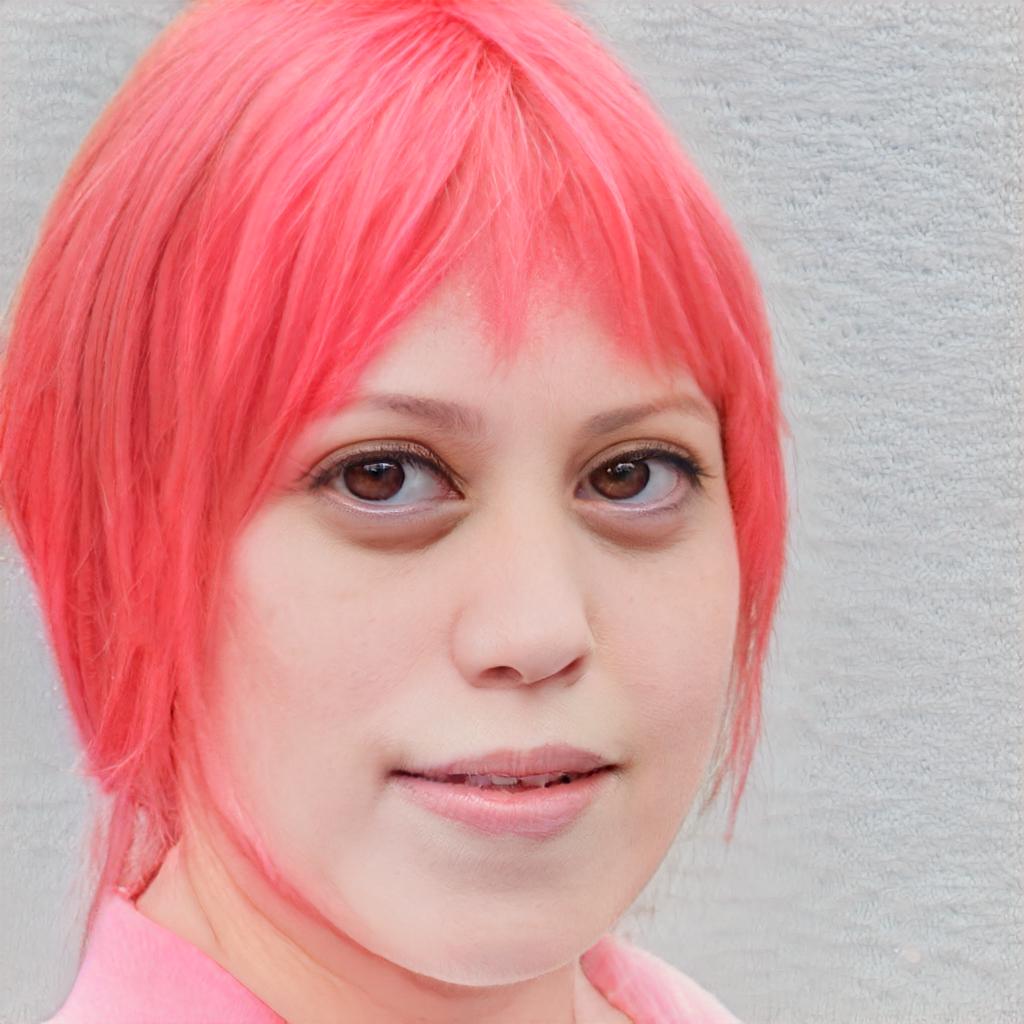} &
 \includegraphics[width=\imsize\linewidth]{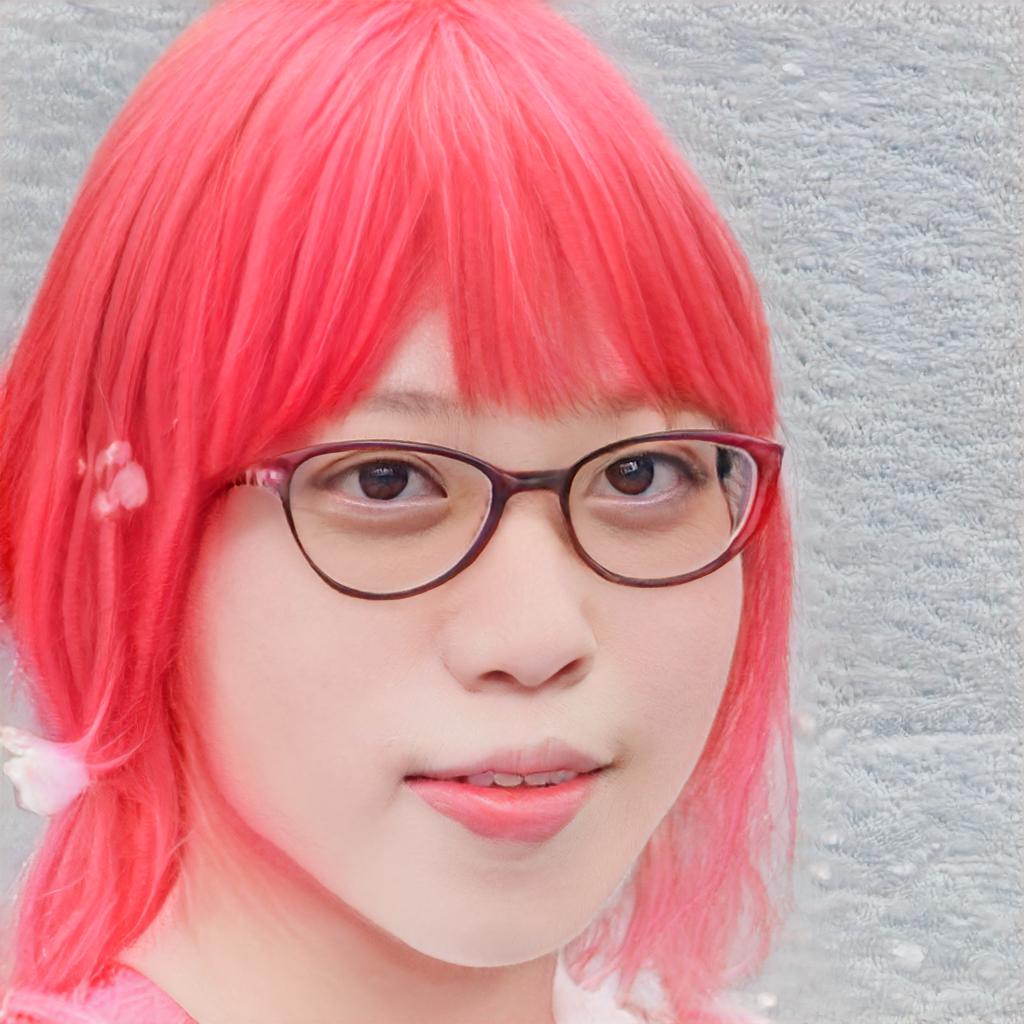} &
 \includegraphics[width=\imsize\linewidth]{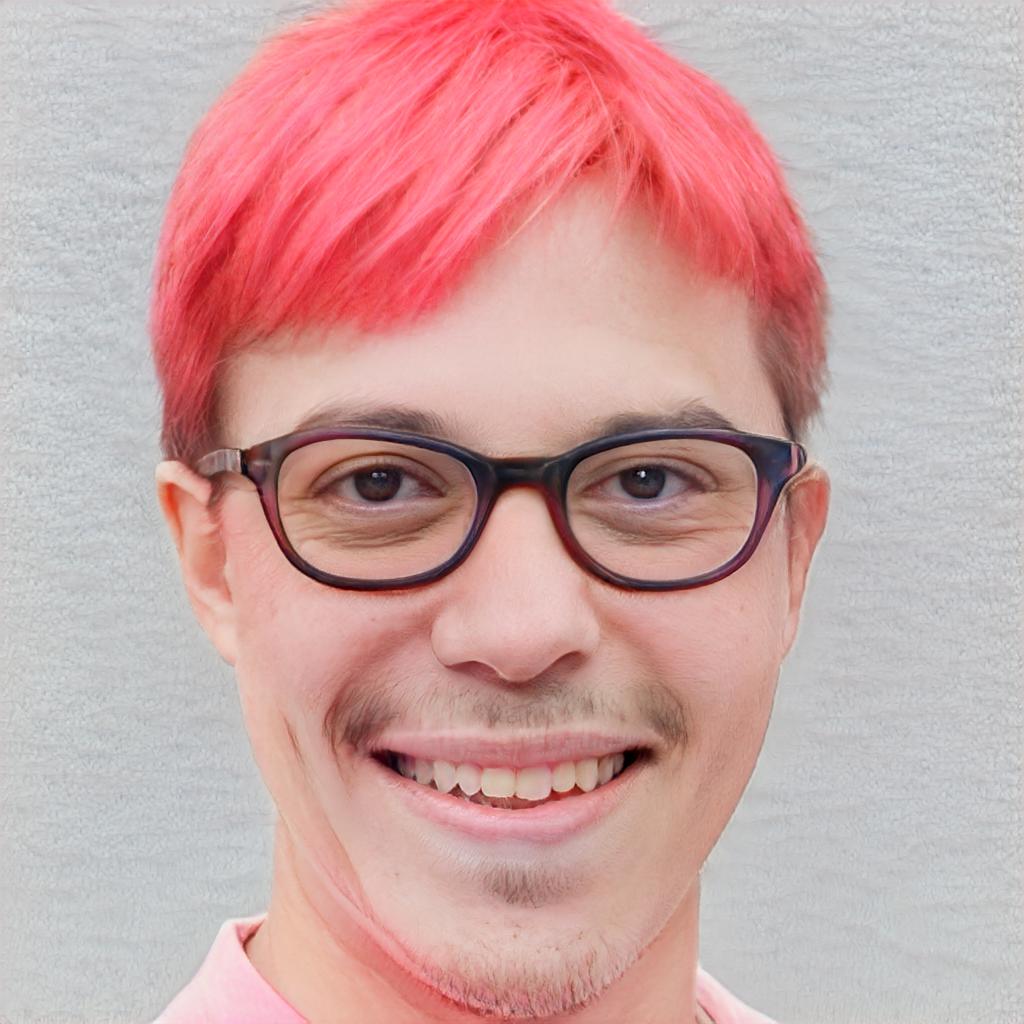} \\ 

 \hspace{0.7in}\rotatebox{90}{\hspace{0.29in}Ours} &
 \includegraphics[width=\imsize\linewidth]{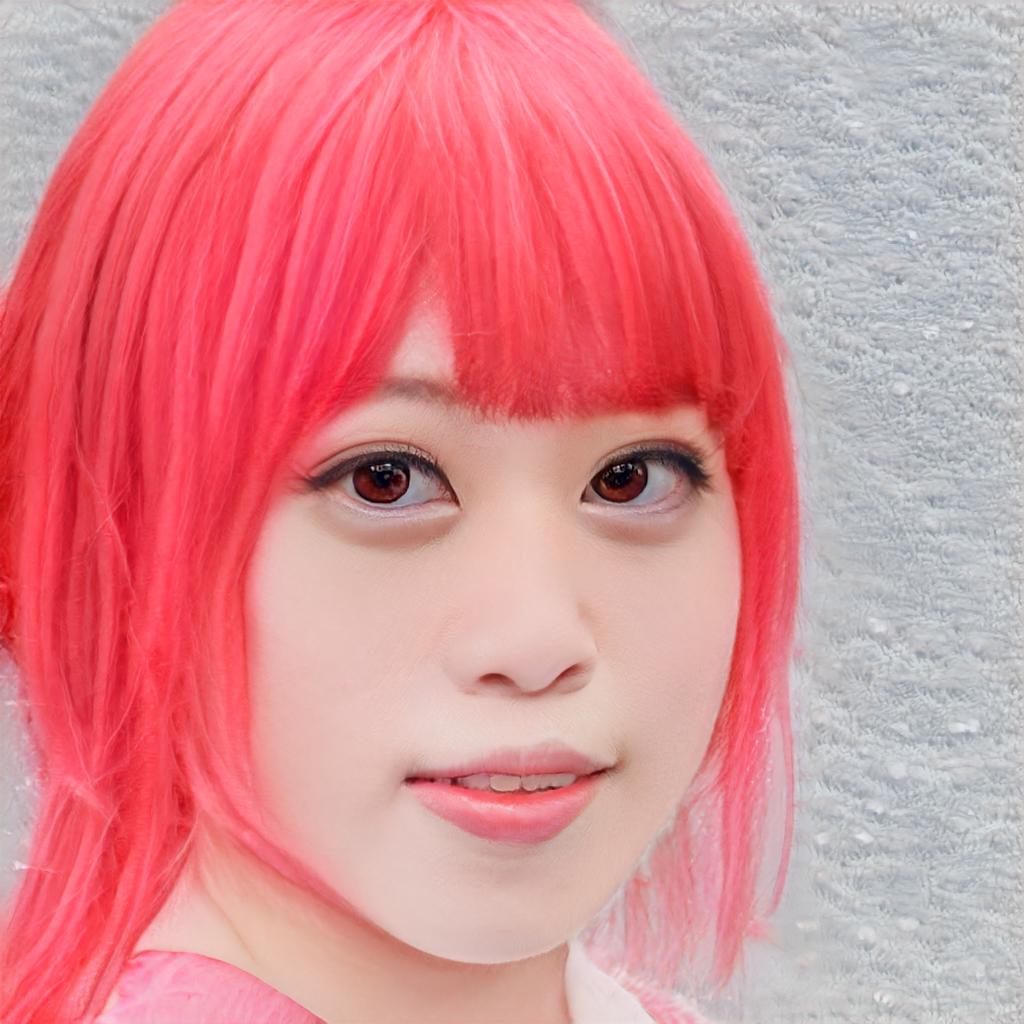} &
 \includegraphics[width=\imsize\linewidth]{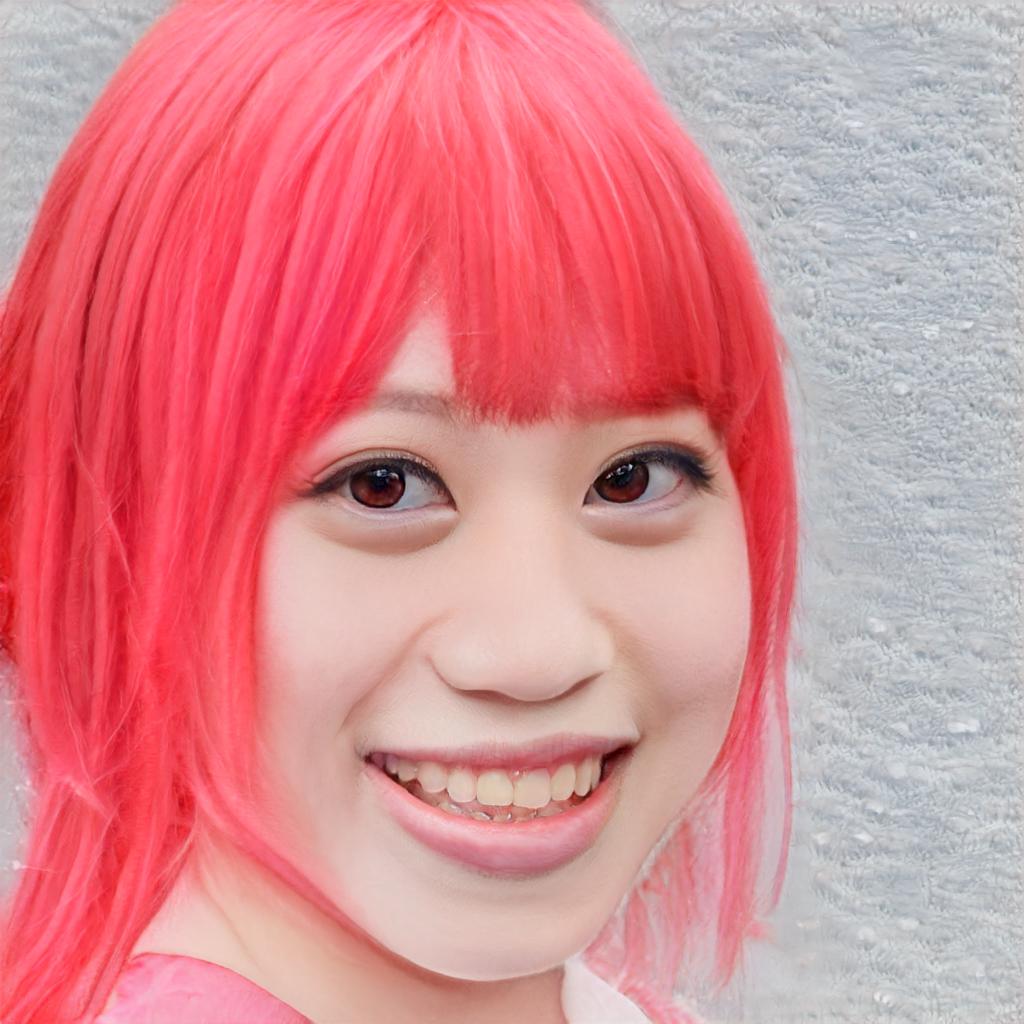} &
 \includegraphics[width=\imsize\linewidth]{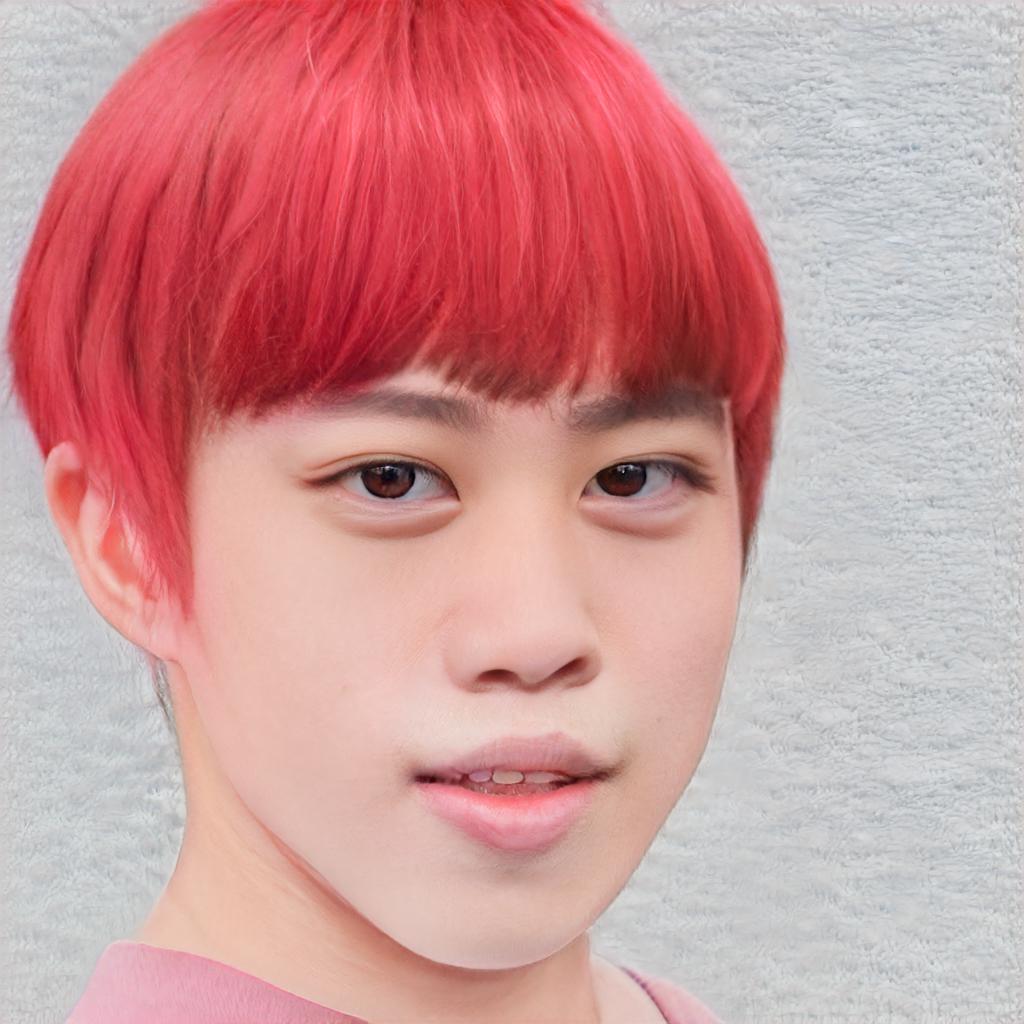} &
 \includegraphics[width=\imsize\linewidth]{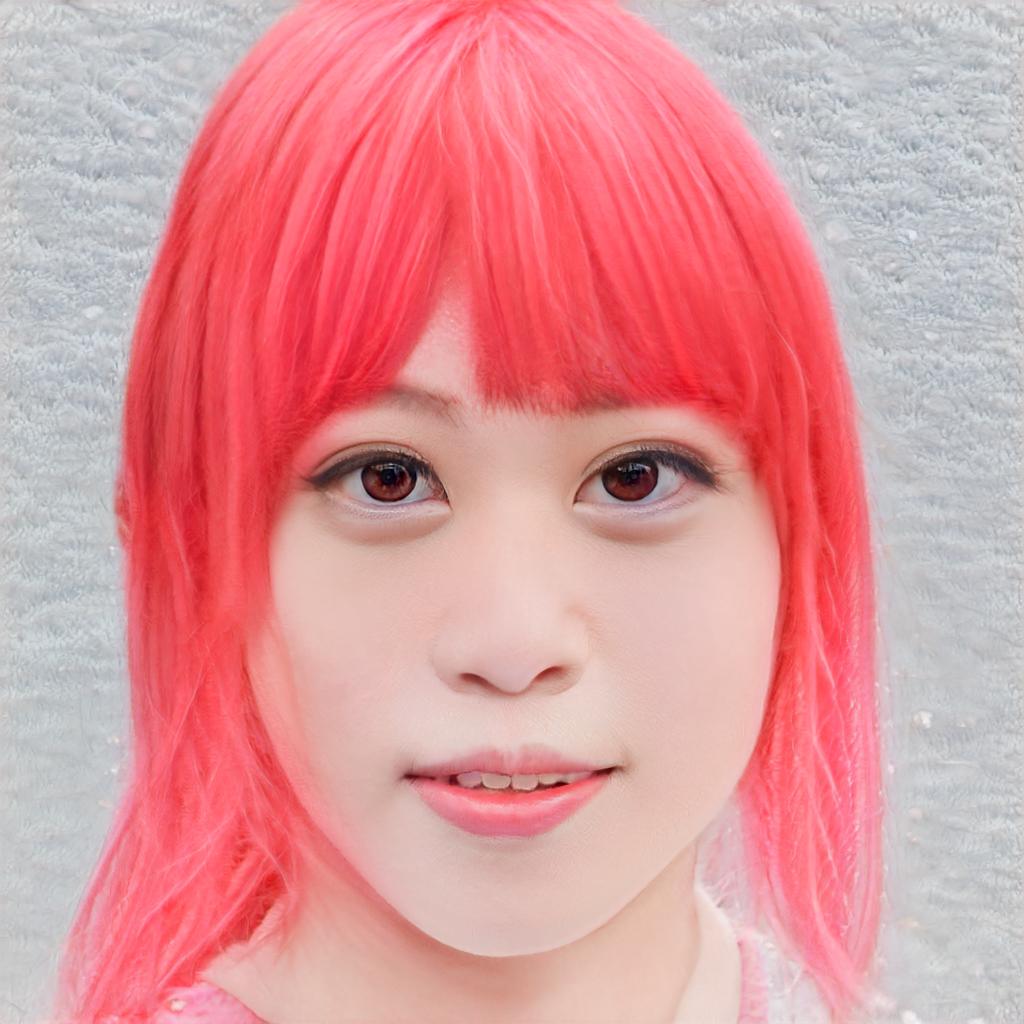} &
 \includegraphics[width=\imsize\linewidth]{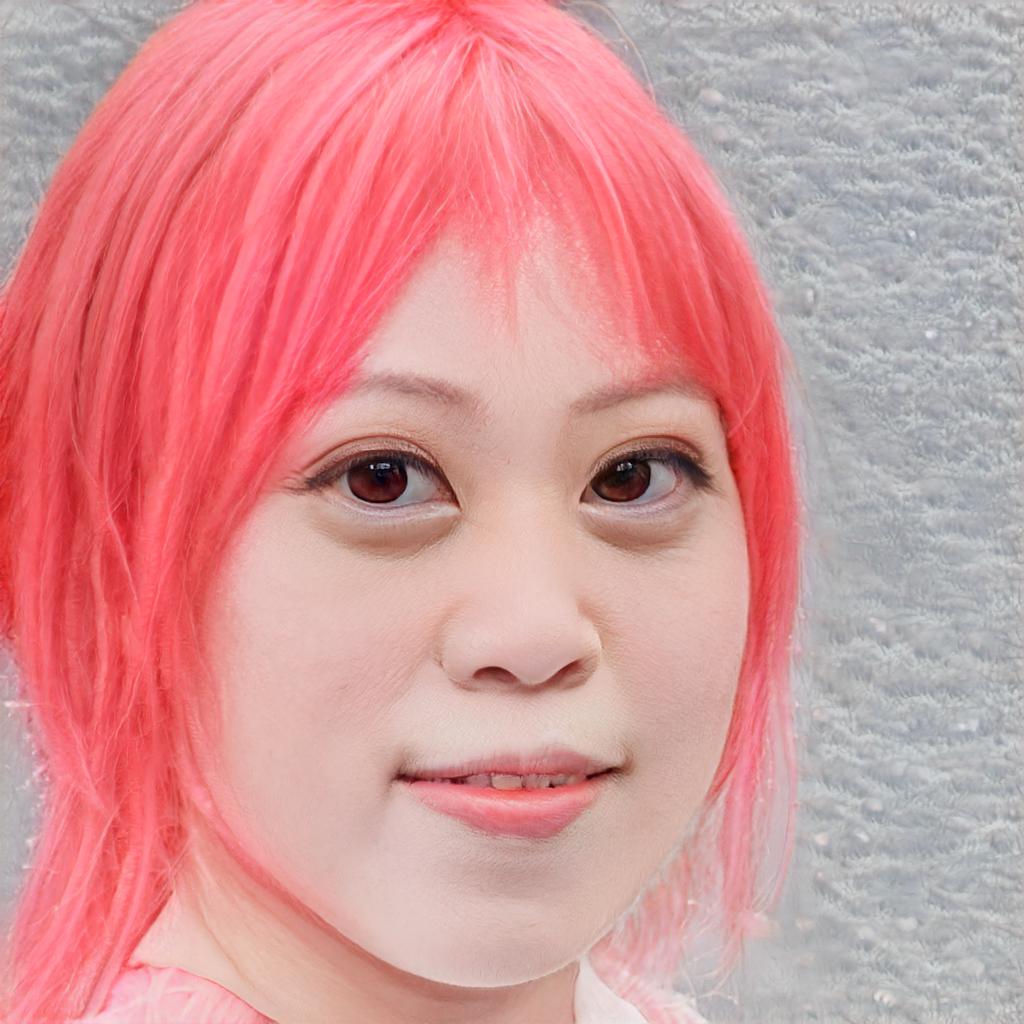} &
 \includegraphics[width=\imsize\linewidth]{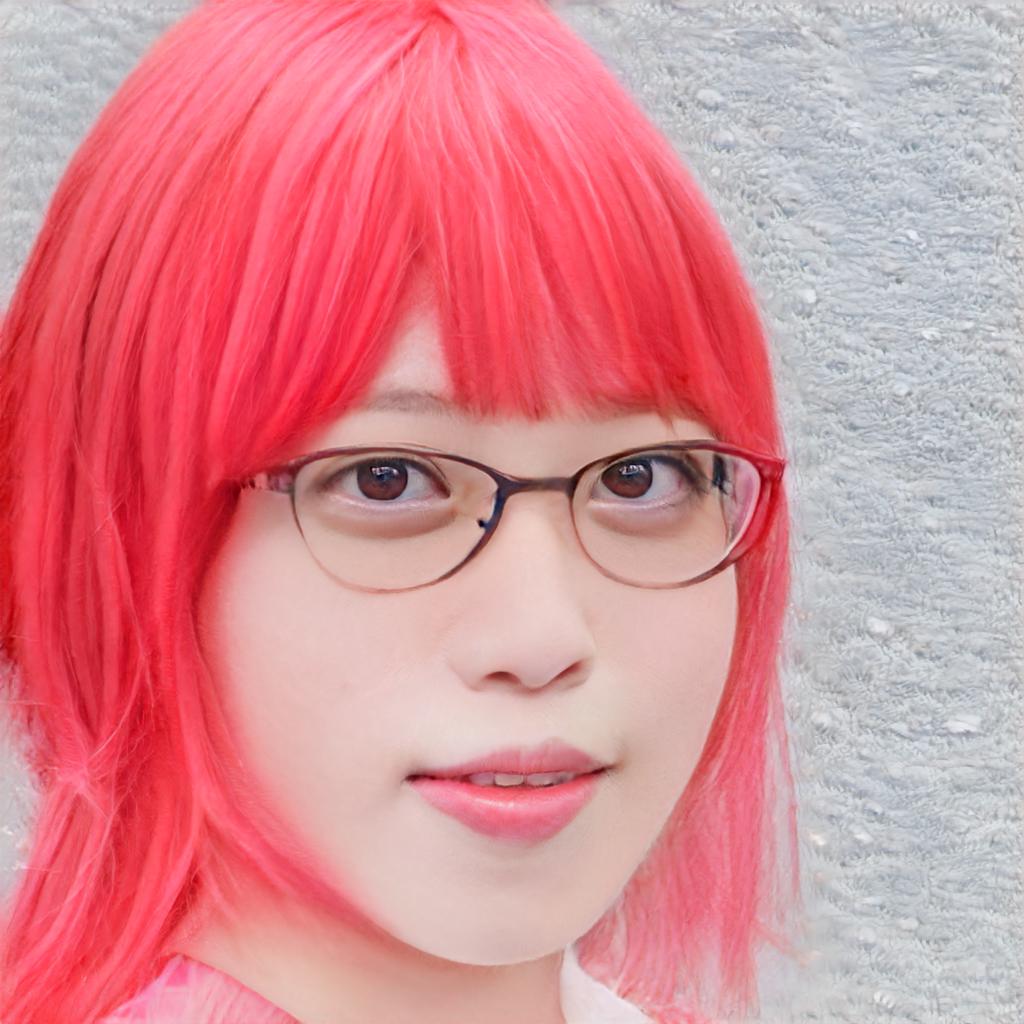} &
 \includegraphics[width=\imsize\linewidth]{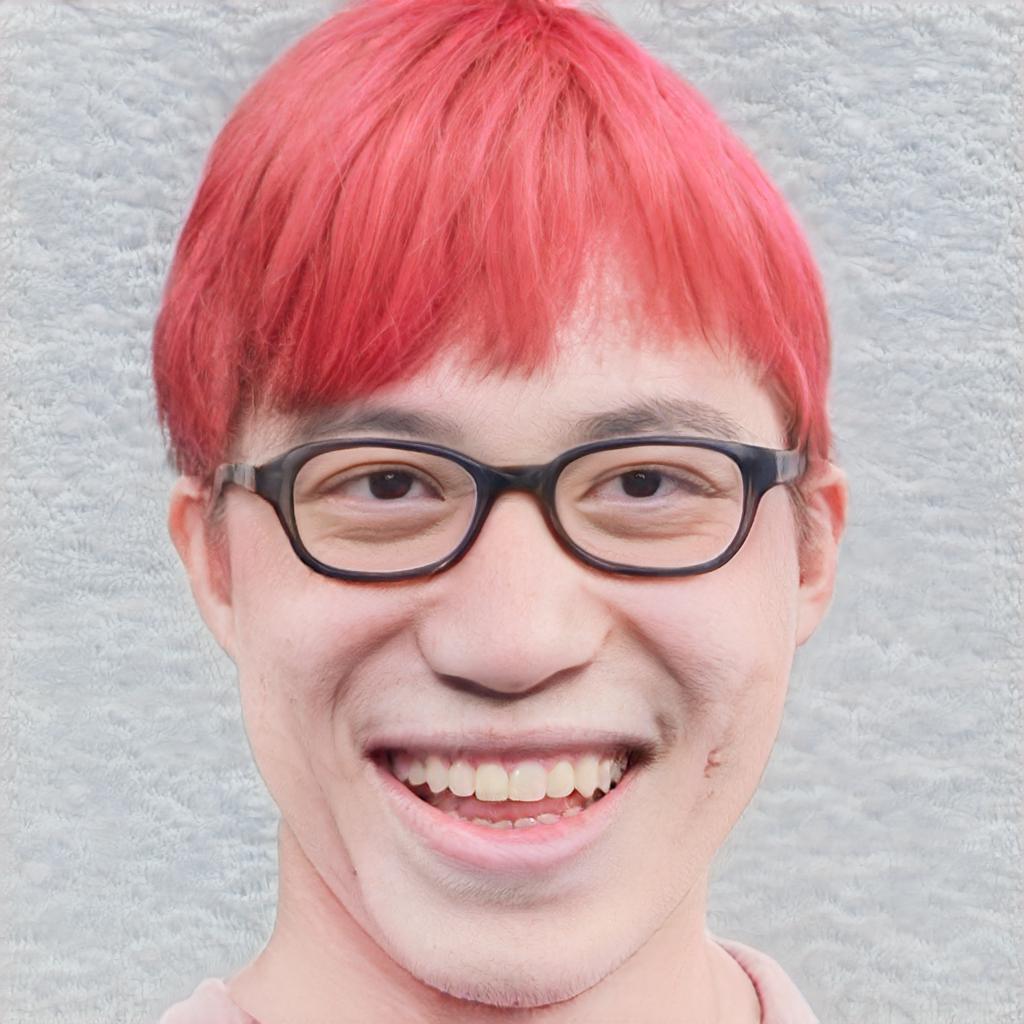} \\
 
\end{tabular}
\caption{Comparison of real image editing between our method and the baseline approaches. The edit direction for each attribute was determined using a linear SVM classifier.}
\label{fig:fig3}
\end{figure*}

\newcommand*{\figsize}{0.16}

\begin{figure}
\centering
\setlength{\tabcolsep}{1pt}
    \begin{tabular}{c|ccccc}
        \multicolumn{6}{c}{$\xrightarrow[]{\text{edit direction}}$} \\
         \includegraphics[width=\figsize\linewidth]{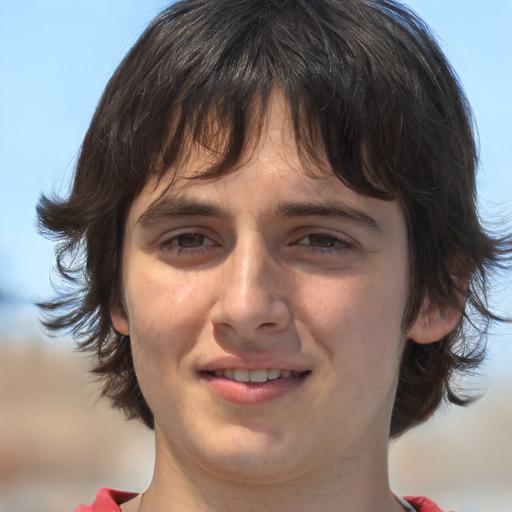} &
         \includegraphics[width=\figsize\linewidth]{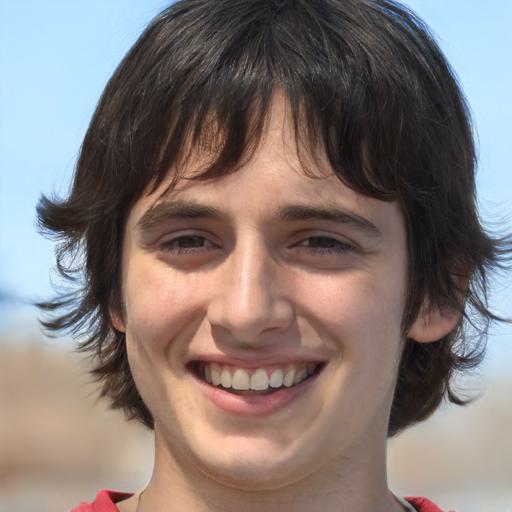} &
         \includegraphics[width=\figsize\linewidth]{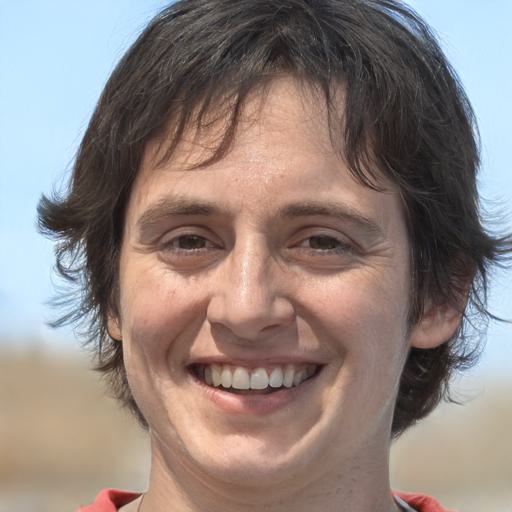} &
         \includegraphics[width=\figsize\linewidth]{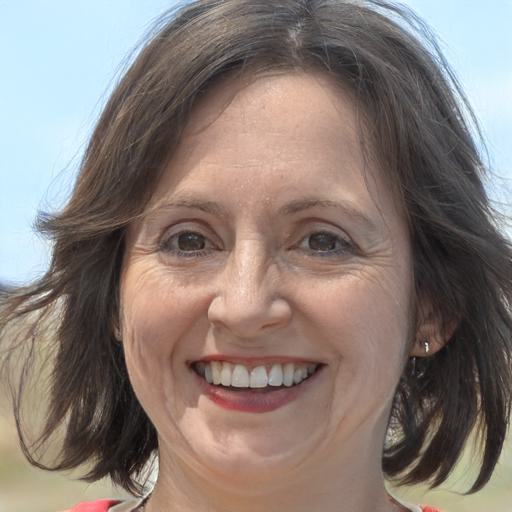} &
         \includegraphics[width=\figsize\linewidth]{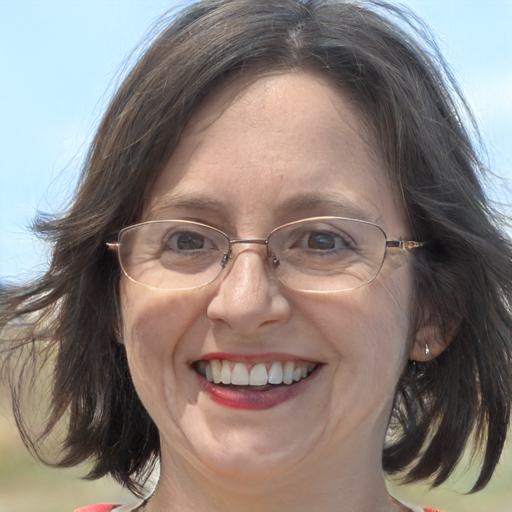} &
         \includegraphics[width=\figsize\linewidth]{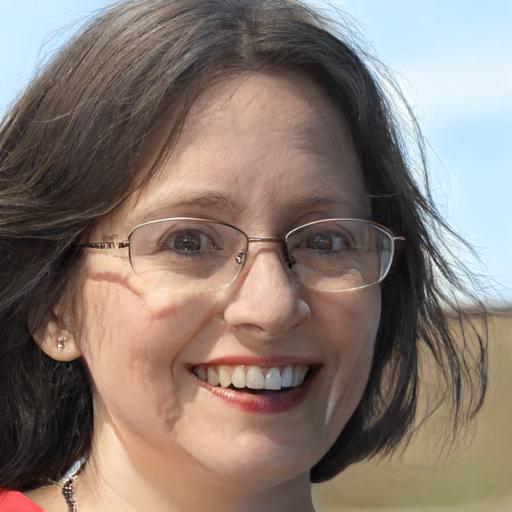} \\
         \includegraphics[width=\figsize\linewidth]{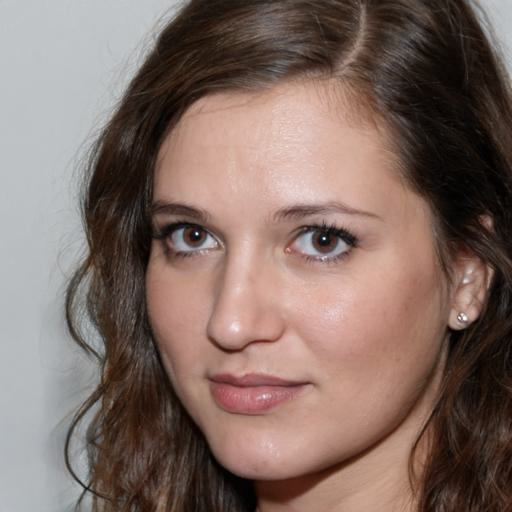} &
         \includegraphics[width=\figsize\linewidth]{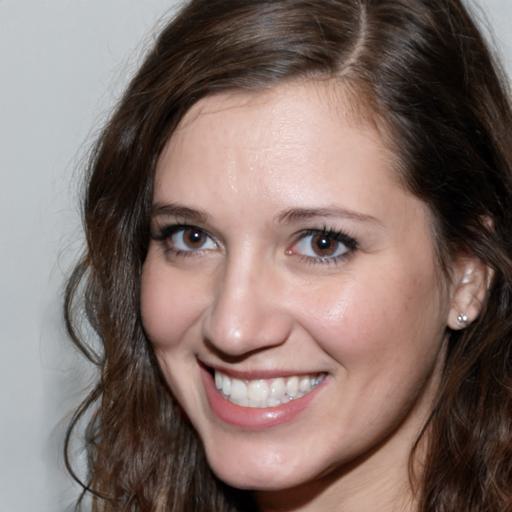} &
         \includegraphics[width=\figsize\linewidth]{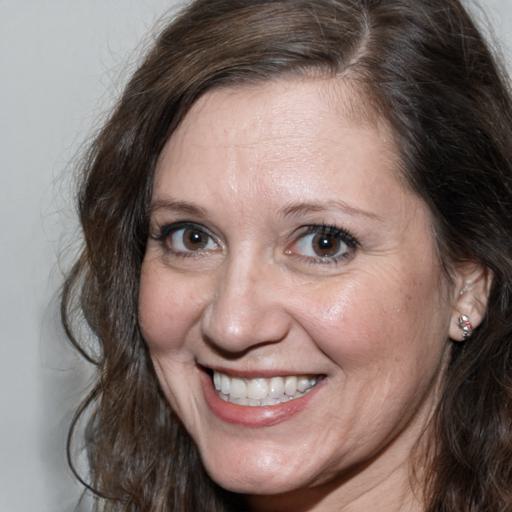} &
         \includegraphics[width=\figsize\linewidth]{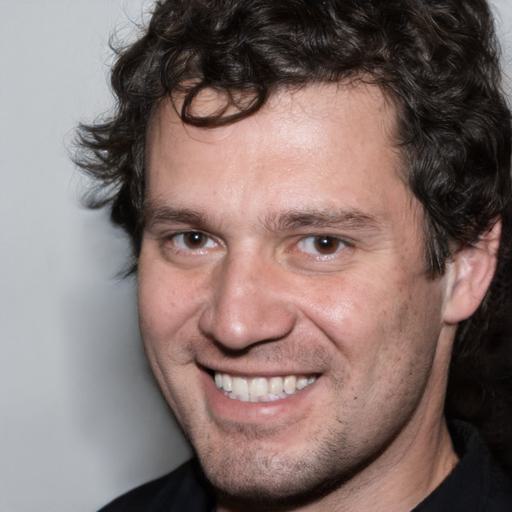} &
         \includegraphics[width=\figsize\linewidth]{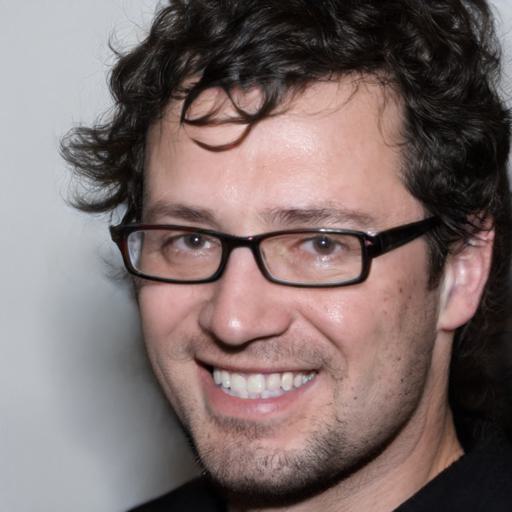} &
         \includegraphics[width=\figsize\linewidth]{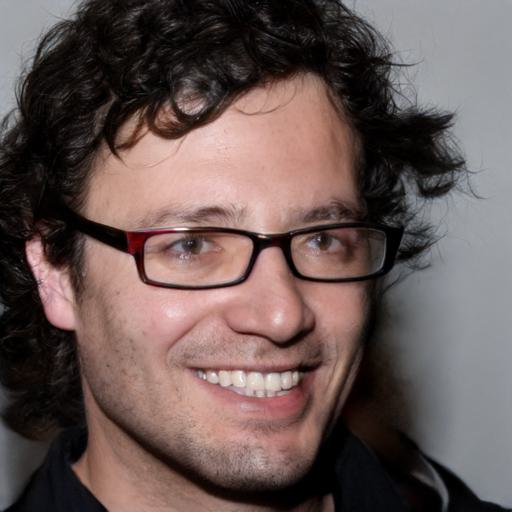} \\
         \includegraphics[width=\figsize\linewidth]{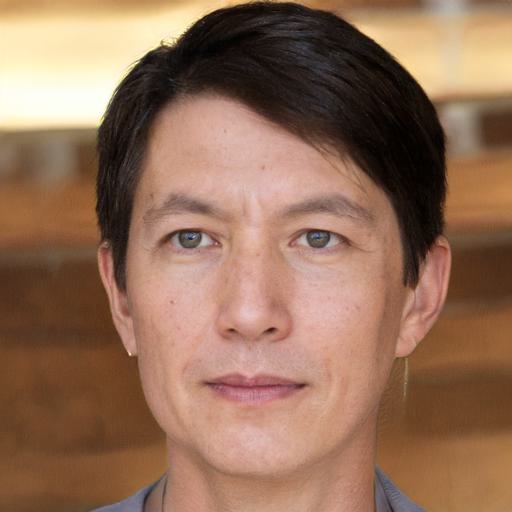} &
         \includegraphics[width=\figsize\linewidth]{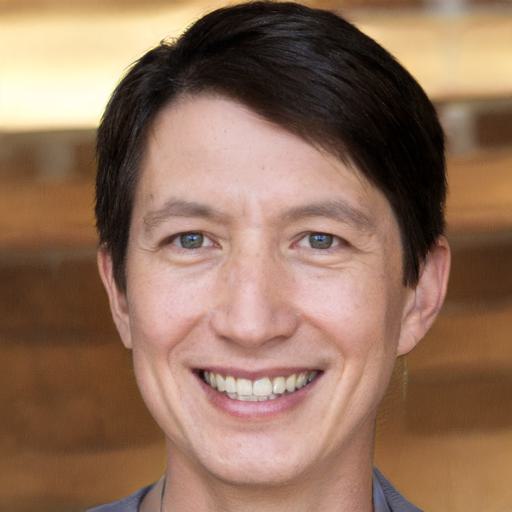} &
         \includegraphics[width=\figsize\linewidth]{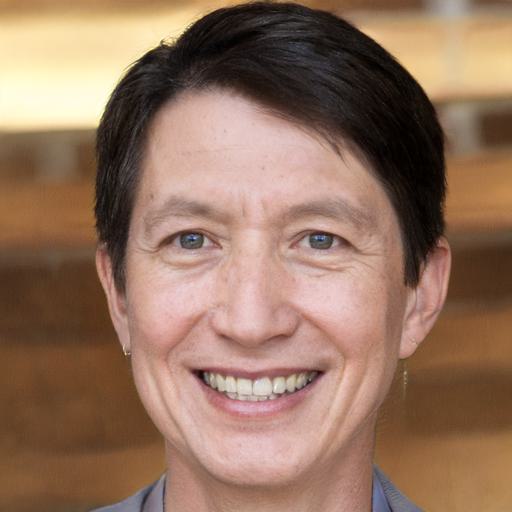} &
         \includegraphics[width=\figsize\linewidth]{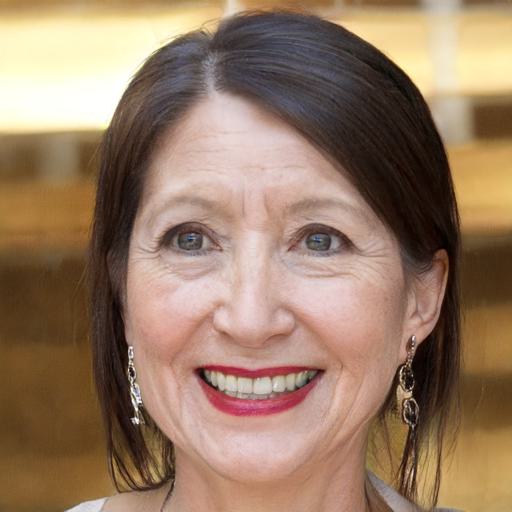} &
         \includegraphics[width=\figsize\linewidth]{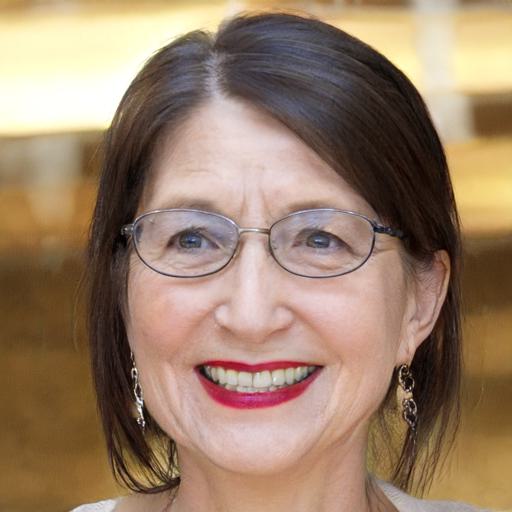} &
         \includegraphics[width=\figsize\linewidth]{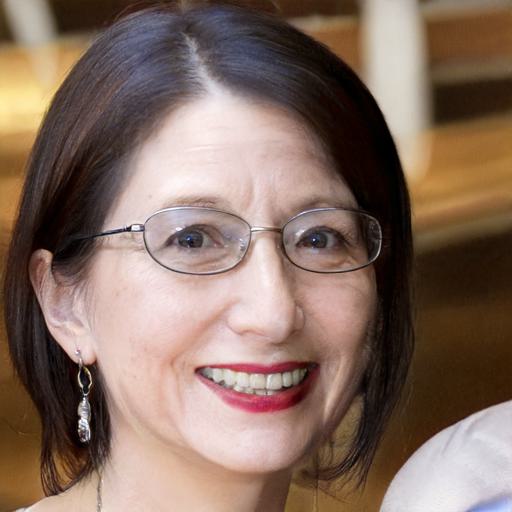} \\
         Input & +Smile & +Age & +Gender & +Glasses & +Pose
         
    \end{tabular}
\caption{Sequential human face editing result.}
\label{fig:fig4}
\end{figure}

\subsection{Quantitative Comparison}
\label{sec:quantitative}
To evaluate our model's disentanglement capabilities, we develop and utilize two different methods: attribute correlation and face preservation. Furthermore, to measure the diversity introduced by multi-directional subspace editing, we utilize the LPIPS score\cite{zhang2018unreasonable} and the Frechet Inception Distance (FID)\cite{heusel2017gans} to evaluate image fidelity.


\newcommand*{\MinNumber}{0.0}%
\newcommand*{\MidNumber}{0.18}%
\newcommand*{\MaxNumber}{1.0}%

\newcommand{\ApplyGradient}[1]{%
        \pgfmathsetmacro{\PercentColor}{max(min(100.0*(#1 - \MidNumber)/(\MaxNumber-\MidNumber),100.0),0.00)} %
        \hspace{-0.33em}\colorbox{green!\PercentColor!white}{#1}
}

\newcolumntype{R}{>{\collectcell\ApplyGradient}c<{\endcollectcell}}
\renewcommand{\arraystretch}{0}
\setlength{\fboxsep}{2mm} 
\setlength{\tabcolsep}{0pt}


\begin{table*}[b]
    \caption{Attribute correlation matrices of edited images (in absolute values).}
    \label{tab:table1}
\centering
\begin{tabular}{cc}
    (a) SeFa & (b) InterFaceGan \vspace{0.04in} \\

    \begin{tabular}{l*{5}{R}}
        &\multicolumn{1}{c}{Pose}&\multicolumn{1}{c}{Smile}&\multicolumn{1}{c}{Age}&\multicolumn{1}{c}{Gender}&\multicolumn{1}{c}{Glasses}\\
 Pose & 1.000 & 0.150 & 0.039 & 0.133 & 0.051 \\
 Smile & 0.150 & 1.000 & 0.001 & 0.170 & 0.233 \\
 Age & 0.039 & 0.001 & 1.000 & 0.533 & 0.367 \\
 Gender & 0.133 & 0.170 & 0.533 & 1.000 & 0.339 \\
 Glasses & 0.051 & 0.233 & 0.367 & 0.339 & 1.000 \\ \hdashline \\
 Avg & 0.093 & 0.115 & 0.235 & 0.293 & 0.247
    \end{tabular} & \hspace{0.1in}

    \begin{tabular}{l*{5}{R}} 
        &\multicolumn{1}{c}{pose}&\multicolumn{1}{c}{Smile}&\multicolumn{1}{c}{Age}&\multicolumn{1}{c}{Gender}&\multicolumn{1}{c}{Glasses}\\
 Pose & 1.000 & 0.098 & 0.090 & 0.017 & 0.079 \\
 Smile & 0.098 & 1.000 & 0.154 & 0.198 & 0.084 \\
 Age & 0.090 & 0.154 & 1.000 & 0.565 & 0.600 \\
 Gender & 0.017 & 0.198 & 0.565 & 1.000 & 0.363 \\
 Glasses & 0.079 & 0.084 & 0.600 & 0.363 & 1.000 \\ \hdashline \\
  Avg & 0.071 & 0.133 & 0.352 & 0.285 & 0.281
        \end{tabular} \vspace{0.1in} \\
        
    (c) StyleFlow & (d) Ours \vspace{0.04in} \\
    
    \begin{tabular}{l*{5}{R}}
        &\multicolumn{1}{c}{Pose}&\multicolumn{1}{c}{Smile}&\multicolumn{1}{c}{Age}&\multicolumn{1}{c}{Gender}&\multicolumn{1}{c}{Glasses}\\
 Pose & 1.000 & 0.113 & 0.126 & 0.103 & 0.056 \\
 Smile & 0.113 & 1.000 & 0.264 & 0.011 & 0.074 \\
 Age & 0.126 & 0.264 & 1.000 & 0.581 & 0.494 \\
 Gender & 0.103 & 0.011 & 0.581 & 1.000 & 0.338 \\
 Glasses & 0.056 & 0.074 & 0.494 & 0.338 & 1.000 \\ \hdashline \\
 Avg & 0.099 & 0.115 & 0.366 & 0.258 & 0.240
    \end{tabular} & \hspace{0.1in}
    
    \begin{tabular}{l*{5}{R}} 
        &\multicolumn{1}{c}{Pose}&\multicolumn{1}{c}{Smile}&\multicolumn{1}{c}{Age}&\multicolumn{1}{c}{Gender}&\multicolumn{1}{c}{Glasses}\\
 Pose & 1.000 & 0.018 & 0.008 & 0.029 & 0.046 \\
 Smile & 0.018 & 1.000 & 0.120 & 0.050 & 0.100 \\
 Age & 0.008 & 0.120 & 1.000 & 0.388 & 0.363 \\
 Gender & 0.029 & 0.050 & 0.388 & 1.000 & 0.203 \\
 Glasses & 0.046 & 0.100 & 0.363 & 0.203 & 1.000 \\ \hdashline \\
  Avg & 0.025 & 0.072 & 0.219 & 0.167 & 0.178
    \end{tabular}
    
\end{tabular}
\end{table*}

\textbf{Correlation between edited attributes:} We can assign an attribute score to an image $\textbf{x}$ using $\mathcal{C}_k$ as a feature extractor. The activation vector $l(\textbf{x})$, derived from the last hidden layer of $\mathcal{C}_k$, is used to compute the distance between the feature vector $l(\textbf{x})$ and the classifying hyperplane. This hyperplane is defined by
$\left \{ \textbf{v} \mid \textbf{c}_k^T\textbf{v} + \textbf{t}_k = 0\right \}$.

The attribute score of an image $\textbf{x}$ is given by the distance of $l(\textbf{x})$ from the separating hyperplane: $dist_k(\textbf{x}) =  \frac{\textbf{c}_k^Tl(\textbf{x})+\textbf{t}_k}{|\textbf{c}_k|}$.

To assess the disentanglement capability of our model, we began by generating an evaluation set comprising $1K$ pairs of images paired with their desired face attributes. These images were directly sampled from the StyleGAN generator, while the attributes were independently generated for each image. For each image, we manipulated its latent vector $\mathbf{w}\in \mathcal{W^+}$ along the learned directions and fed it to the generator to produce an edited version, denoted as $\textbf{x}_{edit}$. It's worth noting that in practice we change all attributes simultaneously.

Next, we measured the difference in attribute score between the original and the edited images. Taking the attribute ``smile" as an example, we defined the perceptual distance as: $\Delta_{smile} = dist_{smile}(\textbf{x}_{edit}) - dist_{smile}(\textbf{x})$.

For the final assessment, we computed the Pearson correlation between the perceptual distances of every pair of attributes. Each entry in \cref{tab:table1} displays the absolute correlation value. For instance, for the attributes ``smile" and ``pose", we determine the perceptual distances $\Delta_{smile}$ and $\Delta_{pose}$ across the entire evaluation set. The correlation value shown in the table is:
\begin{equation}
    corr(pose, smile) = \left | \frac{cov(\Delta_{pose}, \Delta_{smile})}{E[\Delta_{pose}]E[\Delta_{smile}]} \right |
\end{equation}

\Cref{tab:table1} presents the correlation results of various methods. As previously mentioned\cite{shen2020interfacegan,doubinsky2021multi, harkonen2020ganspaceli2021dystyle, shen2021closed}, we observe that certain attributes exhibit stronger correlations than others. For instance, the glasses-age correlation is consistently higher than the smile-pose correlation across all models. The table's final row sums the off-diagonal columns, representing the average correlation score. Our method's lower scores highlight its superior disentanglement capabilities compared to the other techniques.

In our previous assessment, all attributes were adjusted simultaneously. To test the stability of attributes when only one is modified, we conducted an experiment editing one attribute at a time and measured the perceptual distances using the classifiers' final layers.
\cref{fig:fig5} displays results from various methods. The x-axis represents changes in the edited attribute, while the y-axis indicates unintentional changes in other attributes. As edits become more extensive, the inter-attribute effect is more pronounced. Among all models, our approach shows the least deviation, indicating superior attribute disentanglement.

\newcommand*{\fffsize}{0.19}

\begin{figure*}[b]
\centering
\begin{tabular}{ccccc}
    
    \includegraphics[width=\fffsize\linewidth]{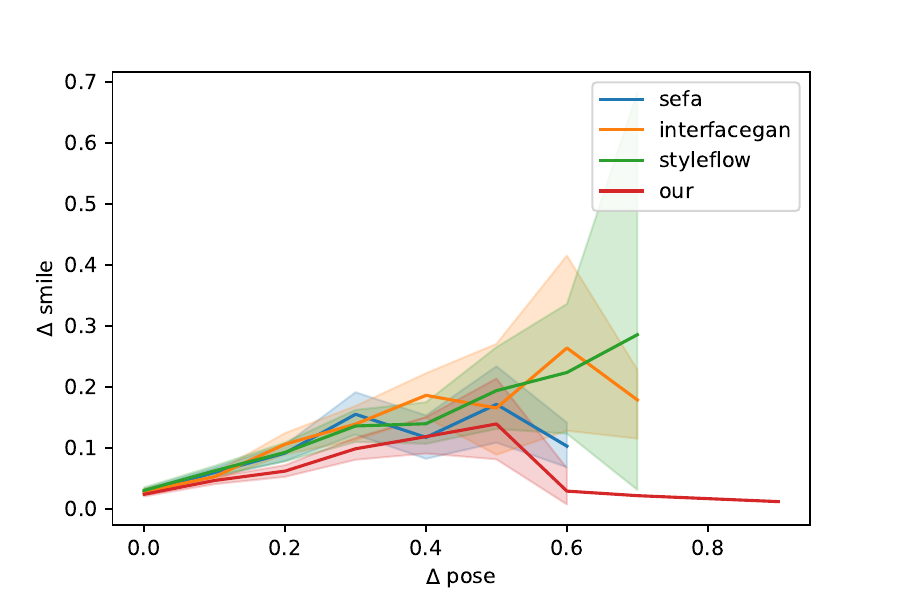} &
    \includegraphics[width=\fffsize\linewidth]{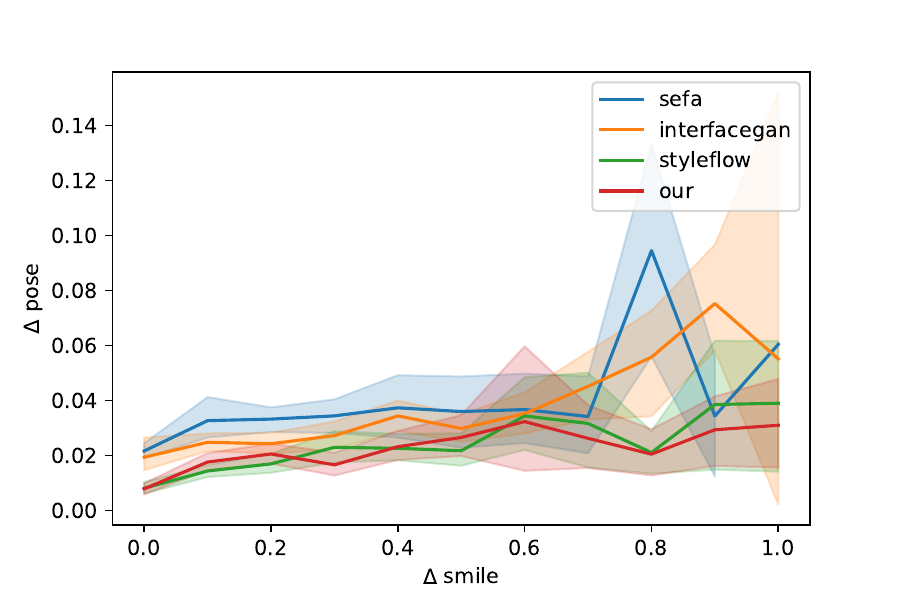} &
    \includegraphics[width=\fffsize\linewidth]{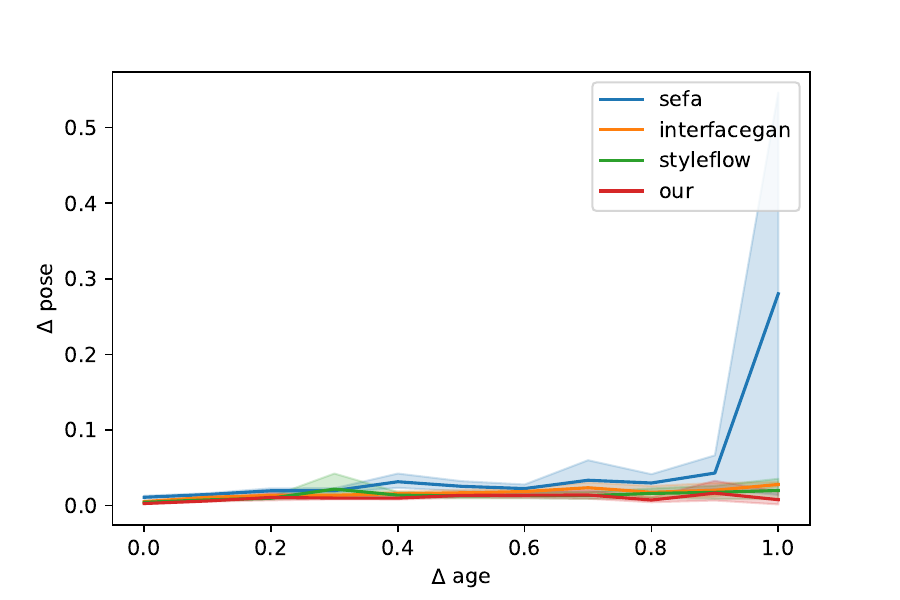} &
    \includegraphics[width=\fffsize\linewidth]{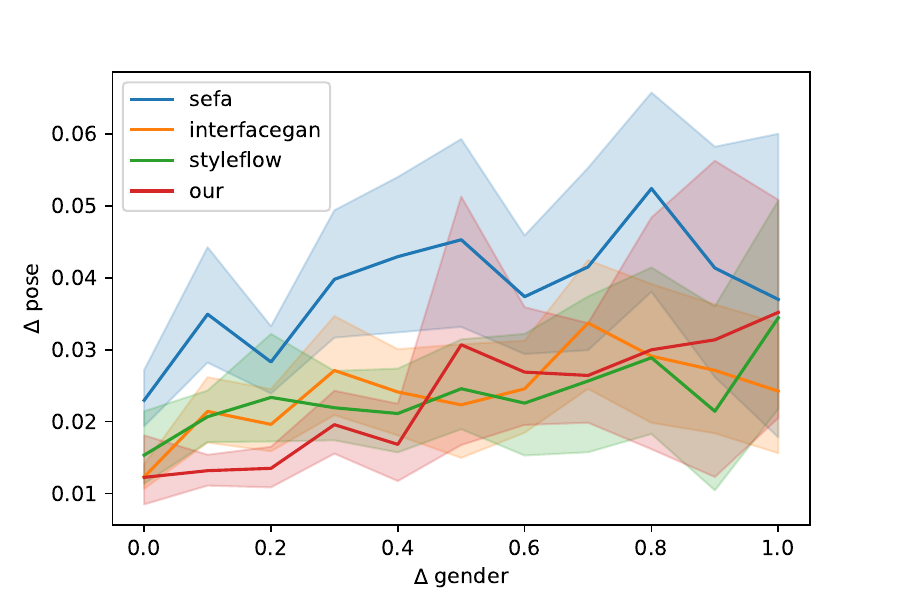} &
    \includegraphics[width=\fffsize\linewidth]{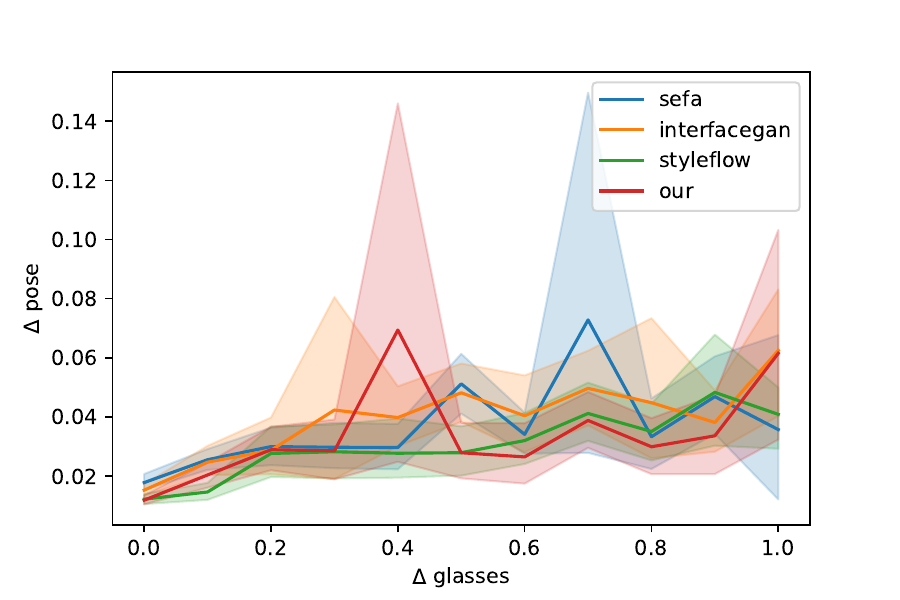} \\
    
    \includegraphics[width=\fffsize\linewidth]{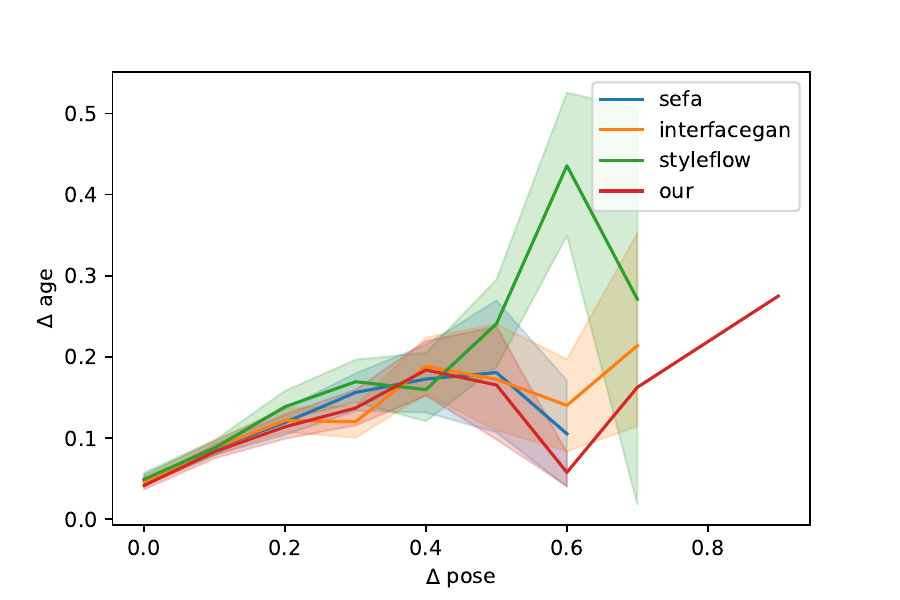} &
    \includegraphics[width=\fffsize\linewidth]{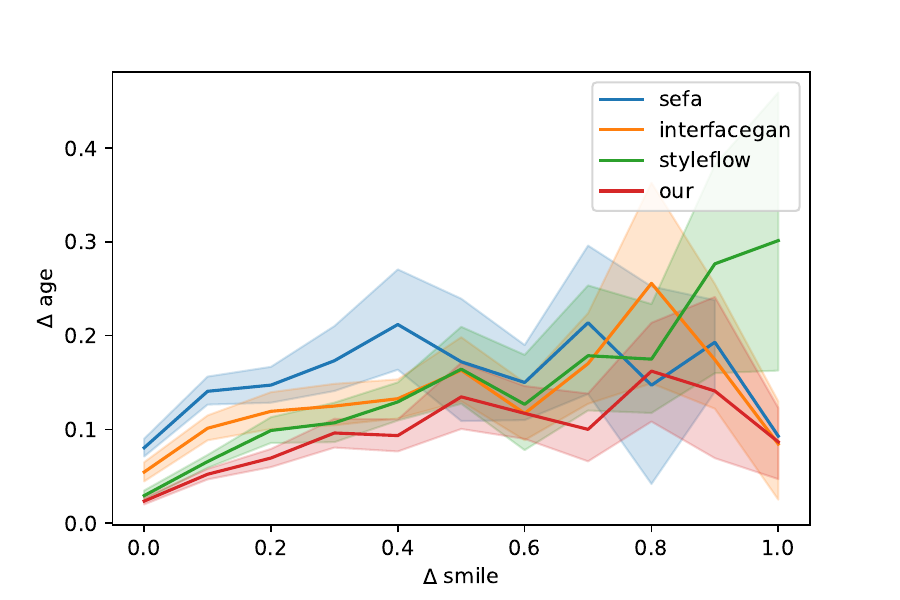} &
    \includegraphics[width=\fffsize\linewidth]{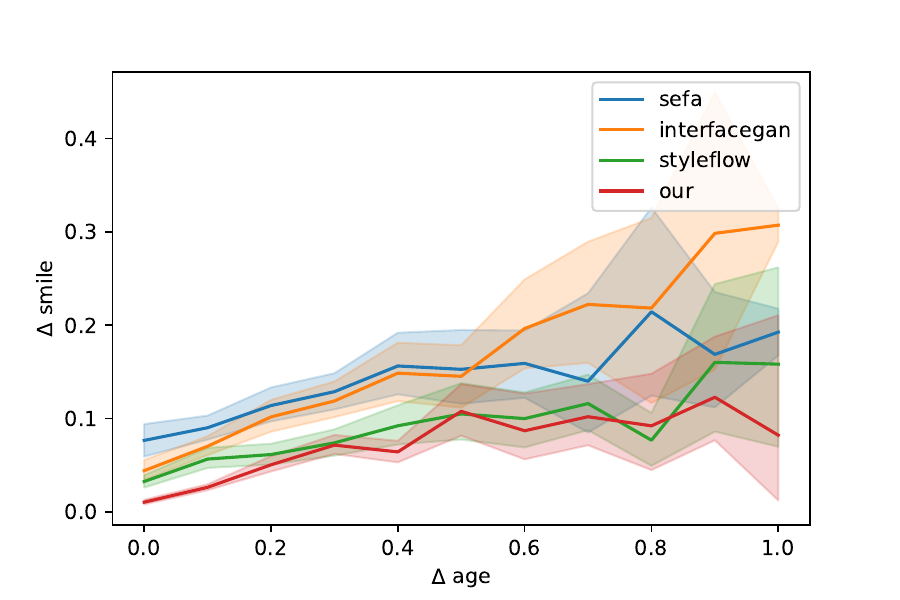} &
    \includegraphics[width=\fffsize\linewidth]{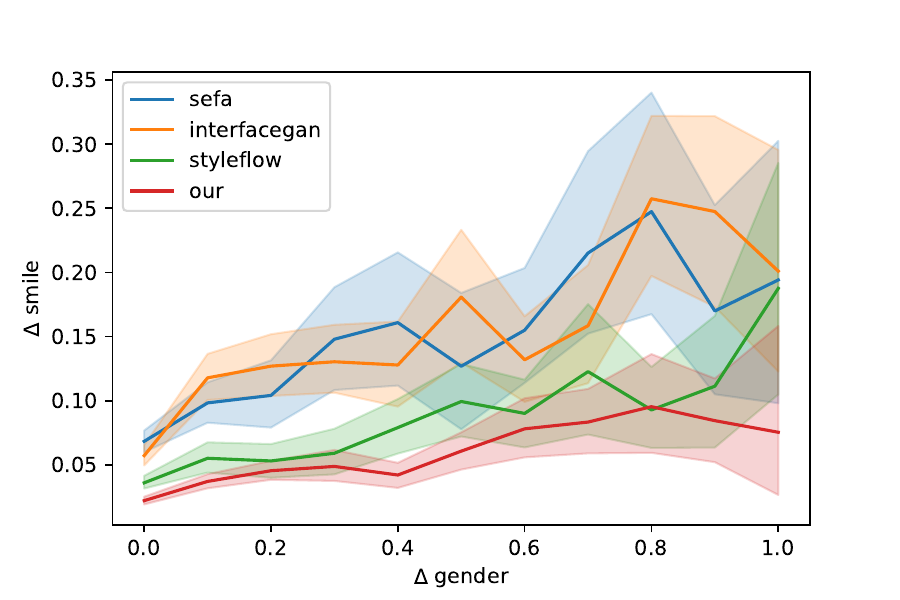} &
    \includegraphics[width=\fffsize\linewidth]{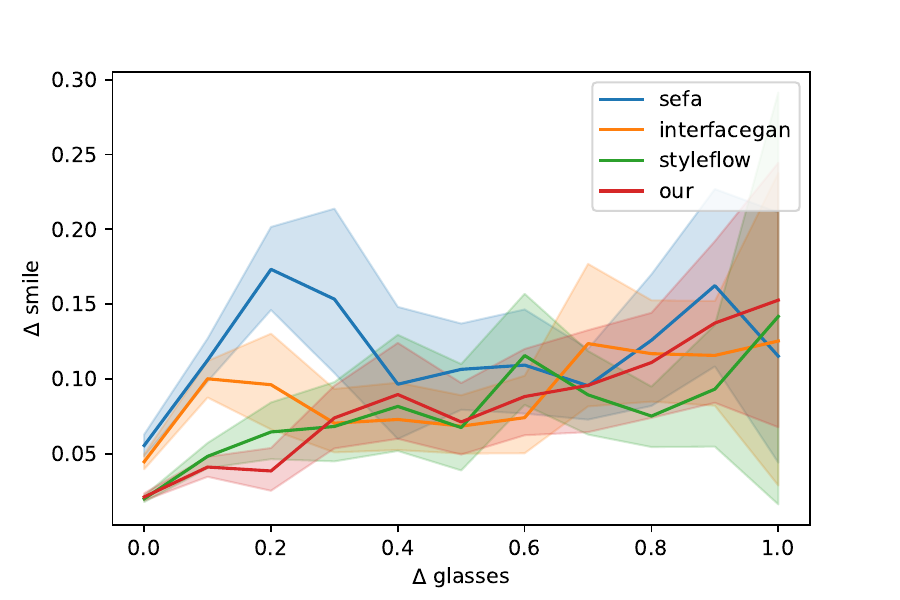} \\
    
    \includegraphics[width=\fffsize\linewidth]{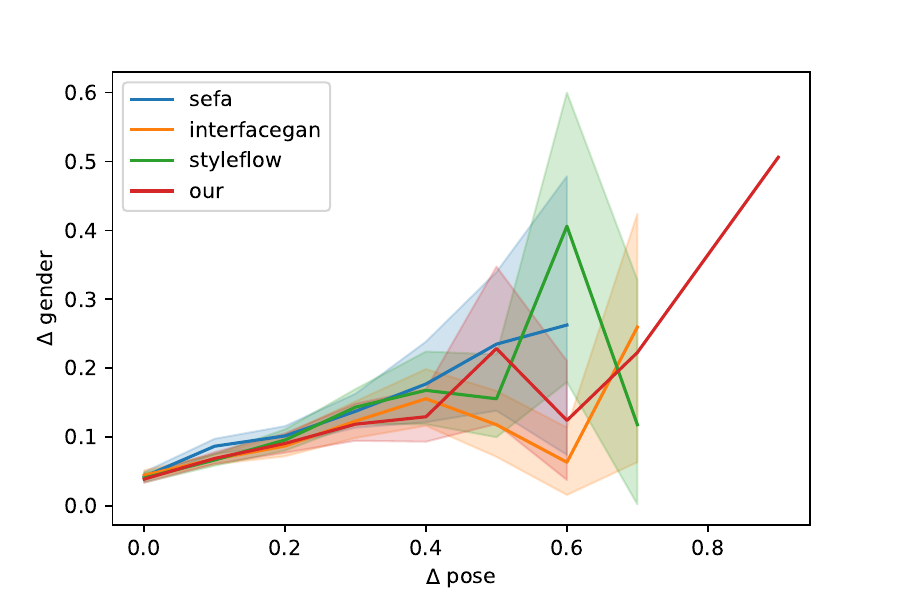} &
    \includegraphics[width=\fffsize\linewidth]{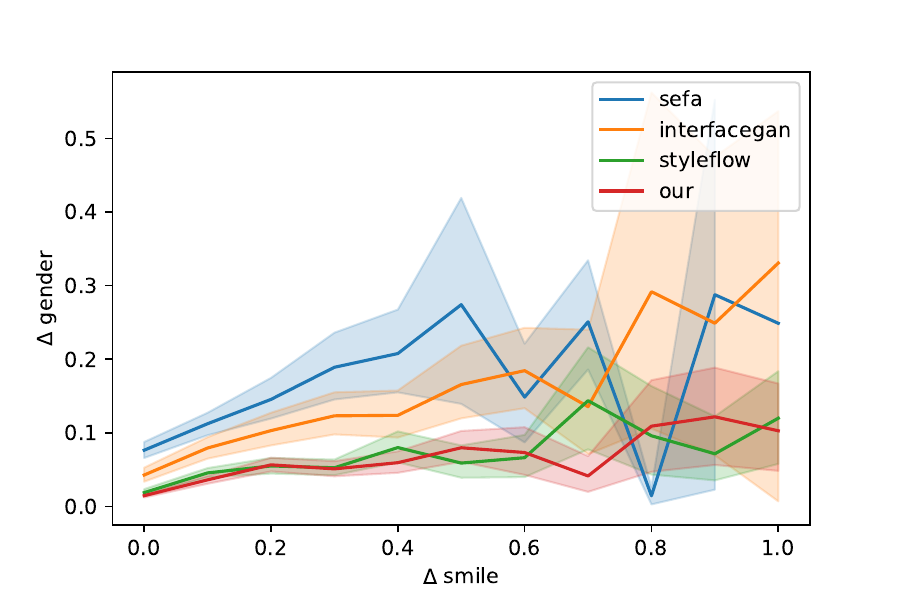} &
    \includegraphics[width=\fffsize\linewidth]{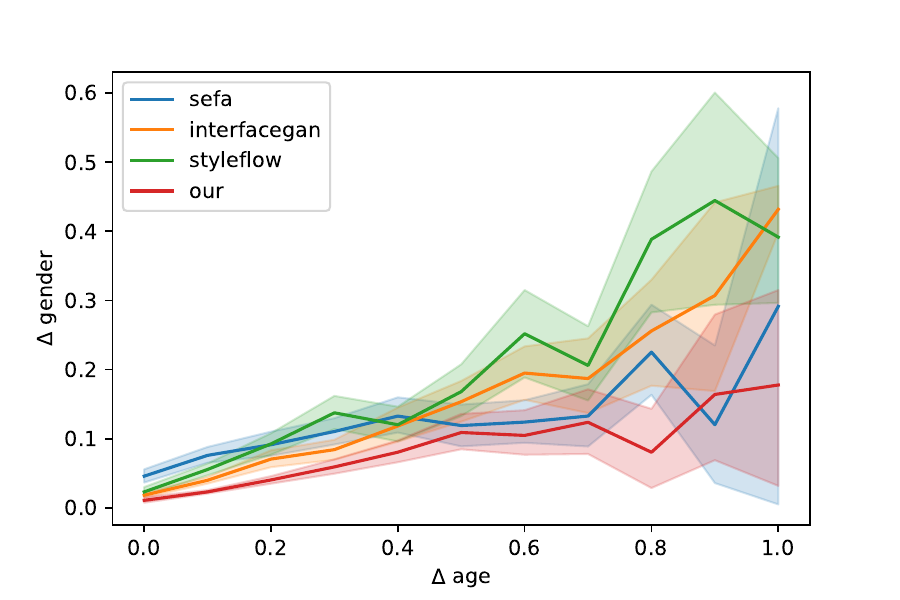} &
    \includegraphics[width=\fffsize\linewidth]{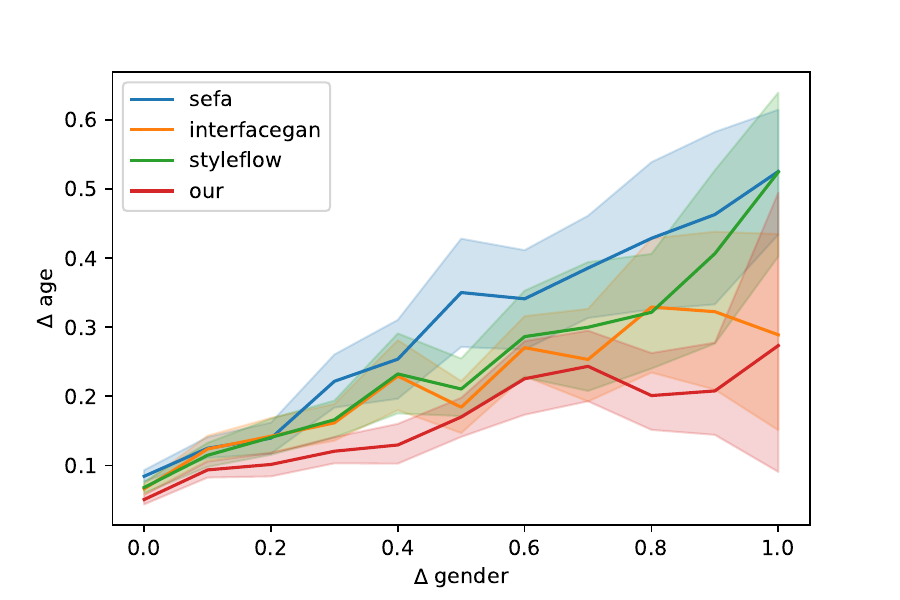} &
    \includegraphics[width=\fffsize\linewidth]{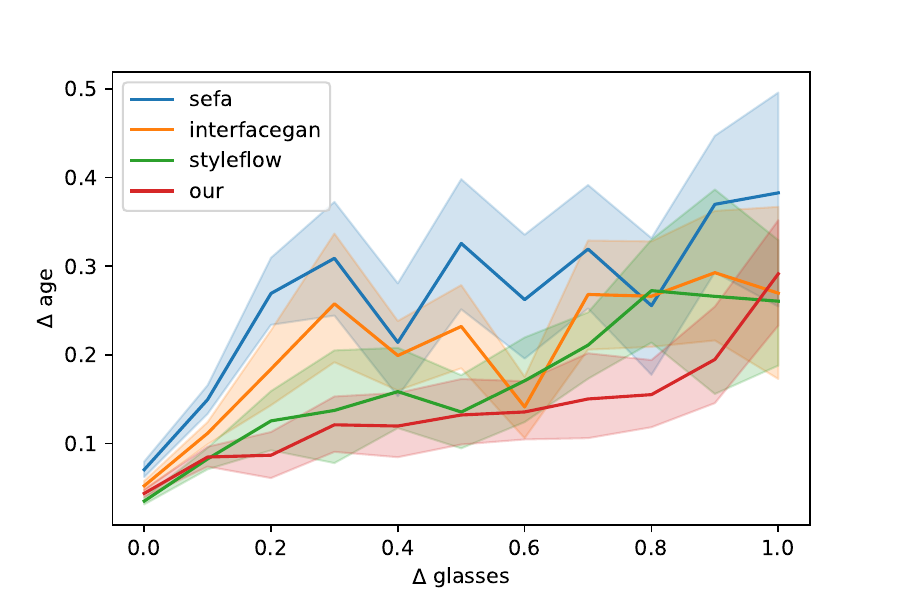} \\
    
    \includegraphics[width=\fffsize\linewidth]{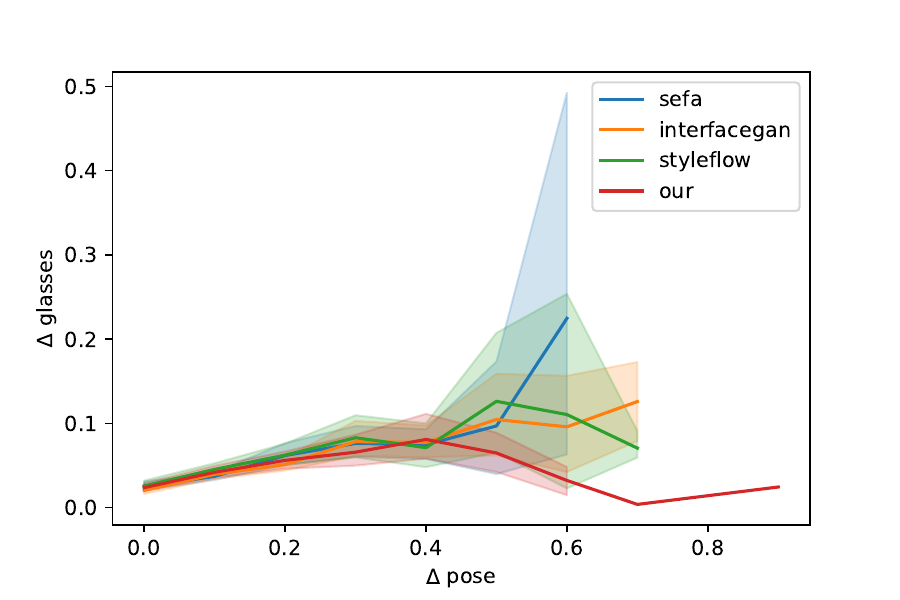} &
    \includegraphics[width=\fffsize\linewidth]{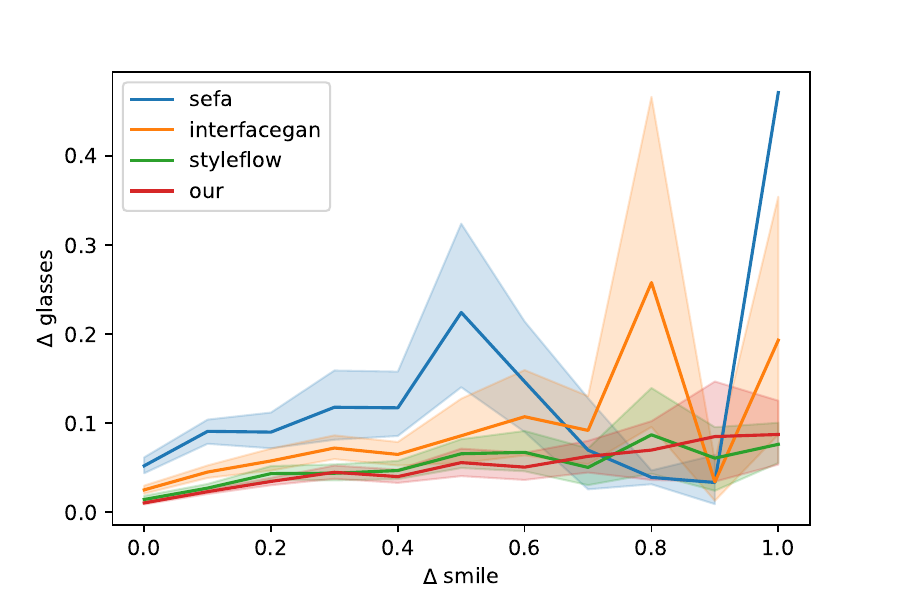} &
    \includegraphics[width=\fffsize\linewidth]{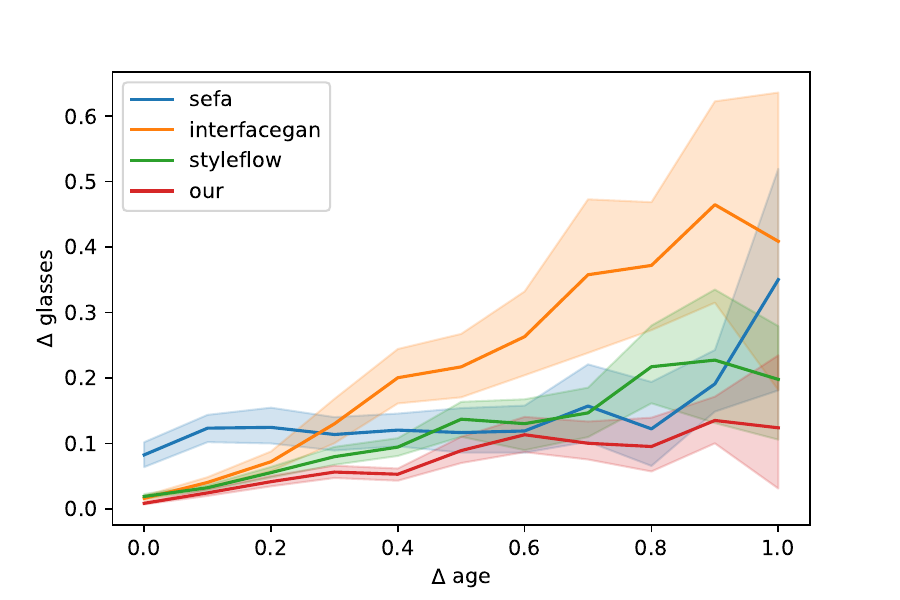} &
    \includegraphics[width=\fffsize\linewidth]{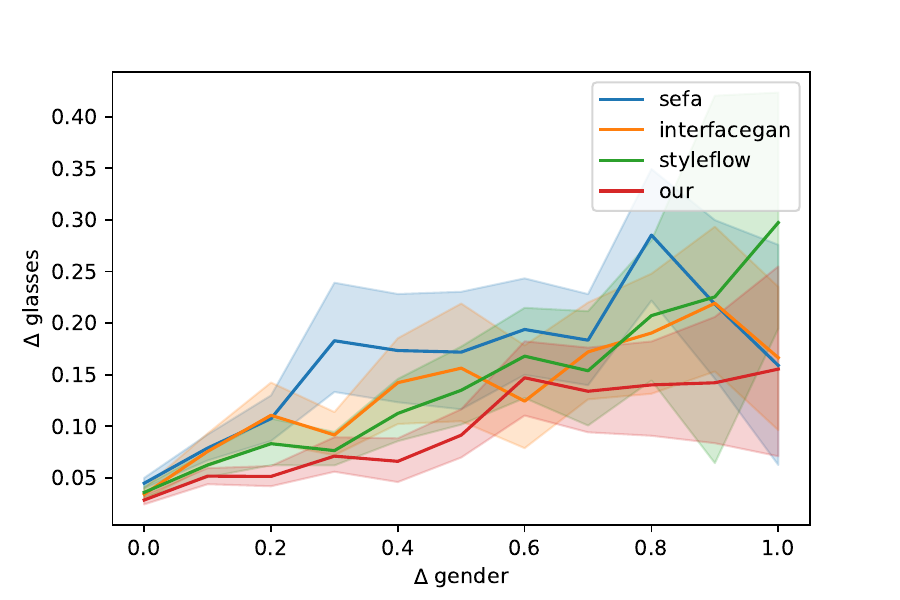} &
    \includegraphics[width=\fffsize\linewidth]{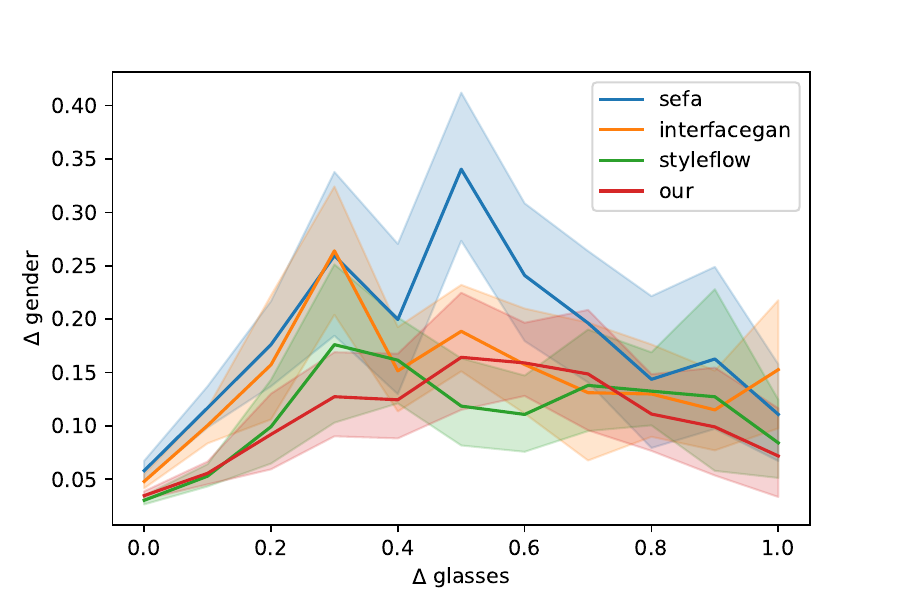} \\
    \hdashline \\
    \includegraphics[width=\fffsize\linewidth]{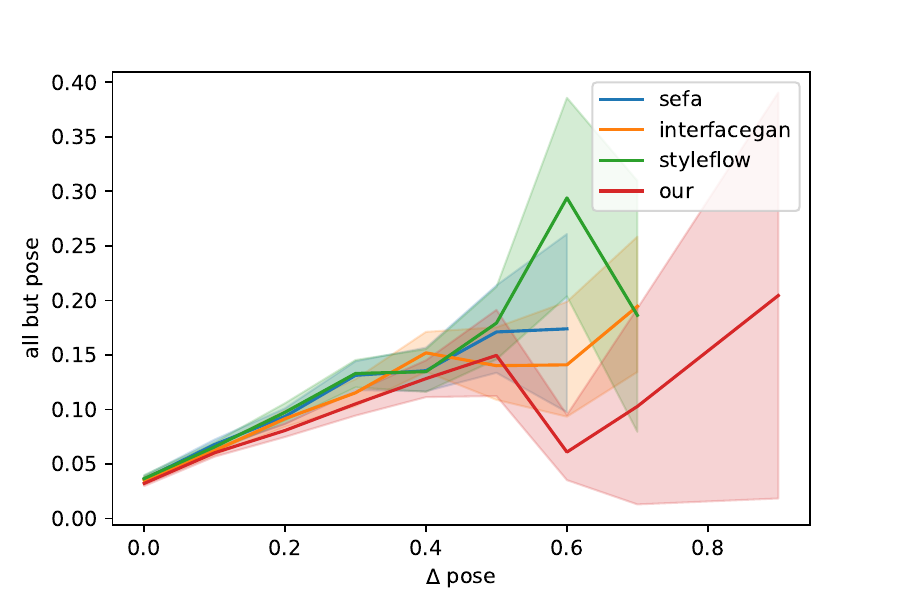} &
    \includegraphics[width=\fffsize\linewidth]{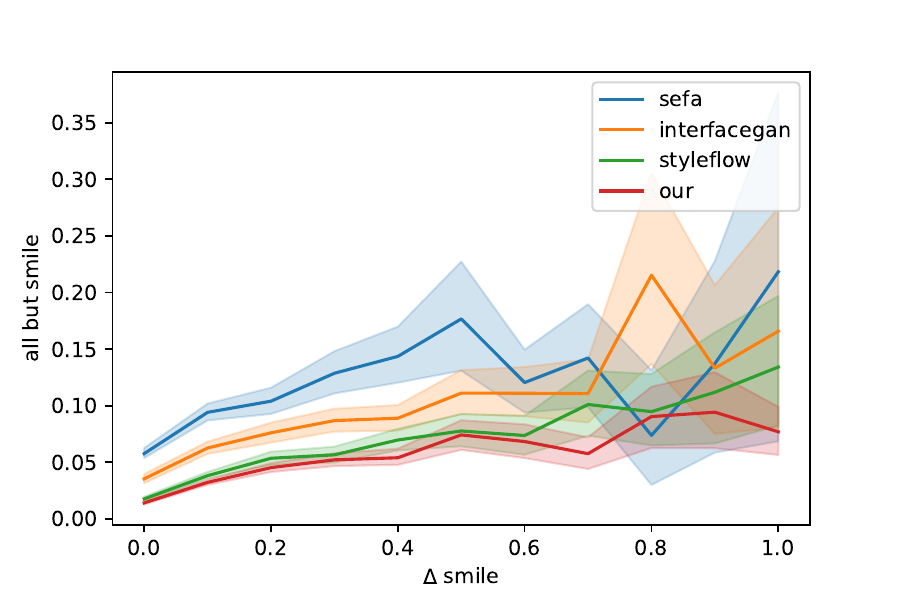} &
    \includegraphics[width=\fffsize\linewidth]{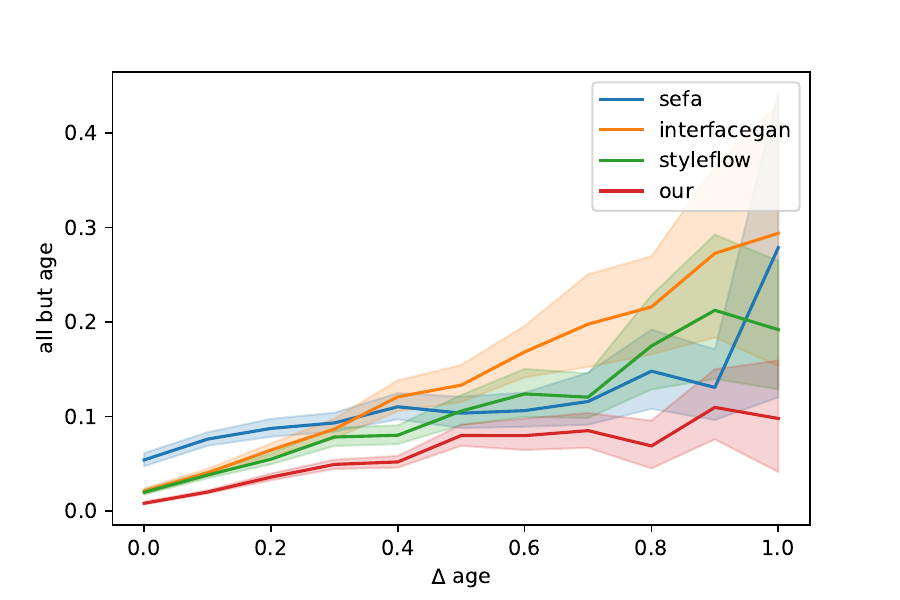} &
    \includegraphics[width=\fffsize\linewidth]{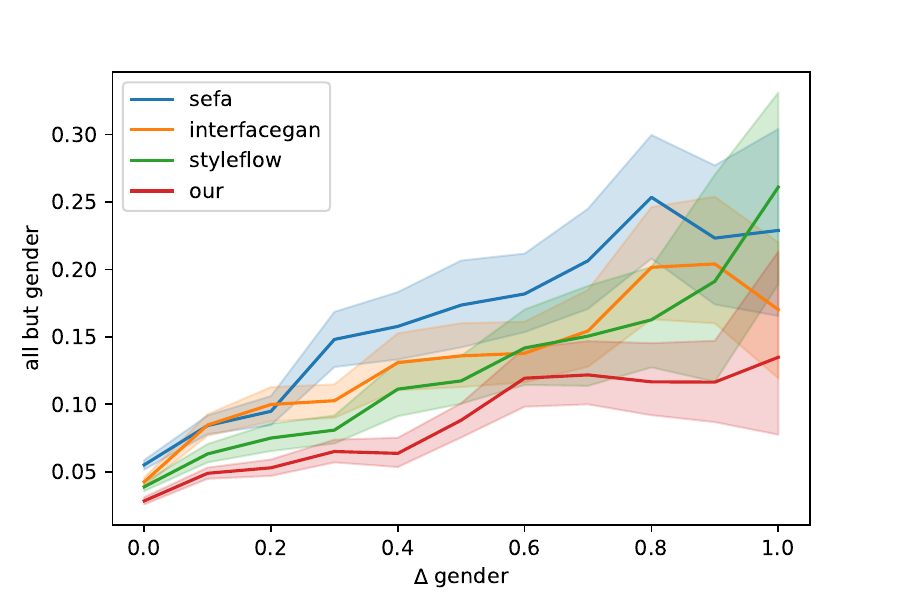} &
    \includegraphics[width=\fffsize\linewidth]{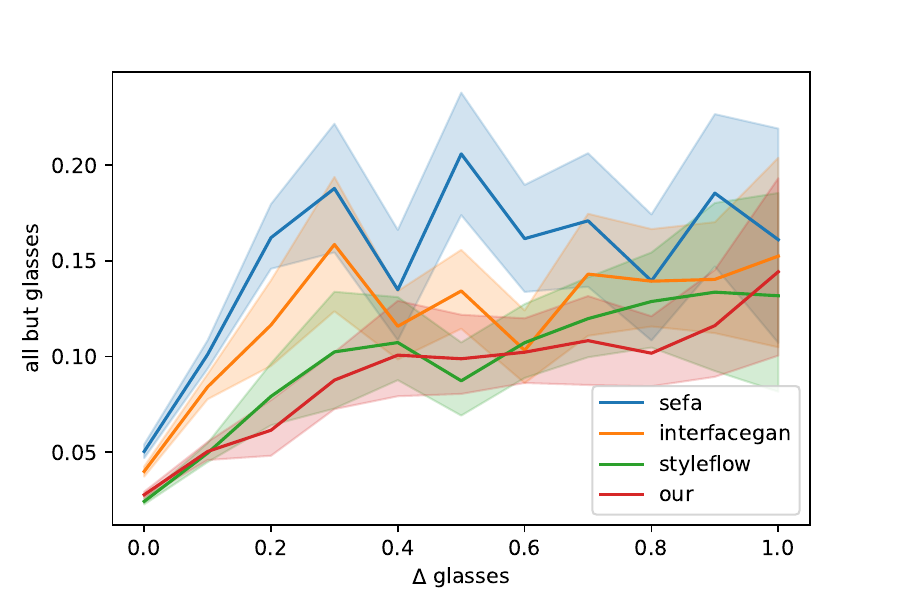}

\end{tabular}
\caption{Comparative analysis of the inter-attribute effects exhibited by different models. The upper rows display how changes in one attribute vary as a function of changes in another attribute. The bottom row provides an average of all the rows above.}
\label{fig:fig5}
\end{figure*}
 

\textbf{Face Identity Preservation:} Following the approach of the face identity score\cite{abdal2021styleflow}, we assess the edited images using a pre-trained face embedding network\cite{Geitgey2021}. We restricted our edits to attributes that should preserve identity, such as pose, glasses, and smile. After editing, we embedded the images into a latent space using the network. We then calculated the Euclidean distance, $E_d$, and the cosine similarity, $C_s$, between the original and edited images.

\Cref{tab:table2} displays the results for face identity preservation. Our method consistently surpasses other techniques across all types of edits. While StyleFlow shows comparable scores for individual attribute edits, our model excels in preserving identity when multiple edits are combined.


\setlength{\tabcolsep}{2pt}
\renewcommand{\arraystretch}{1}
\begin{table}
    \centering
    \caption{Identity preservation scores by different models.}
    \label{tab:table2}
    \begin{tabular}{lccccc}
        \hline
        Edit & Metric & SeFa & InterFaceGan & StyleFlow & Ours \\ \hline
        \multirow{2}{*}{Smile} & $C_s \Uparrow$ & 0.970 & 0.978 & 0.989 & \textbf{0.991}\\
                              & $E_d \Downarrow$ & 0.329 & 0.276 & 0.192 & \textbf{0.174}\\ \hline
        \multirow{2}{*}{Pose} & $C_s \Uparrow$ & 0.978 & 0.979 & 0.981 & \textbf{0.982}\\
                             & $E_d \Downarrow$ & 0.283 & 0.279 & 0.2662 & \textbf{0.253}\\ \hline
        \multirow{2}{*}{Glasses} & $C_s \Uparrow$ & 0.966 & 0.978 & 0.984 & \textbf{0.985}\\
                                & $E_d \Downarrow$ & 0.348 & 0.270 & 0.227 & \textbf{0.215}\\ \hline
        \multirow{2}{*}{All} & $C_s \Uparrow$ & 0.911 & 0.933 & 0.935 & \textbf{0.948}\\
                                & $E_d \Downarrow$ & 0.622 & 0.537 & 0.534 & \textbf{0.467}\\ \hline
    \end{tabular}
    \caption*{\textit{Notation: $C_s$ - Cosine Similarity; $E_d$ - Euclidean Distance.}}
\end{table}


\subsection{Diversity Analysis}
In the previous sections, we introduced a single-direction variant of our model. Now, we will explore the performance of our model when multiple directions in the subspace are utilized for editing. To measure the diversity of the images produced by our model, we employed the LPIPS perceptual similarity score\cite{zhang2018unreasonable}. A lower LPIPS score indicates higher similarity between images, whereas a higher score signifies perceptual differences.

We generated a batch of 1,000 images and applied 5 random edits to each image. For every set of 5 edits, the similarity score between each pair of edited images was computed. Subsequently, all these scores were averaged to derive the final diversity score. We benchmarked our model against StyleFlow and the single-direction variant of our model, referred to as SVM, as detailed in \cref{sec:comp}. The FID score was also calculated to assess the visual quality of the generated images.

The results, summarized in \cref{tab:table3}, reveal that our approach — when multi-directional subspace editing is employed — not only fosters greater image diversity (as indicated by the LPIPS score) but also ensures superior visual quality (as denoted by the FID score). It's noteworthy that there is a marked enhancement in the multi-directional iteration of our model (subspace) when compared against its single-direction counterpart (SVM) or against StyleFlow.

\begin{table}[h]
\centering
\caption{Diversity score results for edited images. Our multi-directional model achieves a higher score which indicates its ability to generate more diverse results while maintaining a lower FID score.}
\label{tab:table3}
\begin{tabular}{lcc} 
 \hline
 Method & LPIPS $\Uparrow$ & FID $\Downarrow$ \\
  \hline 
  StyleFlow & 0.213 & 27.467 \\
  Our (svm) & 0.229 & 26.926 \\
  Our (subspace) & \textbf{0.321} & \textbf{23.648} \\
  \hline
\end{tabular}
\end{table}

\subsection{Ablation Study}
To validate the importance of our orthogonality loss outlined in \cref{sec:train}, we carried out an ablation study. We trained a version of our model with $\lambda_{orth} = 0$, keeping all other configurations, including losses and the training process, consistent. Post-training, we conducted consecutive edits starting from an original image, and the results can be viewed in \cref{fig:fig6}.

We found that images edited with the orthogonality loss exhibit crisper details and fewer visual artifacts when subjected to multiple edits. We also carried out the same quantitative evaluations as detailed in \cref{sec:quantitative}. The results in \cref{tab:table4,tab:table5} indicate that our orthogonality loss leads to better attribute disentanglement in our model.

\newcommand*{\ffsize}{0.2}
\begin{figure}[h]
\centering
\setlength{\tabcolsep}{1pt}
\begin{tabular}{c c}
& $\lambda_{orth}=0$ \vspace{0.05in} \\
\rotatebox[]{90}{Pose} & 
\begin{tabular}{cccc}
     \includegraphics[width=\ffsize\linewidth]{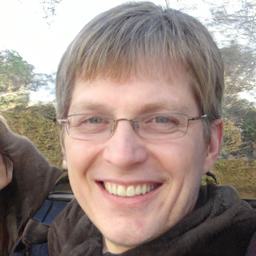} \hspace{0.05in} &  \includegraphics[width=\ffsize\linewidth]{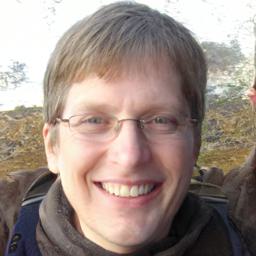} \hspace{0.05in} & 
     \includegraphics[width=\ffsize\linewidth]{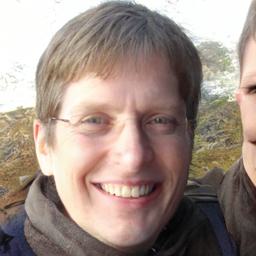} \hspace{0.05in} & 
     \includegraphics[width=\ffsize\linewidth]{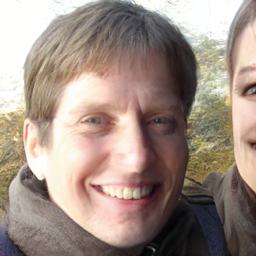}
\end{tabular} \\

\rotatebox[]{90}{Smile} & 
\begin{tabular}{cccc}
     \includegraphics[width=\ffsize\linewidth]{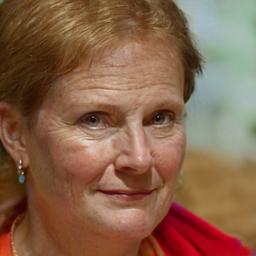} \hspace{0.05in} &  \includegraphics[width=\ffsize\linewidth]{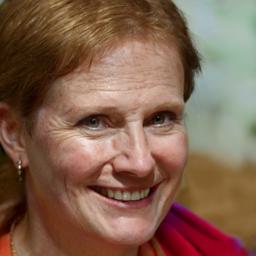} \hspace{0.05in} & 
     \includegraphics[width=\ffsize\linewidth]{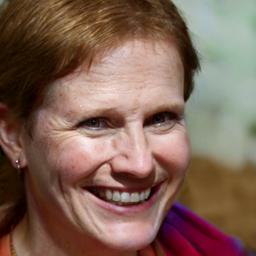} \hspace{0.05in} & 
     \includegraphics[width=\ffsize\linewidth]{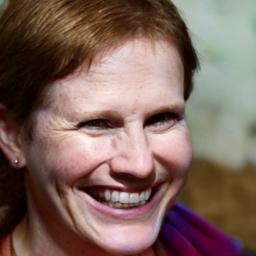}
\end{tabular} \\ 

& $\lambda_{orth}=0.001$ \\
\rotatebox[]{90}{Pose}  & 
\begin{tabular}{cccc}
     \includegraphics[width=\ffsize\linewidth]{images/orth_comp/img_66_orig.jpg} \hspace{0.05in} &  \includegraphics[width=\ffsize\linewidth]{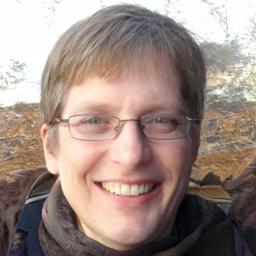} \hspace{0.05in} & 
     \includegraphics[width=\ffsize\linewidth]{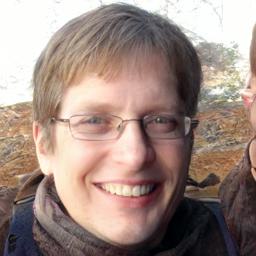} \hspace{0.05in} & 
     \includegraphics[width=\ffsize\linewidth]{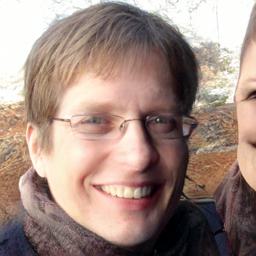}
\end{tabular}  \\

\rotatebox[]{90}{Smile} & 
\begin{tabular}{cccc}
     \includegraphics[width=\ffsize\linewidth]{images/orth_comp/img_73_orig.jpg} \hspace{0.05in} &  \includegraphics[width=\ffsize\linewidth]{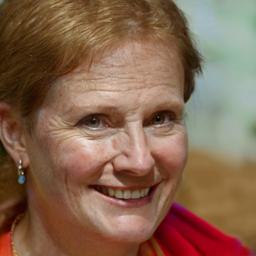} \hspace{0.05in} & 
     \includegraphics[width=\ffsize\linewidth]{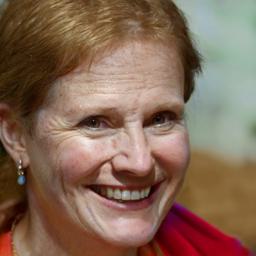} \hspace{0.05in} & 
     \includegraphics[width=\ffsize\linewidth]{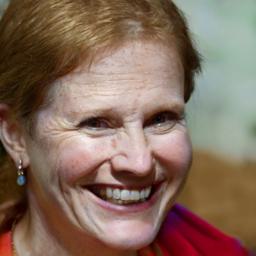}
\end{tabular}

\end{tabular}
\caption{Comparison using orthogonality loss: First and third rows show reduced pose-glasses entanglement; second and fourth rows show better preservation of hair and shirt color.}
\label{fig:fig6}
\end{figure}


\begin{table}[h]
\caption{Attribute correlation matrices with and without orthogonal loss. The values represent the average correlation between an attribute and others.}
\label{tab:table4}
\begin{tabular}{l|c|c|c|c|c} 
 \hline
 Model & Pose & Smile & Age & Gender & Glasses \\
  \hline 
  $\lambda_{orth}=0$ & 0.149 & 0.242 & 0.342 & 0.283 & 0.253 \\
  \hline 
  $\lambda_{orth}=0.001$ & \textbf{0.025} & \textbf{0.072} & \textbf{0.219} & \textbf{0.167} & \textbf{0.178} \\ \hline
\end{tabular}
\end{table}


\setlength{\tabcolsep}{2pt}
\renewcommand{\arraystretch}{1}
\begin{table}[h]
    \centering
    \caption{Comparison of identity preservation using orthogonal loss.}
    \label{tab:table5}
    \begin{tabular}{lccc}
        \hline
        Edit & Metric & $\lambda_{orth}=0$ & $\lambda_{orth}=0.001$  \\ \hline
        \multirow{2}{*}{All} & $C_s \Uparrow$ & 0.932 & \textbf{0.948}\\
                                & $E_d \Downarrow$ & 0.536 & \textbf{0.467}\\ \hline
    \end{tabular}
    \caption*{\textit{Notation: $C_s$ - Cosine Similarity; $E_d$ - Euclidean Distance.}}
\end{table}


\section{Conclusions}
In this work we proposed MDSE, a disentangling generative model for multi-attribute image editing. We introduce the concept of orthogonal subspaces that support multi-directional edits for diverse image generation. Additionally, this model effectively identifies disentangled latent subspaces, allowing precise control over the generated images and the displayed face attributes. We believe that integrating these concepts during the generator's training could further enhance the performance of disentangled models.

{\small
\bibliographystyle{ieee_fullname}
\bibliography{references}
}

\onecolumn
\appendix
\def\cvprPaperID{10190} 
\def\confName{ICCV}
\def\confYear{2023}


\title{Multi-Directional Subspace Editing in Style-Space \\(Supplementary Material)}

\author{Chen Naveh \hspace{5em} Yacov Hel-Or\\
School of Computer Science, Reichman University\\
{\tt\small navehchen1@gmail.com \hspace{5em} toky@runi.ac.il}
}
\maketitle



In this supplementary document we provide some more results of our proposed method, higher resolution images and comparisons against other methods.

\section{Qualitative Comparisons}
\cref{fig:fig1} shows real image editing results of our method with comparison to baseline models. When editing the pose, we can inspect evidence of glasses in Sefa\cite{shen2021closed} and StyleFlow\cite{abdal2021styleflow}. This suggests that those attributes are somewhat entangled. We can also inspect inferior image quality when all attributes are changed at once. Since our latent directions are orthogonal, by definition, changes within a subspace do not affect other subspaces. When performing multiple edits, this results in sharper images with less artifacts.

\newcommand*{\simsize}{0.14}
\begin{figure}[h]
\centering
\setlength{\tabcolsep}{1pt}
\hspace*{-2.0cm}
\begin{tabular}{cccccccc}
 & Reconstruction & Smile & Gender & Pose & Age & Glasses & All \\
 
 \hspace{0.7in}\rotatebox{90}{\hspace{0.26in}Sefa} &
 \includegraphics[width=\simsize\linewidth]{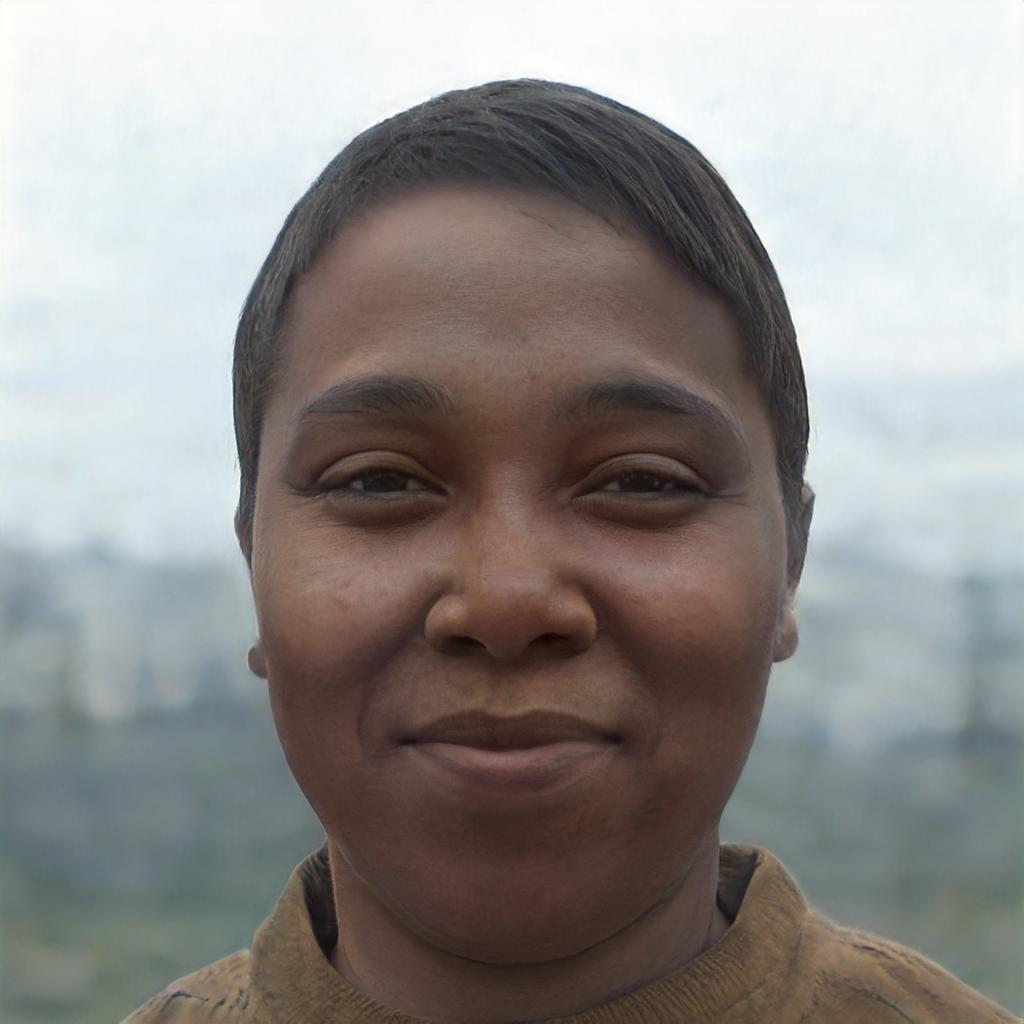} &
 \includegraphics[width=\simsize\linewidth]{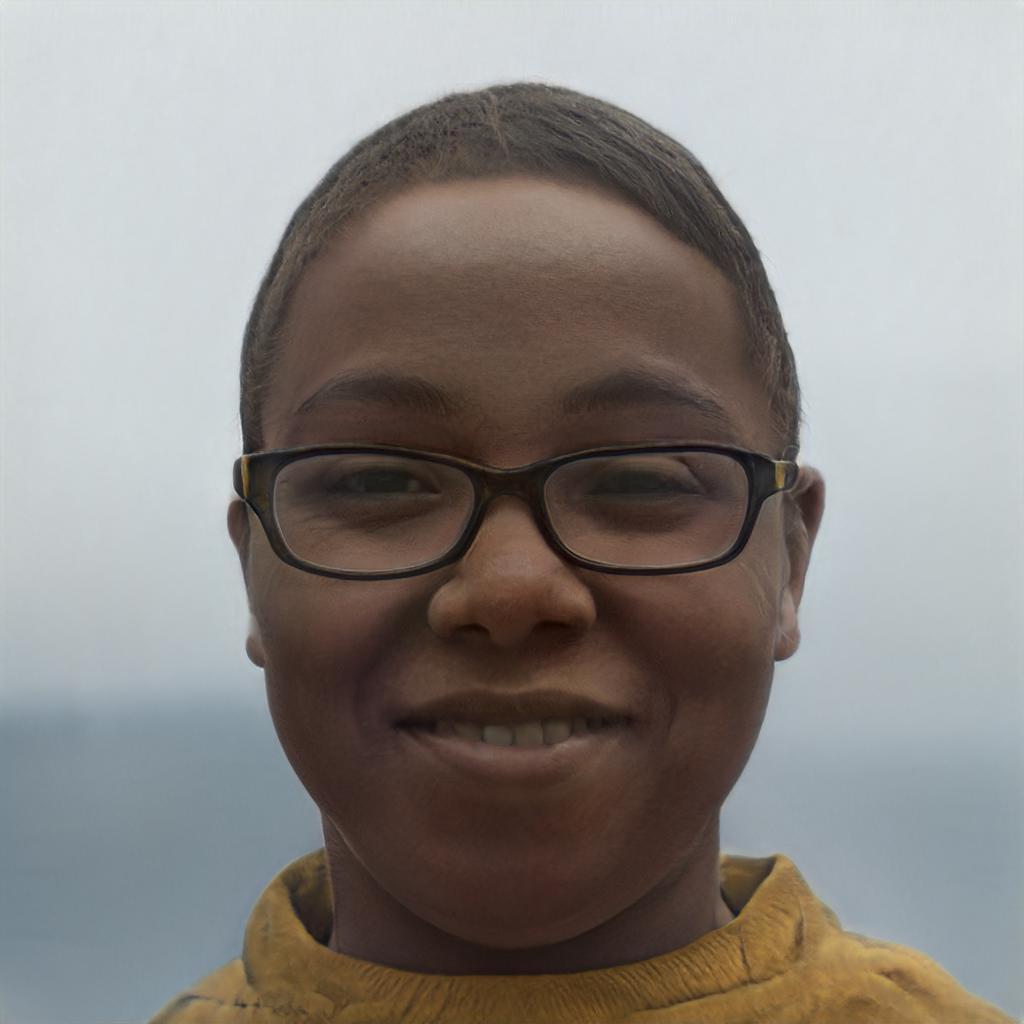} &
 \includegraphics[width=\simsize\linewidth]{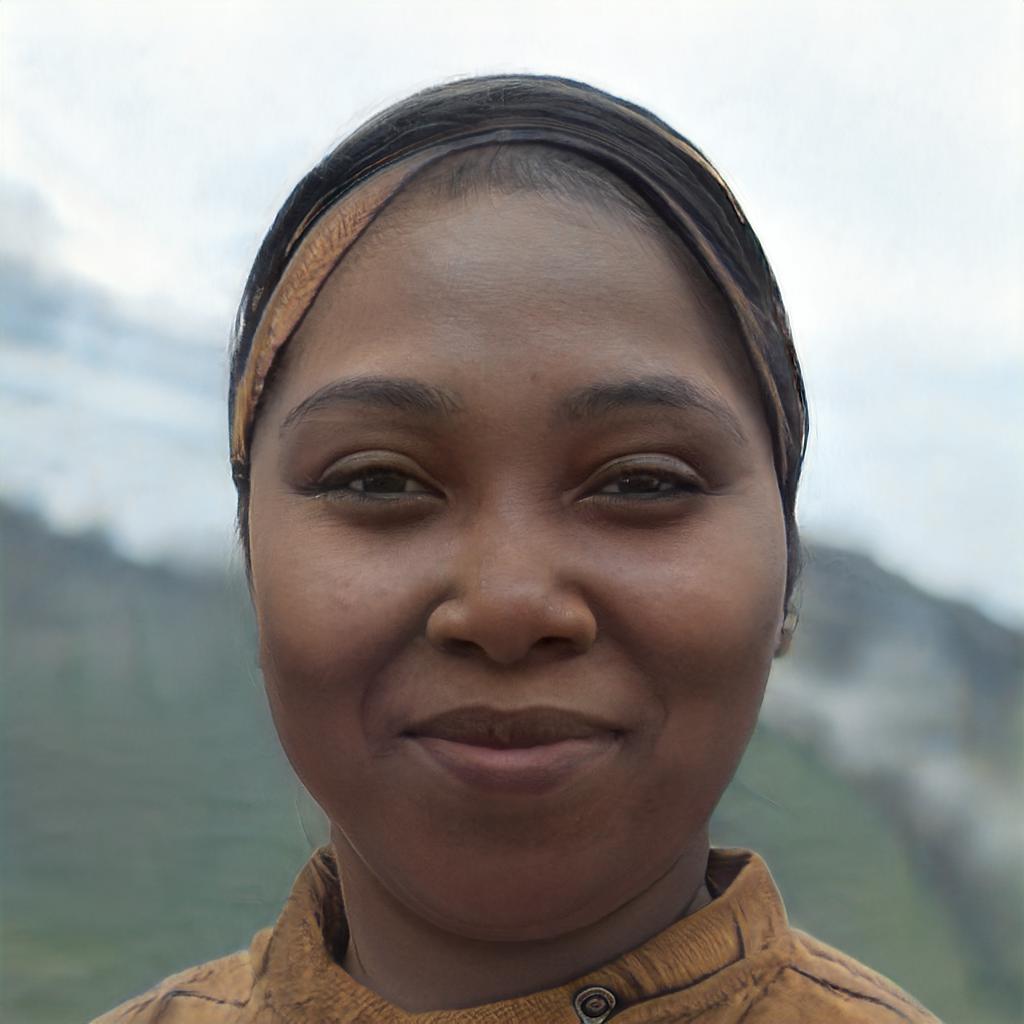} &
 \includegraphics[width=\simsize\linewidth]{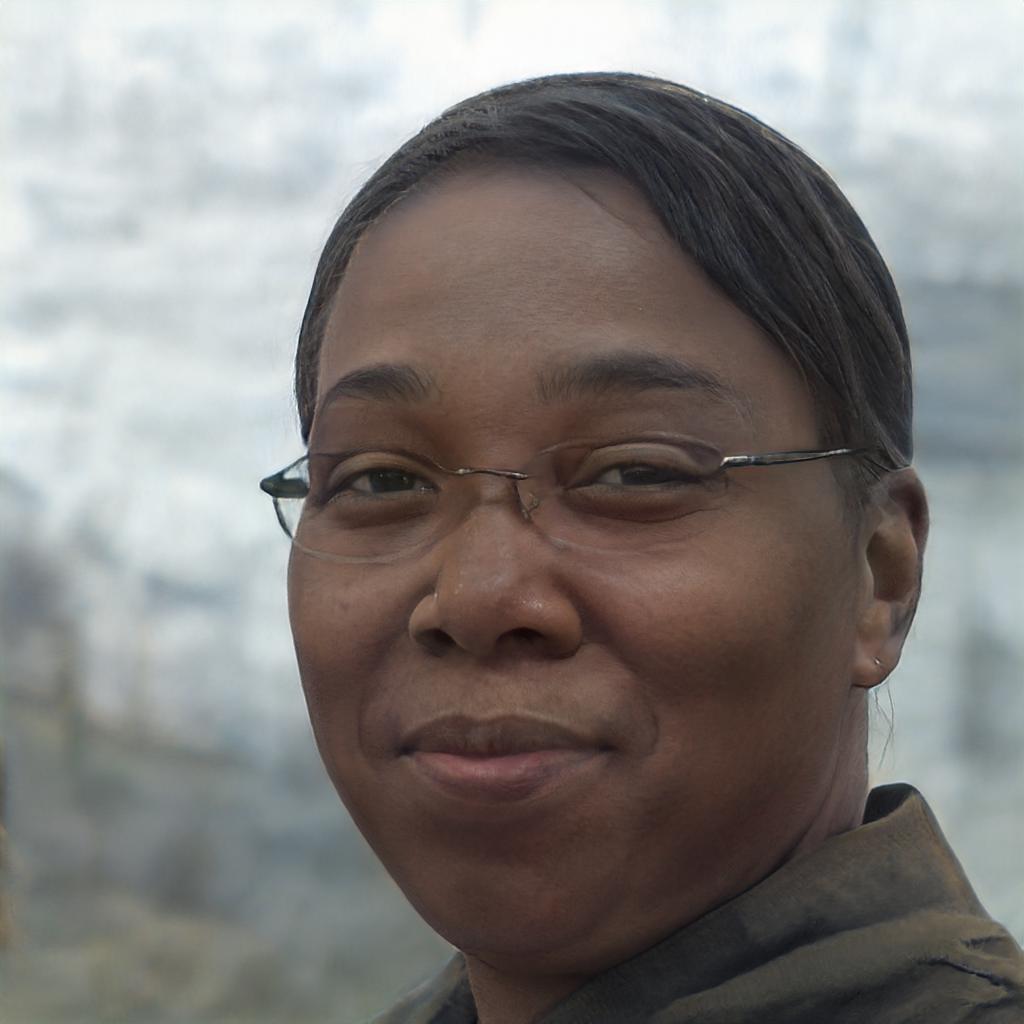} &
 \includegraphics[width=\simsize\linewidth]{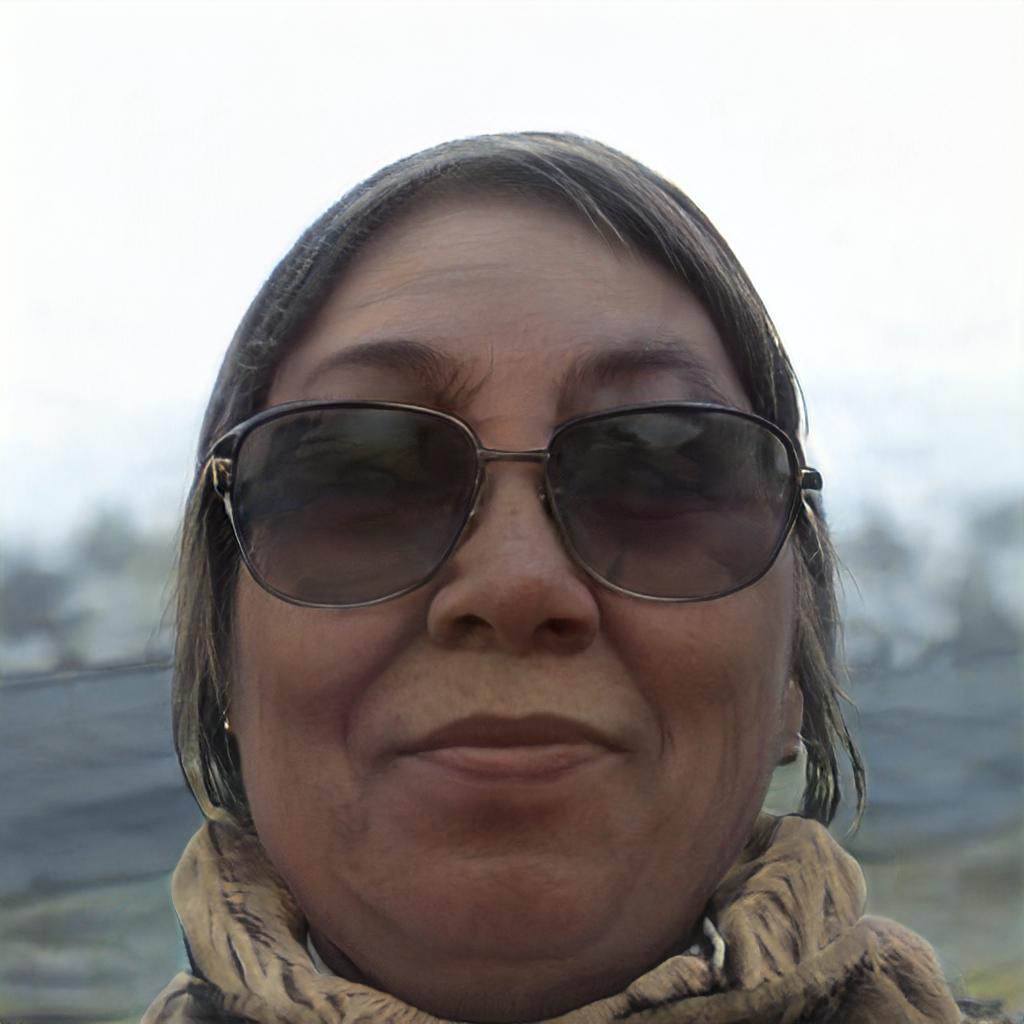} &
 \includegraphics[width=\simsize\linewidth]{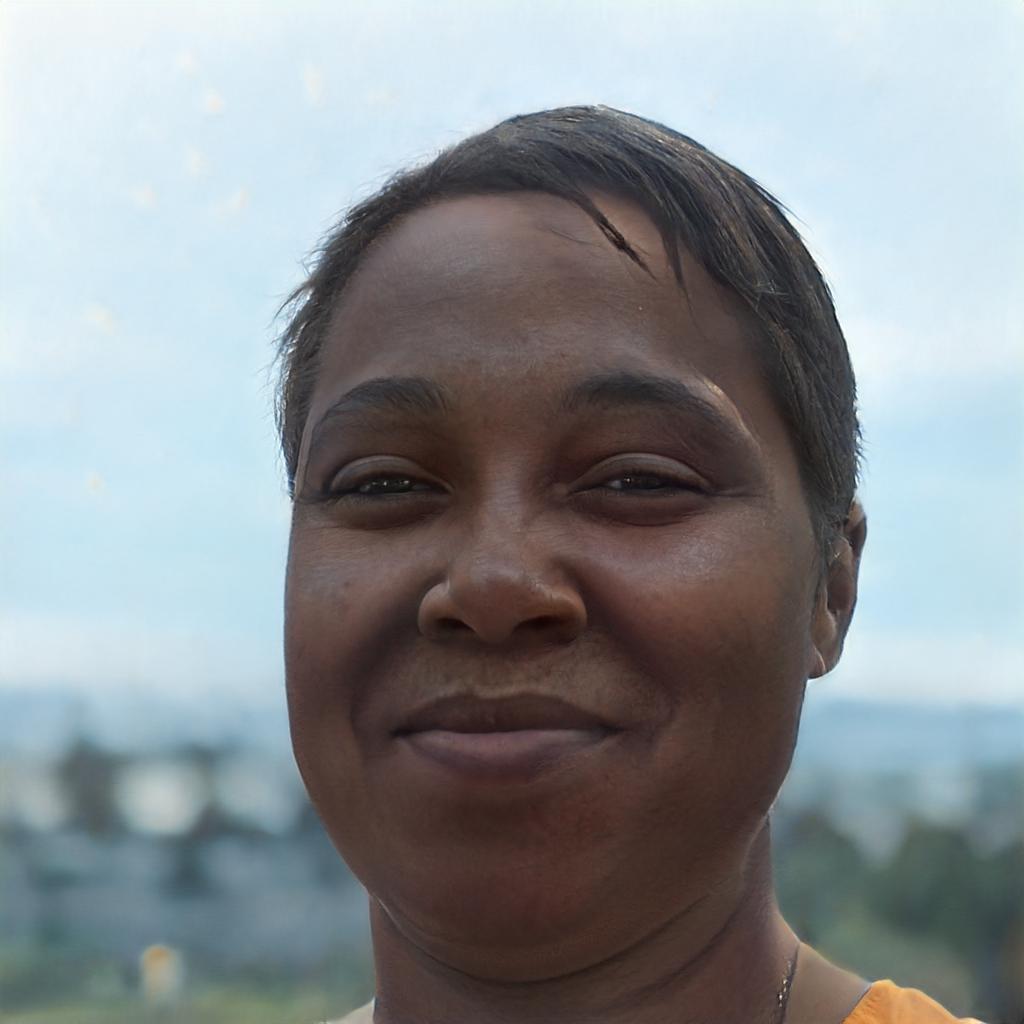} &
 \includegraphics[width=\simsize\linewidth]{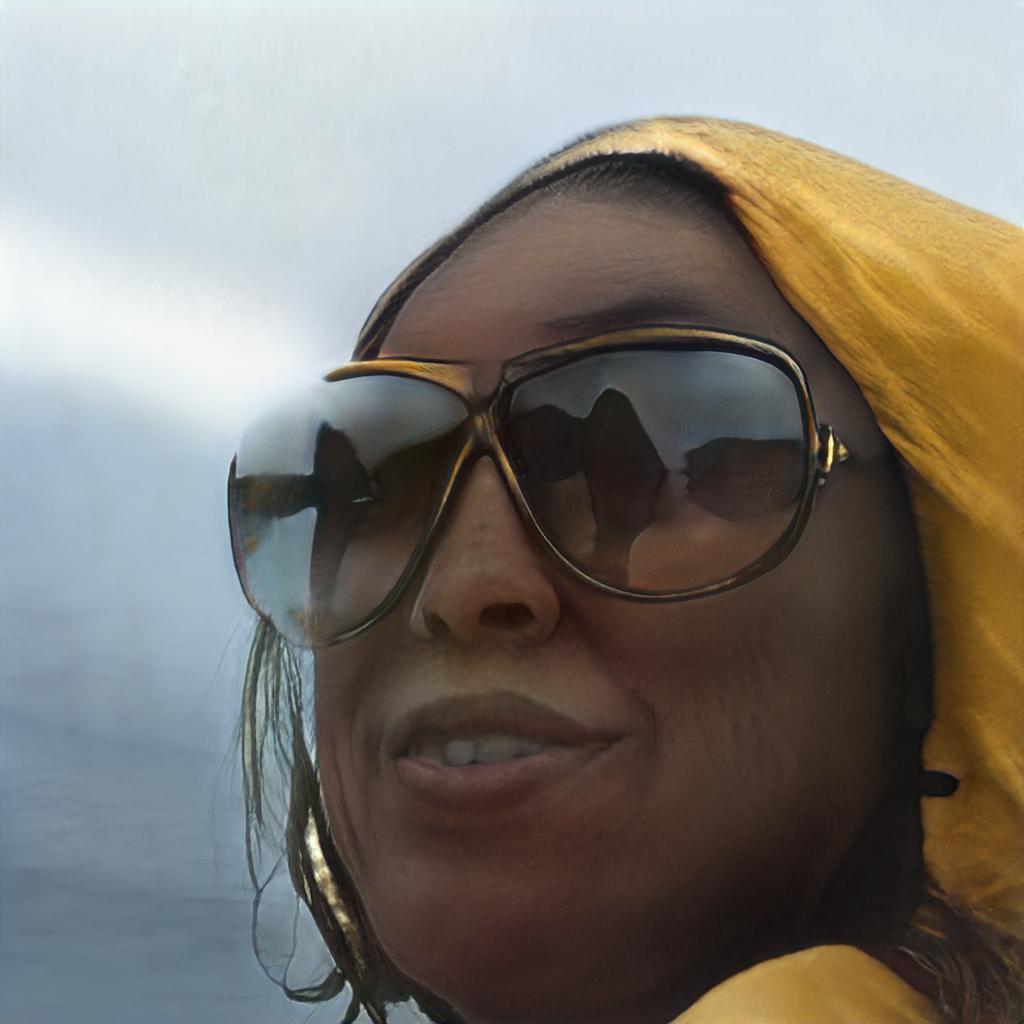} \\ 
 
 \hspace{0.7in}\rotatebox{90}{\hspace{0.0in}InterfaceGan} &
 \includegraphics[width=\simsize\linewidth]{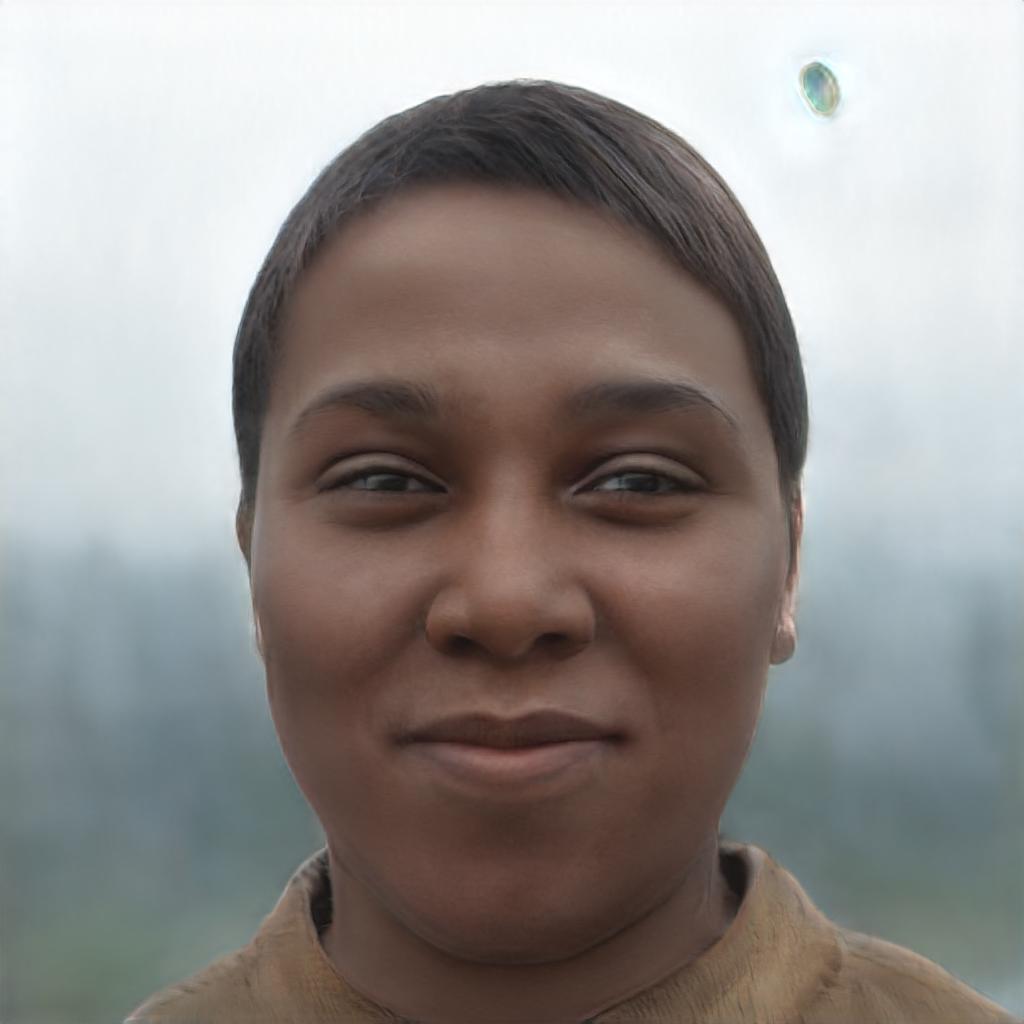} &
 \includegraphics[width=\simsize\linewidth]{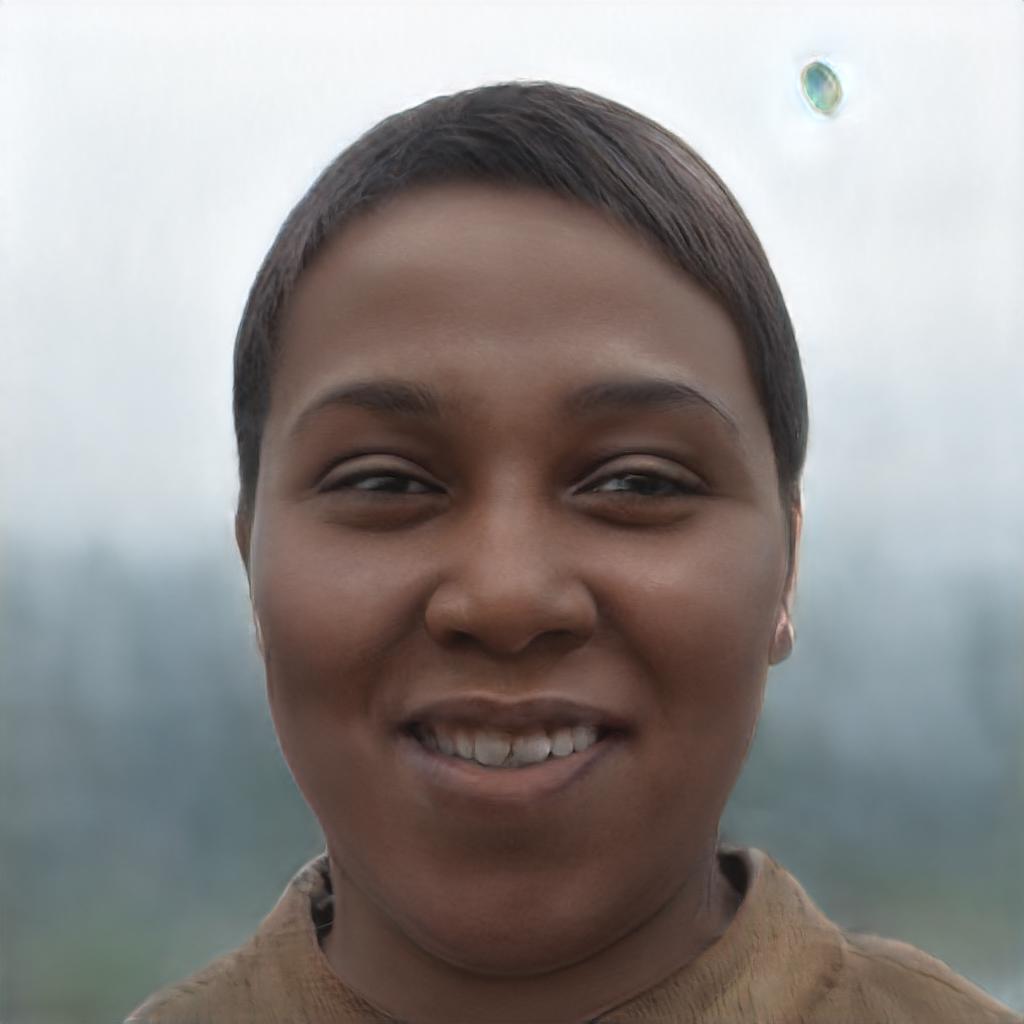} &
 \includegraphics[width=\simsize\linewidth]{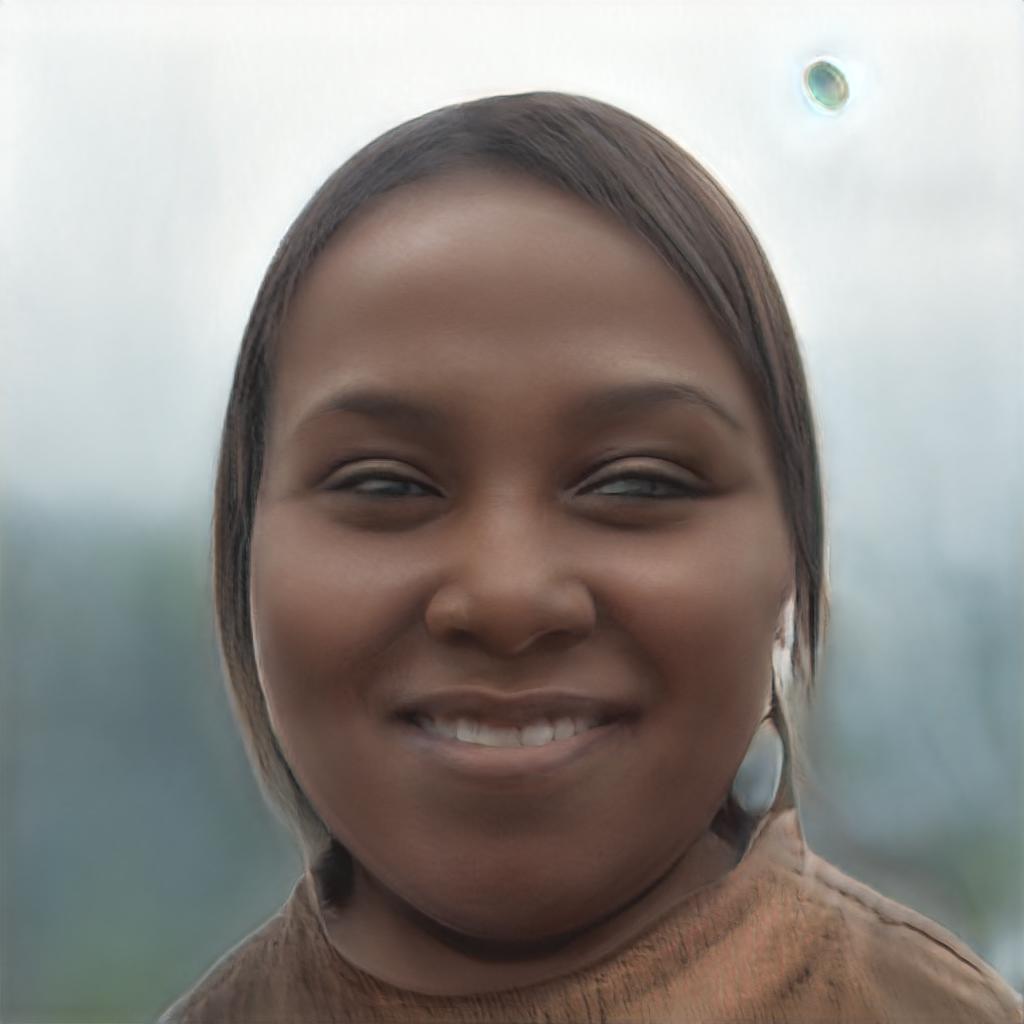} &
 \includegraphics[width=\simsize\linewidth]{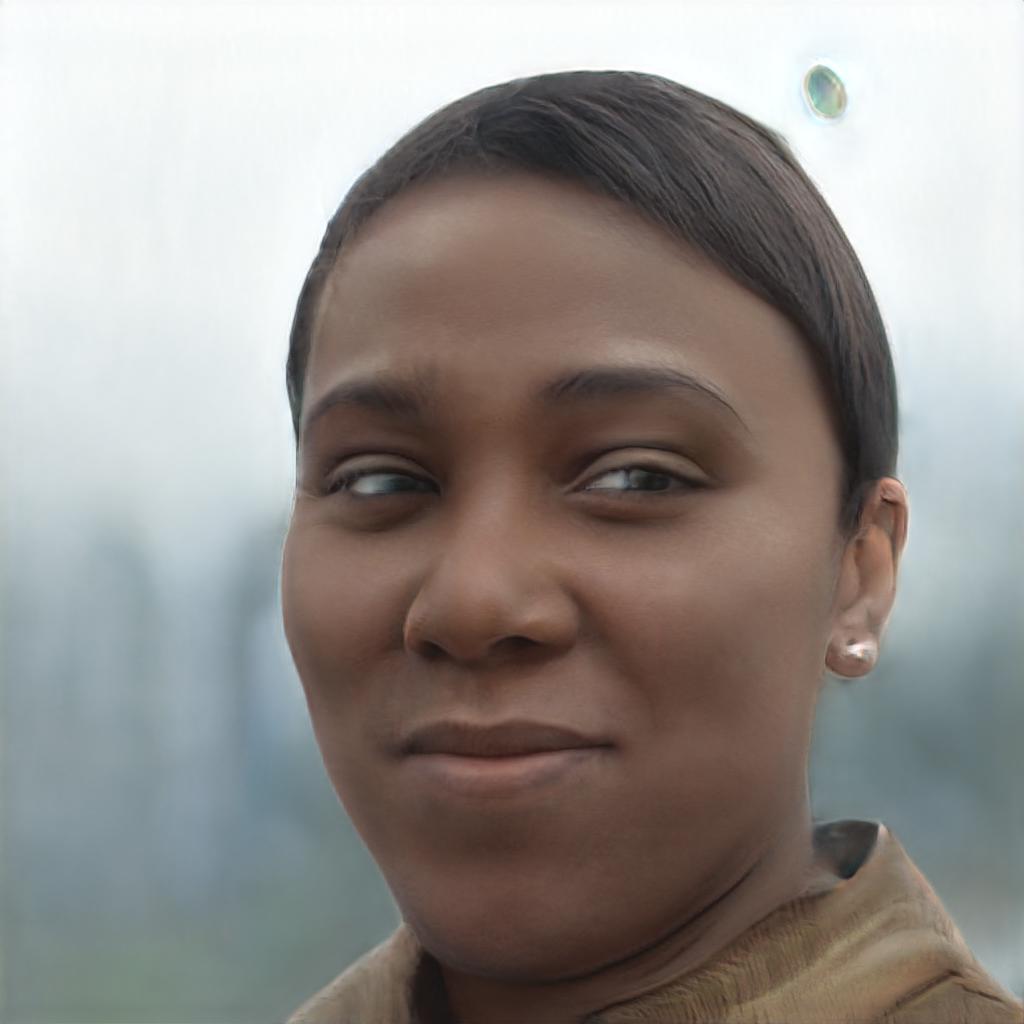} &
 \includegraphics[width=\simsize\linewidth]{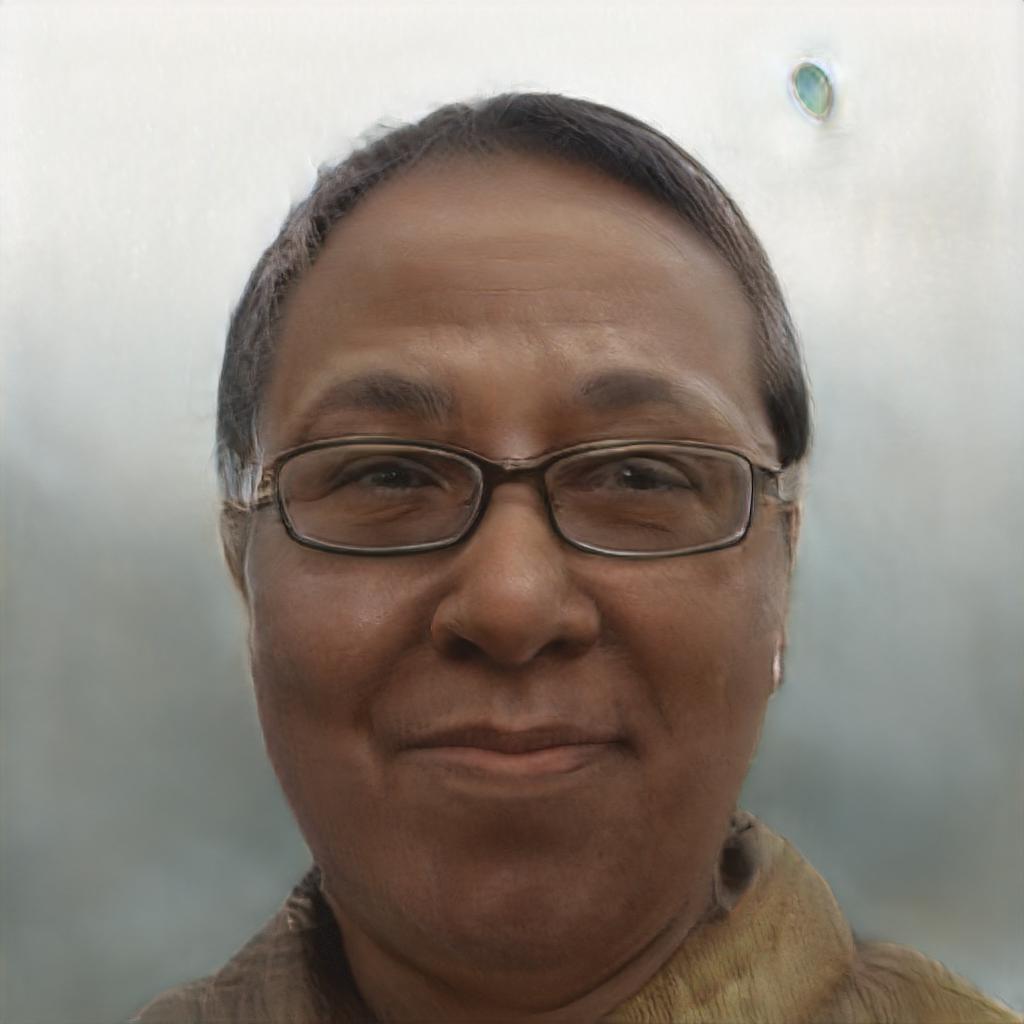} &
 \includegraphics[width=\simsize\linewidth]{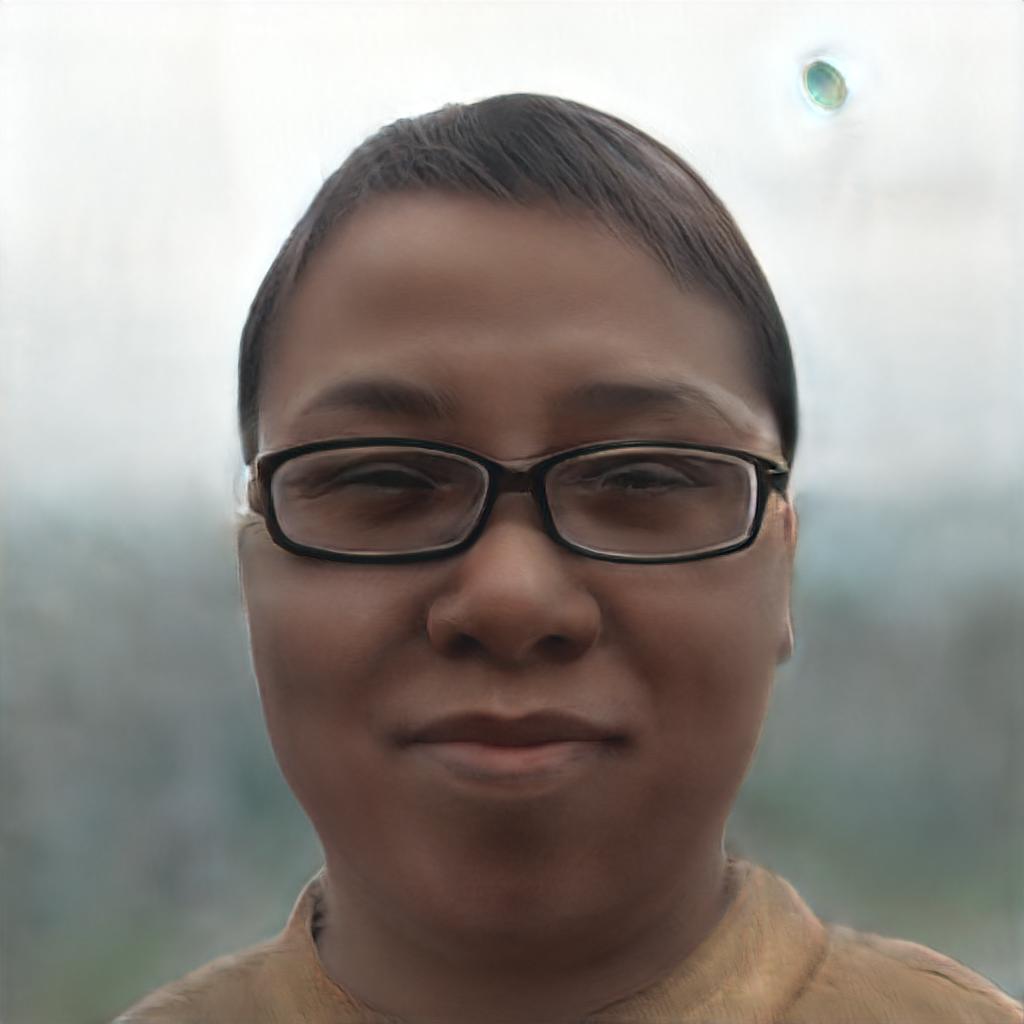} &
 \includegraphics[width=\simsize\linewidth]{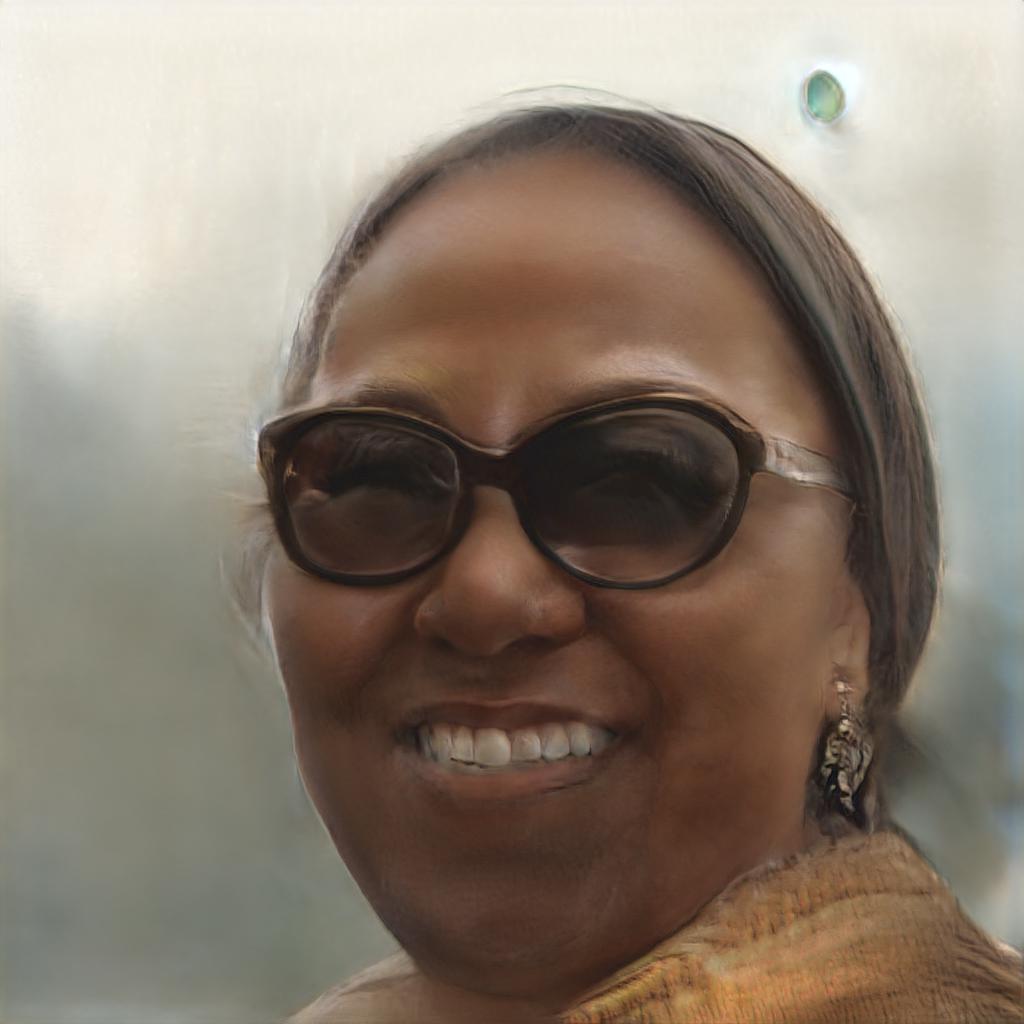} \\
 
   \hspace{0.7in}\rotatebox{90}{\hspace{0.12in}StyleFlow} &
 \includegraphics[width=\simsize\linewidth]{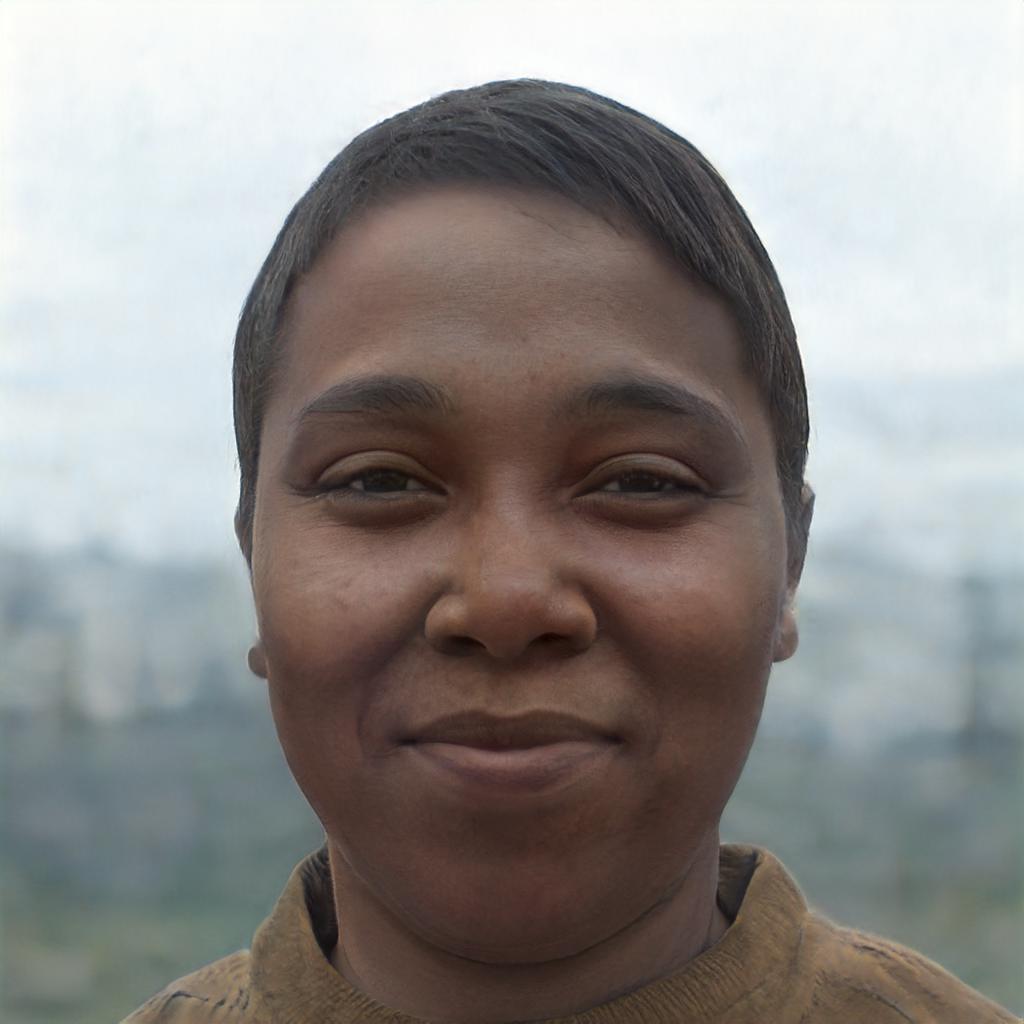} &
 \includegraphics[width=\simsize\linewidth]{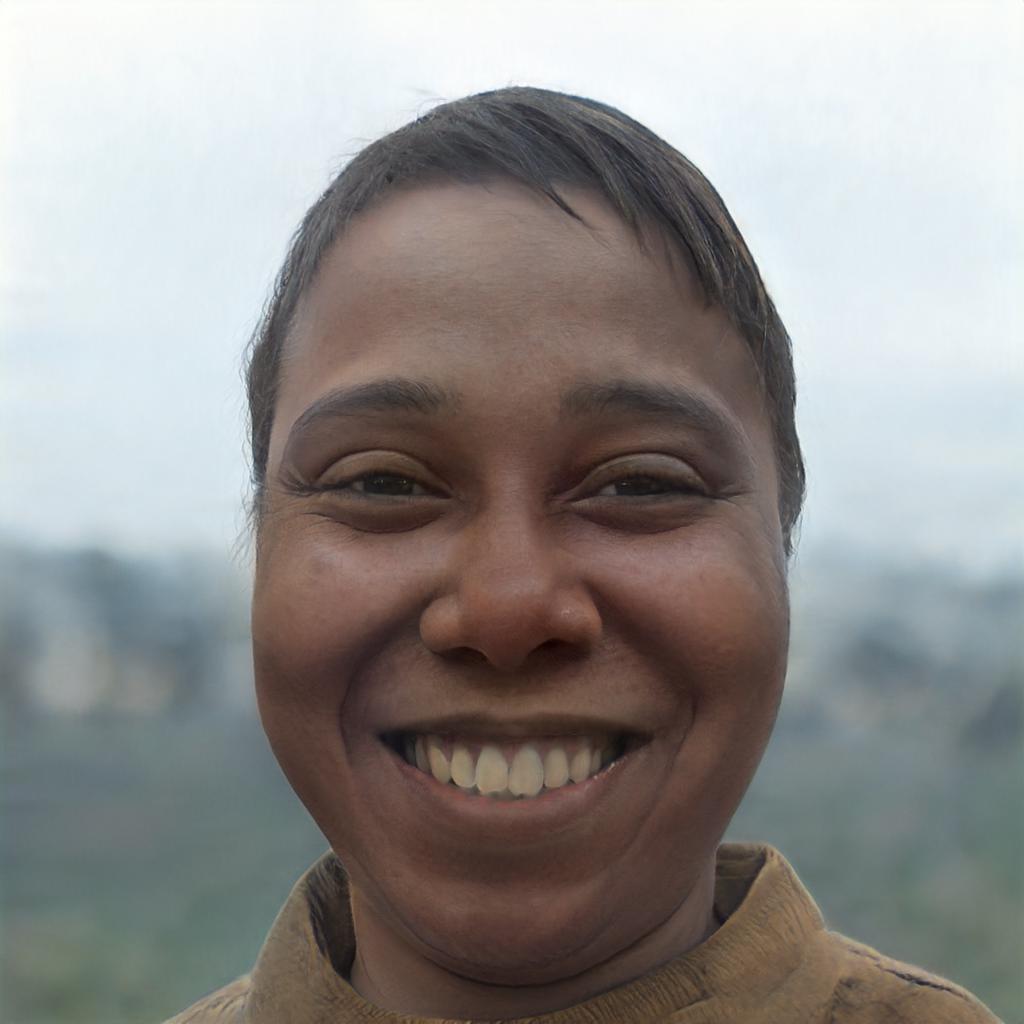} &
 \includegraphics[width=\simsize\linewidth]{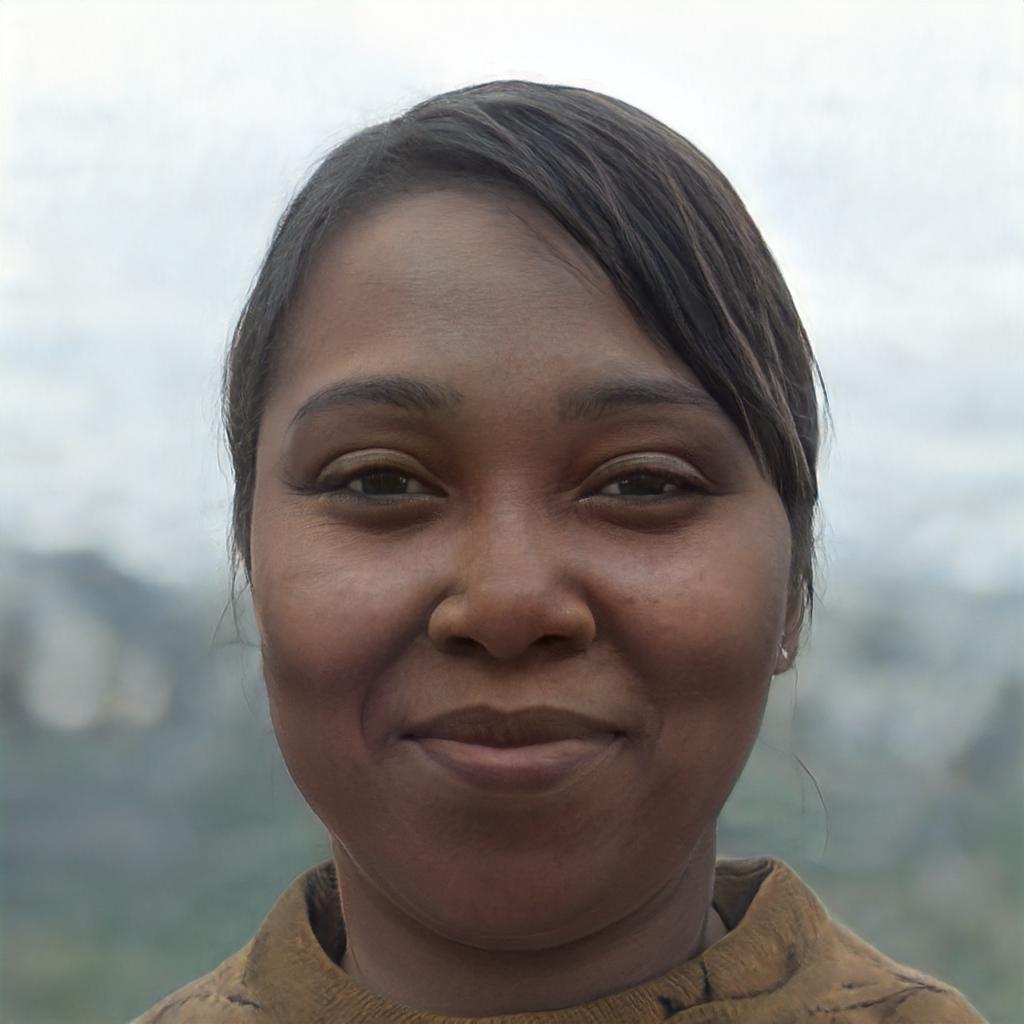} &
 \includegraphics[width=\simsize\linewidth]{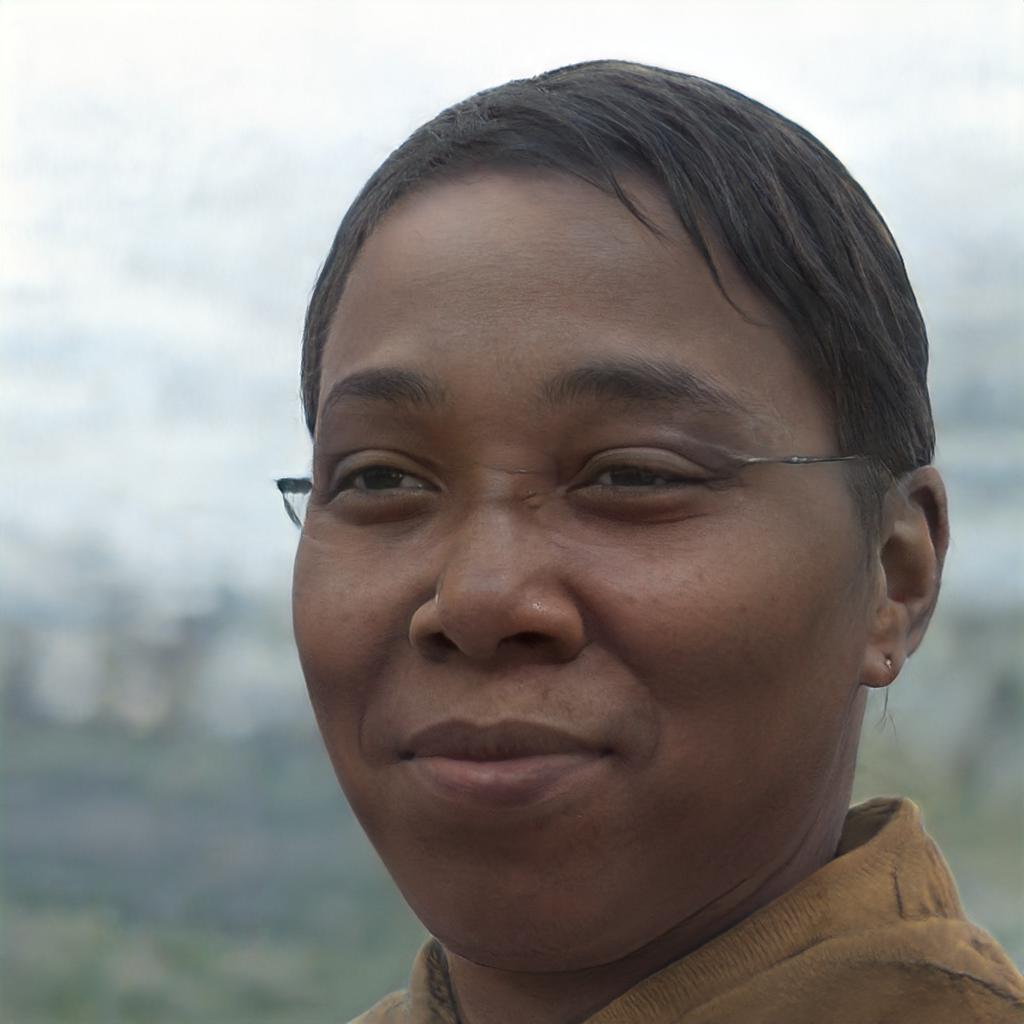} &
 \includegraphics[width=\simsize\linewidth]{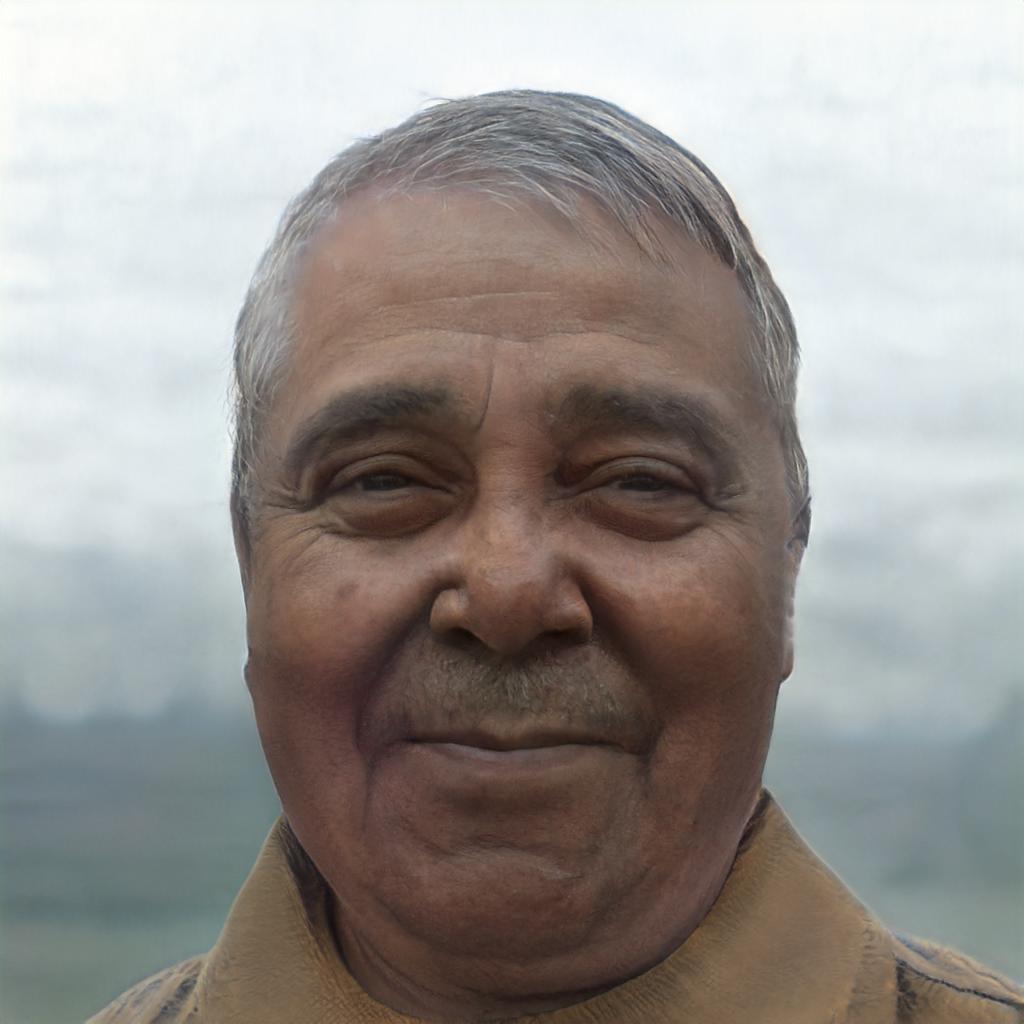} &
 \includegraphics[width=\simsize\linewidth]{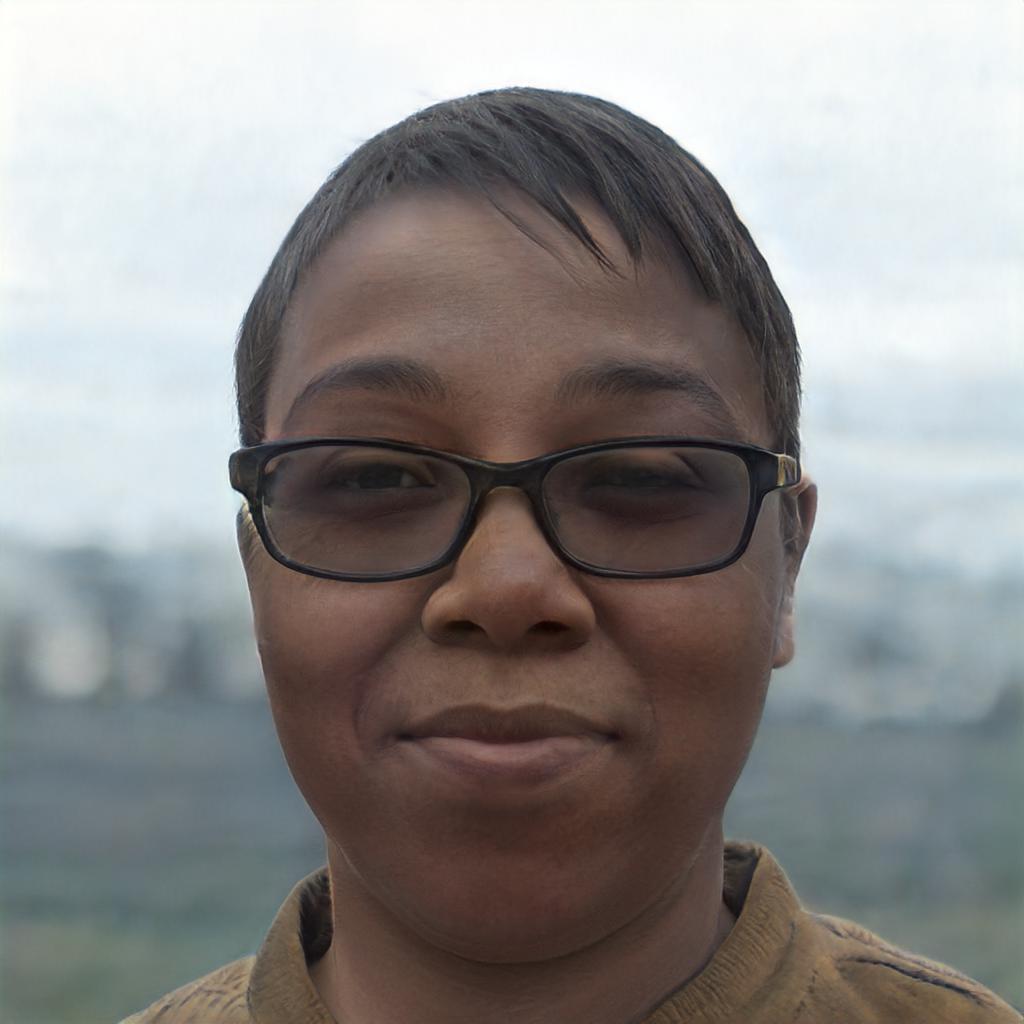} &
 \includegraphics[width=\simsize\linewidth]{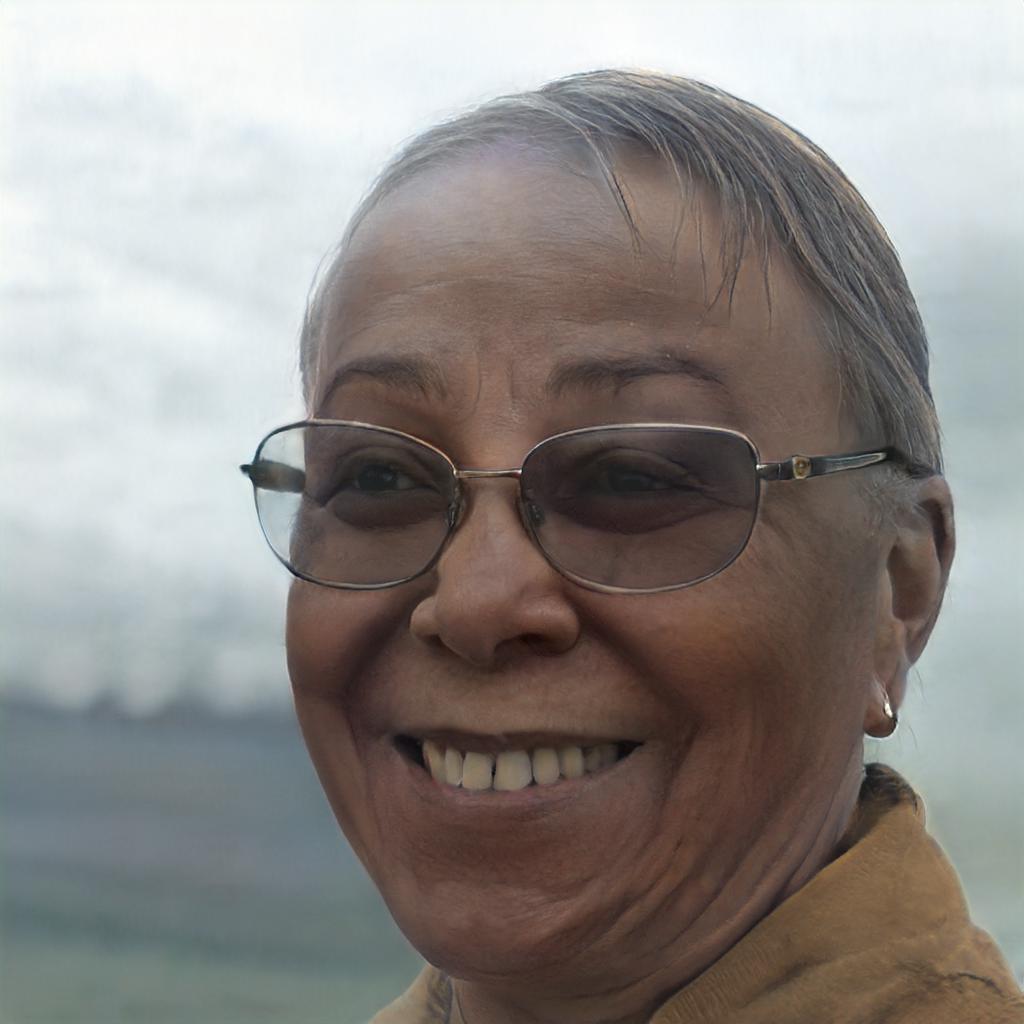} \\ 
 
  \hspace{0.7in}\rotatebox{90}{\hspace{0.29in}Ours} &
 \includegraphics[width=\simsize\linewidth]{images/quality_comp/00001_orig_baseline_2.0.jpg} &
 \includegraphics[width=\simsize\linewidth]{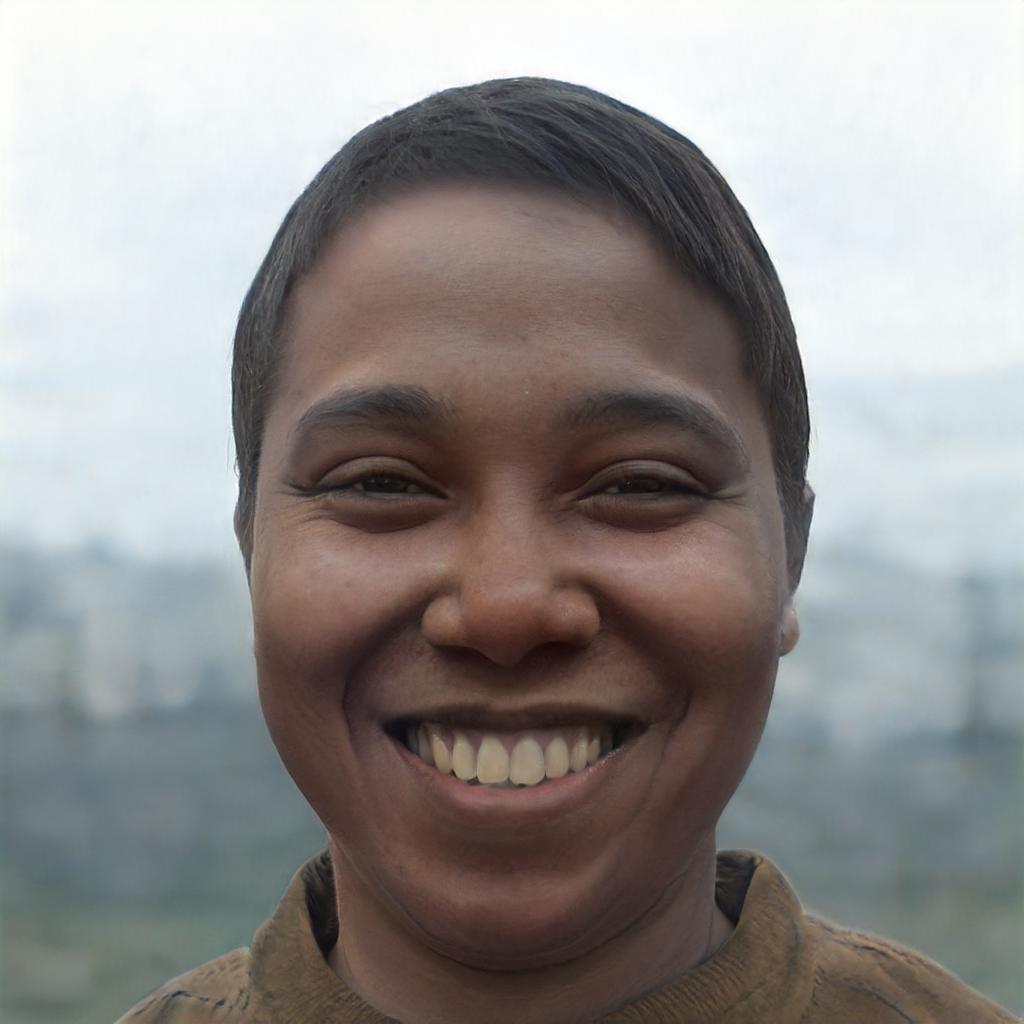} &
 \includegraphics[width=\simsize\linewidth]{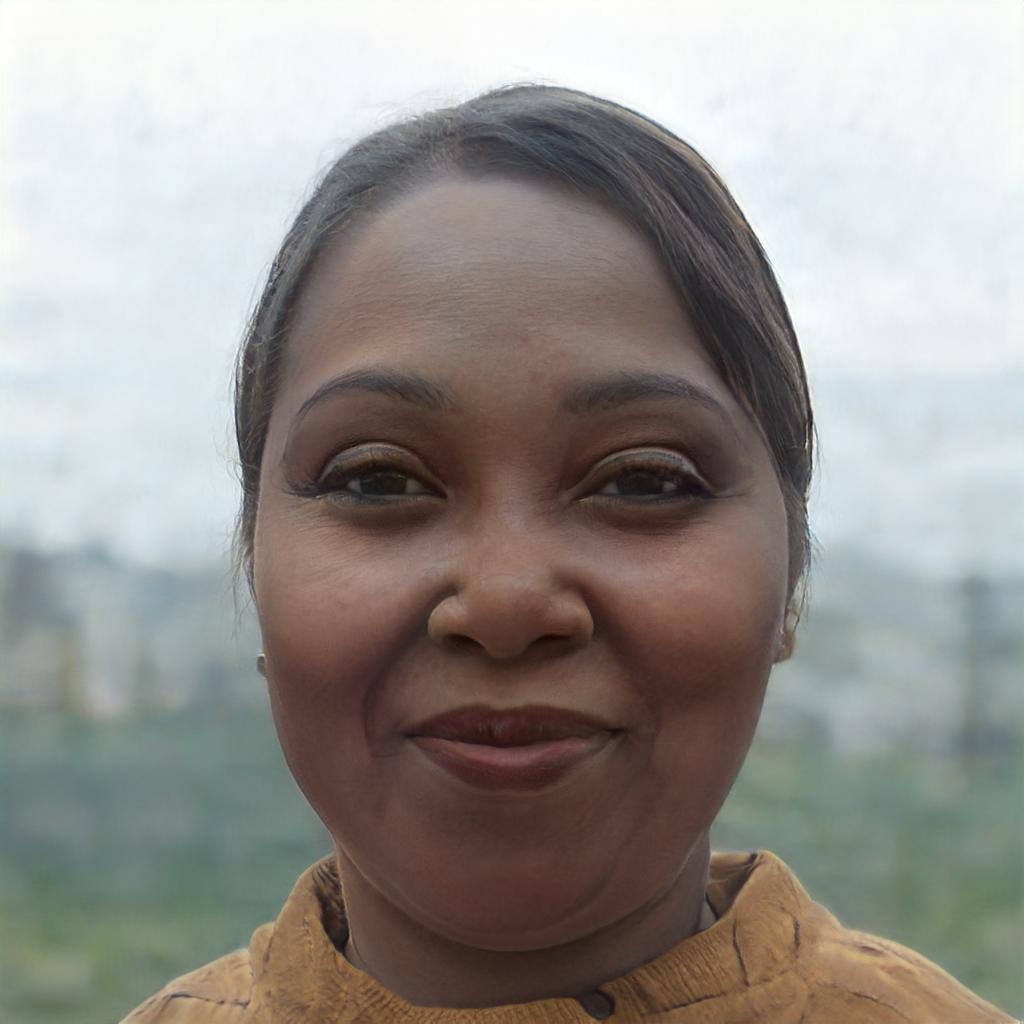} &
 \includegraphics[width=\simsize\linewidth]{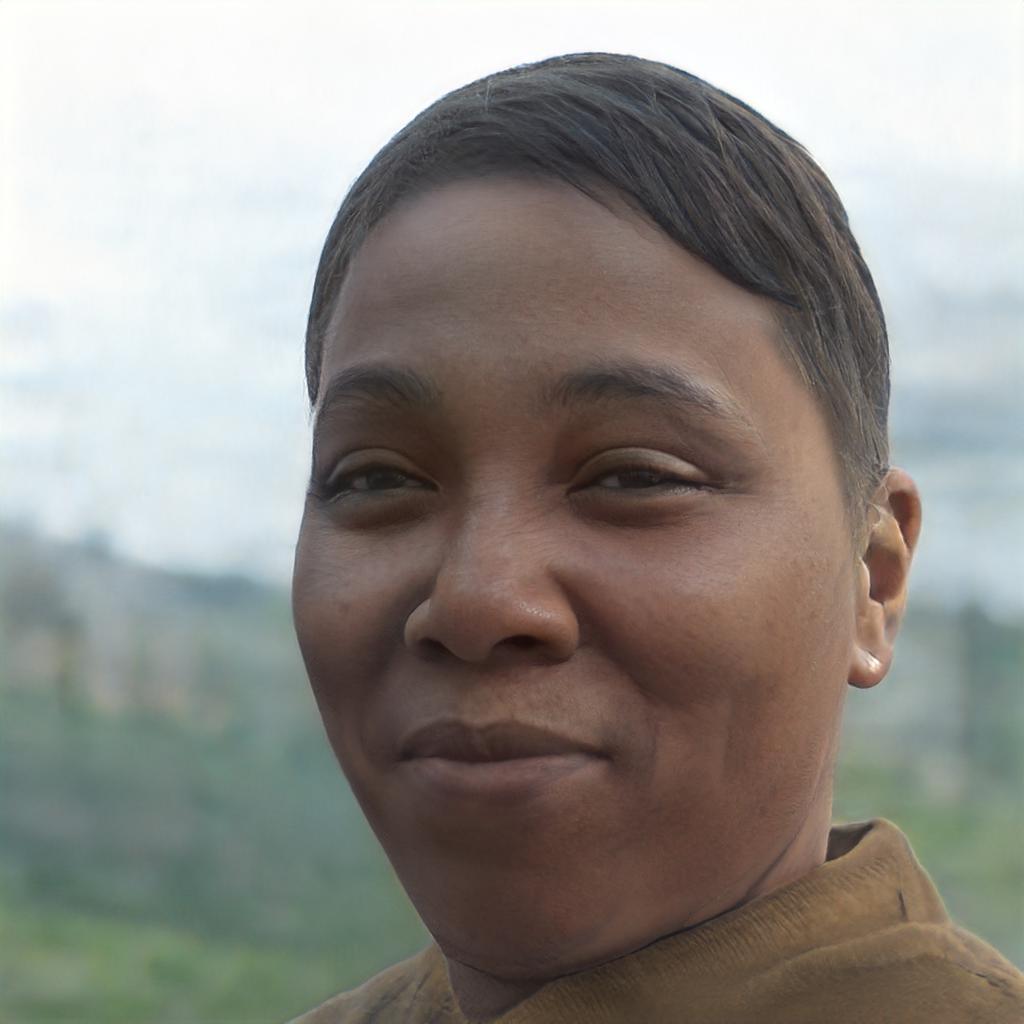} &
 \includegraphics[width=\simsize\linewidth]{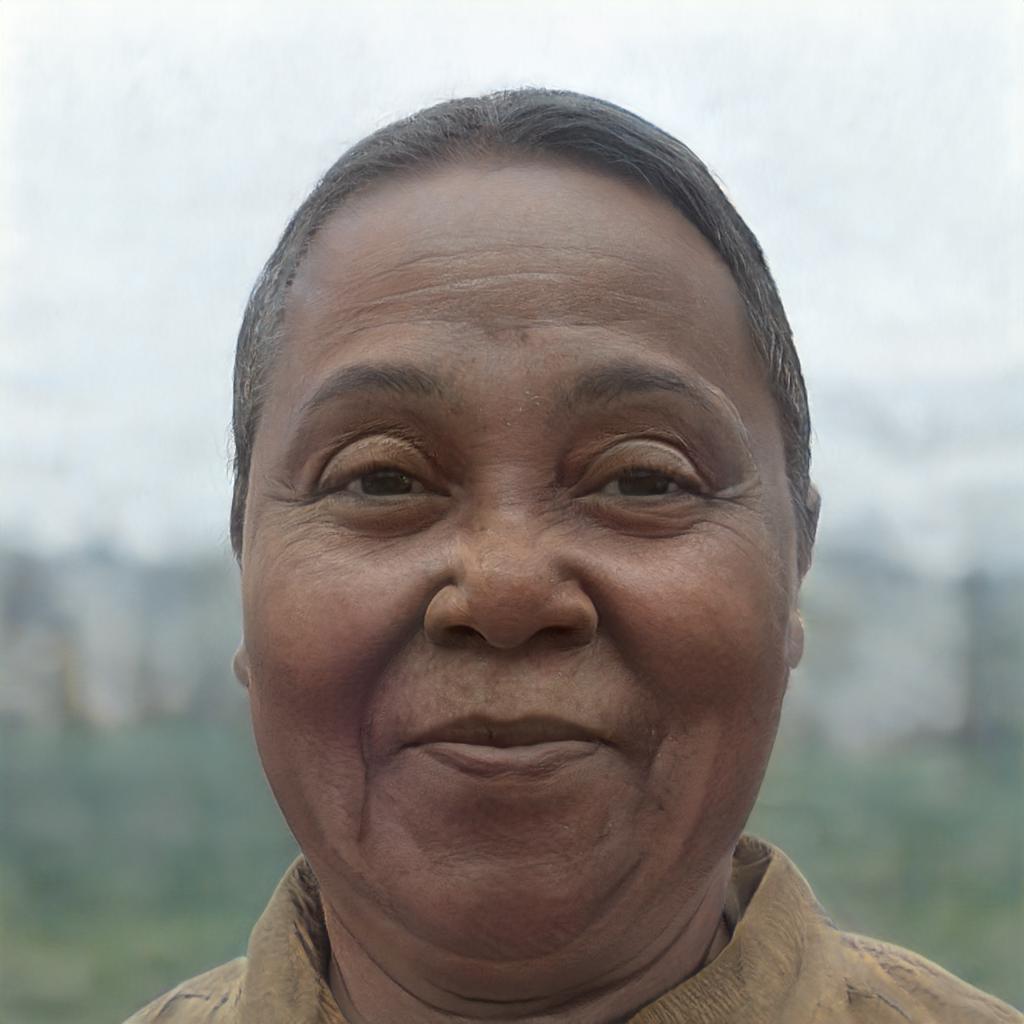} &
 \includegraphics[width=\simsize\linewidth]{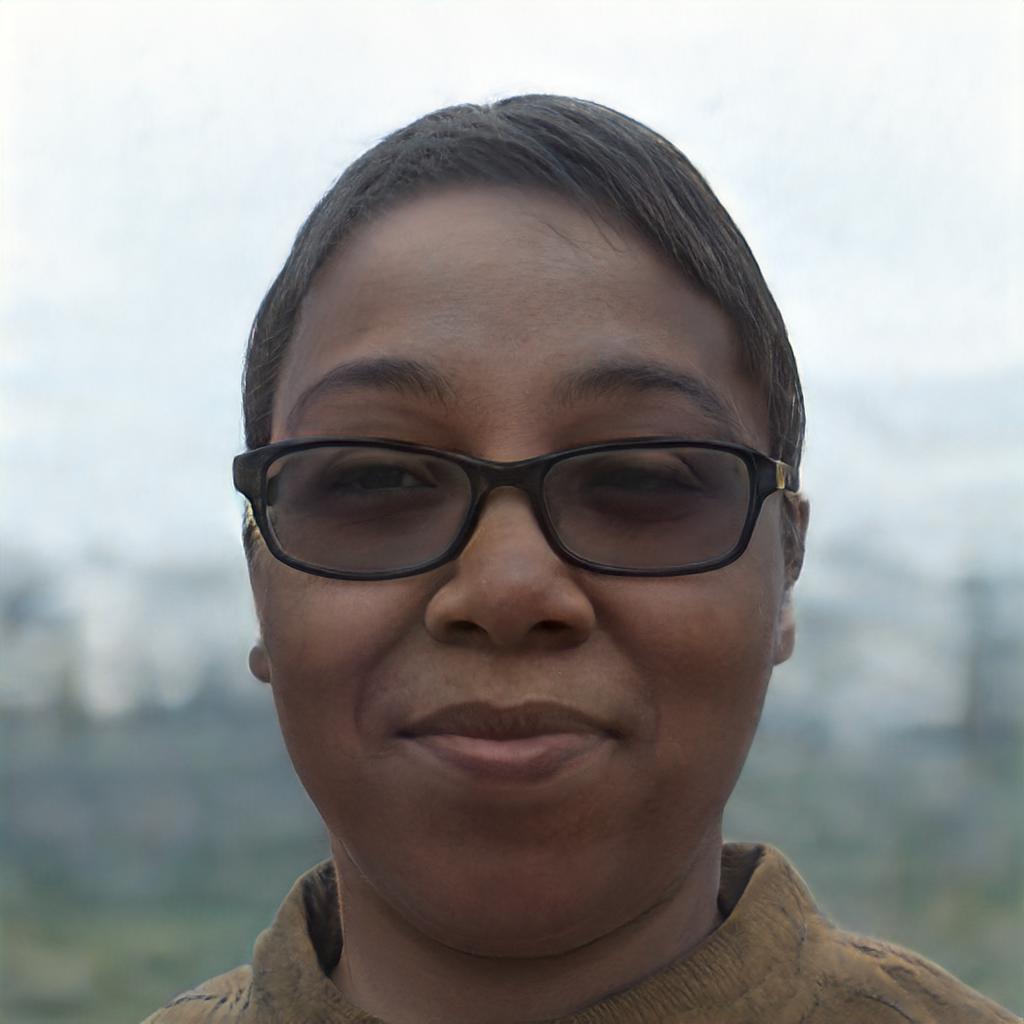} &
 \includegraphics[width=\simsize\linewidth]{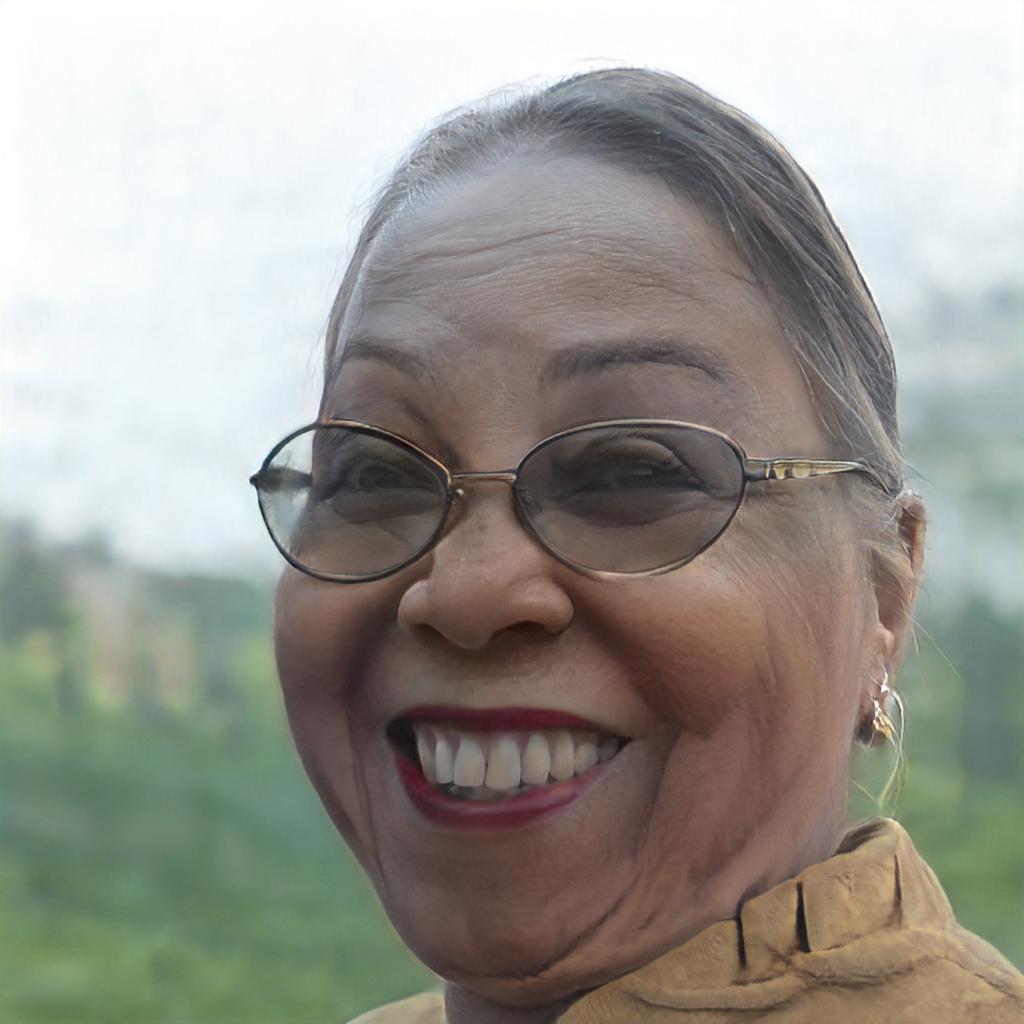} \\
 
\end{tabular}
\caption{Real image editing comparison between our method and InterfaceGAN\cite{shen2020interfacegan}, Sefa\cite{shen2021closed} and StyleFlow\cite{abdal2021styleflow}. On the right-most column we apply all previous edits at once.}
\label{fig:fig1}
\end{figure}


\section{Multi-Directional Edits}
To demonstrate our multi-directional edits we first inverted real images into the latent space of StyleGAN. Then we chose different vectors inside a subspace to edit each image. The results are shown in \cref{fig:fig7,fig:fig77,fig:fig777}.

\newcommand*{\simwidth}{0.17}
\begin{figure}[h]
    \centering
    \begin{tabular}{c|cccc}
        Input & \multicolumn{4}{c}{Edits In Gender} \\
         \includegraphics[width=\simwidth\linewidth]{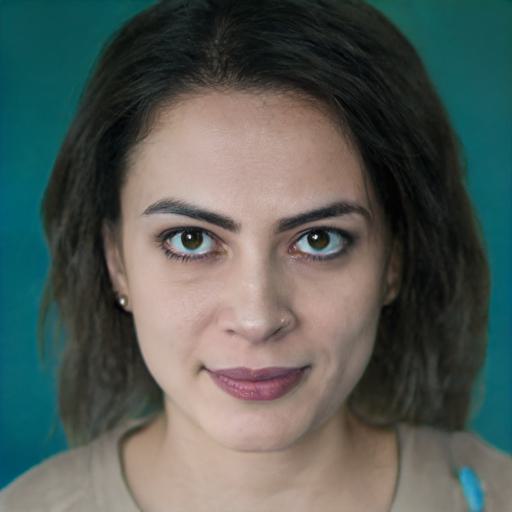} &
         \includegraphics[width=\simwidth\linewidth]{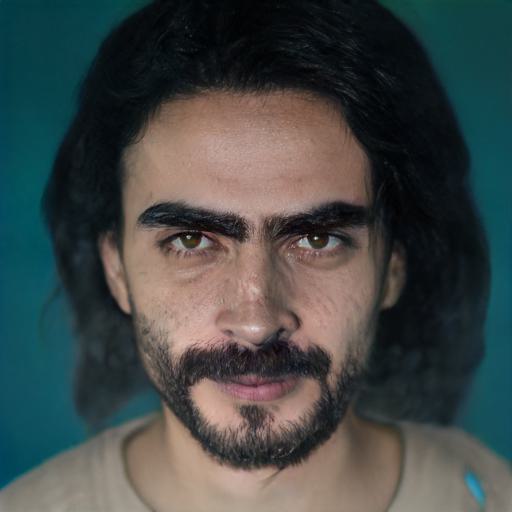} &
         \includegraphics[width=\simwidth\linewidth]{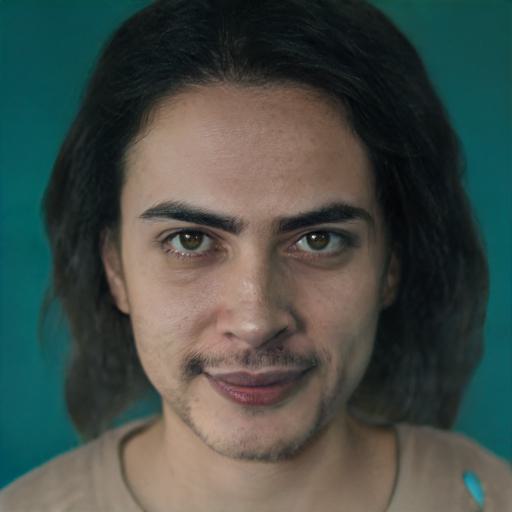} &
         \includegraphics[width=\simwidth\linewidth]{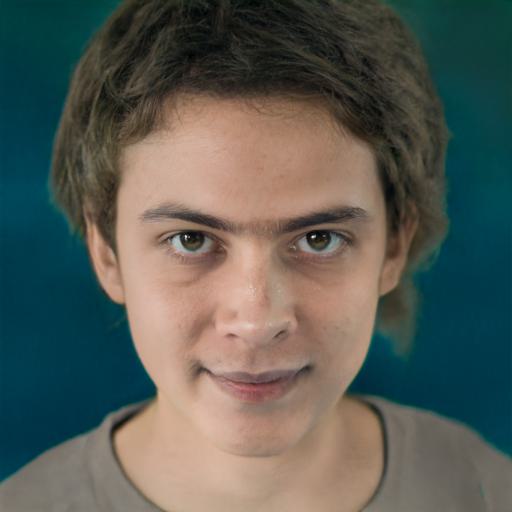} &
         \includegraphics[width=\simwidth\linewidth]{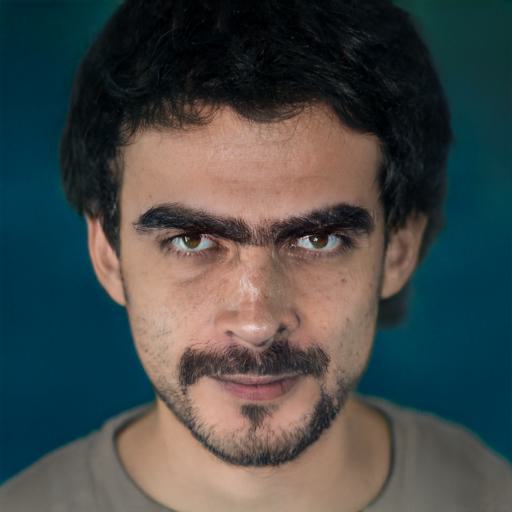} \\
         \includegraphics[width=\simwidth\linewidth]{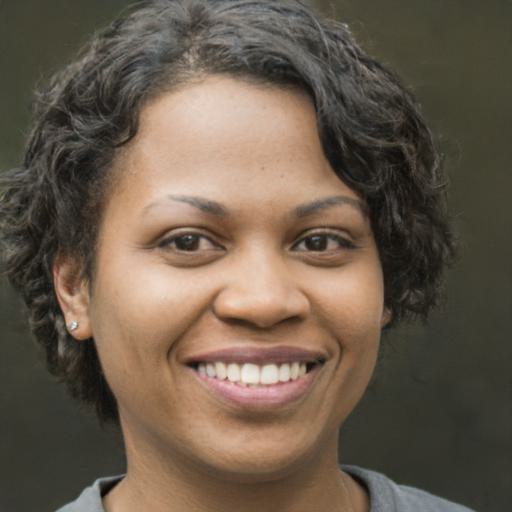} &
         \includegraphics[width=\simwidth\linewidth]{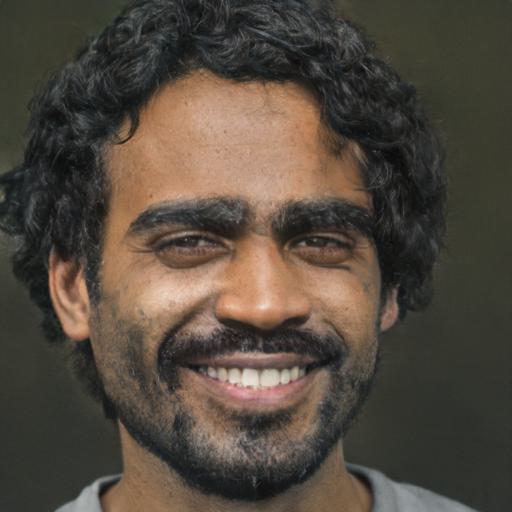} &
         \includegraphics[width=\simwidth\linewidth]{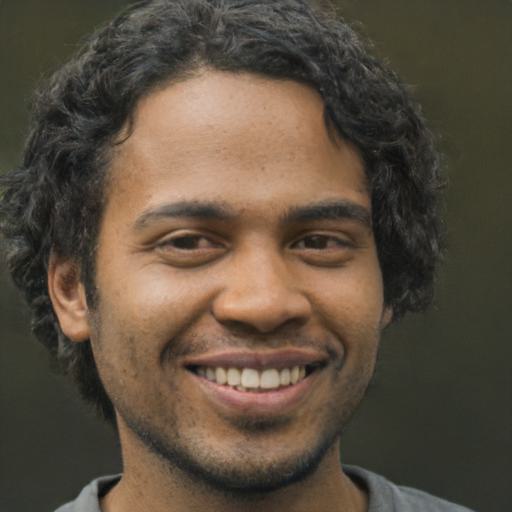} &
         \includegraphics[width=\simwidth\linewidth]{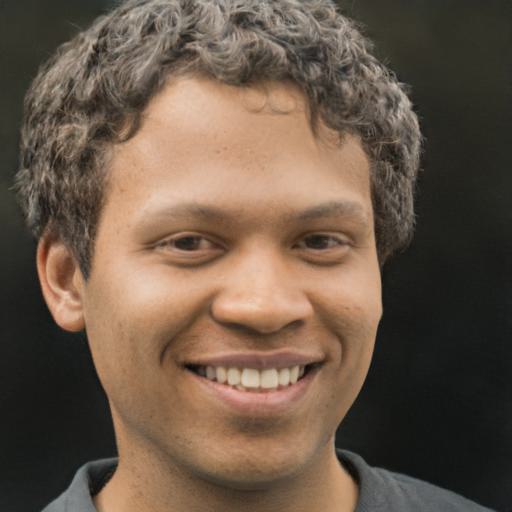} &
         \includegraphics[width=\simwidth\linewidth]{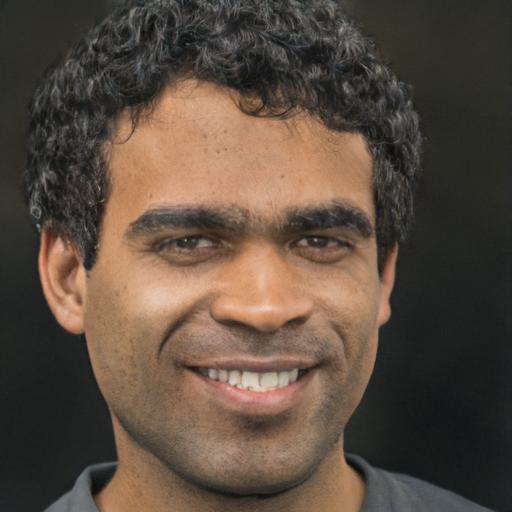} \\
         \includegraphics[width=\simwidth\linewidth]{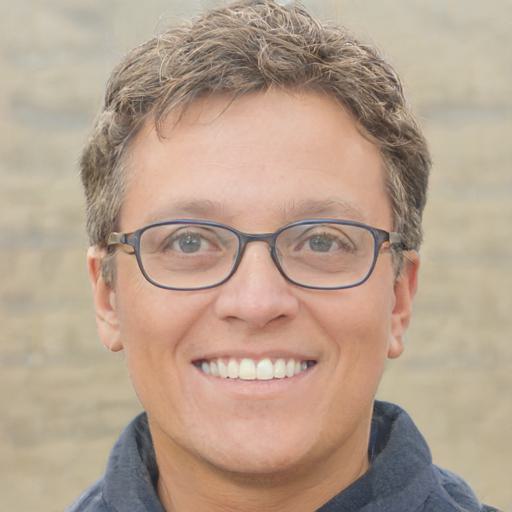} &
         \includegraphics[width=\simwidth\linewidth]{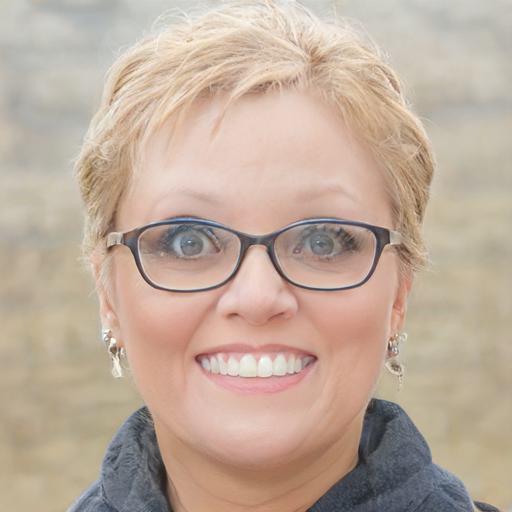} &
         \includegraphics[width=\simwidth\linewidth]{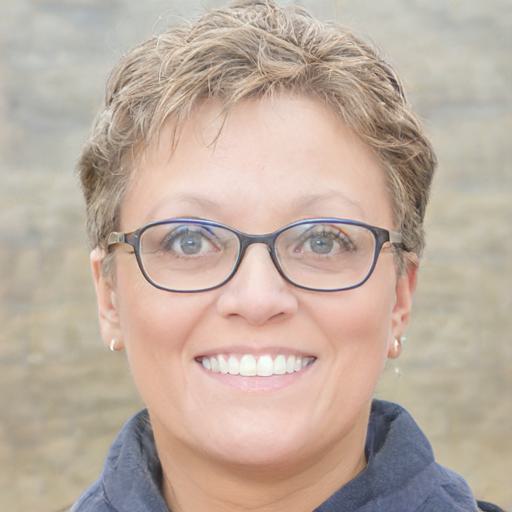} &
         \includegraphics[width=\simwidth\linewidth]{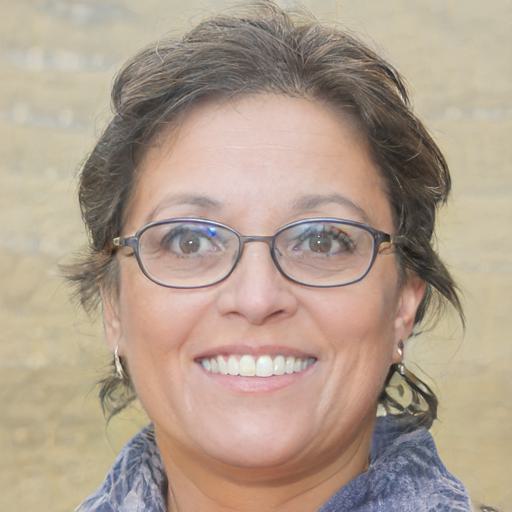} &
         \includegraphics[width=\simwidth\linewidth]{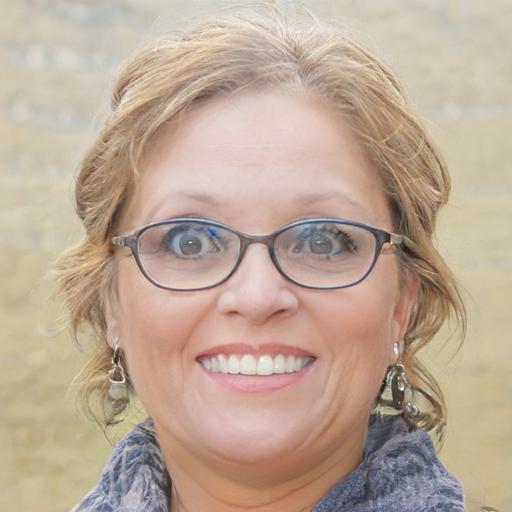} 
        \end{tabular}

    \caption{Multi-directional gender editing of real images. Each row displays different edits in the associated subspace.}
    \label{fig:fig7}
\end{figure}

\begin{figure}[h]
    \centering
    \begin{tabular}{c|cccc}
        Input & \multicolumn{4}{c}{Edits In Age} \\
         \includegraphics[width=\simwidth\linewidth]{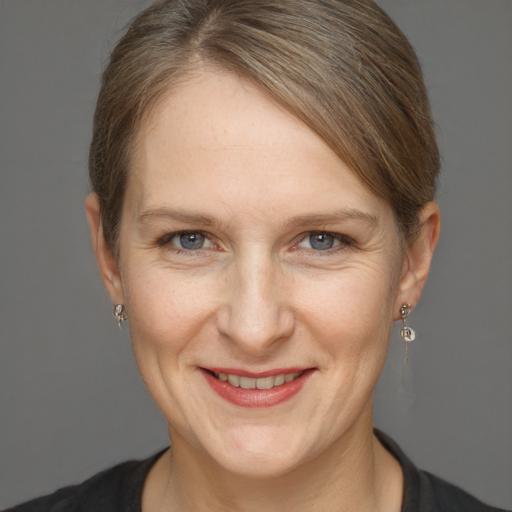} &
         \includegraphics[width=\simwidth\linewidth]{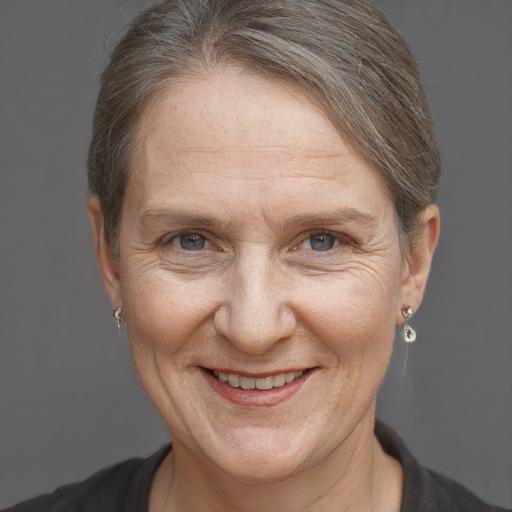} &
         \includegraphics[width=\simwidth\linewidth]{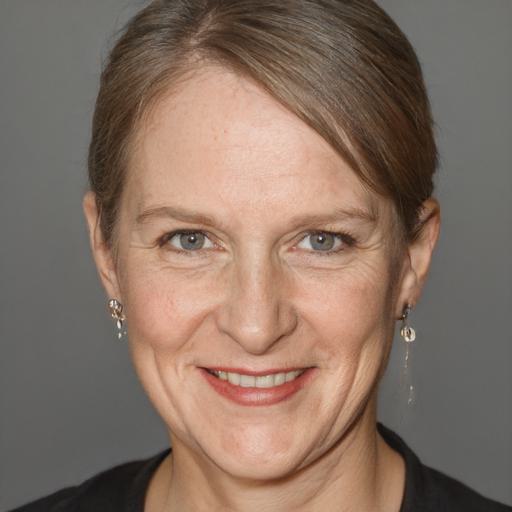} &
         \includegraphics[width=\simwidth\linewidth]{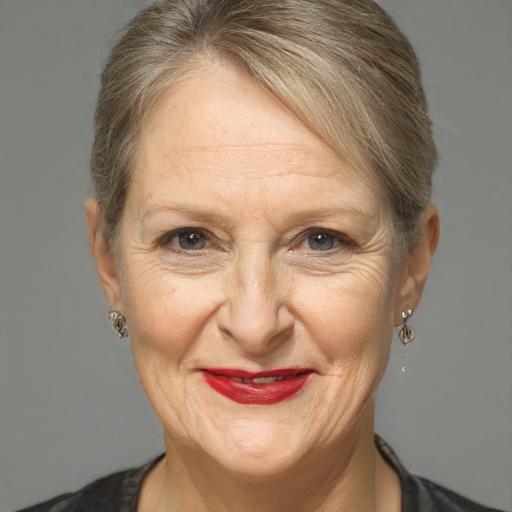} &
         \includegraphics[width=\simwidth\linewidth]{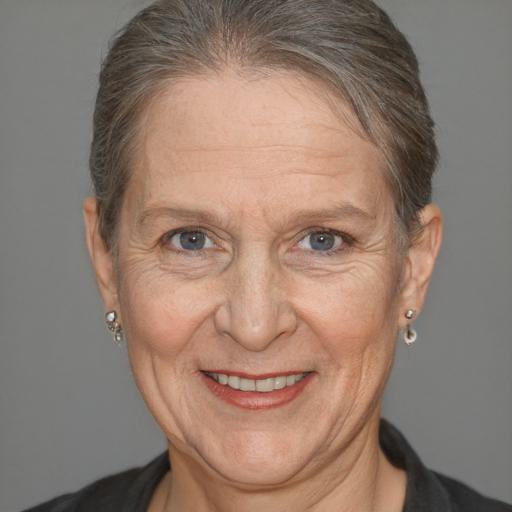} \\
         \includegraphics[width=\simwidth\linewidth]{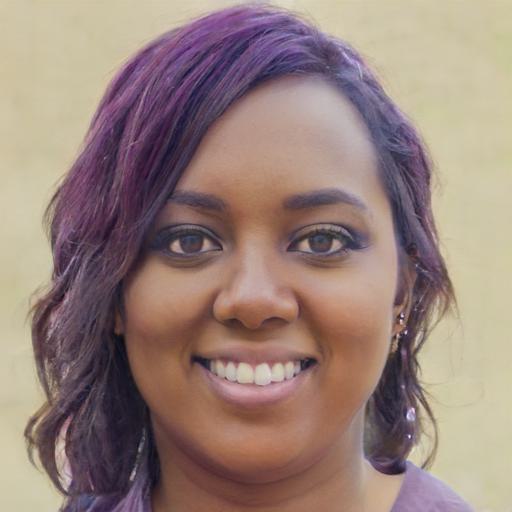} &
         \includegraphics[width=\simwidth\linewidth]{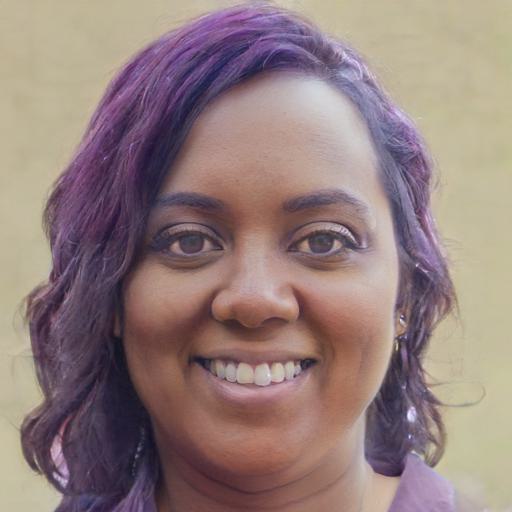} &
         \includegraphics[width=\simwidth\linewidth]{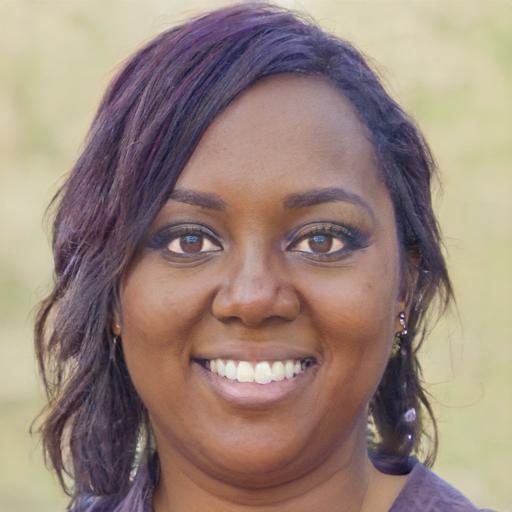} &
         \includegraphics[width=\simwidth\linewidth]{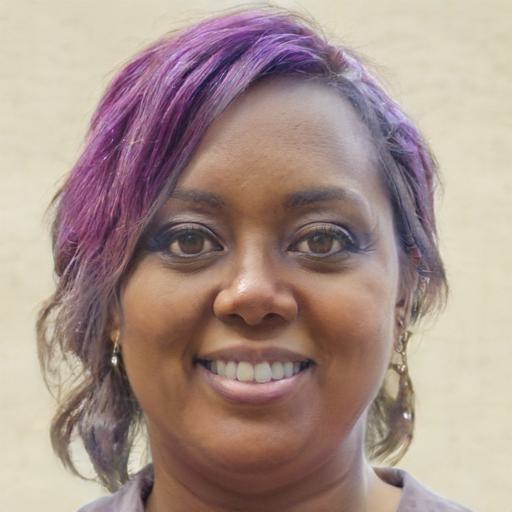} &
         \includegraphics[width=\simwidth\linewidth]{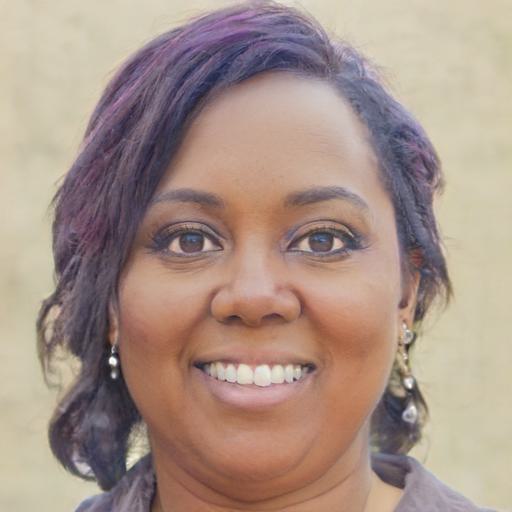} \\
         \includegraphics[width=\simwidth\linewidth]{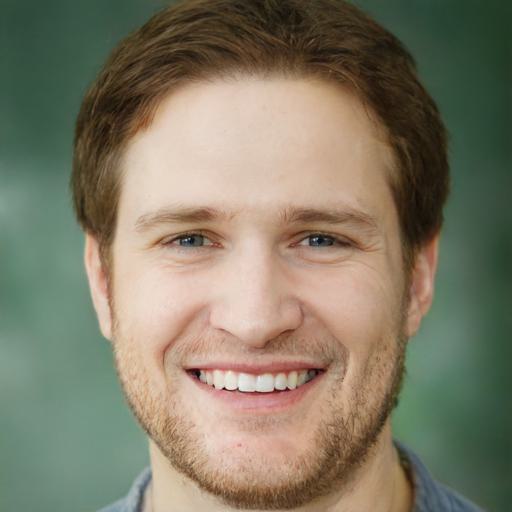} &
         \includegraphics[width=\simwidth\linewidth]{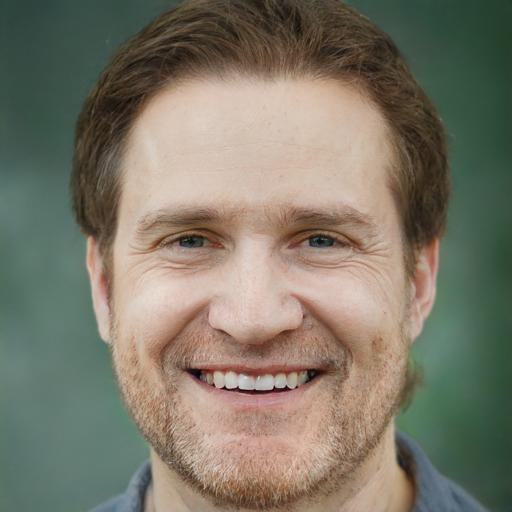} &
         \includegraphics[width=\simwidth\linewidth]{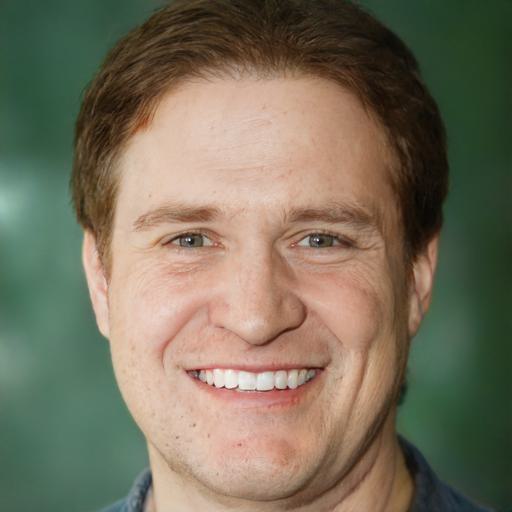} &
         \includegraphics[width=\simwidth\linewidth]{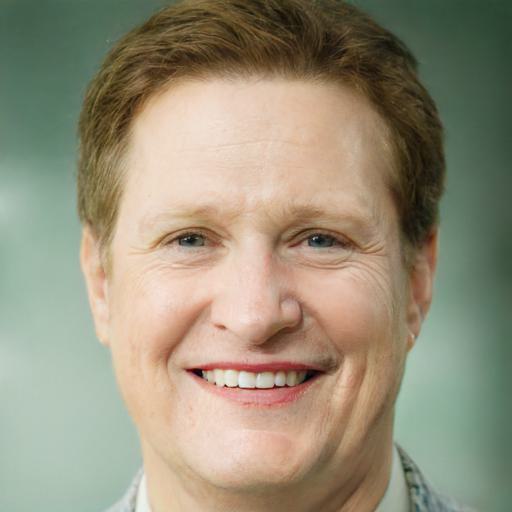} &
         \includegraphics[width=\simwidth\linewidth]{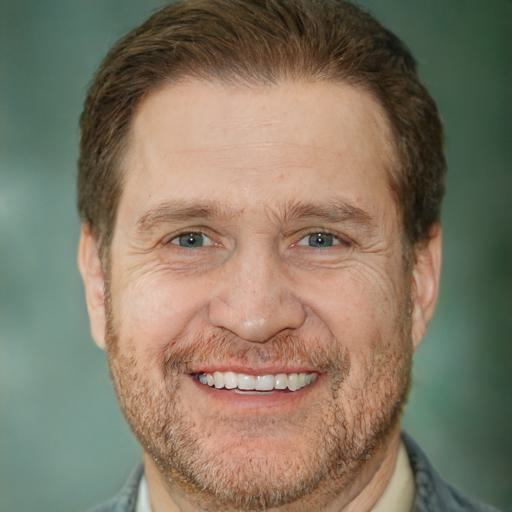} 
        \end{tabular}

    \caption{Multi-directional age editing of real images. Each row displays different edits in the associated subspace.}
    \label{fig:fig77}
\end{figure}

\begin{figure}[H]
    \centering
    \begin{tabular}{c|cccc}
        Input & \multicolumn{4}{c}{Edits In Race} \\
          \includegraphics[width=\simwidth\linewidth]{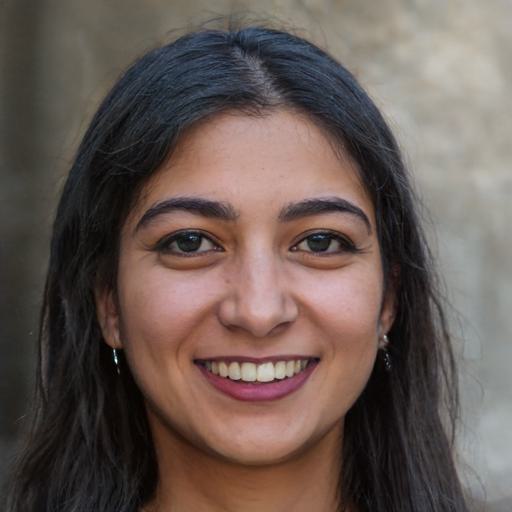} &
         \includegraphics[width=\simwidth\linewidth]{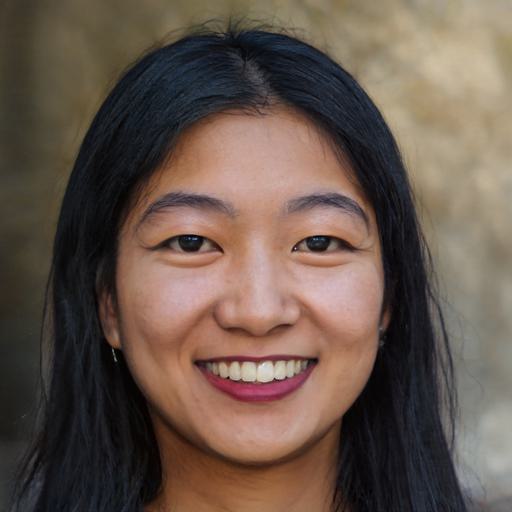} &
         \includegraphics[width=\simwidth\linewidth]{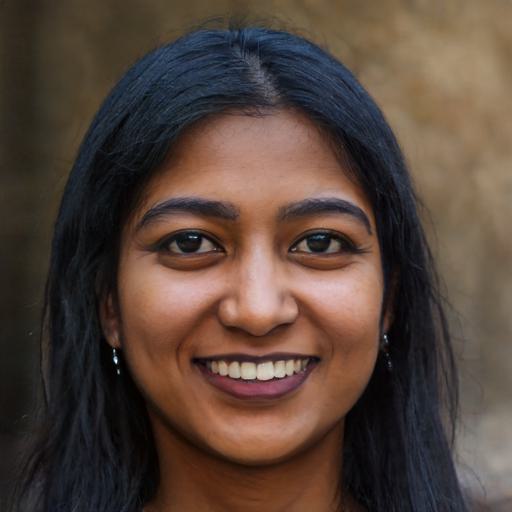} &
         \includegraphics[width=\simwidth\linewidth]{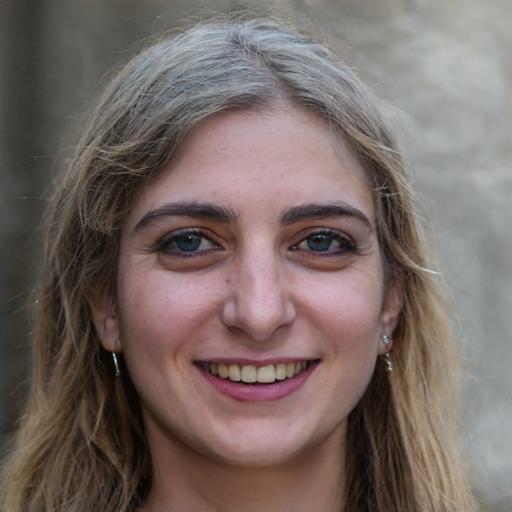} &
         \includegraphics[width=\simwidth\linewidth]{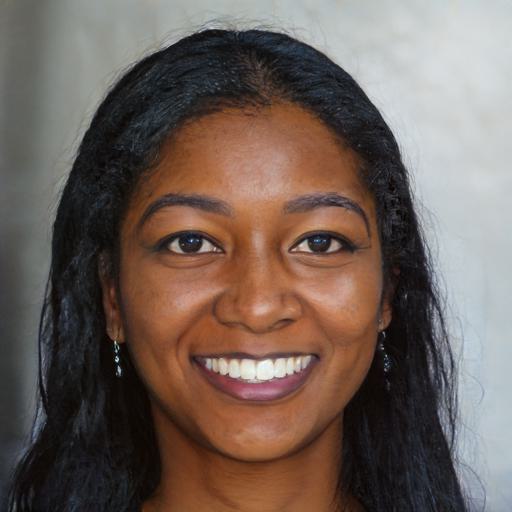} \\
         \includegraphics[width=\simwidth\linewidth]{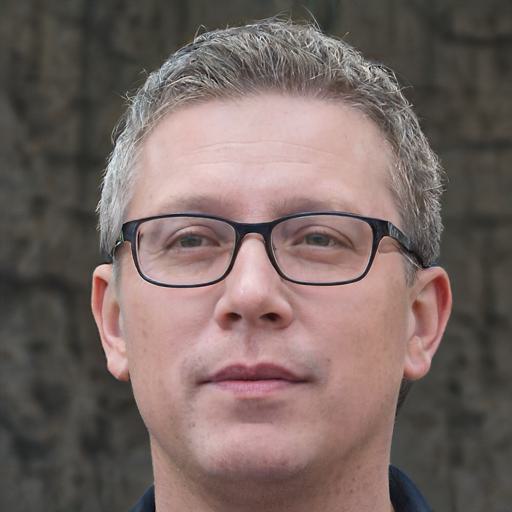} &
         \includegraphics[width=\simwidth\linewidth]{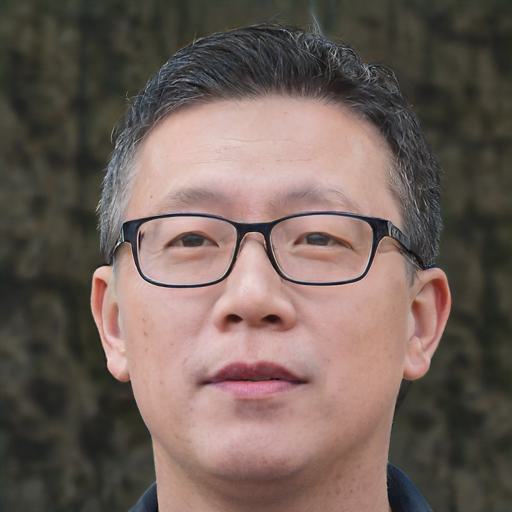} &
         \includegraphics[width=\simwidth\linewidth]{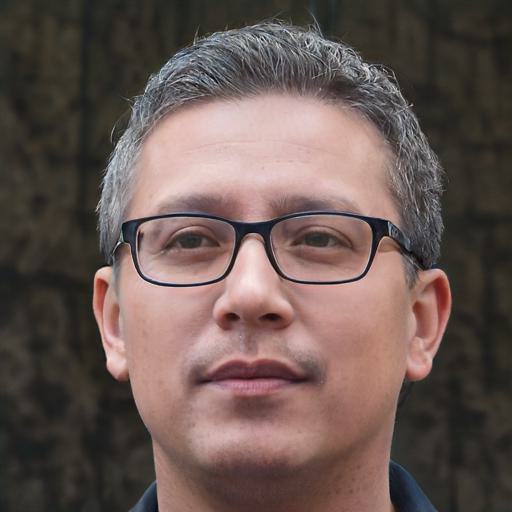} &
         \includegraphics[width=\simwidth\linewidth]{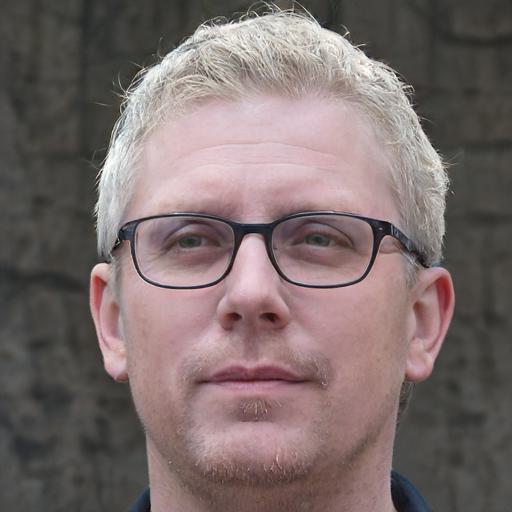} &
         \includegraphics[width=\simwidth\linewidth]{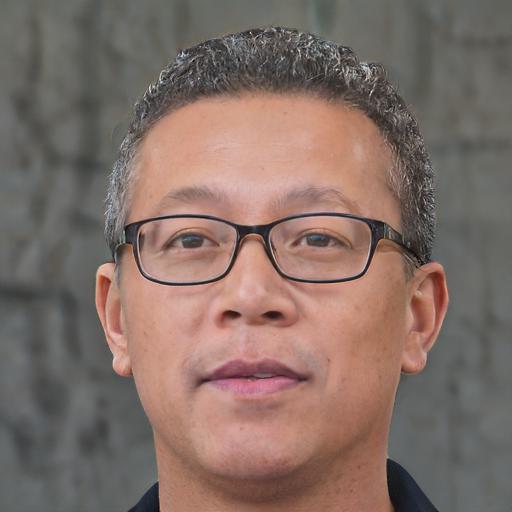} \\
         \includegraphics[width=\simwidth\linewidth]{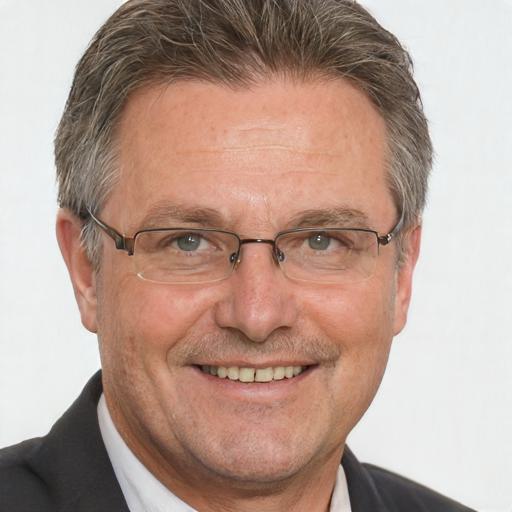} &
         \includegraphics[width=\simwidth\linewidth]{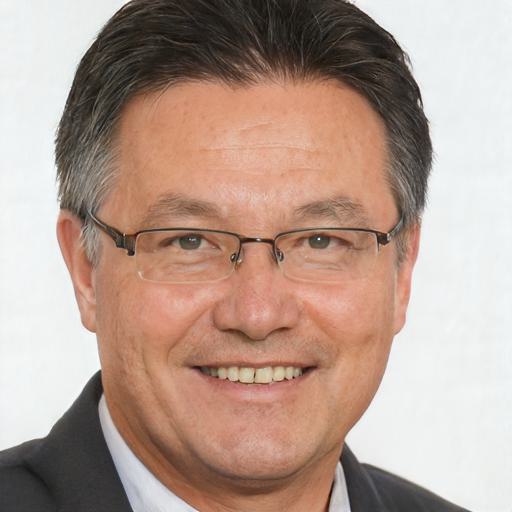} &
         \includegraphics[width=\simwidth\linewidth]{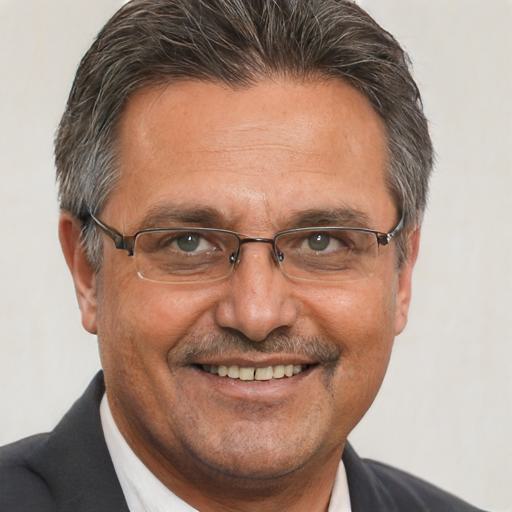} &
         \includegraphics[width=\simwidth\linewidth]{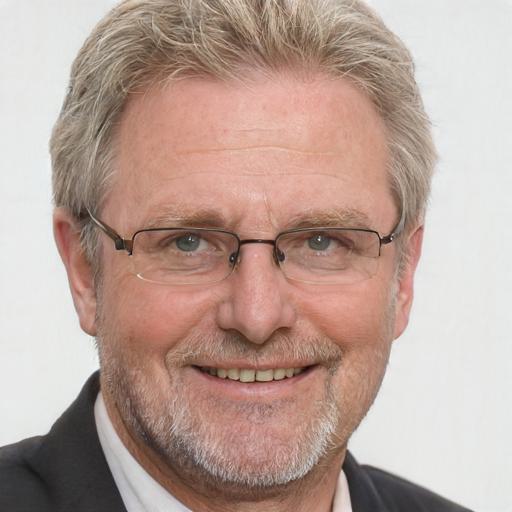} &
         \includegraphics[width=\simwidth\linewidth]{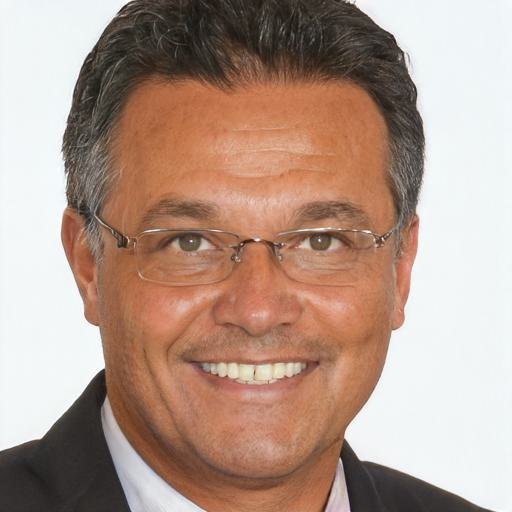} 
        \end{tabular}

    \caption{Multi-directional race editing of real images. Each row displays different edits in the associated subspace.}
    \label{fig:fig777}
\end{figure}

\section{Comparison to StyleFlow }
We compare our diverse edits results to StyleFlow as it outperforms all other baselines. The results are shown in \cref{fig:fig2}.
Different edits of our method are shown in Columns (a)-(c). We can observe the attributes manifest in various ways with each edit. Looking at gender, we can see that (b) and (c) better preserve the original hairstyle, while (c) also control facial hair. Changes in age may be expressed in hair color, hairstyle and facial wrinkles. Additionally, when comparing our results, we notice better preservation of unmodified attributes. For example, in the first two rows, our images preserve the original smile better than StyleFlow.

\newcommand*{\sfigsize}{0.15}

\begin{figure}[h]
\centering
\setlength{\tabcolsep}{1pt}
\begin{tabular}{cccccc}
 Input & StyleFlow & (a) & (b) & (c) \\
 \includegraphics[width=\sfigsize\linewidth]{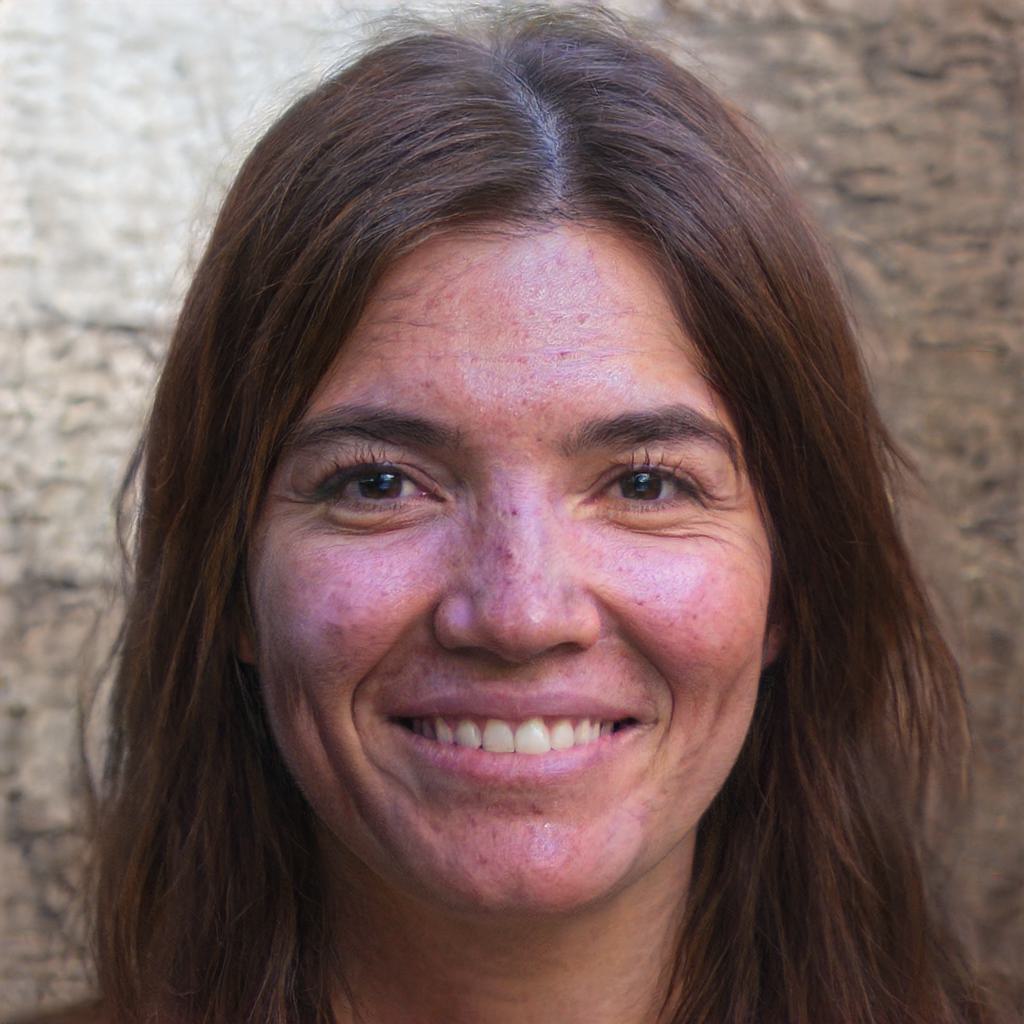} &
 \includegraphics[width=\sfigsize\linewidth]{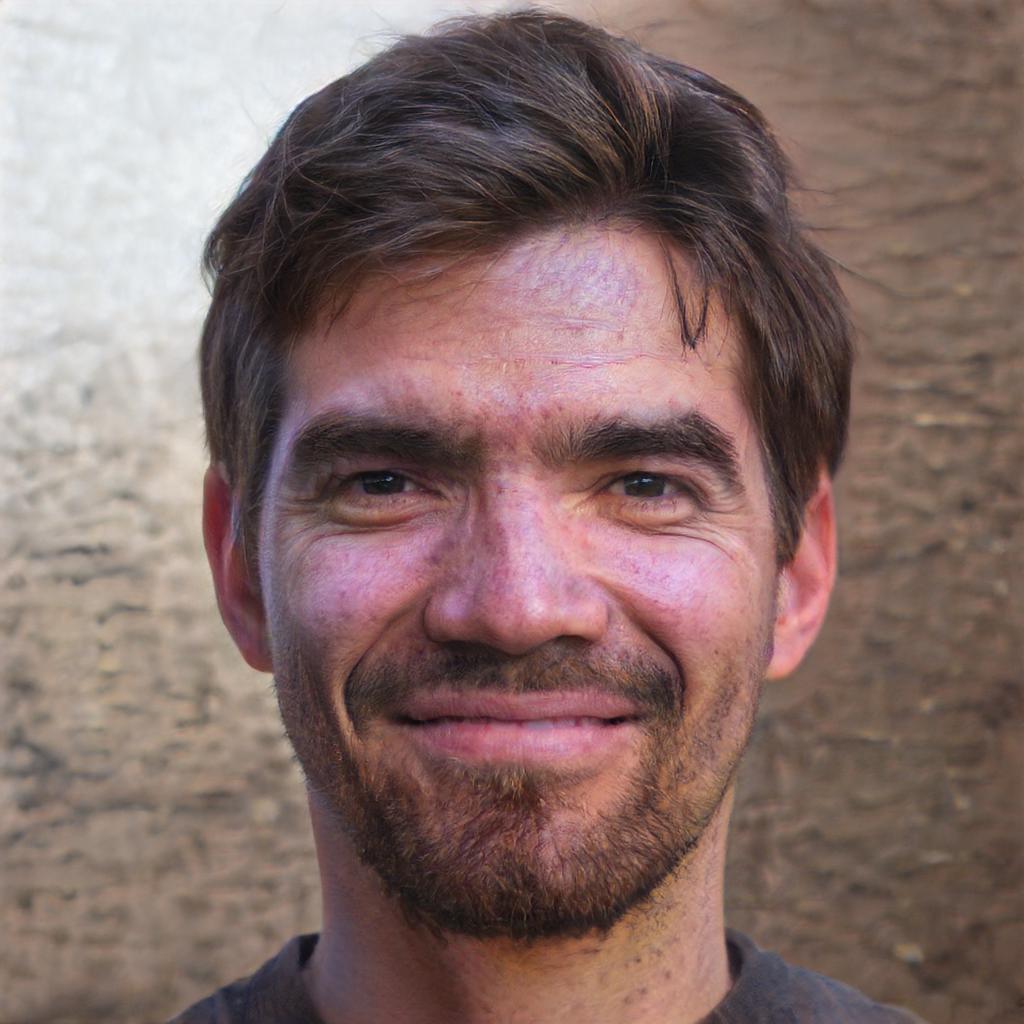} &
 \includegraphics[width=\sfigsize\linewidth]{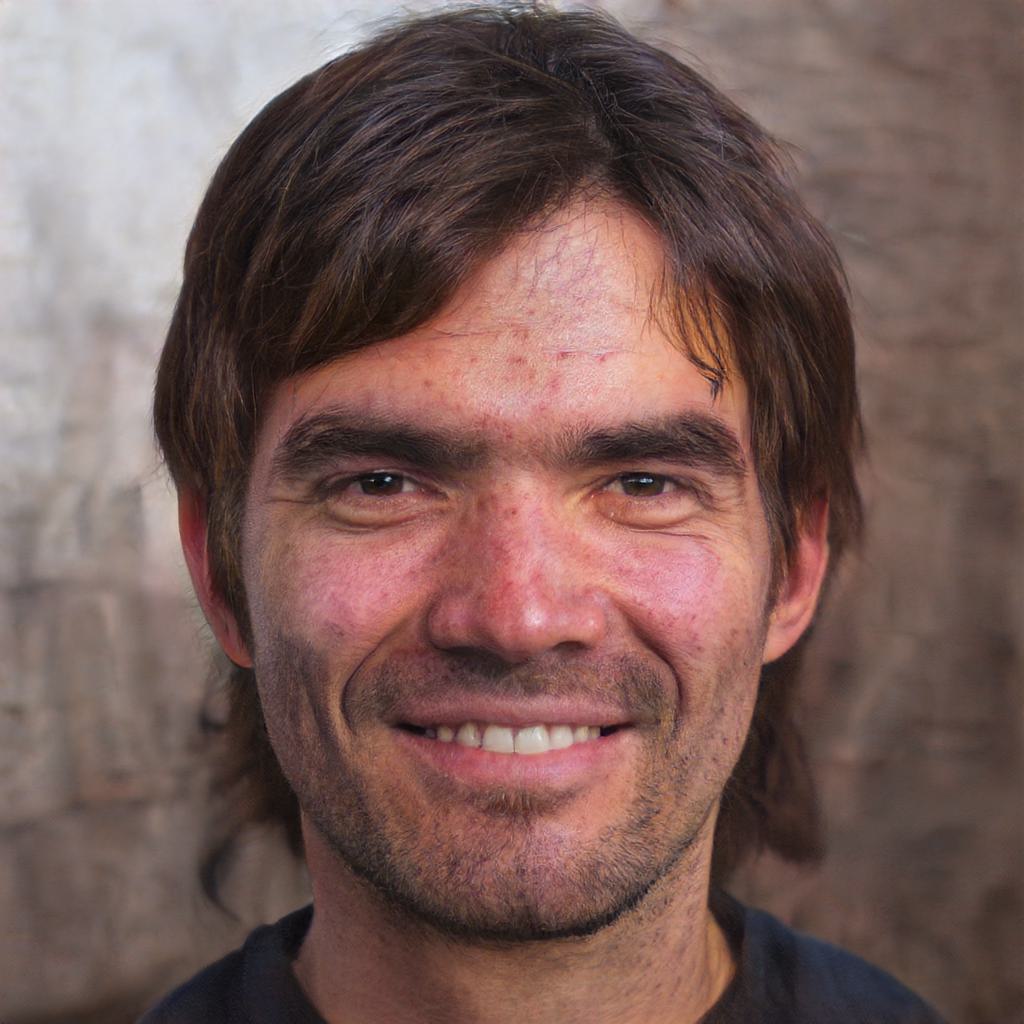} &
 \includegraphics[width=\sfigsize\linewidth]{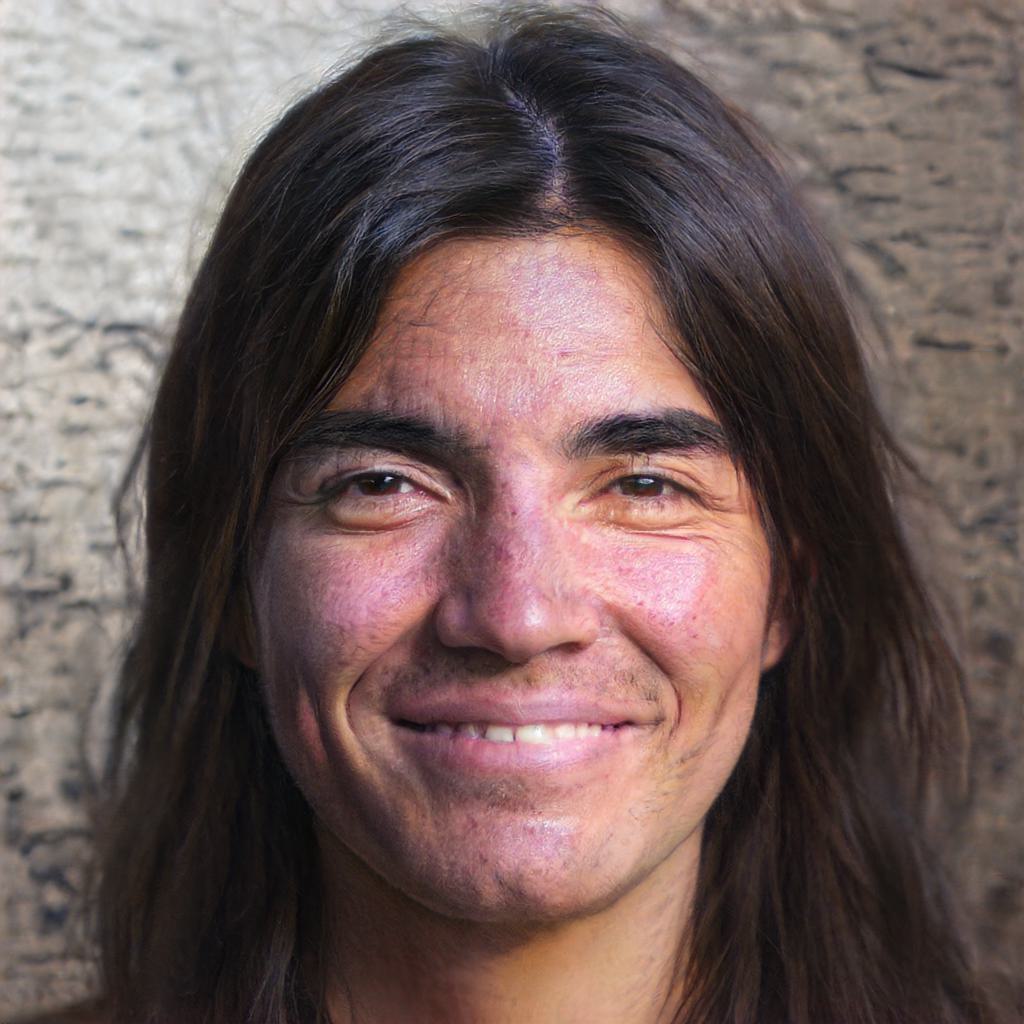} &
 \includegraphics[width=\sfigsize\linewidth]{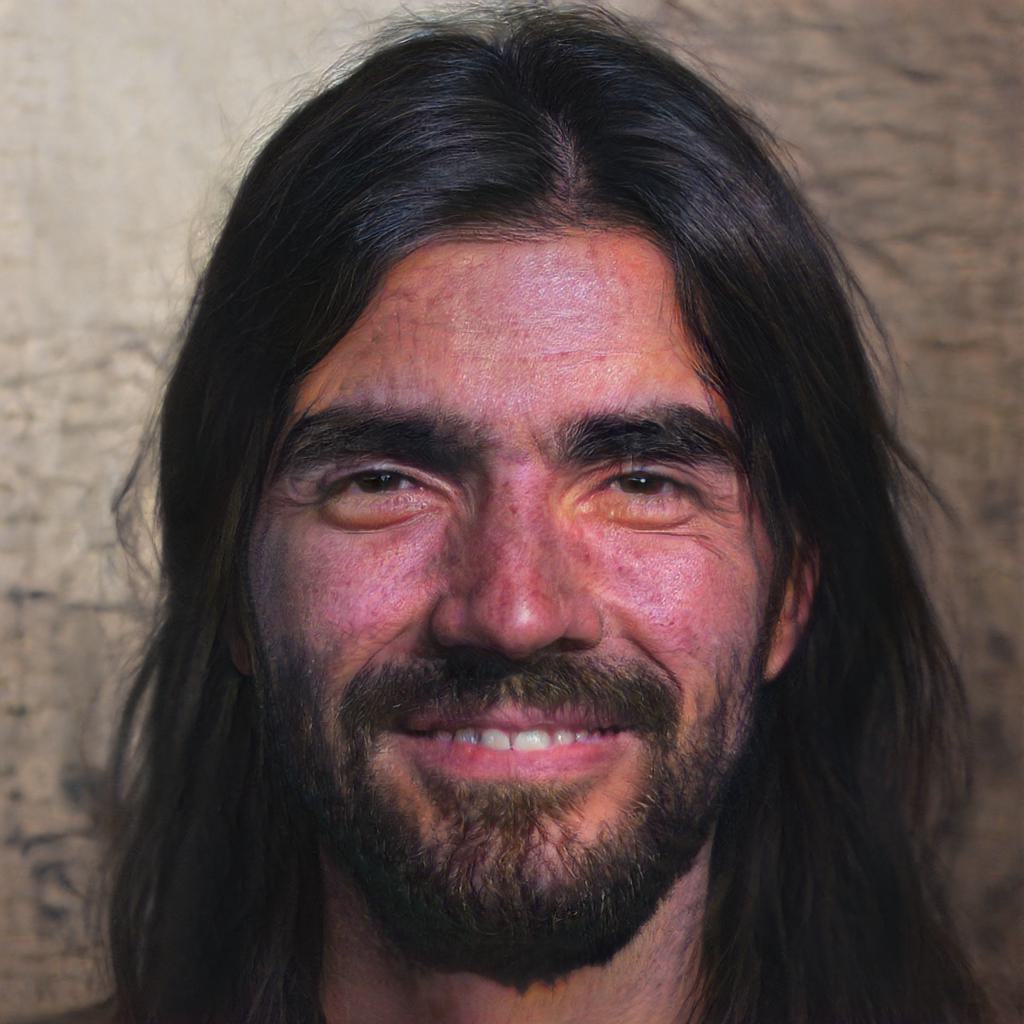} &
 \rotatebox[origin=c]{-90}{\hspace{-0.8in}Gender} \\
 
 & \includegraphics[width=\sfigsize\linewidth]{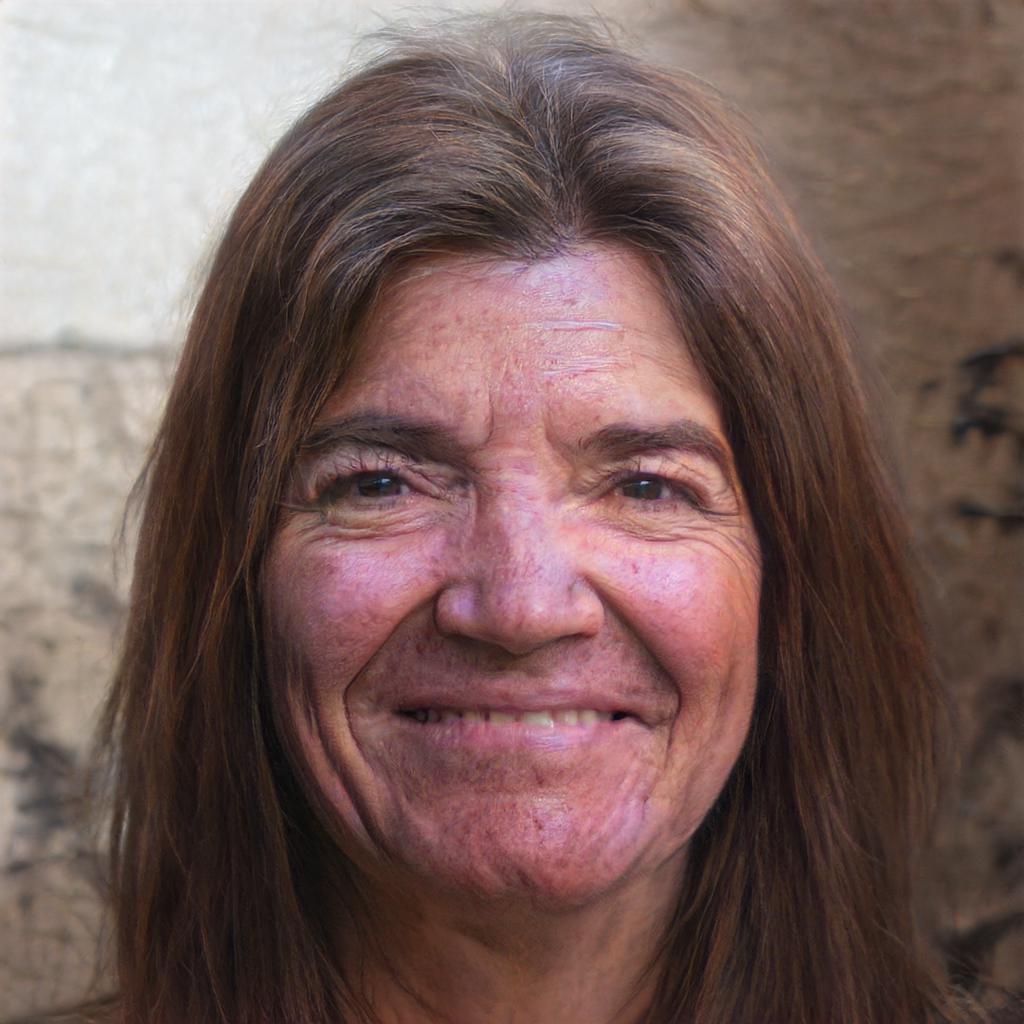} &
 \includegraphics[width=\sfigsize\linewidth]{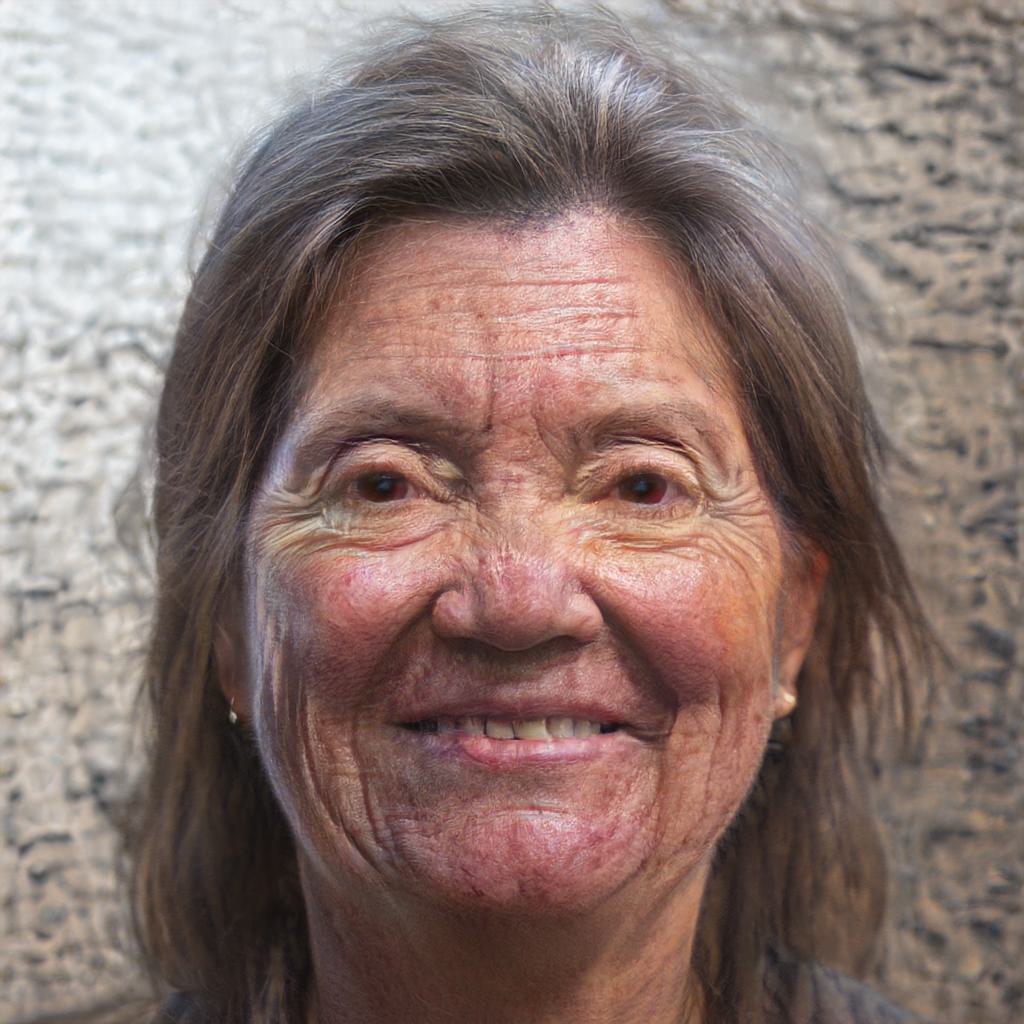} &
 \includegraphics[width=\sfigsize\linewidth]{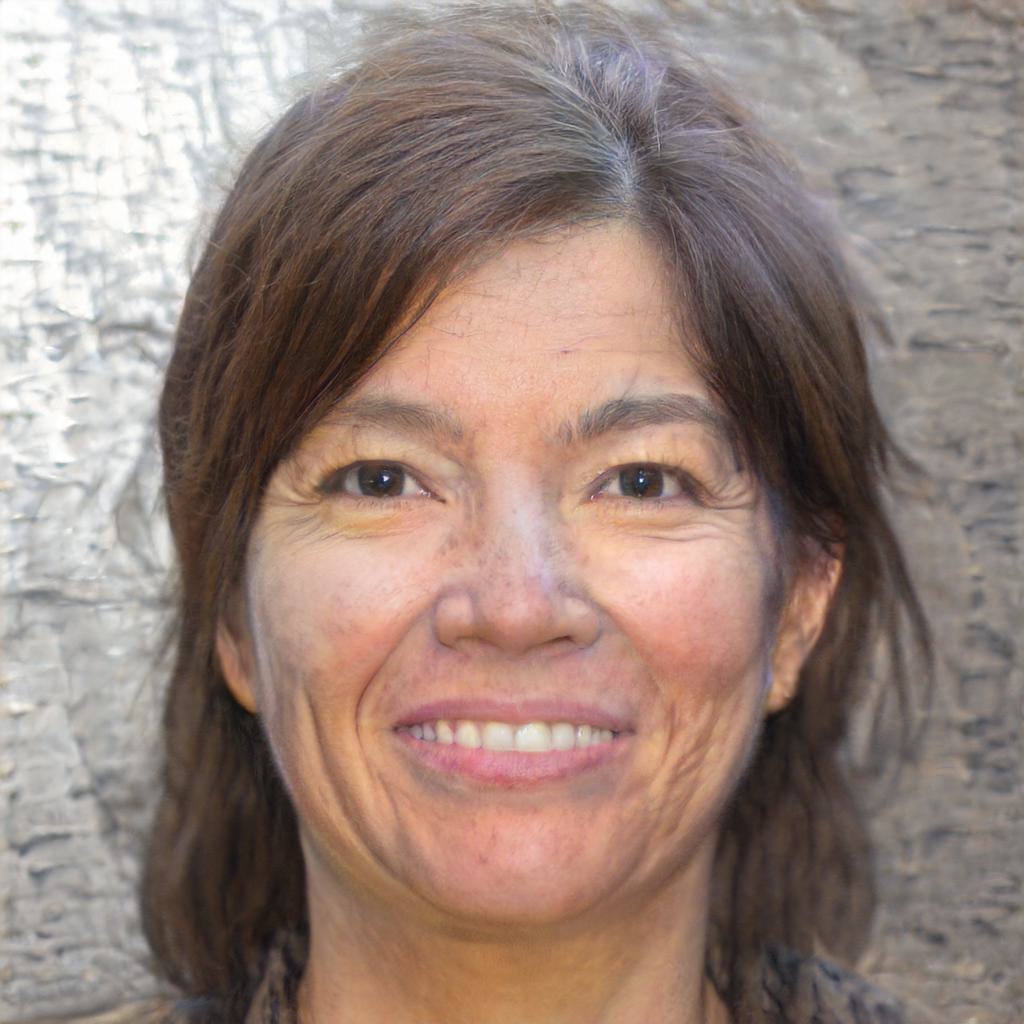} &
 \includegraphics[width=\sfigsize\linewidth]{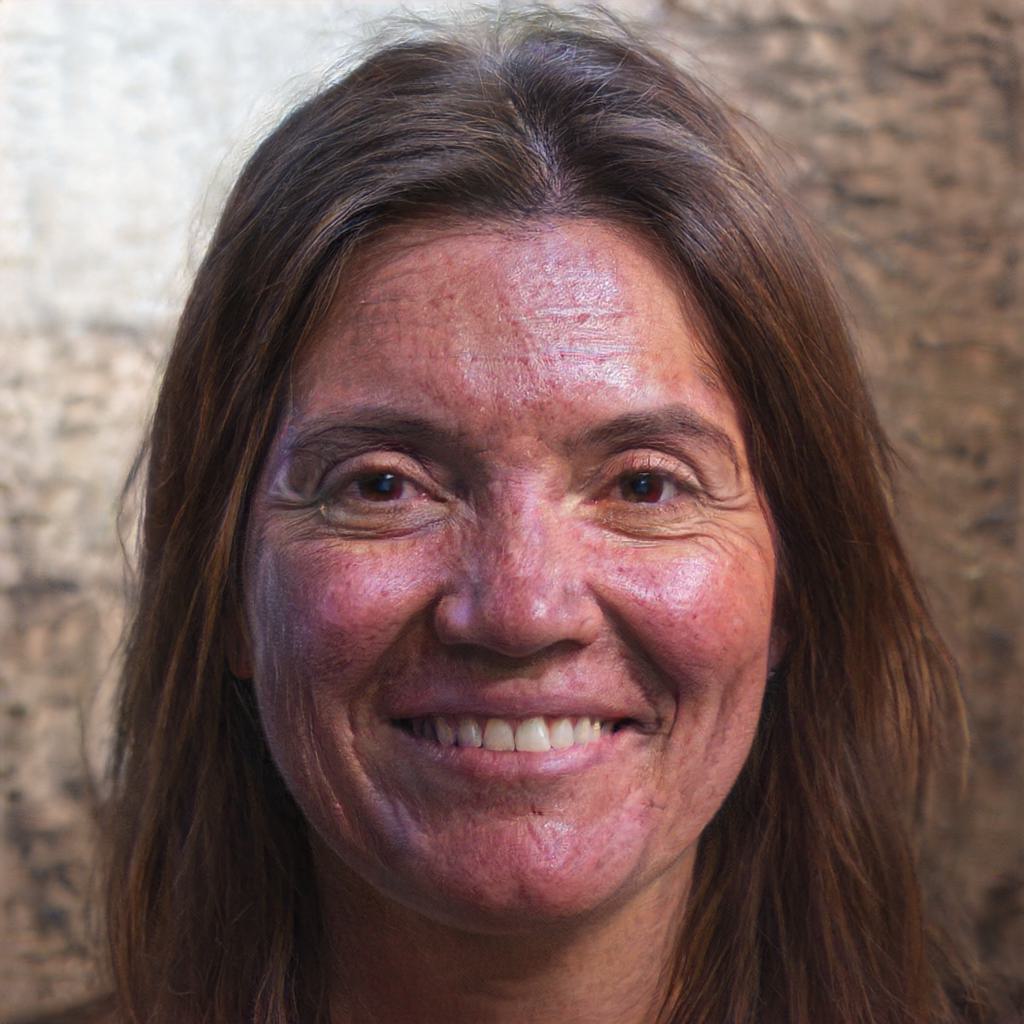} &
 \rotatebox[origin=c]{-90}{\hspace{-0.8in}Age} \\

 \includegraphics[width=\sfigsize\linewidth]{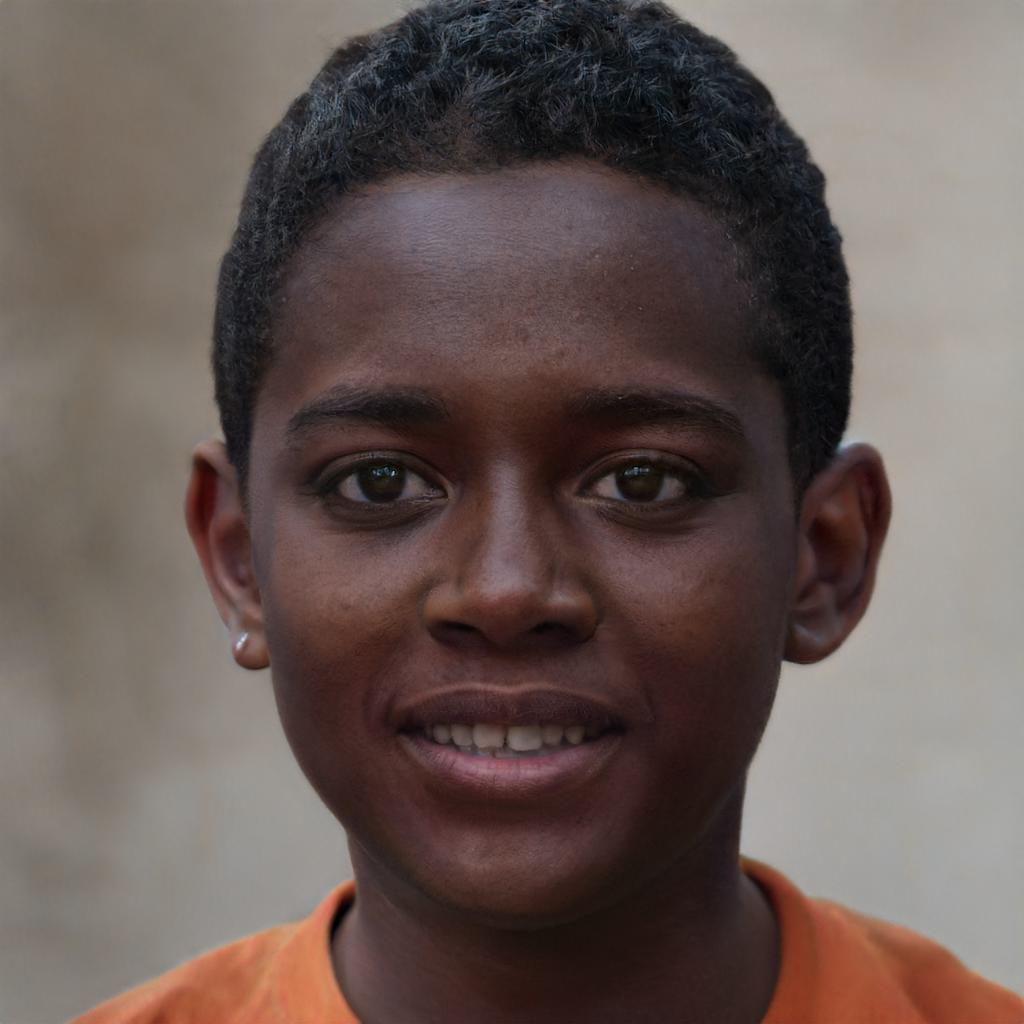} &
  \includegraphics[width=\sfigsize\linewidth]{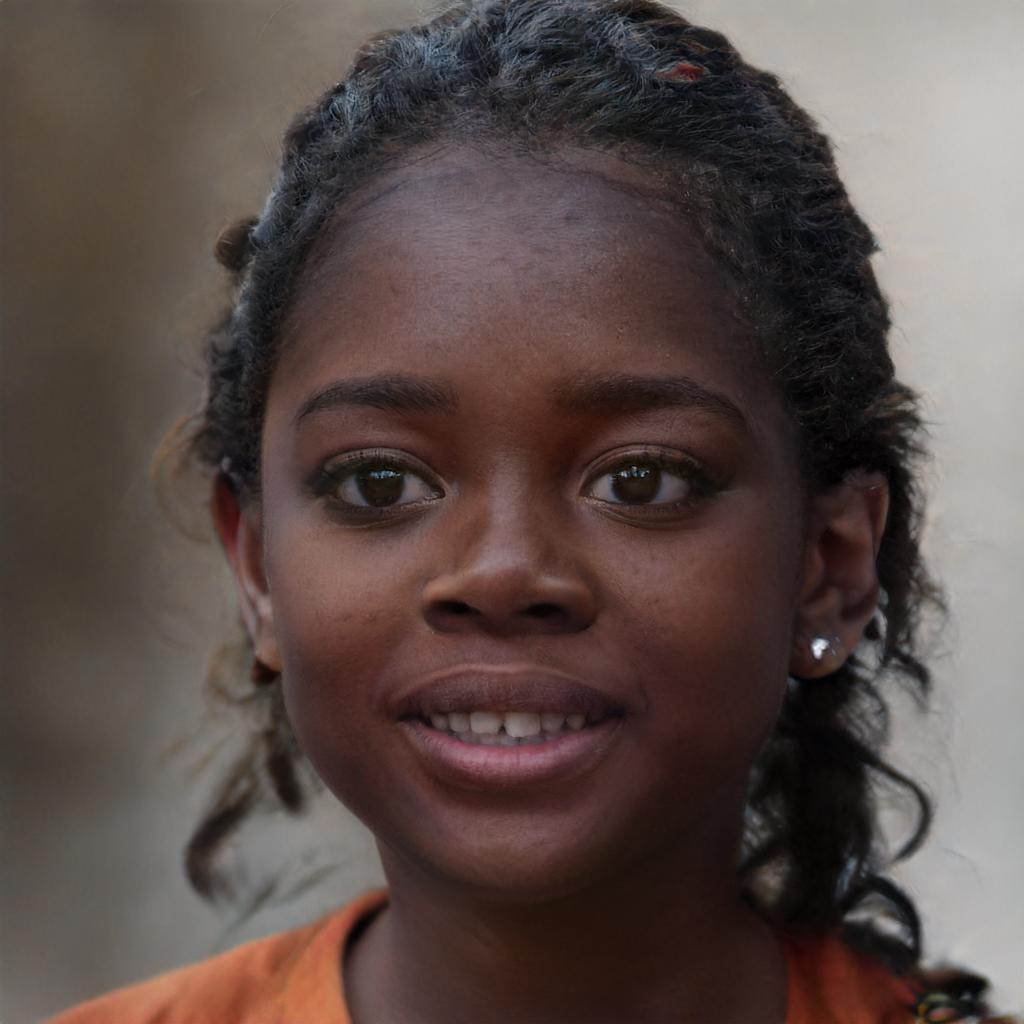} &
 \includegraphics[width=\sfigsize\linewidth]{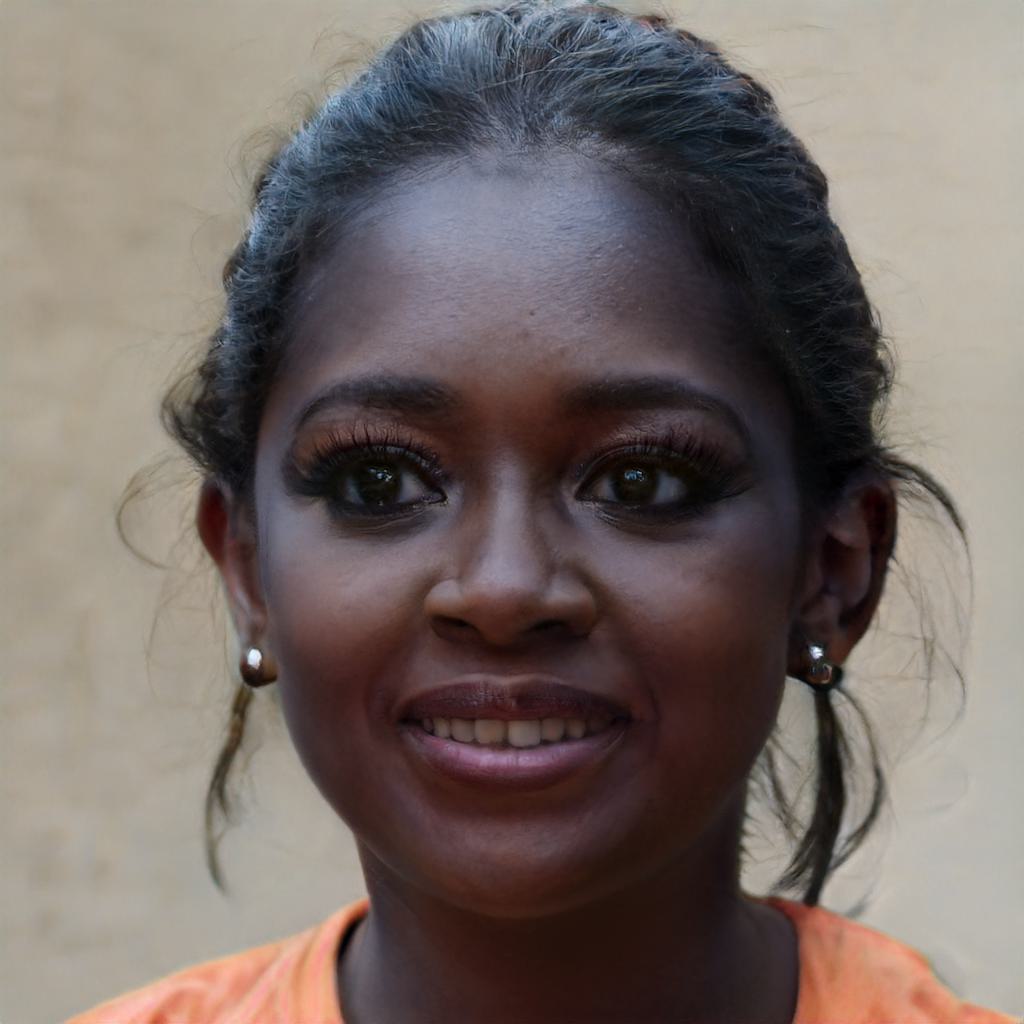} &
 \includegraphics[width=\sfigsize\linewidth]{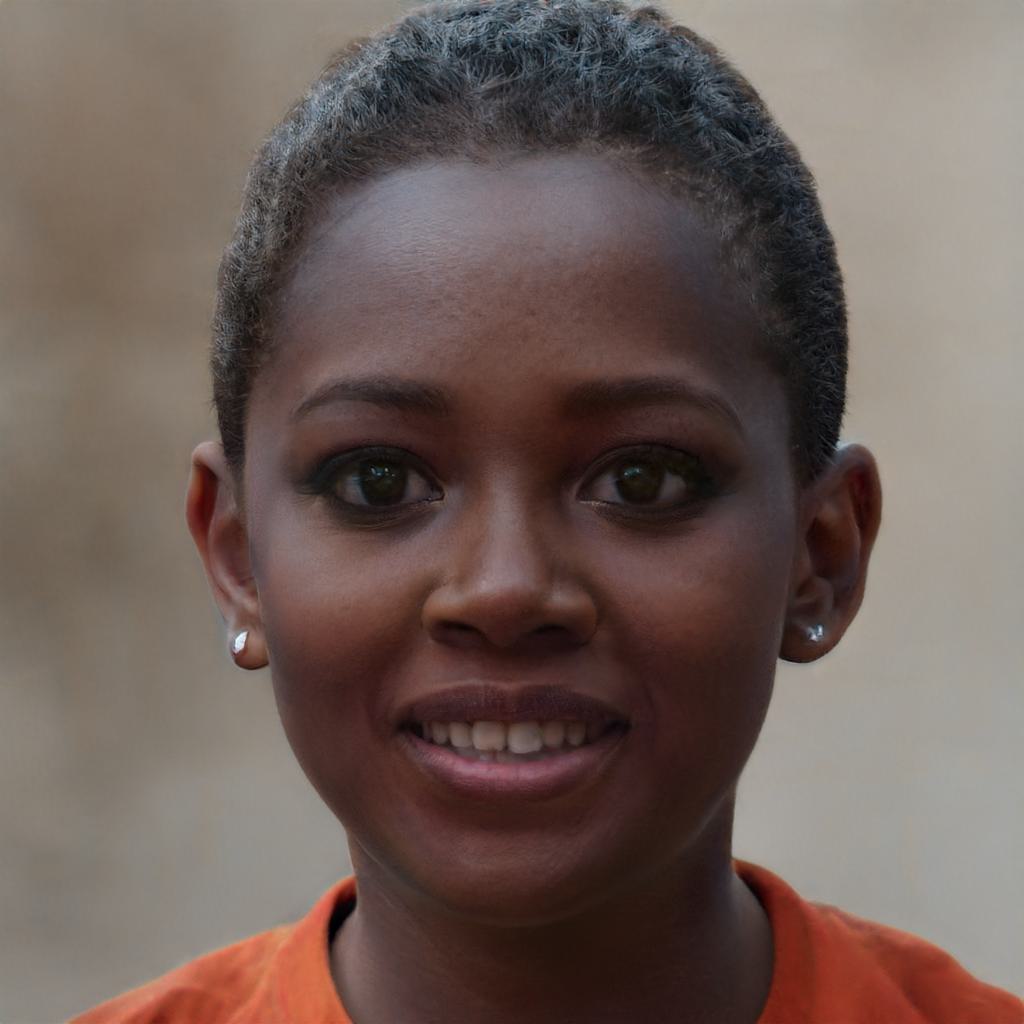} &
 \includegraphics[width=\sfigsize\linewidth]{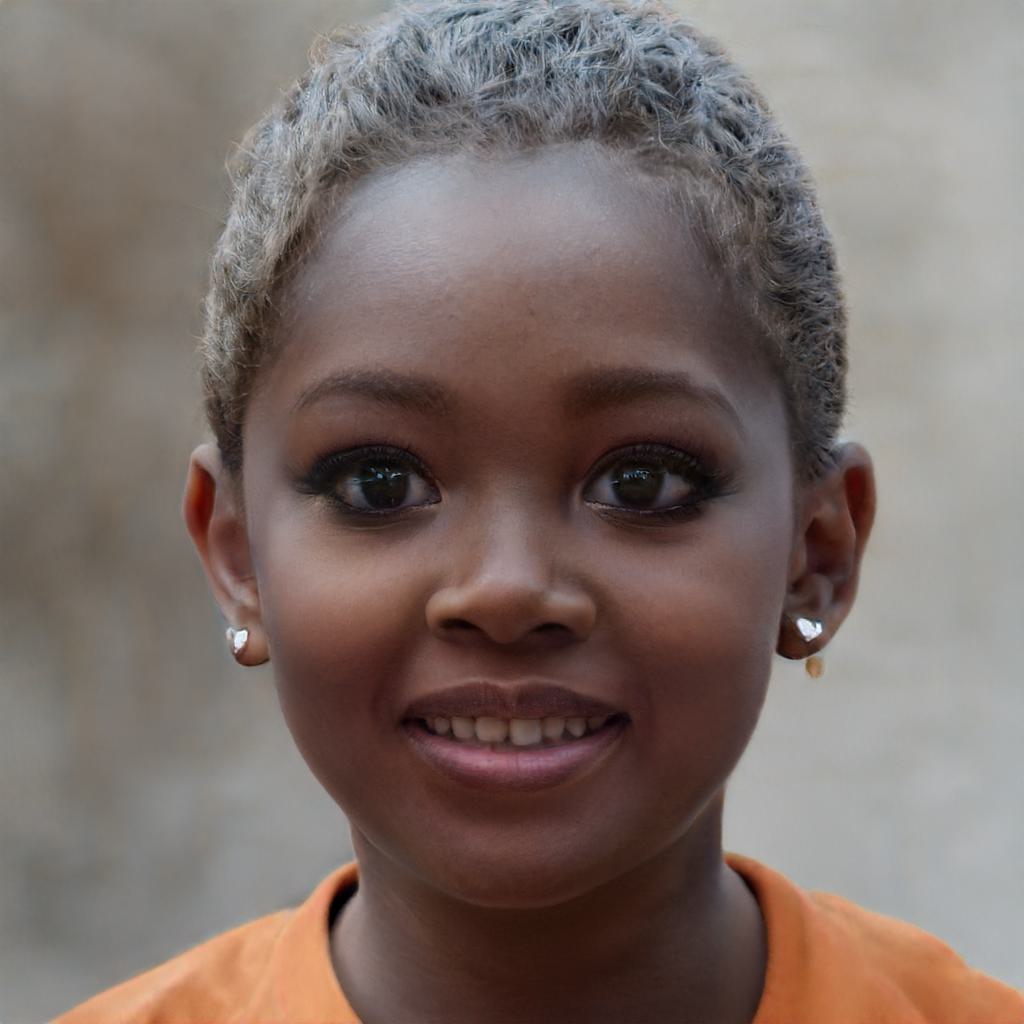} &
  \rotatebox[origin=c]{-90}{\hspace{-0.8in}Gender} \\

  & \includegraphics[width=\sfigsize\linewidth]{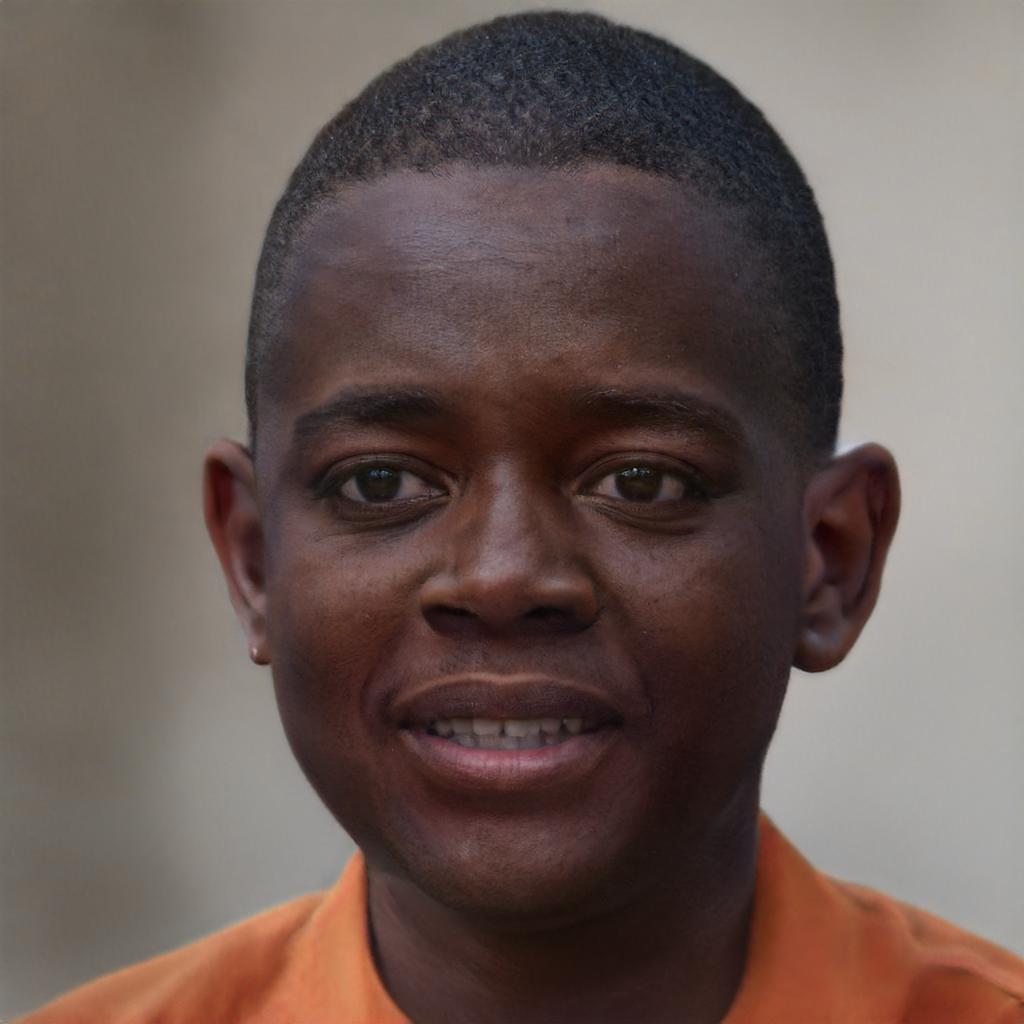} &
 \includegraphics[width=\sfigsize\linewidth]{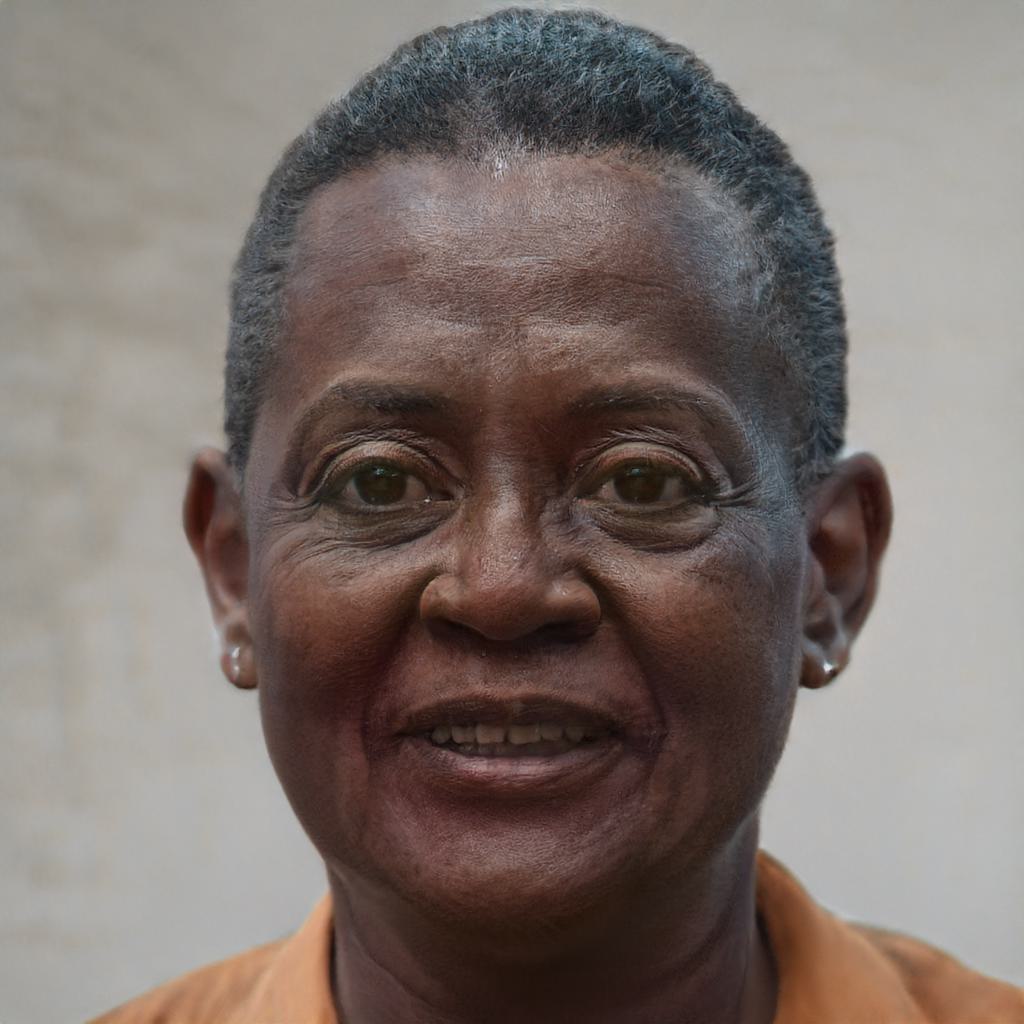} &
 \includegraphics[width=\sfigsize\linewidth]{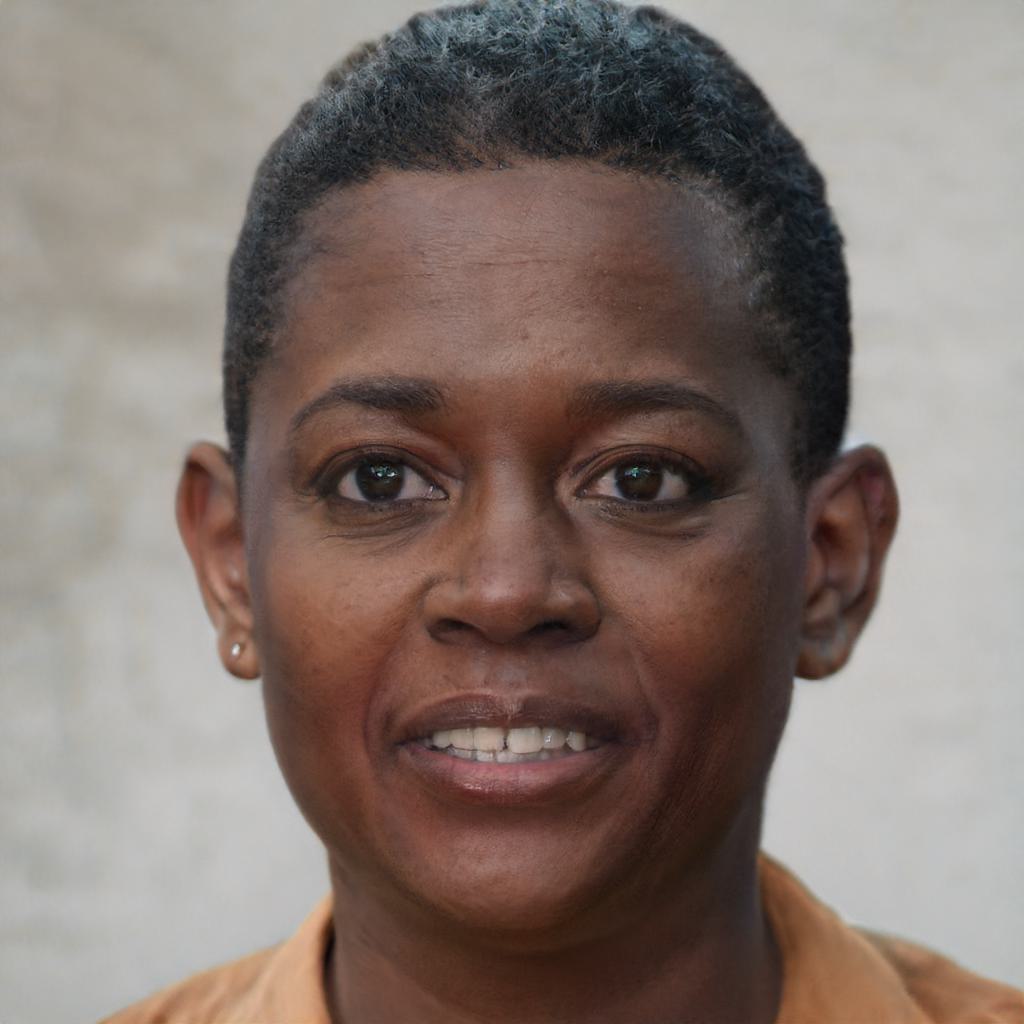} &
 \includegraphics[width=\sfigsize\linewidth]{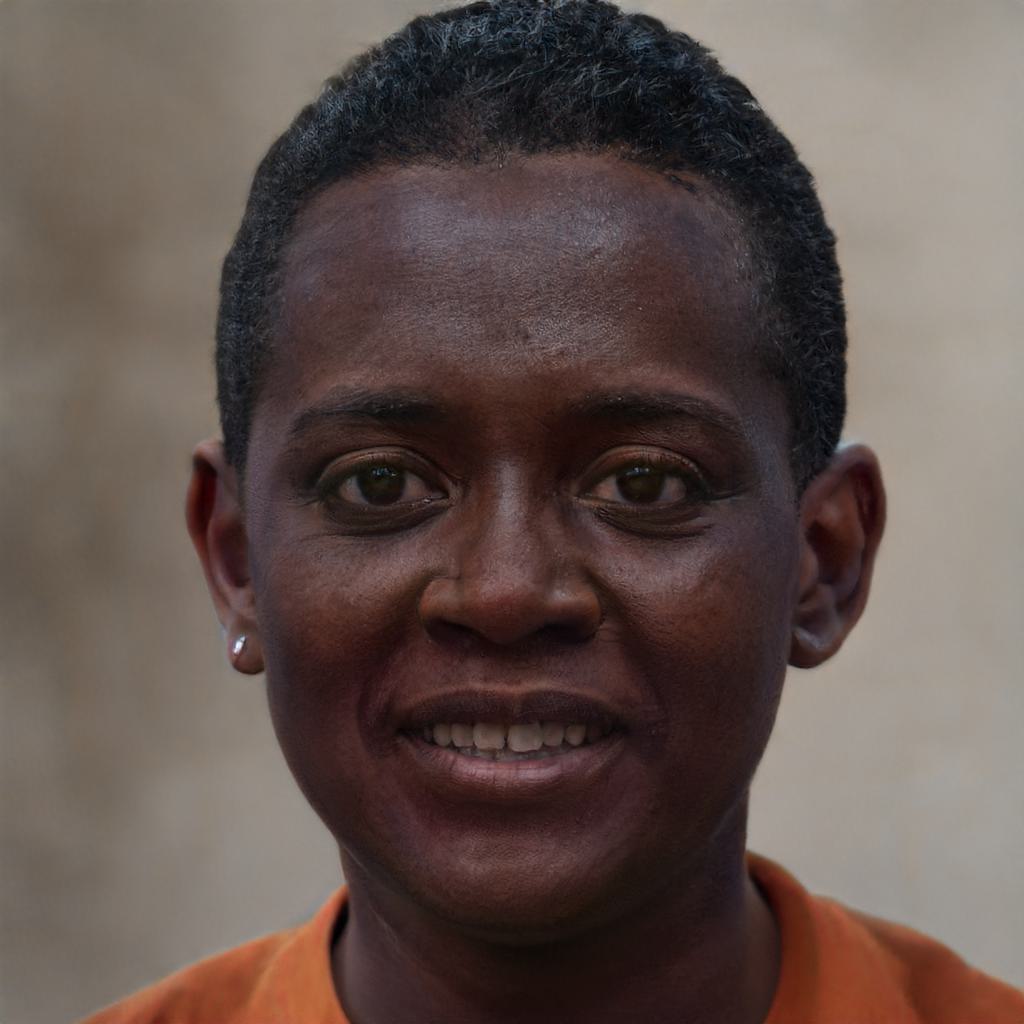} &
  \rotatebox[origin=c]{-90}{\hspace{-0.8in}Age} \\

 \includegraphics[width=\sfigsize\linewidth]{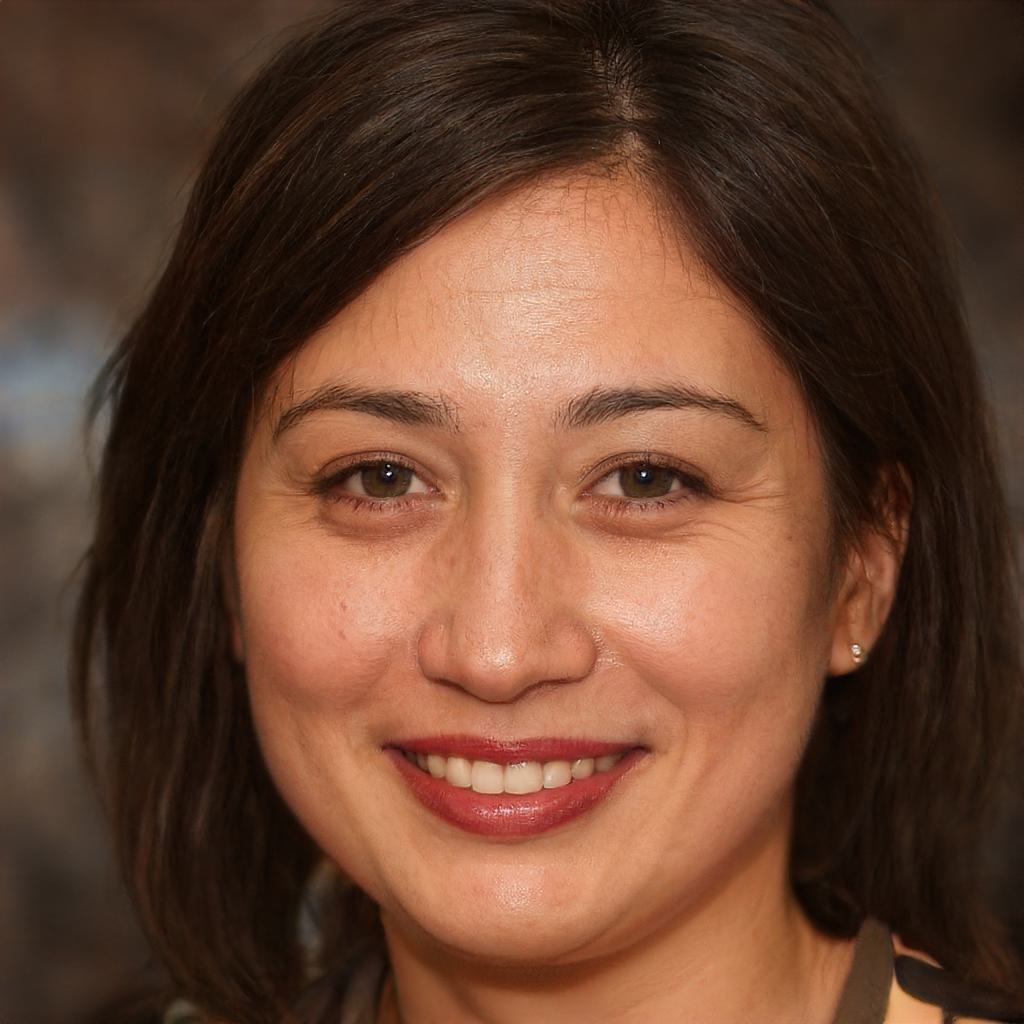} &
 \includegraphics[width=\sfigsize\linewidth]{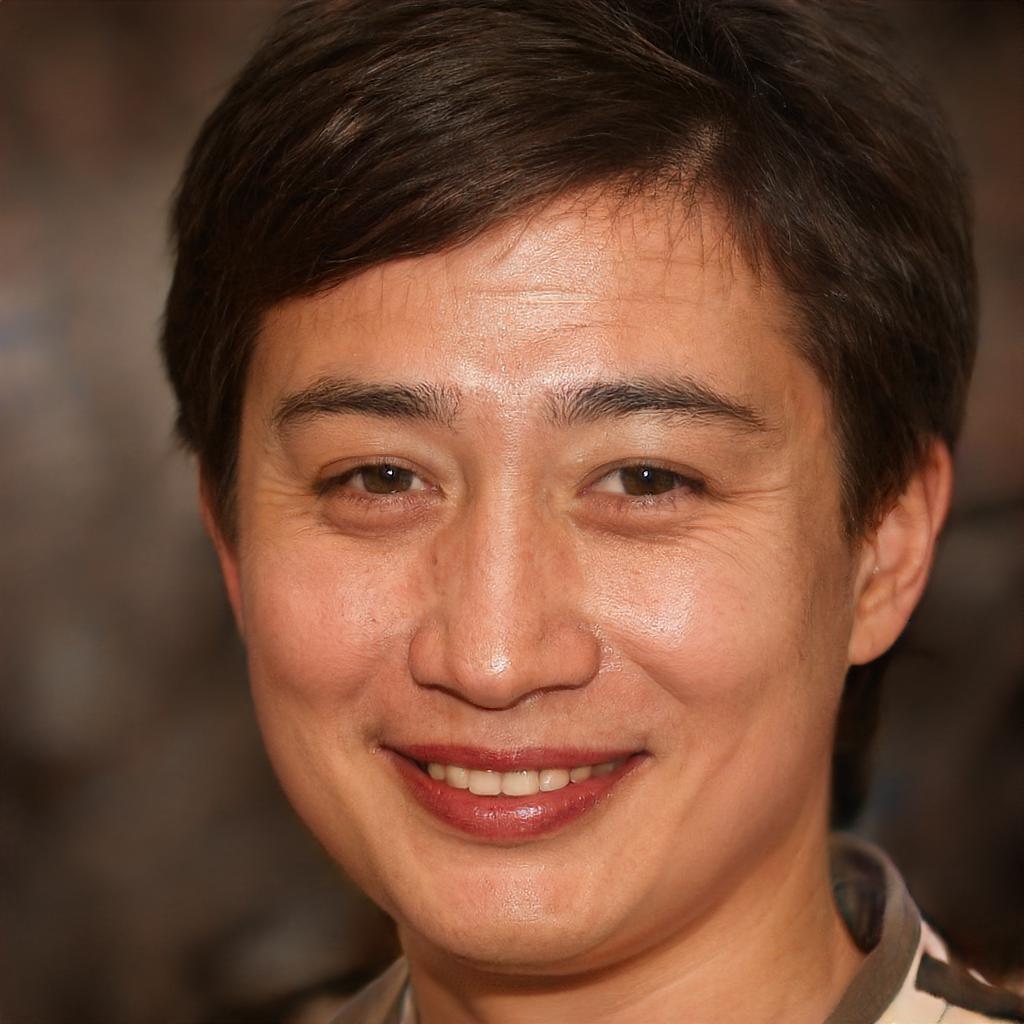} &
 \includegraphics[width=\sfigsize\linewidth]{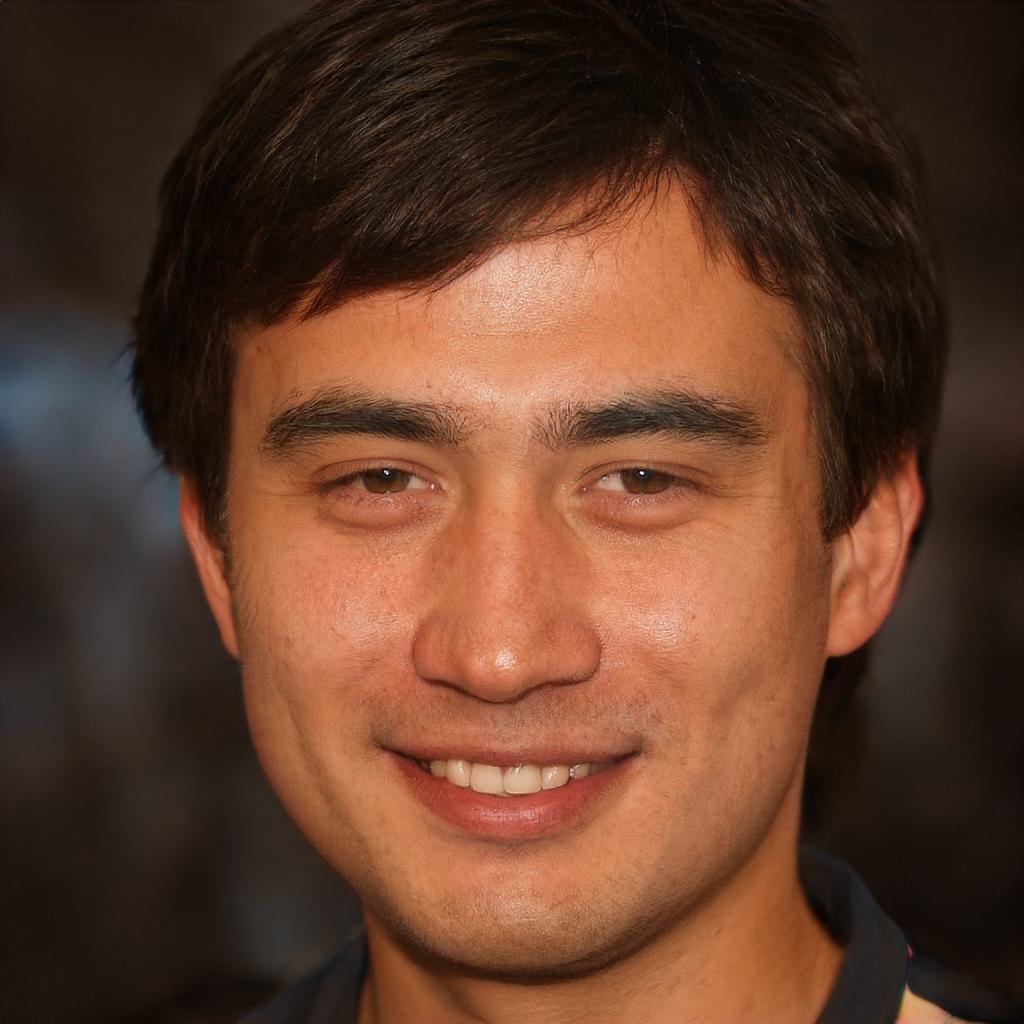} &
 \includegraphics[width=\sfigsize\linewidth]{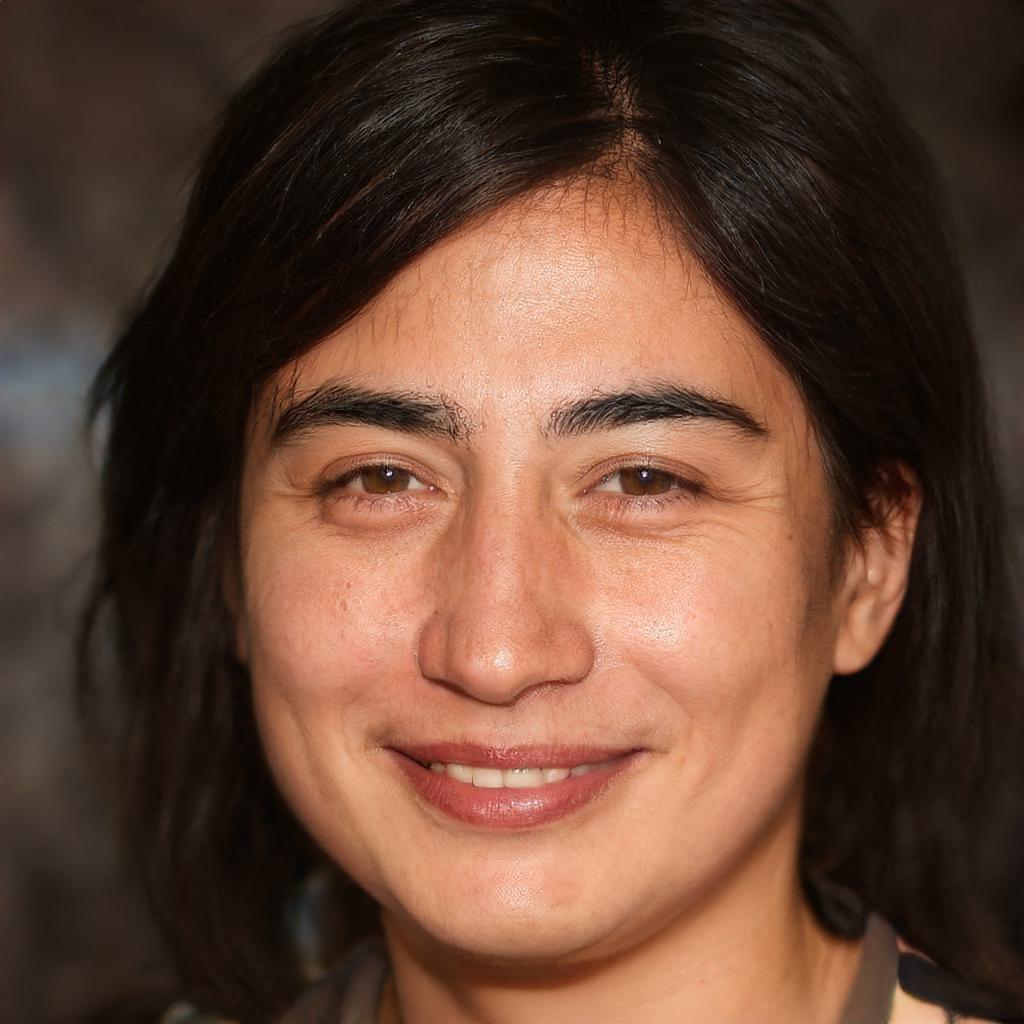} &
 \includegraphics[width=\sfigsize\linewidth]{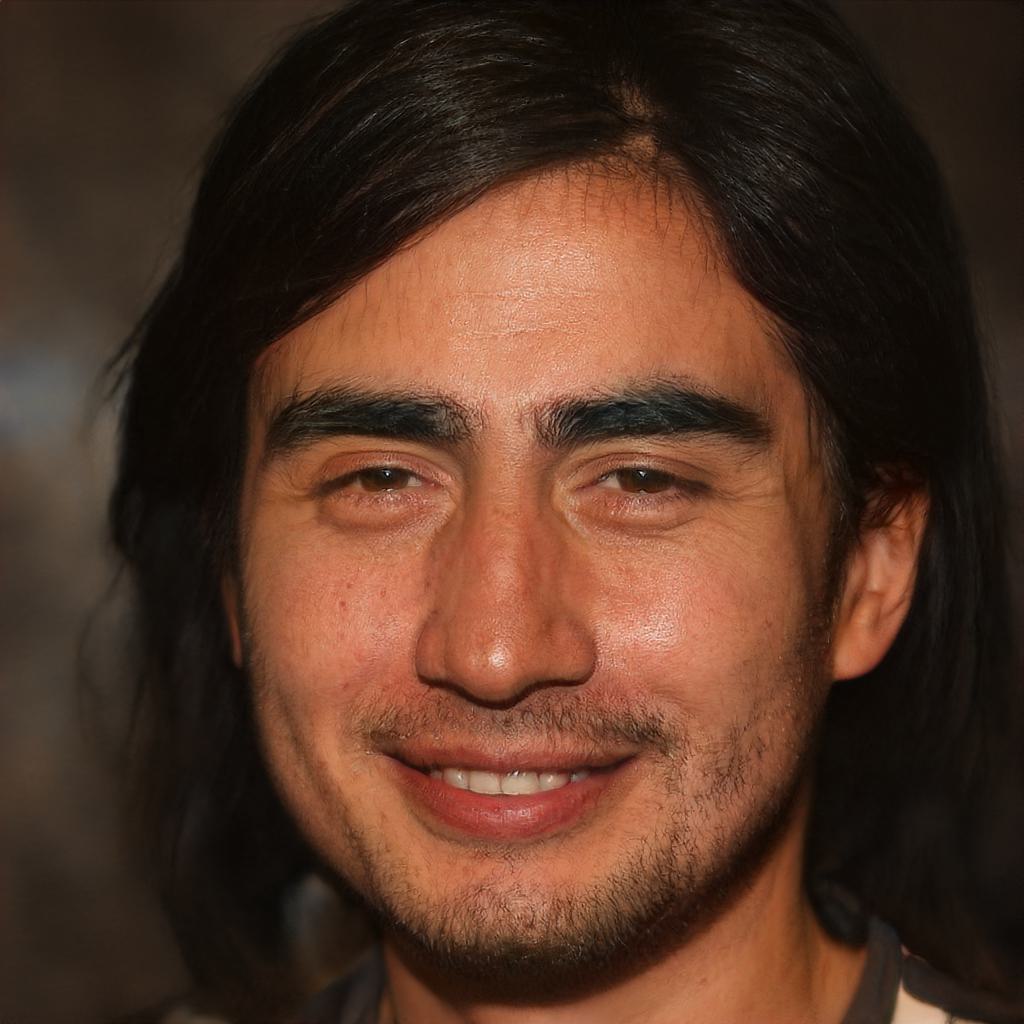} &
  \rotatebox[origin=c]{-90}{\hspace{-0.8in}Gender} \\
 
  & \includegraphics[width=\sfigsize\linewidth]{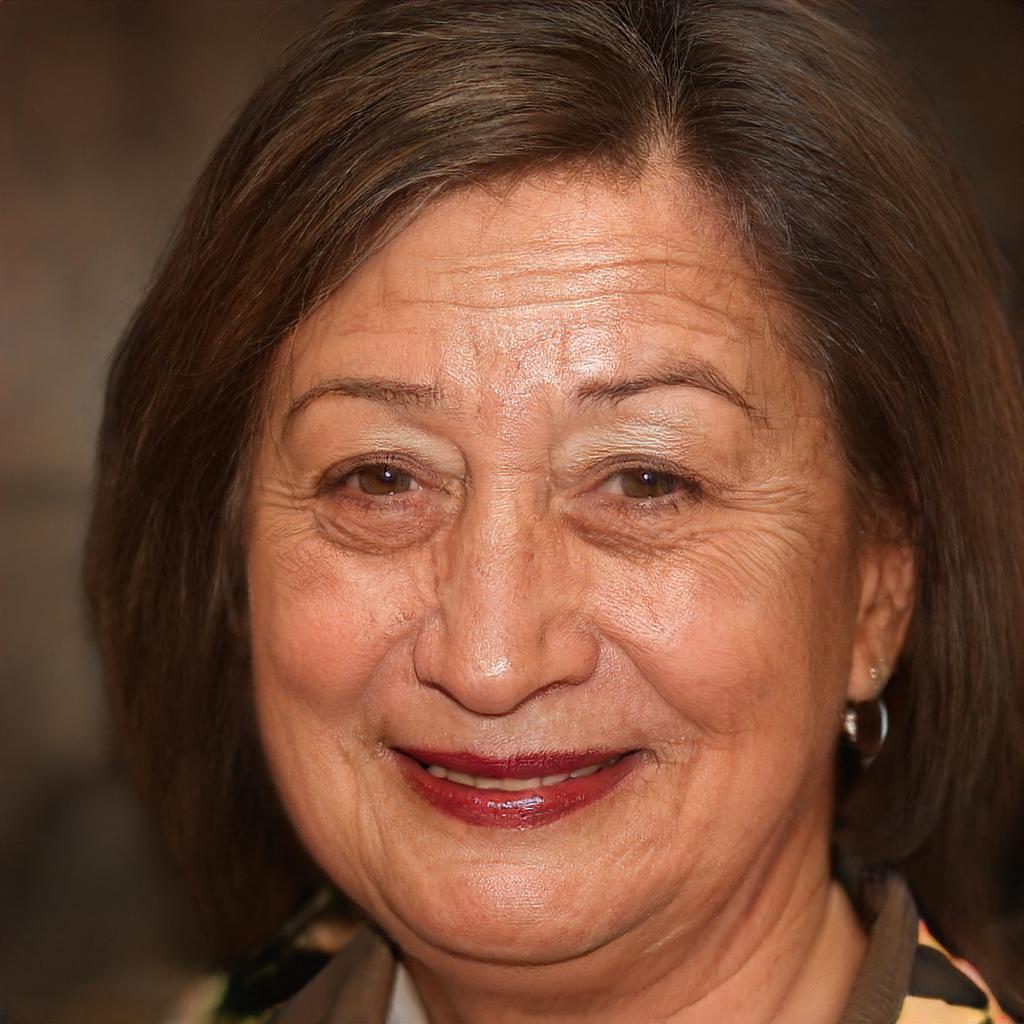} &
 \includegraphics[width=\sfigsize\linewidth]{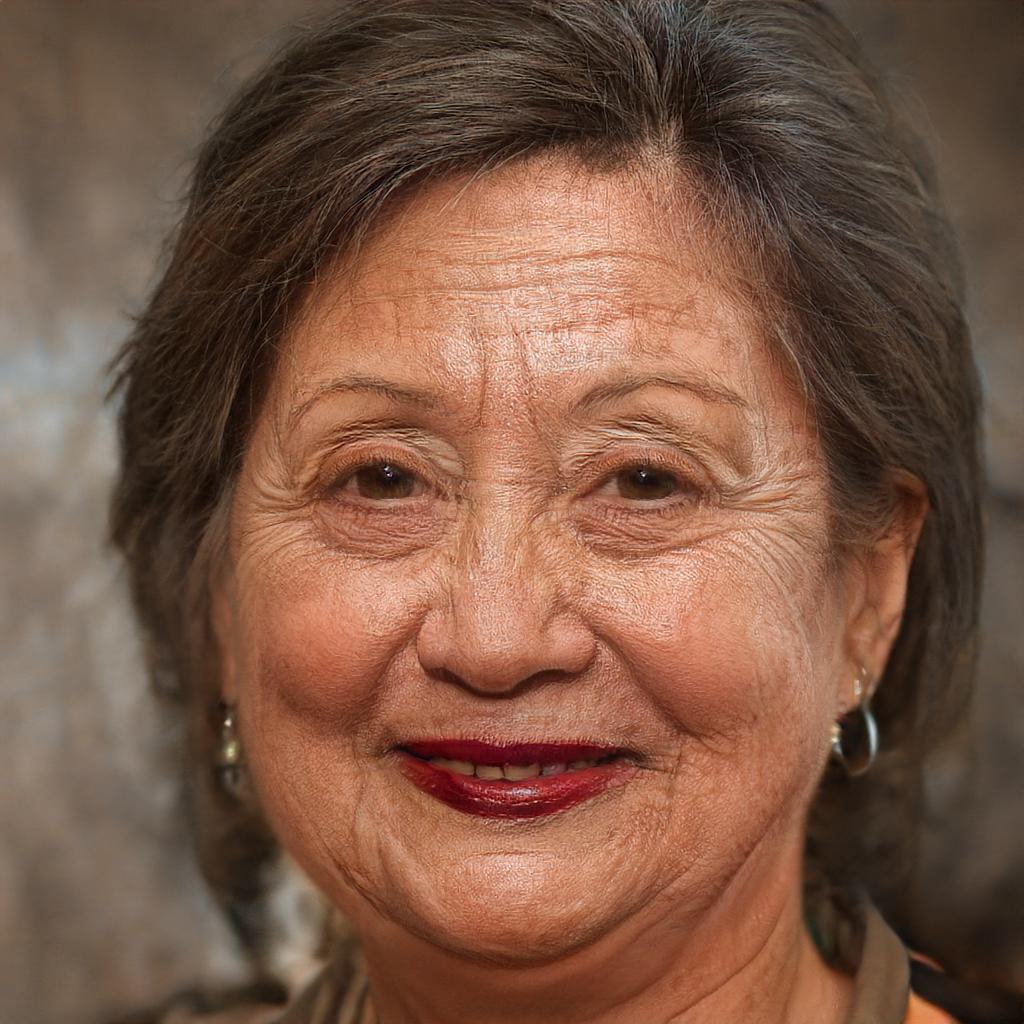} &
 \includegraphics[width=\sfigsize\linewidth]{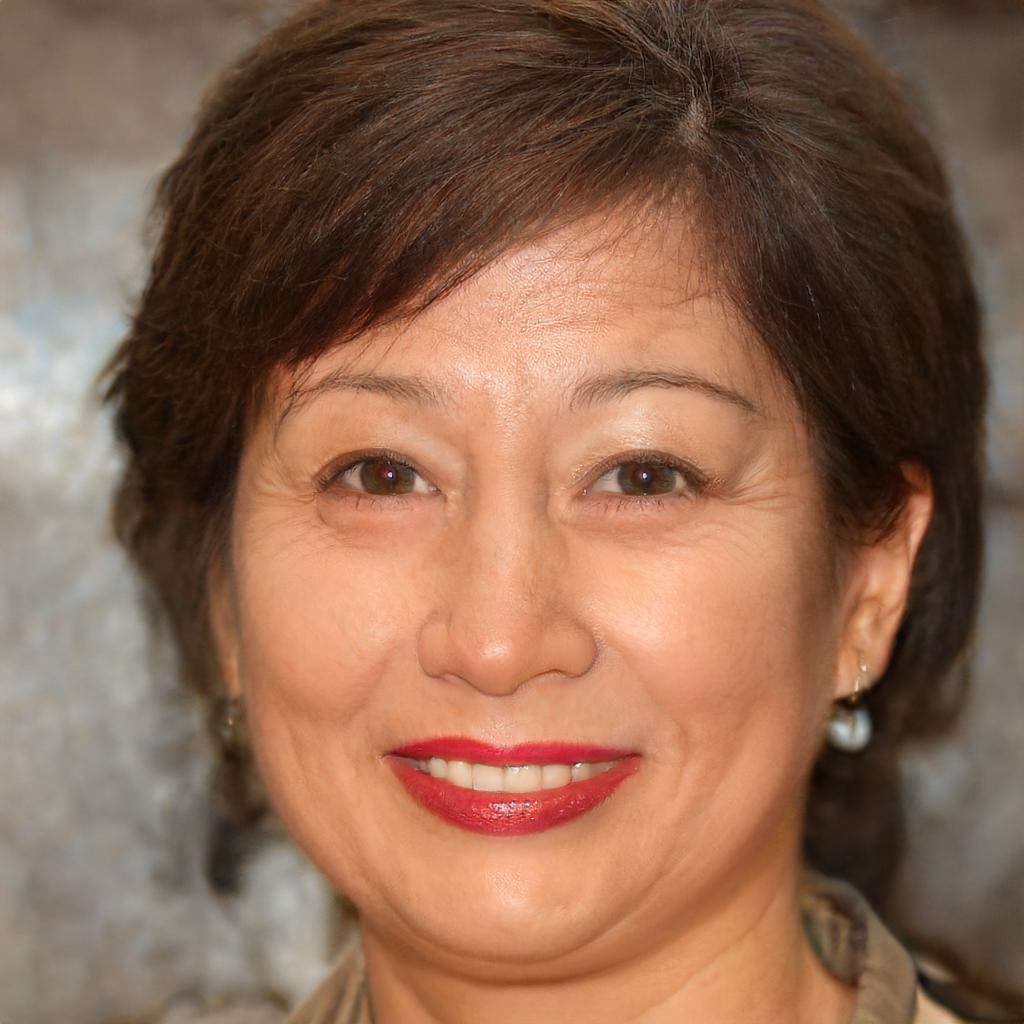} &
 \includegraphics[width=\sfigsize\linewidth]{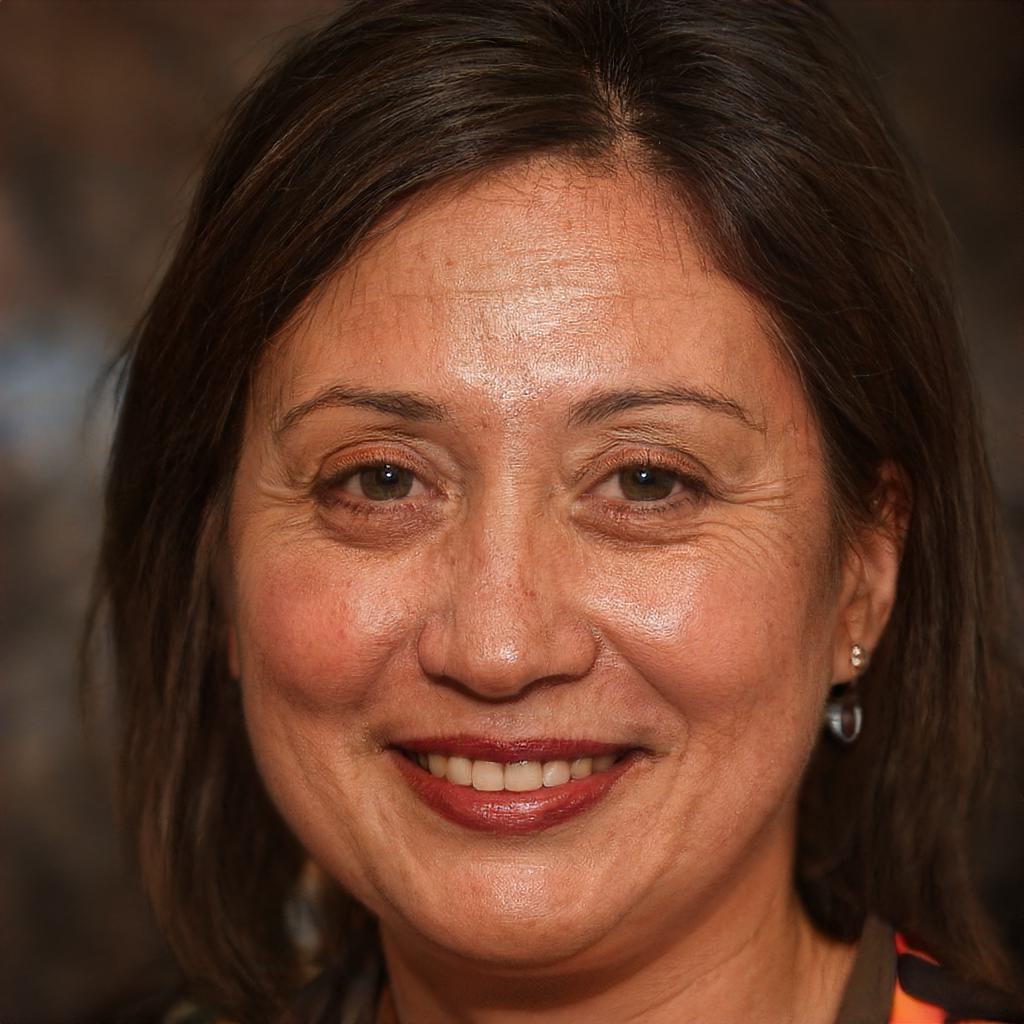} &
  \rotatebox[origin=c]{-90}{\hspace{-0.8in}Age} \\

\end{tabular}
\caption{Multi-directional editing of real images in full resolution (1024$\times$1024) and comparison to StyleFlow. Columns (a)-(c) represent edits in different directions inside the corresponding subspace (gender and age), allowing for rich and diverse edits.}
\label{fig:fig2}
\end{figure}

\newcommand*{\figs}{0.17}
\section{Outside the Domain Editing}
Here we display additional editing results of images outside the FFHQ dataset. These include paintings and old photography. Our model can generalize and perform diverse editing on images outside the training set. The results can be seen in the figure below.

\begin{figure}[h]
\centering
\setlength{\tabcolsep}{1pt}
\begin{tabular}{c|cccccc}
 Input & \multicolumn{4}{c}{Multi-Directional Edits} \\
 \includegraphics[width=\figs\linewidth]{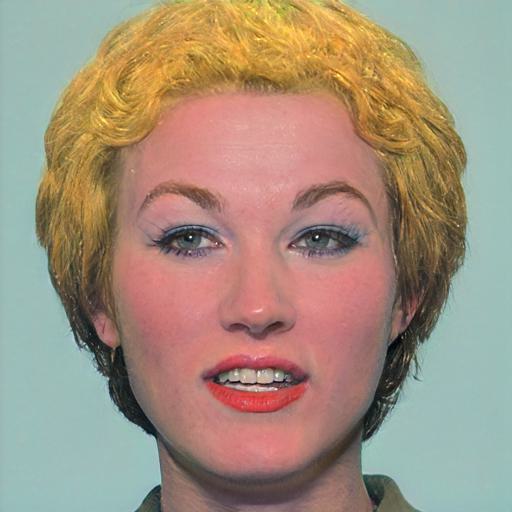} &
  \includegraphics[width=\figs\linewidth]{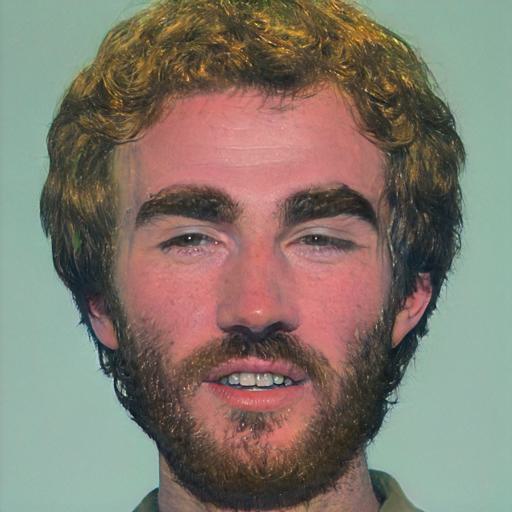} &
 \includegraphics[width=\figs\linewidth]{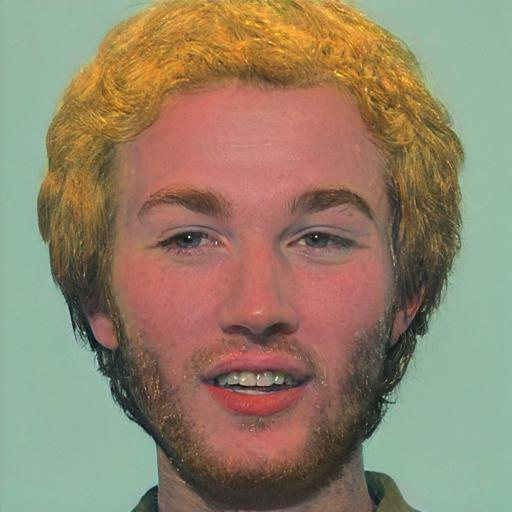} &
 \includegraphics[width=\figs\linewidth]{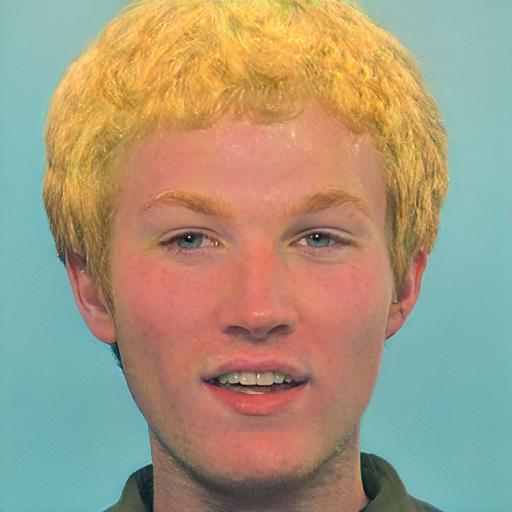} &
 \includegraphics[width=\figs\linewidth]{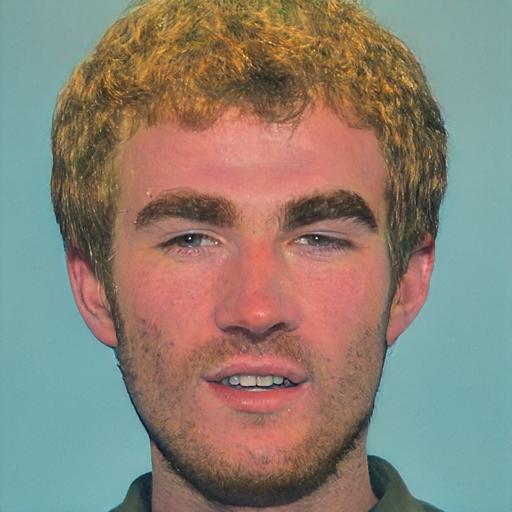} &
 \rotatebox[origin=c]{-90}{\hspace{-0.8in}Gender} \\

  \includegraphics[width=\figs\linewidth]{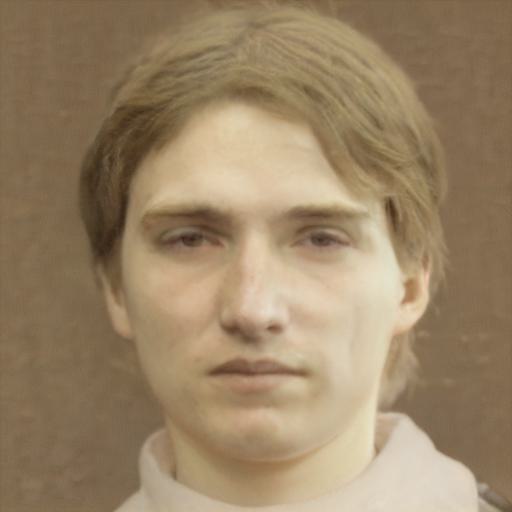} &
  \includegraphics[width=\figs\linewidth]{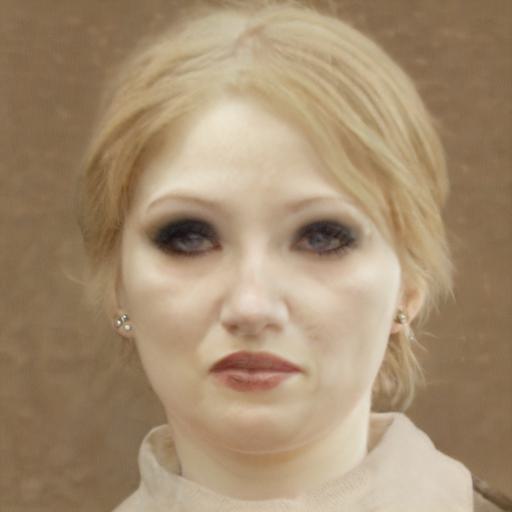} &
 \includegraphics[width=\figs\linewidth]{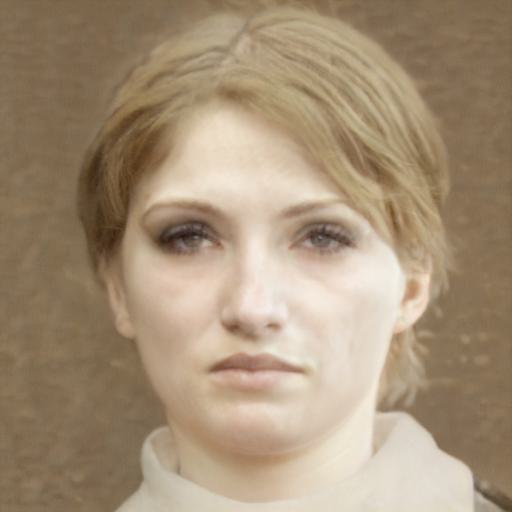} &
 \includegraphics[width=\figs\linewidth]{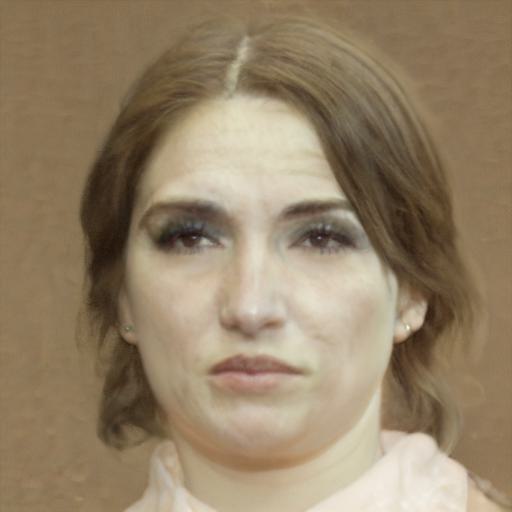} &
 \includegraphics[width=\figs\linewidth]{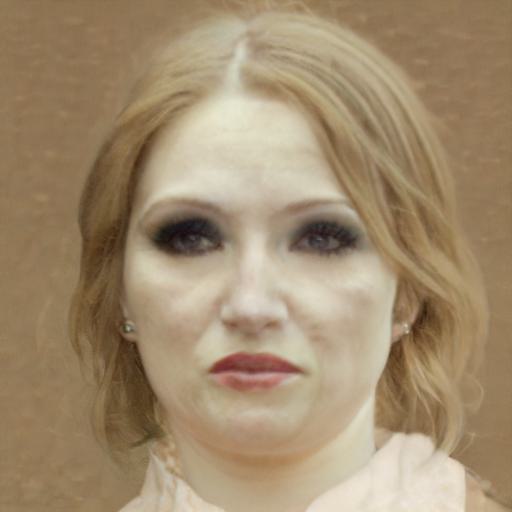} &
 \rotatebox[origin=c]{-90}{\hspace{-0.8in}Gender} \\

   \includegraphics[width=\figs\linewidth]{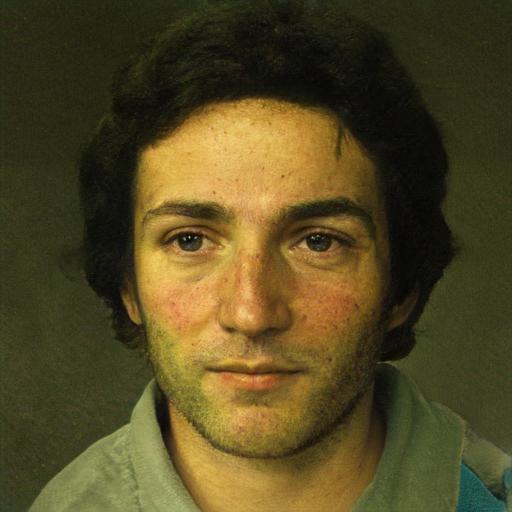} &
  \includegraphics[width=\figs\linewidth]{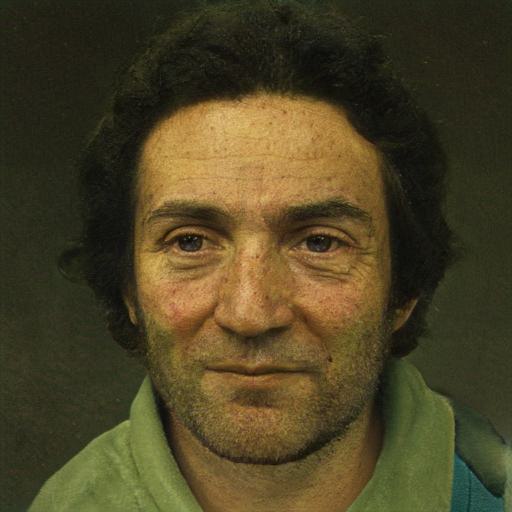} &
 \includegraphics[width=\figs\linewidth]{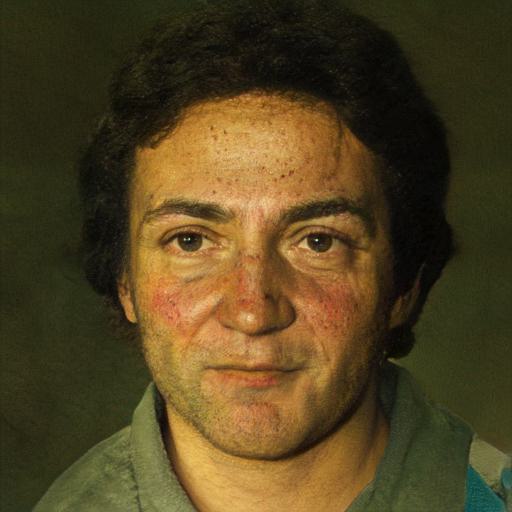} &
 \includegraphics[width=\figs\linewidth]{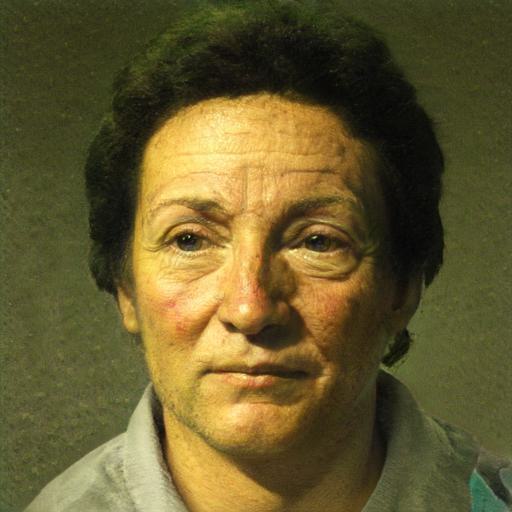} &
 \includegraphics[width=\figs\linewidth]{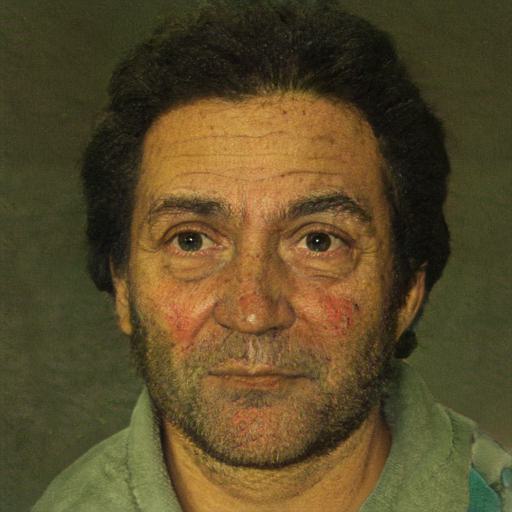} &
 \rotatebox[origin=c]{-90}{\hspace{-0.8in}Age} \\
 
   \includegraphics[width=\figs\linewidth]{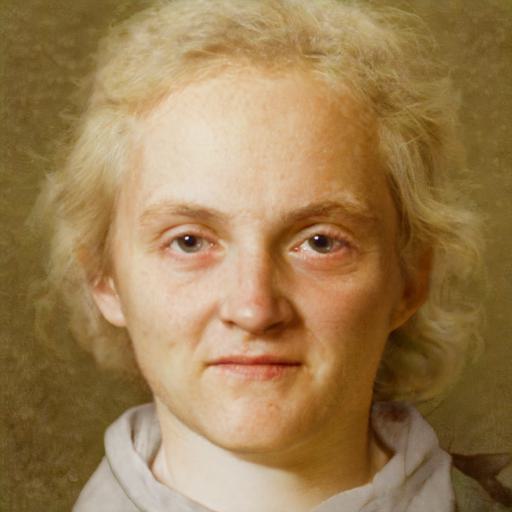} &
  \includegraphics[width=\figs\linewidth]{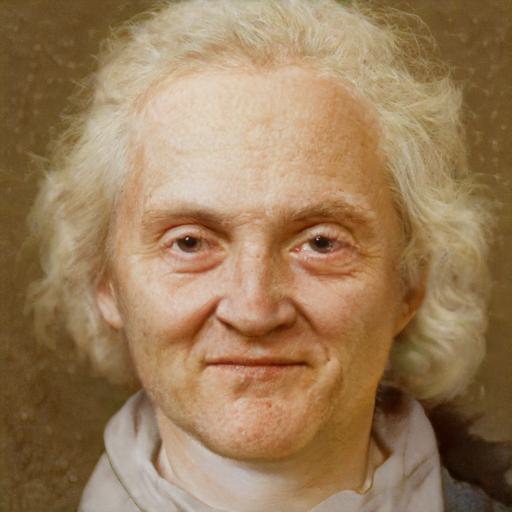} &
 \includegraphics[width=\figs\linewidth]{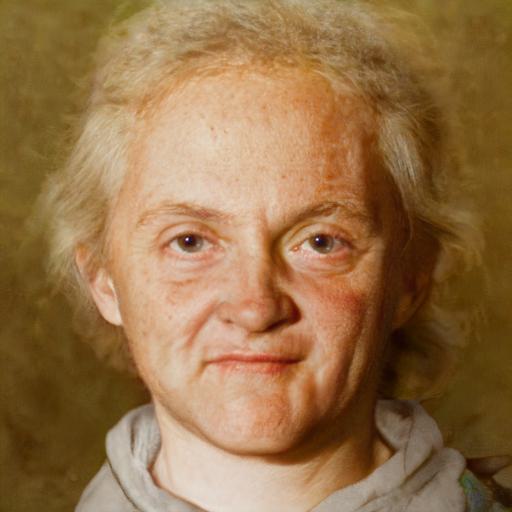} &
 \includegraphics[width=\figs\linewidth]{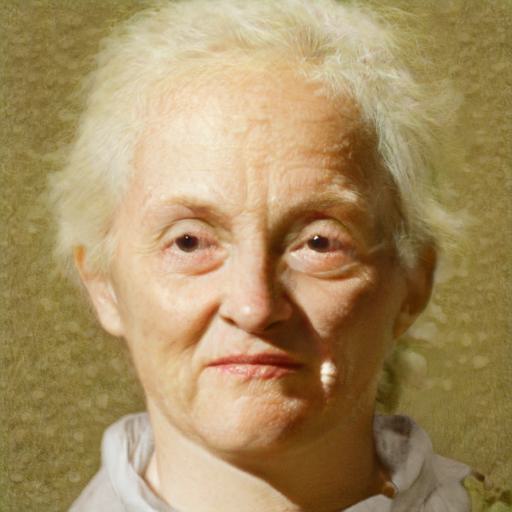} &
 \includegraphics[width=\figs\linewidth]{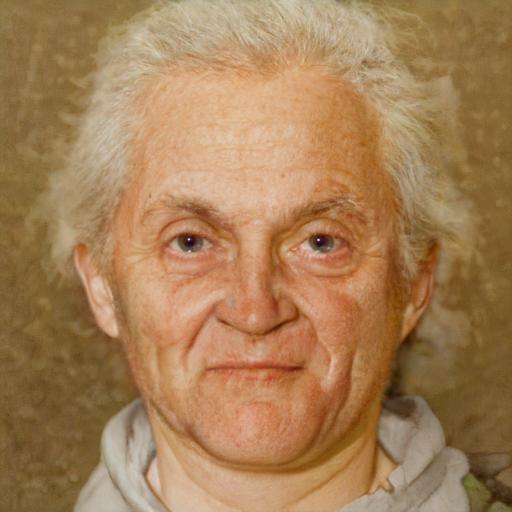} &
 \rotatebox[origin=c]{-90}{\hspace{-0.8in}Age} \\

    \includegraphics[width=\figs\linewidth]{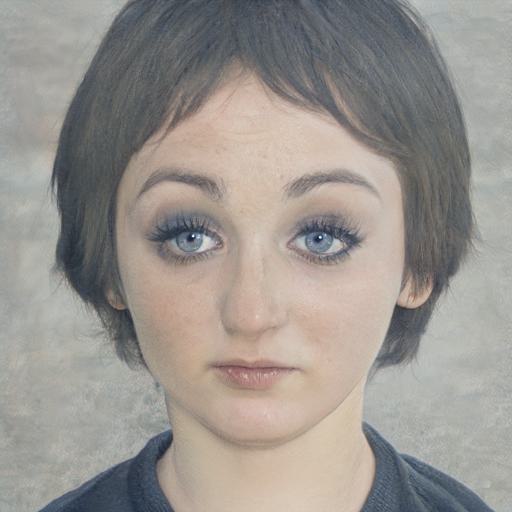} &
  \includegraphics[width=\figs\linewidth]{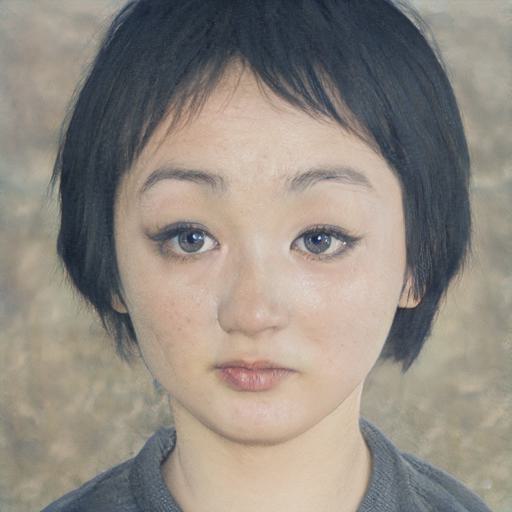} &
 \includegraphics[width=\figs\linewidth]{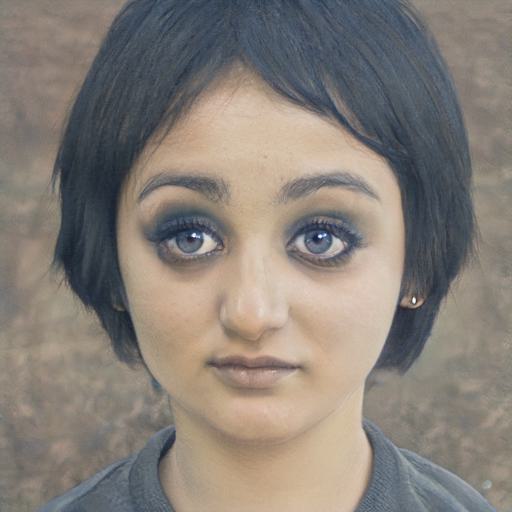} &
 \includegraphics[width=\figs\linewidth]{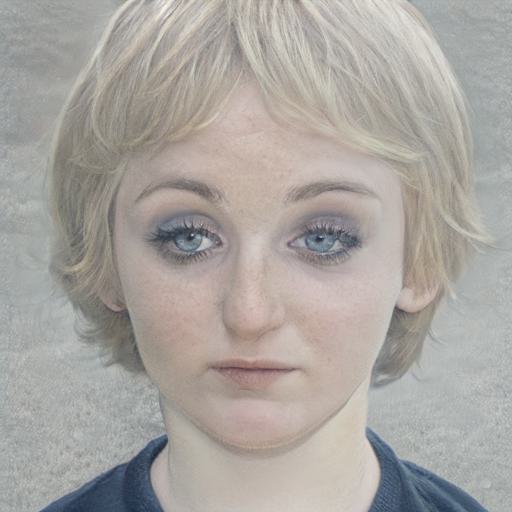} &
 \includegraphics[width=\figs\linewidth]{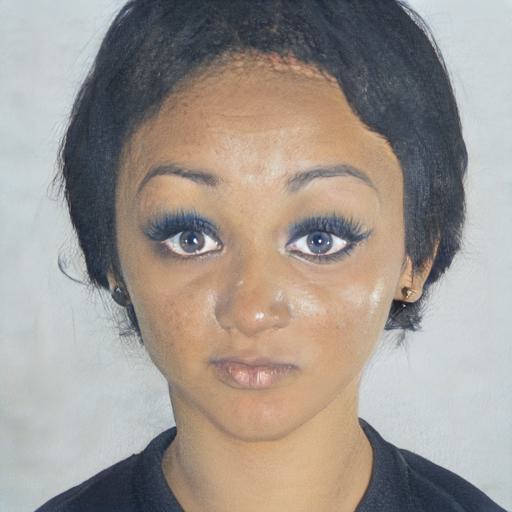} &
 \rotatebox[origin=c]{-90}{\hspace{-0.8in}Race} \\

    \includegraphics[width=\figs\linewidth]{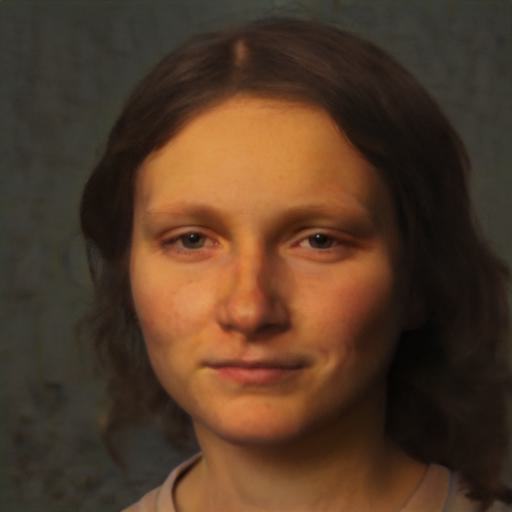} &
  \includegraphics[width=\figs\linewidth]{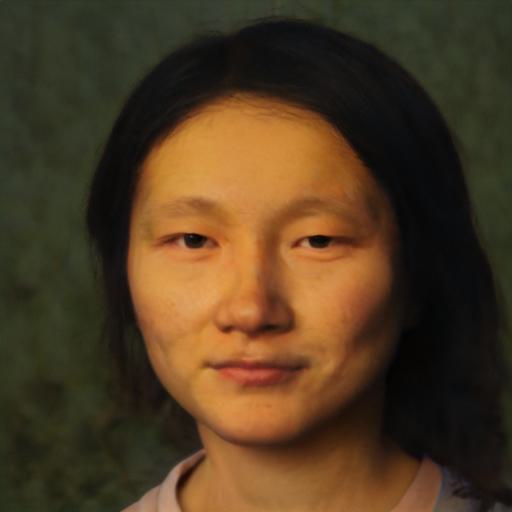} &
 \includegraphics[width=\figs\linewidth]{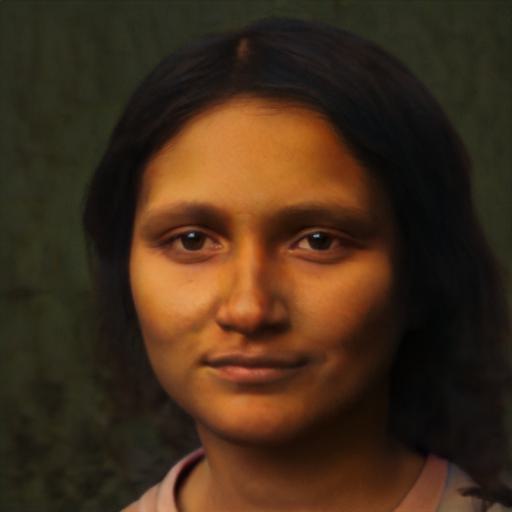} &
 \includegraphics[width=\figs\linewidth]{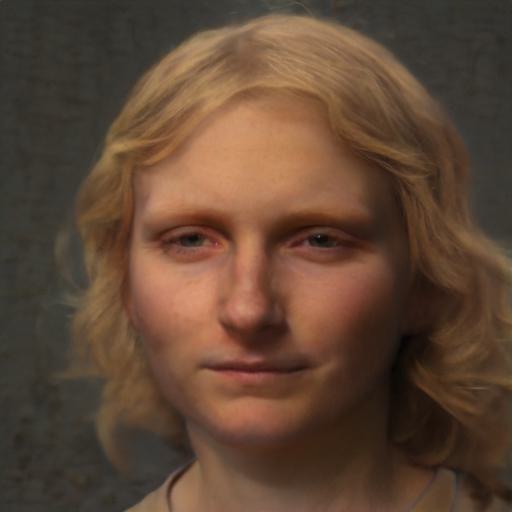} &
 \includegraphics[width=\figs\linewidth]{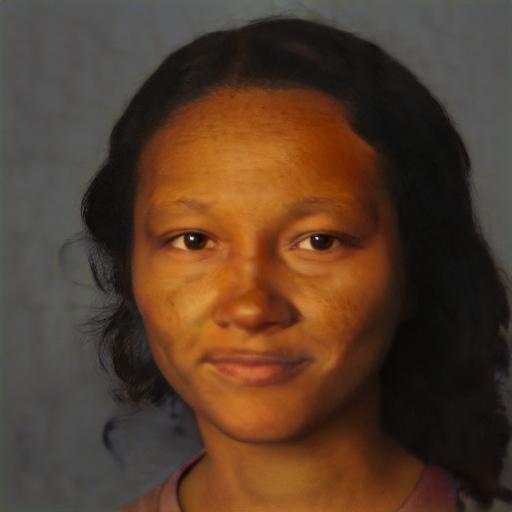} &
 \rotatebox[origin=c]{-90}{\hspace{-0.8in}Race} \\
 
\end{tabular}
\caption{Additional human face editing results. Editing of images outside of the domain of StyleGAN.}
\label{fig:fig3}
\end{figure}



\end{document}